%% file: thesis.tex
\documentclass[reqno,12pt,oneside]{report} 

\usepackage{rac}         
\usepackage{aas_macros}  
\usepackage[intlimits]{amsmath} 
\usepackage{amsxtra}     
\usepackage{amsthm}
\usepackage{amssymb}
\usepackage{amsfonts}
\usepackage{graphicx}    
\usepackage{rotating}
\usepackage{color}
\usepackage{epsfig}

\usepackage{subcaption}

\usepackage{verbatim}
\usepackage[numbers]{natbib}      
\usepackage[printonlyused]{acronym} 
\usepackage{setspace}    
\usepackage[hidelinks]{hyperref}
\usepackage{microtype}
\usepackage{lscape}
\usepackage[normalem]{ulem}

\input{Chapters/algo_setup}

\usepackage{booktabs}   
                        
\usepackage{amsmath}

\usepackage{multirow}
\usepackage{tabularx}
\usepackage{newtxtext,newtxmath, breqn}

\usepackage{rotating}

\theoremstyle{plain}

\theoremstyle{definition}

\theoremstyle{remark}

\newcommand\mybox[1]{\centering \rotatebox{90}{%
    \parbox{\widthof{Performance Impact Per Noise\ }}{\centering #1}}}
    
\newcommand\myboxsm[1]{\centering \rotatebox{90}{%
    \parbox{\widthof{Reverb\ }}{\centering #1}}}
    
\newcommand\myboxmed[1]{\centering \rotatebox{90}{%
    \parbox{\widthof{Perception\ }}{\centering #1}}}

\numberwithin{theorem}{chapter}     

\makeatletter
\def\cleardoublepage{\clearpage\if@twoside \ifodd\c@page\else
\hbox{}
\thispagestyle{empty}
\newpage
\if@twocolumn\hbox{}\newpage\fi\fi\fi}
\makeatother 

\newcommand{\todo}[1]{\vspace{5 mm}\par \noindent
\marginpar{\textsc{To Do}}
\framebox{\begin{minipage}[c]{0.95 \textwidth}
\tt\begin{center} #1 \end{center}\end{mini`page}}\vspace{5 mm}\par}

\begin{document}

\bibliographystyle{plain}    

\titlepage{Implicit Design Choices and Their Impact on Emotion Recognition Model Development and Evaluation}{Mimansa Jaiswal}{Doctor of Philosophy}
{Computer Science and Engineering}{2023}
{Professor Emily Mower Provost, Chair\\
Professor Nikola Bancovic\\
 Professor Benjamin Fish\\
 Douwe Kiela, Contextual AI\\
 Professor V.G. Vinod Vydiswaran}
 

\initializefrontsections



\makeatletter
\if@twoside \setcounter{page}{4} \else \setcounter{page}{1} \fi
\makeatother


\input{FrontPages/orcid}
 
\dedicationpage{Dedicated to my parents and my late granddad, B.P. Bhagat, and my late grandmom, Smt. Radha Devi for their boundless love and support}

\startacknowledgementspage
\input{FrontPages/acknowledgment}
\label{Acknowledgements}


\tableofcontents     
\listoffigures       
\listoftables        


\startabstractpage{}
\input{FrontPages/abstract}
\label{Abstract}

\startthechapters 


 \chapter{Introduction}
 \label{chap:Intro}
 \input{Chapters/chap1}
 

\chapter{Related Work: Modeling Emotions}
\label{chap:relworkmodel}
\input{Chapters/chap2}

 \chapter{Datasets and Pre-processing}
\label{chap:relworkemo}
\input{Chapters/chap3}

\chapter{Emotion Recognition Dataset: MuSE}
\label{chap:muse}
\input{Chapters/chap4}
\chapter{Best Practices for Noise Based Augmentation in Emotion Datasets}
\label{chap:noise_aug}
\input{Chapters/chap5}

\chapter{Context in Crowdsourcing Emotion Annotations}
\label{chap:emotion_annotation}
\input{Chapters/chap6}

\chapter{Controlling for Confounders in Emotion Recognition}
\label{chap:conf}
\input{Chapters/chap7}

\chapter{Reducing Leakage of Demographic and Membership Information Through Adversarial Networks}
\label{chap:leakage}
\input{Chapters/chap8}

 \chapter{Sociolinguistics Based Human-Centered Metric for Emotion Recognition}
 \label{chap:metric}
 \input{Chapters/chap9}

 \chapter{Concluding Remarks}
  \label{chap:conclusion}
  \input{Chapters/chap10}

 
 
  \startbibliography
 \begin{singlespace} 
  \bibliography{thesis}   
 \end{singlespace}

 


 



\end{document}

%% file: Chapters/algo_setup.tex
\usepackage[ruled, noline, fillcomment, ]{algorithm2e}
\usepackage{setspace}
\usepackage{etoolbox}

\usepackage{ amssymb }

\SetCommentSty{mycommfont}
\SetAlgoHangIndent{0.1em}

\SetKwComment{tcp}{$\cdot$\ }{}
\SetKwComment{tcc}{$\cdot$\ }{}

\SetAlCapNameSty{mycapsty}
\makeatletter
\renewcommand{\fnum@algocf}{\AlCapSty{\AlCapFnt}}
\makeatother

\SetAlgoCaptionSeparator{}



\makeatletter
\newcommand{\shortalgrule}{\par \vskip .05\baselineskip \moveright 0.07\columnwidth \vbox{\hrule width0.86\columnwidth}
\par \vskip .05\baselineskip}
\makeatother

\makeatletter
\patchcmd{\@algocf@start}
  {-1.5em}
  {-0.4em}
  {}{}
\makeatother



%% file: FrontPages/orcid.tex
\newpage
\thispagestyle{empty} 
\addtocounter{page}{-1}
\hbox{ }
\vertadjust
\vfill
\begin{center}
Mimansa Jaiswal\\
mimansa@umich.edu\\
ORCID iD: 0009-0001-8290-7743\\
\end{center}
\vfill

%% file: FrontPages/acknowledgment.tex
I would like to start by expressing my profound gratitude towards my PhD advisor, Emily \textit{(Emily Mower Provost)}. She has been my go-to person for any research brainstorming and has shown me tremendous patience, support, and guidance throughout my PhD journey. Without her persistence and suggestions, completing this PhD would not have been possible. I am also incredibly grateful for my thesis committee members: Vinod \textit{(VG Vinod Vydiswaran)}, Nikola \textit{(Nikola Banovic)}, Benjamin \textit{(Benjamin Fish)}, and Douwe \textit{(Douwe Kiela)}. Their valuable insights during my thesis proposal helped shape the final version of the thesis.

For putting up with me during my PhD journey, my immense gratitude goes towards my family, especially my parents, grandparents, and Bert. My parents have been a rock for me over the past six years. Though I could not visit them often or talk to them much, they were always there when I needed someone. They seem to have aged fifteen years in the six years of my PhD, stressed about me, but their support never wavered. My mom \textit{(Archana Kumari)} received her own PhD in 2021, and my dad \textit{(Abhay Kumar)} became the Vice-Chancellor of a new IIIT—two accomplishments that were their lifelong dreams and inspired me immensely. I unfortunately lost two of my grandparents during the PhD program, and I will never forget their blessings and excitement for me embarking on my higher education journey. During early 2020, in the midst of COVID, I adopted a cat named Bert—yes, named after the language model. Without him, I would not have maintained my sanity during the dark, lonely nights and tiring, long work days. His purring loudly into my ear calmed me down on the worst of nights.

I was lucky enough to secure three internships and have amazing research mentors for all of them. Ahmad \textit{(Ahmad Beirami)} taught me how to approach Conversational AI, how to create effective presentations, and how to write research proposals. Adina \textit{(Adina Williams)} taught me how to work with linguistics mixed in with NLP, and how subjectivity can infiltrate seemingly objective parts (like NLI) of NLP. Ana \textit{(Ana Marasović)} was the first person I worked with on really large language models (foundational models), and she taught me how to approach evaluation and benchmarking for generative models—a major part of my current research path.

I want to thank my lab members, starting with Zak \textit{(Zakaria Aldeneh)}. Zak exemplifies what all senior PhD mentors should be, helping me with code, brainstorming, and working with me on papers. He has been an amazing research collaborator. I also want to thank Minxue \textit{(Minxue Sandy Niu)} for being the junior research collaborator anyone would be proud of. She has not only been an amazing collaborator but was also always willing to discuss interesting research problems. I want to thank Matt \textit{(Matthew Perez)} for being the batchmate who has always been there to help, to vent, to advise, and to collaborate, serving as my go-to person for any speech-based research questions. Finally, I want to thank Amrit \textit{(Amrit Romana)} for being an amazing lab member; her observant questions helped me immensely during lab presentations.

I also want to thank my friends, without whom this journey would not have been possible. I will start with Abhinav \textit{(Abhinav Jangda)}, who has been my support system throughout my PhD journey, starting from the application process. Diksha \textit{(Diksha Dhawan)} was the best PhD roommate one could ask for during the first four years of my PhD. She shared laughter and tears with me, cooked with me, and supported me through all the highs and lows. Without her, I could not have survived my PhD. She taught me the value of being proud of my interests in both my personal and professional life, and how friends can sometimes be family, which is the best gift anyone could have given me. Eesh \textit{(Sudheesh Srivastava)}, for all the conversations at the intersection of machine learning, physics, and philosophy, has taught me about areas and theories that I would have otherwise not encountered in any way or format. Conversations with him have always left me rejuvenated, happy, and feeling peppier—a testament to how amazing a best friend he is. Sagarika \textit{(Sagarika Srishti)}, for all her support, both in India and when she came to the US. Her move to the US during my PhD was a major personal highlight. Ariba \textit{(Ariba Javed)}, thanks for all the discussions, talks, and emotional conversations, and for always being up for anything interesting, including a pottery class. Shobhit \textit{(Shobhit Narain)} has been an amazing companion, helping me with job applications and always being the sarcastic, serious, yet most helpful guy I have had the pleasure of calling a friend. And finally, Sai \textit{(Sairam Tabibu)} helped me fill out the PhD application for UMich on the exact deadline, without which, I would not be here at all.

This is probably an unconventional paragraph in acknowledgments, but these were unconventional times during COVID. For the two years of lockdown, I turned to Among Us when I felt lonely or lost in my research. I am really thankful for the streamers whose broadcasts provided some semblance of social interaction. For almost three years, I watched them stream at least 8 hours a day while I worked, to simulate a social environment. And when my research progress stalled, I turned to anonymous Discord communities, playing Among Us and golf for hours, which helped alleviate feelings of depression and sadness, providing a much-needed uplift.

My PhD journey wasn't easy, and a lot happened over the six years, but I made it through. The credit for that goes to all the people mentioned here, to whom I am forever indebted.

%% file: FrontPages/abstract.tex
Emotion recognition is a complex task due to the inherent subjectivity in both the perception and production of emotions. The subjectivity of emotions poses significant challenges in developing accurate and robust computational models. This thesis examines critical facets of emotion recognition, beginning with the collection of diverse datasets that account for psychological factors in emotion production. To address these complexities, the thesis makes several key contributions. 

To handle the challenge of non-representative training data, this work collects the Multimodal Stressed Emotion dataset, which introduces controlled stressors during data collection to better represent real-world influences on emotion production. To address issues with label subjectivity, this research comprehensively analyzes how data augmentation techniques and annotation schemes impact emotion perception and annotator labels. It further handles natural confounding variables and variations by employing adversarial networks to isolate key factors like stress from learned emotion representations during model training. For tackling concerns about leakage of sensitive demographic variables, this work leverages adversarial learning to strip sensitive demographic information from multimodal encodings. Additionally, it proposes optimized sociological evaluation metrics aligned with cost-effective, real-world needs for model testing.

The findings from this research provide valuable insights into the nuances of emotion labeling, modeling techniques, and interpretation frameworks for robust emotion recognition. The novel datasets collected help encapsulate the environmental and personal variability prevalent in real-world emotion expression. The data augmentation and annotation studies improve label consistency by accounting for subjectivity in emotion perception. The stressor-controlled models enhance adaptability and generalizability across diverse contexts and datasets. The bimodal adversarial networks aid in generating representations that avoid leakage of sensitive user information. Finally, the optimized sociological evaluation metrics reduce reliance on extensive expensive human annotations for model assessment.

This research advances robust, practical emotion recognition through multifaceted studies of challenges in datasets, labels, modeling, demographic and membership variable encoding in representations, and evaluation. The groundwork has been laid for cost-effective, generalizable emotion recognition models that are less likely to encode sensitive demographic information.

%% file: Chapters/chap1.tex
In human communication, perceiving and responding to others' emotions in interpersonal conversations play a crucial role~\cite{fussell2002verbal}. To create systems that can aid in human-centered interpersonal situations, it is necessary for these systems to possess the capability to recognize emotions effectively~\cite{varghese2015overview}. 
Robust Emotion Recognition (ER) models can be beneficial in various situations, such as crisis text lines or passive mental health monitoring~\cite{mohammad2022ethics}. However, these ML models often lack robustness when faced with unseen data situations, making deploying them in high-risk situations or healthcare a challenging task~\cite{zhao2019differential}.

Recongizing emotion is a challenging task because it is subjective in both perception and production~\cite{poria2019emotion}. 
The labels used to train emotion recognition models are perceptually subjective~\cite{bianchi2002modeling}. The same emotion can be perceived differently by different people, depending on their cultural background, personal experiences, and other factors~\cite{matsumoto1989cultural}.  Additionally, there is production subjectivity.  The same emotion can be expressed differently by different people, depending on their individual personality, cultural background, physiological and other factors~\cite{averill1980constructivist}. The subjectivity of emotion recognition makes it difficult to develop accurate and robust models that account for these numerous variations~\cite{vastfjall2003subjective}.

In addition to the challenges posed by subjectivity, there are challenges that relate to the information that is learned in addition to and beyond the expression of emotion itself.  The manner in which emotions are expressed are correlated with a person's demographic and identifying features. Hence, systems trained to recognize emotion can often learn implicit associations between an individual's demographic factors and emotion~\cite{stromfelt2017emotion}. When used as a component in larger systems, these implicit associations can  lead to either the leakage of demographic information, or can bias the larger system's output based on demographic information, even when not explicitly trained to do so.

Training any robust machine learning model necessitates having access to large amounts of diverse and labelled data. Training models for emotion recognition faces the challenge of not having access to large quantities of diverse data. 
Scraping data over the internet, as is done for other areas, leads to a dataset that is often demographically biased, and, often exaggerated for entertainment purposes. 
On the other hand, data collected in laboratory environments is intentionally cleaner and often exaggerated in case of scripted sessions. Therefore, both of these data collection methods do not encapsulate possible environmental and personal factors, which leads to models often being trained on either highly skewed or non-representative data. 
The resulting models are either fragile or biased, and ultimately unable to handle real-world variability.

In this dissertation, critical facets of emotion recognition are thoroughly explored, beginning with the collection of datasets, which take into account psychological factors in producing emotions. This is followed closely by examining the influence that alterations in data augmentation processes have on emotion labels, while also challenging and interrogating the validity of previously established labels. Alterations in labeling techniques and the resulting effects on annotator-assigned labels are also scrutinized. Simultaneously, the research develops robust models specifically trained to disregard certain physiological emotion production factors. Integral to the research is the creation of bimodal models that generate representations aiming to tackle the reduction of leakage of sensitive demographic variables. The concluding portion of the study involves an in-depth evaluation of the robustness and impartiality of these models, carried out in a human-centric manner, ensuring an emphasis on minimal costs for data annotation. From this extensive research, valuable insights are gained into the complexities of emotion recognition, which pave the way for more nuanced and robust labeling, modeling, and interpretation techniques. It also lays the groundwork for future efforts in the development of robust and cost-effective emotion recognition models.

\section{Emotion Theories and the Impact on Emotion Recognition Model Development}

To better understand the subjectivity inherent in emotion recognition and its correlation with the research gaps and challenges, we must first explore the contrasts between emotion production and emotion perception theories. These theories elucidate the distinct factors related to the subjectivity of emotions in both production and recognition processes and offer valuable insights for developing robust and unbiased emotion recognition models.

\subsection{Emotion Production and Emotion Perception}

Emotion production refers to experiencing and generating emotional responses, encompassing several factors, including cognitive appraisal, physiological response, behavior and expression, and subjective experience. These components work together to create the unique process of producing emotions within each person.

Emotion perception, conversely, focuses on recognizing and interpreting others' emotional signals, influenced by factors such as emotional cues, context and environment, past experiences and learning, and individual differences. This process involves making sense of others' emotions based on various internal and external factors.

\subsection{Emotion Theories, Research Challenges, and Implications for Emotion Recognition}

Various theories of emotion provide insights into the challenges faced in developing computational models for emotion recognition in speech or text. Below, we discuss the relevance and implications of some prominent theories in the context of speech or text-based (bimodal) emotion recognition.

\begin{itemize}

\item James-Lange Theory and Cannon-Bard Theory~\cite{staats1990paradigmatic}: Both theories emphasize physiological responses' importance in emotion. In speech or text-based recognition, it is vital to consider correlations between observable features (e.g., vocal tonality, speech patterns) and underlying physiological responses. Accounting for these correlations can help capture emotions, even though the relationship might be subjective due to personal and cultural differences.

\item Schachter-Singer Two-Factor Theory~\cite{staats1990paradigmatic}: This theory stresses the importance of both physiological arousal and cognitive appraisal for experiencing emotions. In speech or text-based emotion recognition, cognitive appraisal aspects such as semantic content, contextual factors, and discourse patterns can be extracted. However, the subjectivity of cognitive appraisal processes presents challenges given personal experiences' impact on interpretation.

\item Lazarus Cognitive-Mediational Theory~\cite{staats1990paradigmatic}: Centered around the role of cognitive appraisal, this theory highlights the need for emotion recognition systems to account for individuals' interpretations of situations through cues that may suggest appraisal (e.g., word choice, phrase structure, conversational context). Advanced models might need to factor in users' personal and demographic features to better understand cognitive appraisal processes. This approach introduces more subjectivity and potential privacy concerns, as individual perspectives and experiences can vary significantly.

\end{itemize}

Integrating insights from these theories can aid unraveling the complexities and subjective nature of emotions expressed through language, as speech or text-based emotion recognition relies primarily on linguistic patterns, tone, and content analysis.

\subsection{Addressing Challenges Through Thesis Contributions}

The thesis contributions align with and address the subjectivity challenges in emotion production and perception, thus tackling the complexities involved in developing robust and unbiased emotion recognition models.

\begin{itemize}

\item Collecting datasets that account for psychological factors in emotion production: By considering psychological factors influencing unique emotional experiences, more diverse datasets are created, allowing models to account for subjectivity in emotion production and generalize across emotions.

\item Examining the influence of data augmentation processes on emotion perception labels: This contribution seeks to understand data augmentation's impact on ground truth labels, creating better representations of emotions in the datasets, accounting for subjectivity in emotion perception.

\item Analyzing labeling setups' impact on annotators' emotion perception labels: This investigates how labeling setups influence emotion perception, aiming to improve label consistency and reduce inter-annotator disagreement, thus better representing subjectivity in emotion perception.

\item Training robust models by explicitly disregarding emotion production factors: This minimizes the impact of subjective elements associated with emotion production, enabling models to focus on core emotional cues.

\item Developing bimodal models for generation of emotion representations that are debiased and reduce encoding of demographic and membership information: This creates models that consider multiple emotional cues while disregarding sensitive features, addressing subjectivity challenges in both emotion production and perception.

\item Evaluating models in a human-centric manner: Designing evaluation methods aligned with real-world expectations and without incurring significant annotation costs ensures the models effectively tackle subjectivity challenges in a practical way.

By focusing on these contributions, the thesis emphasizes the connection between emotion production and perception's subjectivity and its influence on model development, advancing the creation of more robust and unbiased emotion recognition models.
\end{itemize}

\section{Emotion Recognition}
Emotion recognition models are customarily trained using laboratory-collected data encompassing video, audio, and corresponding text. These algorithms strive to capture the speaker's underlying emotional state either autonomously or as part of a larger pipeline, such as response generation. Supervised learning techniques predominantly train these models. Obtaining ground truth labels for the dataset samples is crucial for successfully training a supervised learning model. The emotion theories presented earlier are intrinsically linked with the complexity of emotion recognition. Understanding the interplay between these theories and model development is essential.

\subsection{Emotion Labels}
Emotion labels typically fall into two categories: categorical and dimensional. Categorical variables aim to discretely categorize emotion attributes, such as excitement, happiness, anger, or sadness. These labels' limitations align with the James-Lange and Cannon-Bard theories—emotions are subjective, making it difficult to define universal emotions across cultures. This subjectivity is intensified by both personal physiological responses to stimuli and cultural context.

Dimensional emotional labels describe emotions across two dimensions, valence (sad to happy) and arousal (calm to excited). The dimensional approach is more consistent with the James-Lange and Cannon-Bard theories, addressing the physiological components of emotions, as well as the cognitive components emphasized by the Schachter-Singer Two-Factor Theory and Lazarus Cognitive-Mediational Theory. However, these dimensional labels also face the challenge of cultural and personal influences on the perception and expression of emotions.

\subsection{Emotion Features}
Three primary modalities are used in combination to train emotion recognition models: text, audio, and video. This thesis focuses predominantly on audio and its corresponding text as the feature set for these models.

Mel-filterbanks (MFBs) are often used as inputs to neural network models in speech. MFBs can capture correlations between vocal tonality, speech patterns, and underlying physiological responses. Nevertheless, factors like pitch, volume, or other nuances of speech may be affected by cultural and linguistic contexts. Furthermore, personal characteristics can influence these features, further complicating emotion recognition in cross-cultural or highly diverse settings.

Language features, which provide contextualized representations for words, capture the cognitive appraisal aspects (semantic content, contextual factors, and discourse patterns). The Lazarus Cognitive-Mediational Theory further highlights the need for models that account for user demographics. More advanced models may need to balance the understanding of individual emotions with ethical considerations.

\subsection{Emotion Recognition Models}
Audio-based emotion recognition models initially relied on Hidden Markov Models (HMMs) or Gaussian Mixture Models (GMMs) and later shifted focus to LSTMs and RNNs. These models aim to capture the dynamic and time-varying nature of speech, reflecting the James-Lange Theory and Cannon-Bard Theory's emphasis on physiological responses. However, these models must also account for the inherent cultural and linguistic differences in the way emotions are expressed through speech.

Language-based models, like recent advances in transformer architectures, address long and indirect contextual information challenges, in line with the Schachter-Singer Two-Factor Theory's cognitive appraisal aspects. These models strive to understand the nuances of language, cultural expressions, and individual semantic and contextual differences in recognizing emotions.

Multi-modal models exploit relevant information from text, audio, or video to form powerful emotion recognition models. Informed by the emotion theories, these models take into account the subjectivity of emotions by leveraging different modalities to discern the nuances of emotion expression. By combining these modes, models can better account for the emotional complexity that arises from intercultural and personal differences in perception, expression, and context.

\section{Challenges in Emotion Recognition}
The variable and subjective nature of emotions make it challenging to train models that can accurately identify emotion in any given scenario. Addressing three major challenges is necessary for any emotion recognition model deployed in a real-world setting: (a) Non-representative training data, (b) Subjective labels, (c) Unintentional encoding and leakage of sensitive information. Previous work has looked at varying ways to counter these challenges, talked about in detail in Chapter~\ref{chap:relworkemo}, Section~\ref{sec:chaprelworkemo:concernswithemorecog}.

\subsection{Non-representative data}
Emotion production in real-world settings is influenced by various factors, including data collection settings, demographics, and personal factors. Addressing these confounding factors aligns with the implications of the earlier-discussed emotion theories. Researchers can tackle this challenge by developing more robust models, incorporating real-world variability through dataset augmentation or mitigating confounding factors.

\subsection{Label Subjectivity}
As highlighted in the emotion theories, emotions are inherently subjective and deeply influenced by personal experiences, culture, and context. This subjectivity leads to difficulty in pinpointing an objective and universal ground truth for training emotion recognition models. Researchers should account for label subjectivity by using diverse and representative datasets, annotations from multiple sources, and considering multiple emotion theories during the model design process.

\subsection{Unintentional encoding and leakage of sensitive information}
Variability can lead unintentional encoding and leakage of sensitive information concerns, specifically in human centered tasks, such as emotion recognition models, as the associative nature of the task and sensitive demographic variables may inadvertently lead to encoding personal information.

\section{Proposed Methods}

A robust and effective emotion recognition system must successfully navigate a range of challenges, including addressing subjectivity in emotion production and perception, handling natural variations and confounding variables, reducing encoded sensitive information, and providing relevant evaluation metrics. Here, we present a series of proposed methods aligned with the outlined contributions to address these challenges.

\subsection{Dataset Collection for Emotion Recognition}

Tackling the challenge of subjectivity in emotion production, it's essential that we consider the issues in widely used emotion recognition datasets that arise due to design choices, methodology of data collection, and inherent subjectivity. Emotion datasets traditionally aim for minimal variation to ensure generalizability. However, this can result in non-robust models that struggle with unexpected variability. We propose the construction and validation of a new dataset called Multimodal Stressed Emotion (MuSE), which introduces a controlled situational confounder (stress) to better account for subjectivity. In addition, we discuss the use of domain adversarial networks to achieve more stable and reliable cross-corpus generalization while avoiding undesired characteristics in encodings.

\subsection{Data Augmentation with Noise in Emotion Datasets}

Addressing the challenge of subjectivity in emotion perception, we examine data augmentation with noise in emotion datasets, focusing on the Interactive Emotional Dyadic Motion Capture (IEMOCAP) dataset, which features dyadic interactions with text, video, and audio modalities. Introducing realistic noisy samples through environmental and synthetic noise, we evaluate how ground truth and predicted labels change due to noise sources. We discuss the effects of commonly used noisy augmentation techniques on human emotion perception, potential inaccuracies in model robustness testing, and provide recommendations for noise-based augmentation and model deployment.

\subsection{Annotations of Emotion Datasets}

To further address subjectivity in emotion perception, we investigate how design choices in the annotation collection process impact the performance of trained models. Focusing on contextual biasing, we examine how annotators perceive emotions differently in the presence or absence of context. Commonly-used emotion datasets often involve annotators who have knowledge of previous sentences, but models are frequently evaluated on individual utterances. We explore the implications of this discrepancy on model evaluation, and its potential for generating errors.

\subsection{Methods for Handling Natural Variations and Confounding Variables}

As mentioned earlier, we collect a dataset of differences in similar emotion production under varying levels of stress. Emotion recognition models may spuriously correlate these stress-based factors to perceived emotion labels, which could limit generalization to other datasets. Consequently, we hypothesize that controlling for stress variations can improve the models' generalizability. To achieve this, we employ adversarial networks to decorrelate stress modulations from emotion representations, examining the impact of stress on both acoustic and lexical emotion predictions. By isolating stress-related factors from emotion representations, we aim to enhance the model's ability to generalize across different stress conditions. Furthermore, we analyze the transferability of these refined emotion recognition models across various domains, assessing their adaptability to evolving contexts and scenarios. Ultimately, our approach aims to improve emotion recognition model robustness by addressing the inherent variability of emotional expression due to stress and ensuring greater applicability across multiple domains.

\subsection{Approaches for Tackling Sensitive Information Leakage in Trained Emotion Recognition Models}

Emotions are inherently related to demographic factors such as gender, age, and race. Consequently, emotion recognition models often learn these latent variables even if they are not explicitly trained to do so. This learning behavior poses a risk to user privacy, as the models inadvertently capture sensitive demographic information. Storing representations instead of raw data does not fully mitigate this issue, as latent variables can still compromise user privacy. To address this challenge, we present approaches for mitigating the learning of certain demographic factors in emotion recognition embeddings. Furthermore, we tackle the issue of user-level membership identification by employing an adversarial network that strips this information from the final encoding, reduced leakage of sensitive information from generated representations.

\subsection{Methods for Model Evaluation and Perception}

Large language models face limitations in subjective tasks like emotion recognition due to inadequate annotation diversity and data coverage. Acquiring comprehensive annotations and evaluations is often costly and time-consuming. To address these challenges, we propose cost-effective sociological metrics for emotion generalization and reduced demographic vairable leakage. These metrics reduce reliance on expensive human-based feedback while still capturing the nuances of human emotions. By evaluating model performance and demographic variables encoded in generated representations, the proposed metrics improve cross-corpus results and allow for the development of accurate, relevant emotion recognition models in a more economic manner.

\section{Contributions}
This dissertation proposes several investigations and novel solutions to address various concerns related to real-world emotion recognition model deployment. 

The contributions of the works in this dissertation can be summarized as follows:

\begin{itemize}

\item  Chapter~\ref{chap:muse}:
    \begin{itemize}
       \item  Introduction of Multimodal Stressed Emotion (MuSE) dataset.
       \item  Detailed data collection protocol.
       \item  Potential uses and emotion content annotations.
       \item  Performance measuring baselines for emotion and stress classification.
    \end{itemize}

\item  Chapter~\ref{chap:noise_aug}:
    \begin{itemize}
       \item  Speech emotion recognition's impact under influence of various factors such as noise.
       \item  Investigation of noise-altered annotation labels and their aftermath.
       \item  Consequences on evaluation of ML models considering noise.
       \item  Specific recommendations for noise augmentations in emotion recognition datasets.
    \end{itemize}

\item  Chapter~\ref{chap:emotion_annotation}:
    \begin{itemize}
       \item  Crowdsourced experiments to study the subjectivity in emotion expression and perception.
       \item  Contextual and randomized annotation schemes of the MuSE dataset.
       \item  Comparative analysis revealing contextual scheme's closeness to speaker's self-reported labels.
    \end{itemize}

\item  Chapter~\ref{chap:conf}:
    \begin{itemize}
       \item  Examination of emotion expressions under stress variations.
       \item  Utilization of adversarial networks to separate stress modulations from emotion representations.
       \item  Exploration of stress’s impact on acoustic and lexical emotional predictions.
       \item  Evidence of improved generalizability with stress control during model training.
    \end{itemize}

\item  Chapter~\ref{chap:leakage}:
    \begin{itemize}
       \item  Highlighting the unintentional leak of sensitive demographic information in multimodal representations.
       \item  Use of adversarial learning paradigm to improve sensitive information reduction metric.
       \item  Maintenance of primary task performance, despite improvements to privacy.
    \end{itemize}

\item  Chapter~\ref{chap:metric}:
    \begin{itemize}
       \item  New template formulation to derive human-centered, optimizable and cost-effective metrics.
       \item  Correlation establishment between emotion recognition performance, biased representations and derived metrics.
       \item  Employment of metrics for training an emotion recognition model with increased generalizability and decreased bias.
       \item  Finding of positive correlation between proposed metrics and user preference.
    \end{itemize}

\end{itemize}

\section{Outline of the dissertation}
Initiating with Chapter~\ref{chap:relworkemo}, it delves into a comprehensive review of pertinent literature spanning from emotion recognition and privacy preservation to adversarial networks, model interpretability, and crowdsourcing designs. Moving forward, Chapter~\ref{chap:relworkmodel} provides an introduction to the common datasets, and features employed throughout this research. Subsequent chapters, from Chapter~\ref{chap:muse}  to~\ref{chap:metric}, engage in a thorough exploration and discussion of the research work undertaken, characterized in the Contributions section. Lastly, Chapter~\ref{chap:conclusion} serves as a conclusive summary encapsulating the primary contributions made, elaborating on the proposed future works.

%% file: Chapters/chap2.tex
    Emotion recognition is a complex, multifaceted field drawing on various research areas. This chapter explores the various methods and considerations in this field, from the use of crowdsourcing to the importance of context, and from handling confounding factors to the impact of noise on machine learning models. We explore the ethical considerations of unintentional encoding of sensisitive variables in data collection and neural networks, the role of interpretability in model trustworthiness, and the importance of automating human in the loop feedback. We also delve into the challenge of generalizability in emotion recognition.

    \section{Concerns with Emotion Recognition Datasets}
    \label{sec:chaprelworkemo:concernswithemorecog}

    Some aspects of the above mentioned datasets limit their applicability, including: a lack of naturalness, unbalanced emotion content, unmeasured confounding variables, small size, small number of speakers, and presence of background noise. These datasets are also limited in the number of modalities they use, usually relying on visual and acoustic/lexical information. 
    
    \subsection{Recorded Modalities}
    As shown in Table~\ref{all:datasets}, the most common modalities are video, acoustics, and text. In addition to these modalities, we chose to record two more modalities: thermal and physiological.  Previous research has shown that thermal recordings perform well as non-invasive measurement of physiological markers like, cardiac pulse and skin temperature \cite{pavlidis2000thermal,pavlidis2002thermal,garbey2007contact}.  They have been shown to be correlated to stress symptoms, among other physiological measures.
    We used the physiological modality to measure stress responses \cite{yaribeygi2017impact,Activity-Aware-Mental-Stress-Detection-Using-Physiological-Sensors} to psychological stressors. This modality has been previously noted in literature for measuring stress \cite{horvath1978experimental}, usually measured in polygraph tests. 
    We perform baseline experiments to show that the modalities collected in the dataset are indeed informative for identifying stress and emotion.
    
    \subsection{Lack of Naturalness}
    A common data collection paradigm for emotion is to ask actors to portray particular emotions. These are usually either short snippets of information~\cite{busso2008iemocap}, a single sentence in a situation~\cite{busso2017msp}, or obtained from sitcoms and rehearsed broadcasts~\cite{friends_dataset}. A common problem with this approach is that the resulting emotion display is not natural~\cite{jurgens2015effect}. These are more exaggerated versions of singular emotion expression rather than the general, and messier, emotion expressions that are common in the real world~\cite{audibert2010prosodic,batliner1995can,fernandez2013emotion}.  Further, expressions in the real world are influenced by both conversation setting and psychological setting. 
    While some datasets have also collected spontaneous data~\cite{busso2008iemocap,busso2017msp}, these utterances, though emotionally situated, are often neutral in content when annotated.
    The usual way to get natural emotional data is to either collect data using specific triggers that have been known to elicit a certain kind of response or to completely rely on in-the wild data, which however often leads to unbalanced emotional content in the dataset~\cite{ringeval2013introducing}. 
    
    \subsection{Unbalanced Emotion Content}
    In-the-wild datasets are becoming more popular~\cite{friends_dataset,khorram2018priori,li2016towards}. The usual limitation to this methodology is that, firstly, for most people, many conversations are neutral in emotion expression. This leads to a considerable class imbalance~\cite{ringeval2013introducing}. 
    To counter this issue, MSP-Podcast~\cite{lotfian2017building} deals with unbalanced content by pre-selecting segments that are more likely to have emotional content.
    Secondly, data collected in particular settings, e.g., therapy~\cite{nasir2017predicting}, or patients with clinical issues~\cite{Lassalle2019} comprise mostly of negative emotions because of the recruitment method used in the collection protocol.
    
    \subsection{Presence of Interactional Variables}
    The common way of inducing emotions involves either improvisation prompts or scripted scenarios.
    Emotion has been shown to vary with a lot of factors that are different from the intended induction~\cite{siedlecka2019experimental,zhang2014does,mills2014validity}. These factors in general can be classified into: (a) recording environment confounders and (b) collection confounders. Recording environment-based variables hamper the models' ability to to learn the emotion accurately. These can be environment noise~\cite{banda2011noise}, placement of sensors or just ambient temperature~\cite{bruno2017temperature}. 
    
    \begin{landscape}
    \begin{table*}[t]
    \addtolength{\tabcolsep}{-3.2pt}
        \centering
            \caption{Summary of some of the existing emotion corpora. Lexical modality is  mentioned for manually transcribed datasets. A - Audio, L - Lexical, T- Thermal, V- Visual, P - Physiological.\newline}
        \begin{tabular}{llcccccc}
        \hline
            &Corpus & Size & Speakers & Rec. Type & Language & Modality & Annotation Type \\
            \hline
            1.&IEMOCAP & 12h26m & 10 & improv/acted & English & A, V, L & Ordinal, Categorical\\
            2.&MSP-Improv & 9h35m & 12 & improv/acted & English & A, V & Ordinal\\
            3.&VAM & 12h & 47 & spontaneous & German & A, V & Ordinal\\
            4.&SEMAINE & 6h21m & 20 & spontaneous & English & A, V & Ordinal, Categorical\\
            5.&RECOLA & 2h50m & 46 & spontaneous & French & A, V, P & Ordinal\\
            6.&FAU-AIBO & 9h12m & 51 & spontaneous & German & A, L & Categorical\\
            7.&TUM AVIC & 0h23m & 21 & spontaneous & English & A, V, L & Categorical \\
            8.&Emotion Lines & 30k samples & - & spont/scripted & English & A, L & Categorical\\
            9.&OMG-Emotion & 2.4k samples & - & spontaenous & English & A, V, L & Ordinal\\
            10.&MSP-Podcast & 27h42m & 151 & spontaenous & English & A & Ordinal, Categorical\\
            11.&MuSE & 10h & 28 & spontaneous & English & A, V, L, T, P & Ordinal (Random, Context)\\
            \hline
        \end{tabular}
        \label{all:datasets}
    \end{table*}
    \end{landscape}
    
    \subsection{Demographics in Dataset Collection Recruitment}
    The data collection variations influence both the data generation and data annotation stages.  The most common confounders are gender, i.e., ensuring an adequate mix of male vs female, and culture, i.e., having a representative sample to train a more general classifier. Another confounding factor includes personality traits~\cite{zhao2018personality}, which influence how a person both produces~\cite{zhao2018personality} and perceives~\cite{mitchell2006relationship} emotion. Another confounder that can occur at the collection stage is the familiarity between the participants, like RECOLA~\cite{ringeval2013introducing}, which led to most of the samples being mainly positive due to the colloquial interaction between the participants. They also do not account for the psychological state of the participant. Psychological factors such as 
    stress~\cite{lech2014stress}, anxiety~\cite{werner2011assessing} and fatigue~\cite{berger2012acute} have been shown previously to have significant impact on the display of emotion. But the relation between these psychological factors and the performance of models trained to classify emotions in these situations has not been studied. 

\section{Crowdsourcing and Context in Emotion Recognition}

Crowdsourcing has emerged as a highly efficient approach for gathering dependable emotion labels, as extensively investigated by Burmania et al.~\cite{burmania2016tradeoff}. In addition to this, previous studies have concentrated on enhancing the dependability of annotations by employing quality-control methods. For instance, Soleymani et al.~\cite{soleymani2010crowdsourcing} have proposed the utilization of qualification tests to weed out spammers from the crowd, thereby ensuring the quality of collected data. Furthermore, Burmania et al.~\cite{burmania2016increasing} have explored the use of gold-standard samples to continuously monitor the reliability and fatigue levels of annotators.

The interpretation of emotions is heavily influenced by the context in which they are expressed. Various factors such as tone, choice of words, and facial expressions can significantly impact how individuals perceive and understand emotions~\cite{laplante2003things}. It is noteworthy that this contextual information is implicitly incorporated in the labeling schemes of commonly used emotion datasets like IEMOCAP~\cite{busso2008iemocap} and MSP-Improv~\cite{busso2017msp}. However, a notable disparity often exists between the information available to human annotators and that accessible to emotion classification systems. This discrepancy arises because emotion recognition systems are typically trained on individual utterances~\cite{aldeneh2017using,abdelwahab2017incremental,mirsamadi2017automatic,sarma2018emotion}.

\section{Handling Confounding Factors}

\subsection{Singularly Labeled or Unlabeled Factors}

To address confounding factors that are either labeled singularly or cannot be labeled, researchers have devised specific methods. For instance, Ben-David et al.~\cite{ben2010theory} conducted a study wherein they showed that a sentiment classifier, trained to predict the sentiment expressed in reviews, could also implicitly learn to predict the category of the products being reviewed. This finding highlights the potential of classifiers to capture additional information beyond their primary task. In a similar vein, Shinohara~\cite{shinohara2016adversarial} employed an adversarial approach to train noise-robust networks for automatic speech recognition. By leveraging this technique, Shinohara aimed to enhance the network's ability to handle noisy and distorted speech signals.
\subsection{Explicitly Labeled Factors}

In addition to addressing confounding factors that are singularly or unlabeled, researchers have also developed methods to handle confounding factors that are explicitly labeled during the data collection process. One such approach involves the use of adversarial multi-task learning, which aims to mitigate variances caused by speaker identity~\cite{meng2018speaker}. By incorporating this technique, researchers can reduce the influence of speaker-specific characteristics on the emotion recognition system, thereby enhancing its generalizability. Furthermore, a similar approach has been employed to prevent networks from learning publication source characteristics, which could introduce biases in the classification process~\cite{satire-adv}

\section{Noise and Approaches to Dealing with it in Machine Learning Models}

The impact of noise on machine learning models has been the subject of extensive research, which can be broadly classified into three main directions: robustness in automatic speech recognition, noise-based adversarial example generation, and performance improvement through model augmentation with noise.

One area of focus is the robustness of models in automatic speech recognition (ASR) when exposed to noisy environments. Researchers have explored various techniques to enhance the performance of ASR systems in the presence of noise. This includes the development of noise-robust feature extraction methods, such as mel-frequency cepstral coefficients (MFCCs) and perceptual linear prediction (PLP) features~\cite{li2014overview}. These techniques aim to minimize the impact of noise on the accuracy of speech recognition systems, enabling them to effectively operate in real-world, noisy conditions.

Another line of research involves the generation of noise-based adversarial examples, which are intentionally crafted to deceive machine learning models. Adversarial attacks exploit vulnerabilities in models by adding imperceptible noise to input samples, causing the models to misclassify or produce incorrect outputs. Carlini and Wagner~\cite{carlini2018audio} and Gong et al.~\cite{gong2017crafting} have proposed methodologies for generating adversarial audio examples that can fool ASR systems. These techniques highlight the importance of understanding and addressing the susceptibility of machine learning models to adversarial noise.

Furthermore, researchers have explored the potential benefits of incorporating noise during the training and augmentation process of machine learning models. By augmenting the training data with various types of noise, models can become more robust and adaptable to real-world conditions. For instance, Sohn et al.~\cite{sohn2020fixmatch} and Wallace et al.~\cite{wallace2019allennlp} have investigated the effectiveness of noise augmentation techniques in improving the performance of models across different tasks. These methods aim to enhance model generalization and reduce overfitting, ultimately leading to better model performance in noise-affected scenarios.

While evaluating model robustness to noise or adversarial attacks, researchers commonly introduce noise into the dataset and assess the model's performance~\cite{abdullah2019practical}. However, when it comes to emotion recognition, introducing noise while ensuring that the perception of emotions remains intact can be highly challenging. It is crucial to strike a balance between adding noise for robustness evaluation purposes and preserving the original emotional content. This ensures that the introduced noise does not distort or alter the true emotional expression, enabling accurate and reliable emotion recognition systems.

\section{Unintentional Sensisitve Variable Encoding, and Ethical Considerations in Data Collection and Neural Networks}

The preservation of privacy in data collection has been a key area of focus in early research. Various methods such as rule-based systems and the introduction of background noise have been explored in order to achieve this goal~\cite{gomez2010data, evfimievski2002randomization}. However, more recent studies have shifted their attention towards privacy preservation in the context of neural networks. In particular, researchers have primarily concentrated on ensuring that the input data used in these networks are not memorized and cannot be retrieved even when the model is deployed~\cite{carlini2019secret, abadi2016deep}.

Another crucial consideration in the field of privacy preservation is fair algorithmic representation. The objective here is to develop networks that are invariant to specific attributes, often related to demographic information, in order to ensure fairness~\cite{bolukbasi2016man, corbett2018measure, davidson2019racial}. Although certain methods have demonstrated promise in achieving fairness, they may still inadvertently lead to privacy violations~\cite{jaiswal2020privacy}.

\section{The Role of Interpretability in Model Trustworthiness}

The aspect of interpretability plays a crucial role in establishing trustworthiness of models. Studies have indicated that individuals are more inclined to trust the decisions made by a model if its explanations align with their own decision-making processes~\cite{sperrle2019human, ferrario2019ai, schmidt2019quantifying}. In addition, interpretability methods can be employed by model designers to evaluate and debug a trained model~\cite{engler2021auditing}. These methods provide insights into the inner workings of the model and facilitate a better understanding of its decision-making process.
\section{Automating Human in the Loop Feedback}

In order to automate human in the loop feedback, several approaches have been proposed. One such approach involves the utilization of a teacher-student feedback model, where feedback from human teachers is used to improve the performance of the model~\cite{Pruthi2022EvaluatingEH}. Another avenue of research focuses on enhancing active learning techniques, which aim to select the most informative data points for annotation by human experts, thereby reducing the overall labeling effort required~\cite{Kaushik2021OnTE}.

These methods often incorporate a combination of fine-tuning and prompt-based learning techniques, which further enhance the model's ability to learn from human feedback and adapt its performance accordingly~\cite{Valente2021ImprovingTC}. By fine-tuning the model based on the feedback received and utilizing prompts as guiding cues, these approaches enable the model to continually improve its performance, making it more effective in addressing the specific task or problem at hand.

\section{Generalizability in Emotion Recognition}

Achieving generalizability in emotion recognition poses a significant challenge for researchers. To address this challenge, various methods have been explored in order to obtain models that can generalize well across different datasets and scenarios. One approach is the use of combined and cross-dataset training, where multiple datasets are combined during the training process to create a more comprehensive and diverse training set. This helps the model learn a wider range of emotion patterns and improves its ability to generalize to unseen data~\cite{Li2021ControllableET}.

Another technique that has been investigated is transfer learning, which involves leveraging knowledge acquired from pre-trained models on a related task and applying it to the emotion recognition task. By transferring the learned representations and weights from a pre-trained model, the model can benefit from the general knowledge and feature extraction capabilities it has acquired, leading to improved generalizability in emotion recognition~\cite{Li2021ControllableET}.

Furthermore, researchers have also explored the concept of generalizability from the perspective of noisy signals. Emotion recognition often deals with noisy data, such as speech with background noise or facial expressions with occlusions. By developing models that are robust to such noise and can effectively extract emotion-related information from imperfect signals, the generalizability of the models can be enhanced~\cite{Hansen2021AGS}.

\section{Conclusion}

    The field of emotion recognition is complex, with many factors and considerations influencing the development and deployment of effective models. This chapter has explored some of the key areas in this field, highlighting the importance of crowdsourcing, context, handling confounding factors, dealing with noise, and ensuring that the representations don't inadverdently encode sensitive demographic or membership information. The role of interpretability in model trustworthiness and the challenge of automating human in the loop feedback were also discussed. Although progress has been made in many of these areas, the challenge of generalizability in emotion recognition remains, and future research will need to continue to address this issue.

%% file: Chapters/chap3.tex
This thesis focusses on emotion recognition as a task. For this purpose, we use a standard set of datasets and features as described in this chapter. This allows us to perform experiments with a set of known and commonly used datasets, keeping them uniform across experimental variables.

\section{Datasets Used In Thesis}
\label{sec:data}
In the past years, there have been multiple emotional databases collected and curated to develop better emotion recognition systems. 
Table~\ref{all:datasets} shows the major corpora that are used for emotion recognition.
\subsection{IEMOCAP}
The IEMOCAP dataset was created to explore the relationship between emotion, gestures, and speech. Pairs of actors, one male and one female (five males and five females in total), were recorded over five sessions. Each session consisted of a pair performing either a series of given scripts or improvisational scenarios. 
The data were segmented by speaker turn, resulting in a total of 10,039 utterances (5,255 scripted turns and 4,784 improvised turns). 

\subsection{MSP-Improv}
The MSP-Improv dataset was collected to capture naturalistic emotions from improvised scenarios. It partially controlled for lexical content by including target sentences with fixed lexical content that are embedded in different emotional scenarios.
The data were divided into 652 target sentences, 4,381 improvised turns (the remainder of the improvised scenario, excluding the target sentence), 2,785 natural interactions (interactions between the actors in between recordings of the scenarios), and 620 read sentences 
for a total of 8,438 utterances.

\subsection{MSP-Podcast}
The MSP-Podcast dataset was collected to build a naturlisitic emotionally balanced speech corpus by retrieving emotional speech from existing podcast recordings. This was done using machine learning algorithms, which along with a cost-effective annotation process using crowdsourcing, led to a vast and balanced dataset. We use a pre-split part of the dataset which has been identified for gender of the speakers which comprises of 13,555 utterances.  The dataset as a whole contains audio recordings.

\subsection{MuSE}
The MuSE dataset consists of recordings of 28 University of Michigan college students, 9 female and 19 male, in two sessions: one in which they were exposed to an external stressor (final exams period at University of Michigan) and one during which the stressor was removed (after finals have concluded).  Each recording is roughly 45-minutes.
We expose each subject to a series of emotional stimuli, short-videos and emotionally evocative monologue questions. These stimuli are different across each session to avoid the effect of repetition, but capture the same emotion dimensions.  
At the start of each session, we record a short segment of the user in their natural stance without any stimuli, to establish a baseline.  
We record their behavior using four main recording modalities: 1) video camera, both close-up on the face and wide-angle to capture the upper body, 2) thermal camera, close-up on the face, 3) lapel microphone, 4) physiological measurements, in which we choose to measure heart rate, breathing rate, skin conductance and skin temperature (Figure \ref{fig:exp-protocol}).  
The data include self-report annotations for emotion and stress (Perceived Stress Scale, PSS) \cite{cohen1988perceived,cohen1994perceived}, as well as emotion annotations obtained from Amazon Mechanical Turk (AMT).
To understand the influence of personality on the interaction of stress and emotion, we obtain Big-5 personality scores~\cite{goldberg1992development}, which was filled by 18 of the participants, due to the participation being voluntary.

\section{Data Pre-Processing}
We use these features consistently across the thesis to have a standardized set of inputs, aiming to avoid variability that comes from different labelling or pre-processing schemas. Our preprocessing corresponds to converting Likert scale emotion annotations to classes based on quartiles. The feature processing has 2 components, acoustic and lexical, for training, testing or fine-tuning speech-only, text-only or bimodal models.

\subsection{Emotion Labels}  Each utterance in the MuSE dataset was labeled for  
\label{sec:labels}
\textit{activation} 
and \textit{valence} on a nine-point Likert scale by eight crowd-sourced 
annotators~\cite{jaiswal2019muse}, who observed the data in random order across subjects. We average the annotations to obtain a mean score for 
each utterance, and then bin the mean score into one of three classes, defined as, \{``\textit{low}'': [min, 4.5], 
``\textit{mid}'': (4.5, 5.5], ``\textit{high}'': (5.5, max]\}. 
The resulting distribution for activation is: \{``\textit{high}'': $24.58\%$, ``\textit{mid}'': $40.97\%$ and ``\textit{low}'': $34.45\%$\} and for valence is \{``\textit{high}'': $29.16\%$,  ``\textit{mid}'': $40.44\%$ and ``\textit{low}'': $30.40\%$\}.
Utterances in IEMOCAP and MSP-Improv were annotated for valence and activation on a five-point Likert scale.
The annotated activation and valence values were averaged for an utterance and binned as: \{``\textit{low}'': [1, 2.75], ``\textit{mid}'': (2.75, 3.25], ``\textit{high}'': (3.25, max]\}

\subsection{Stress Labels} Utterances in the the MuSE dataset include stress annotations, in addition to the activation and valence annotations.
The stress annotations for each session were self-reported by the participants using the Perceived Stress Scale (PSS)~\cite{cohen1983global}. 
We perform a paired t-test for subject wise PSS scores, and find that the scores are significantly different for both sets (16.11 vs 18.53) at $p<0.05$. This especially true for question three 
(3.15 vs 3.72), and hence, we double the weightage of the score for this question while obtaining the final sum.
We bin the original nine-point adjusted stress scores into three classes, \{``\textit{low}'': (min, mean$-2$], ``\textit{mid}'': (mean$-2$, mean$+2$], ``\textit{high}'': (mean$+2$, max]\}.
We assign the same stress label to all utterances from the same session. The distribution of our data for stress is {``\textit{high}'': $40.33\%$, ``\textit{mid}'': $25.78\%$ and ``\textit{low}'': $38.89\%$}

\textbf{Improvisation Labels.} Utterances in the IEMOCAP dataset were recorded in either a scripted scenario or an improvised one. We label each utterance with a binary value \{``\textit{scripted}'', ``\textit{improvised}''\} to reflect this information.

\section{Lexical and Acoustic Feature Extraction}
\label{sec:features}

\subsection{Acoustic}  We use Mel Filterbank (MFB) features, which are frequently used in speech  processing applications, including speech recognition, 
and emotion recognition~\cite{khorram2017capturing, krishna2018study}.  We extract the 40-dimensional MFB features using a 25-millisecond Hamming window with a step-size of 10-milliseconds. As a result, each utterance is represented as a sequence of 40-dimensional feature vectors. We $z$-normalize the acoustic features by session for each speaker.

\subsection{Lexical} We have human transcribed data available for MuSE and IEMOCAP. We use the word2vec representation based on these transcriptions, which has shown success in sentiment and emotion analysis tasks~\cite{kim2014convolutional}. We represent each word in the text input as a 300-dimensional vector using a pre-trained word2vec model~\cite{mikolov2013distributed}, replacing out-of-vocab words with the $\langle unk\rangle$ token. Besides, we also incorporate BERT embeddings for enhanced contextual understanding. These embeddings, generated from the pre-trained BERT model, provide deep, bidirectional representations by understanding the text context from both directions. Each utterance is eventually represented as a sequence of 768-dimensional feature vectors. We use just acoustic inputs for MSP-Improv because human transcriptions are not available.

%% file: Chapters/chap4.tex
\section{Motivation and Contributions}
Endowing automated agents with the ability to provide support, entertainment and interaction with human beings requires sensing of the users' affective state. These affective states are impacted by a combination of emotion inducers, current psychological state, and various contextual factors. Although emotion classification in both singular and dyadic settings is an established area, the effects of these additional factors on the production and perception of emotion is understudied. This chapter presents a dataset, Multimodal Stressed Emotion (MuSE), to study the multimodal interplay between the presence of stress and expressions of affect. We describe the data collection protocol, the possible areas of use, and the annotations for the emotional content of the recordings. The chapter also presents several baselines to measure the performance of multimodal features for emotion and stress classification.
\section{Introduction}

Virtual agents have become more integrated into our daily lives than ever before \cite{lucas2014s}. For example, Woebot is a chatbot developed to provide cognitive behavioral therapy to a user~\cite{fitzpatrick2017delivering}. For this chatbot agent to be effective, it needs to respond differently when the user is stressed and upset versus when the user is calm and upset, which is a common strategy in counselor training~\cite{thompson2013students}. While virtual agents have made successful strides in understanding the task-based intent of the user, social human-computer interaction 
can still benefit from further research~\cite{clark2019makes}.
Successful integration of virtual agents into real-life social interaction requires machines to be emotionally intelligent~\cite{bertero2016real,yuan2015approach}.

But humans are complex in nature, and emotion is not expressed in isolation~\cite{griffiths2003iii}. Instead, it is affected by various external factors. These external factors lead to interleaved user states, which are a culmination of situational behavior, experienced emotions, psychological or physiological state, and personality traits.
One of the external factors that affects psychological state is stress. Stress can affect everyday behavior and emotion, and in severe states, is associated with delusions, depression and anxiety due to impact on emotion regulation mechanisms~\cite{kingston2016life,schlotz2011perceived,tull2007preliminary,wang2011emotion}.
Virtual agents can respond in accordance to users' emotions only if the machine learning systems can recognize these complex user states and correctly perceive users' emotional intent.
We introduce a dataset designed to elicit spontaneous emotional responses in the presence or absence of stress to observe and sample complex user states.



There has been a rich history of visual~\cite{you2016building,jiang2014predicting}, speech~\cite{lotfian2017building}, linguistic \cite{Strapparava07}, and multimodal emotion datasets~\cite{busso2017msp,busso2008iemocap,ringeval2013introducing}.  Vision datasets have focused both on facial movements~\cite{jiang2014predicting} and body movement \cite{lazarus1977environmental}. Speech datasets have been recorded to capture both stress and emotion separately but do not account for their inter-dependence~\cite{rothkrantz2004voice,horvath1982detecting,kurniawan2013stress,zuo2011cross}. Stress datasets often include physiological data \cite{yaribeygi2017impact,Activity-Aware-Mental-Stress-Detection-Using-Physiological-Sensors}.

Existing datasets are limited because they are designed to elicit emotional behavior, while neither monitoring external psychological state factors nor minimizing their impact by relying on randomization. However, emotions produced by humans in the real world are complex. Further, our natural expressions are often influenced by multiple factors (e.g., happiness \textit{and} stress) and do not occur in isolation, as typically assumed under laboratory conditions. The primary goal of this work is to collect a multimodal stress+emotion dataset -- Multimodal Stressed Emotion (MuSE) -- to promote the design of algorithms that can recognize complex user states. 

For MuSE,
The extracted features for each modality, and the anonymized dataset (other than video) will be released publicly along with all the corresponding data and labels.
We present baseline results for recognizing both emotion and stress in the chapter, in order to validate that the presence of these variables can be computationally extracted from the dataset, hence enabling further research.

\section{MuSE Dataset}
\subsection{Experimental Protocol}
\begin{figure}[tb]
    \centering
    \includegraphics[scale=0.5]{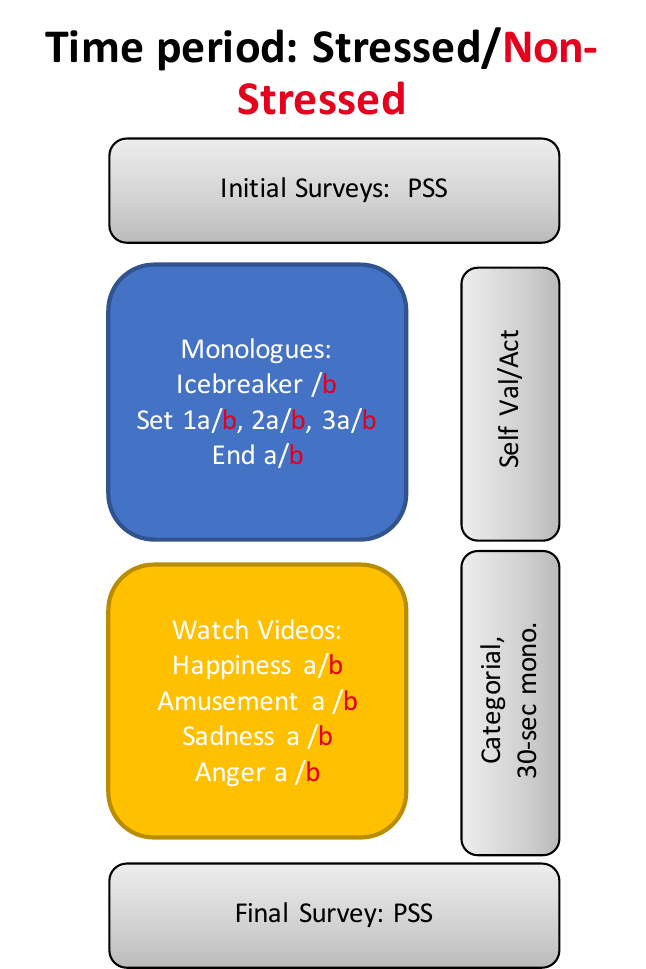}
    \caption{Experimental Protocol For Recording}
    \label{fig:exp-protocol}
\end{figure}

 We collect a dataset that we refer to as Multimodal Stressed Emotion (MuSE) to facilitate the learning of the interplay between stress and emotion. The protocol for data collection is shown in Figure~\ref{fig:exp-protocol}. There were two sections in each recording:  monologues and watching emotionally evocative videos. We measure the stress level at the beginning and end of each recording.
The monologue questions and videos were specifically chosen to cover all categories of emotions. At the start of each recording, we also recorded a short one-minute clip without any additional stimuli to register the baseline state of the subject.

Previous research has elicited situational stress such as public speaking \cite{kirschbaum1993trier,giraud2013multimodal,aguiar2014voce}, mental arithmetic tasks \cite{liao2015laboratory} or use Stroop Word Test \cite{tulen1989characterization}. However, these types of stress are often momentary and fade rapidly in ~two minutes~\cite{liao2015laboratory}. We alleviate this concern by recording both during and after final exams (we anticipate that these periods of time are associated with high stress and low stress, respectively) in April 2018. We measure stress using Perceived Stress Scale~\cite{cohen1994perceived} for each participant.  We measure their self-perception of the emotion using Self-Assessment Manikins (SAM)~\cite{bradley1994measuring}. 
The recordings and the survey measures were coordinated using Qualtrics\footnote{umich.qualtrics.com
}
enabling us to ensure minimal intervention and limit the effect of the presence of another person on the emotion production.

Each monologue section comprised of five questions broken into sections meant to elicit a particular emotion (Table~\ref{dataset_detail_mon}). These questions were shown to elicit thoughtful and emotional responses in their data pool to generate interpersonal closeness~\cite{aron1997experimental}. We include an icebreaker and ending question to ensure cool off periods between change in recording section, i.e., from neutral to monologues, and from monologues to videos, hence decreasing the amount of carry-over emotion from the previous monologue to the next. Each subject was presented with a different set of questions over the two recordings to avoid repetition effect. We also shuffle the order of the other three questions to account for order effects~\cite{lee2011apparatus}.  Each subject was asked to speak for a minimum of two minutes. 
After their response to each question, the subjects marked themselves on two emotion dimensions: activation and valence on a Likert Scale of one to nine using self-assessment manikins~\cite{bradley1994measuring}.

For the second part of the recording, the subjects were asked to watch videos in each of the four quadrants i.e., the combination of \textit{\{low, high\} $\times$ \{activation, valence\}} of emotion. These clips were selected from the corpus~\cite{lichtenauer2011mahnob,bartolini2011eliciting}, which tested for the emotion elicited from the people when watching these clips (Table~\ref{dataset_detail_clips}). The subjects were monitored for their reaction to the clips. 
 After viewing a clip, subjects are asked to speak for thirty seconds about how the video made them feel. After their response, they marked a emotion category, e.g., angry, sad, etc. for the same clip. When switching videos, the subjects were asked to view a one-minute neutral clip to set their physiological and thermal measures back to the baseline~\cite{samson2016eliciting}.

The 28 participants were also asked to fill out an online survey used for personality measures on the big-five scale \cite{goldberg1992development}, participation being voluntary. This scale has been validated to measure five different dimensions named OCEAN (openness, conscientiousness, extraversion, agreeableness, and neuroticism) using fifty questions and has been found to correlate with passion~\cite{dalpe2019personality}, ambition~\cite{barrick1991big}, and emotion mechanisms~\cite{querengasser2014sad}. We received responses for this survey from 18 participants. These labels can be used in further work to evaluate how these personality measures interact with the affects of stress in emotion production, as previously studied in~\cite{zhao2018personality}.

\begin{table}[t]

\caption{Emotion elicitation questions.}
\label{dataset_detail_mon}
\small 
\renewcommand{\arraystretch}{0.8} 
\begin{tabular}{p{\dimexpr\textwidth-2\tabcolsep}} 
\hline
\textbf{Icebreaker}\\
\hline
\begin{enumerate}
\itemsep0em 
\item Given the choice of anyone in the world, whom would you want as a dinner guest?
\item Would you like to be famous? In what way?
\end{enumerate}\\
\hline
\textbf{Positive}\\
\hline
\begin{enumerate}
\itemsep0em 
\item For what in your life do you feel most grateful?
\item What is the greatest accomplishment of your life?
\end{enumerate}\\
\hline
\textbf{Negative}\\
\hline
\begin{enumerate}
\itemsep0em 
\item If you could change anything about the way you were raised, what would it be?
\item Share an embarrassing moment in your life.
\end{enumerate}\\
\hline
\textbf{Intensity}\\
\hline
\begin{enumerate}
\itemsep0em 
\item If you were to die this evening with no opportunity to communicate with anyone, what would you most regret not having told someone?
\item Your house, containing everything you own, catches fire. After saving your loved ones and pets, you have time to safely make a final dash to save any one item. What would it be? Why?
\end{enumerate}\\

\hline
\textbf{Ending}\\
\hline
\begin{enumerate}
\itemsep0em 
\item If you were able to live to the age of 90 and retain either the mind or body of a 30-year old for the last 60 years of your life, which would you choose?
\item If you could wake up tomorrow having gained one quality or ability, what would it be?
\end{enumerate}\\

\hline
\end{tabular}

\end{table}

\begin{table}[t]

\caption{Emotion elicitation clips.}
\label{dataset_detail_clips}
\centering
\begin{tabular}{ll}
\toprule
\textbf{Movie} & \textbf{Description}\\
\midrule
\multicolumn{2}{c}{\textbf{Low Valence, Low Activation (Sad)}}\\
\midrule

City of Angels & Maggie dies in
Seth's arms\\
Dangerous Minds & Students find that one of
their classmates
has died\\

\midrule
\multicolumn{2}{c}{\textbf{Low Valence, High Activation (Anger)}}\\
\midrule

Sleepers & Sexual abuse of children \\
Schindler's List: & Killing of Jews during WWII \\

\midrule
\multicolumn{2}{c}{\textbf{High Valence, Low Activation (Contentment)}}\\
\midrule

Wall-E & 
Two robots dance and
fall in love \\
Love Actually & Surprise orchestra at the wedding \\

\midrule
\multicolumn{2}{c}{\textbf{High Valence, High Activation (Amusement)}}\\
\midrule
Benny and Joone & Actor plays
the fool in a
coffee shop\\
Something About Mary & Ben Stiller fights
with a dog\\

\midrule
\multicolumn{2}{c}{\textbf{Neutral}}\\
\midrule

\multicolumn{2}{c}{A display of zig-zag lines across the screen }\\
\multicolumn{2}{c}{Screen-saver pattern of changing colors}\\

\bottomrule
\end{tabular}

\end{table}

\subsection{Equipment Setup}

The modalities considered in our setup are: thermal recordings of the subject's face, audio recordings of the subject, color video recording of the subject's face, a wide-angle color video recording the subject from the waist up and physiological sensors measuring skin conductance, breathing rate, heart rate and skin temperature. For these modalities we have set up the following equipment:

\begin{enumerate}
\itemsep 0pt
    \item \textbf{FLIR Thermovision A40 thermal camera} for recording the close-up thermal recording of the subject's face. This camera provides a 640x512 image in the thermal infrared spectrum.
    \item \textbf{Raspberry Pi with camera module V2 with wide-angle lens} used for the waist up shot of the subject. We have chosen Raspberry Pi's due to its low price and support for Linux OS, which integrates easily into a generic setup. 
    \item \textbf{Raspberry Pi with camera module V2} used to record the subject from the waist up. 
    \item \textbf{TASCAM DR-100 mk II} used to record audio. We chose this product for its high fidelity. It can record 24-bit audio at 48kHz.
    \item \textbf{ProComp}$^\infty$\textbf{-8 channel biofeedback and neurofeedback system v6.0} used to measure blood volume pulse (BVP sensor), skin conductance (SC sensor), skin temperature (T sensor), and abdominal respiration (BR sensor)
    
\end{enumerate}

\begin{figure}[tb]
    \centering
    \includegraphics[width = 0.7\columnwidth]{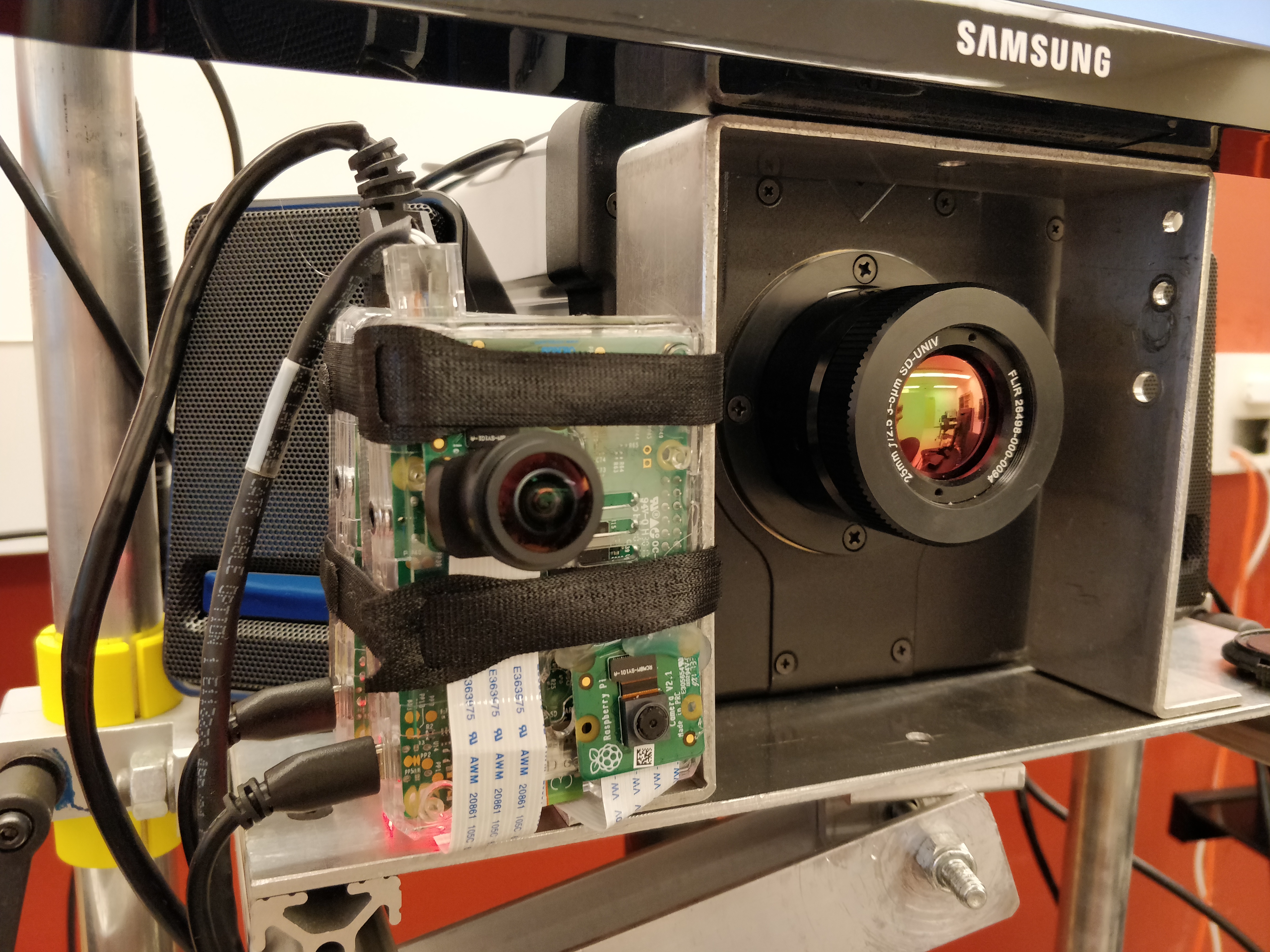}
    \caption{Close-up view of the thermal and video recording equipment.\\}
    \label{fig:thermal-video}
\end{figure}

The equipment operator started and marked the synchronization point between video and audio recordings using a clapper. Subsequent time stamps are recorded by the qualtrics survey using subject click timings.


\subsection{Post-processing}
\textbf{Splitting of the Recordings.} 
Each modality is split into neutral recordings of one-minute, five questions and four video recordings with associated monologues, resulting in fourteen recordings for emotional content, thus 28 recordings per subject. In total we have 784 distinct recordings over five modalities, 28 subjects and two stress states, for a total of 3920 recording events. 
Temperatures are clamped to between $0^o$C and $50^o$C. This helps reduce the size of the thermal recording files after being zipped.

\textbf{Utterance Construction.} 
The five monologues extracted above were divided into utterances. However, since the monologues are a form of spontaneous speech, there are no clear sentence boundaries marking end of utterance.
We manually created utterances by identifying prosodic or linguistic boundaries in spontaneous speech as defined by~\cite{kolavr2008automatic}. The boundaries used for this work are: (a) clear ending like a full stop or exclamation, (b) a change in context after filler words or completely revising the sentence to change meaning, or (c) a very long pause in thought. This method has been previously shown to be effective in creating utterances that mostly maintain a single level of emotion~\cite{khorram2018priori}.

The dataset contains 2,648 utterances with a mean duration of 12.44 $\pm$ 6.72 seconds (Table~\ref{data_stat}). The mean length of stressed utterances ($11.73\pm5.77$ seconds) is significantly different (using two-sample t-test) from that of the non-stressed utterances ($13.30\pm6.73$ seconds).  
We remove utterances that are shorter than $3$-seconds and longer than $35$-seconds and end up retaining $97.2\%$ of our dataset.  This allows us to to avoid short segments that may not have enough information to capture emotion, and longer segments that can have variable emotion, as mentioned in~\cite{khorram2018priori}.
Because our dataset is comprised of spontaneous utterances, the mean length of utterance is larger than those in a scripted dataset~\cite{busso2017msp} due to more corrections and speech overflow.

\textbf{Stress State Verification.} We perform a paired t-test for subject wise PSS scores, and find that the mean scores are significantly different for both sets (16.11 vs 18.53) at $p<0.05$. This implied that our hypothesis of exams eliciting persistently more stress than normal is often true. In our dataset, we also provide levels of stress which are binned into three categories based on weighted average (using questions for which the t-test score was significant).


\section{Emotional Annotation}
\subsection{Crowdsourcing}
Crowdsourcing has previously been shown to be an effective and inexpensive method for obtaining multiple annotations per segment~\cite{hsueh2009data,burmania2017stepwise}. 
We posted our experiments as Human Intelligence Tasks (HITs) on \textit{Amazon Mechanical Turk} and used selection and training mechanisms to ensure quality \cite{muse-ing-icassp}. HITs were defined as sets of utterances in a monologue. 
The workers were presented with a single utterance and were asked to annotate the activation and valence values of that utterance using Self-Assessment Manikins~\cite{bradley1994measuring}. Unlike the strategy adopted in~\cite{friends_dataset}, the workers could not go back and revise the previous estimate of the emotion. We did this to ensure similarity to how a human listening into the conversation might shift their perception of emotion in real time.
These HITs were presented in either the contextual or the random presentation condition defined below.

In the contextual experiment, we posted each HIT as a collection of ordered utterances from each section of a subject's recording. Because each section's question was designed to elicit an emotion, to randomize the carry-over effect in perception, we posted the HITs in a random order over the sections from all the subjects in our recording. 
For example, a worker might see the first HIT as \textit{Utterance 1...N from Section 3 of Subject 4's stressed recording} and see the second HIT as \textit{Utterance 1...M from Section 5 of Subject 10's non-stressed recording} where {\it N, M} are the number of utterances in those sections respectively. This ensures that the annotator adapts to the topic and fluctuations in speaking patterns over the monologue being annotated.

In the randomized presentation, each HIT is an utterance from any section, by any speaker, in random order. So, a worker might see the first HIT as \textit{Utterance 11 from Section 2 of Subject 1's stressed recording monologue} and see the second HIT as \textit{Utterance 1 from Section 5 of Subject 10's non-stressed monologue recording}. We use this method of randomization to ensure lack of adaptation to both speaker specific style and the contextual information. 
The per-utterance and the contextual labels can be used to train different machine learning models that are apt for either singular one-off instances or for holding multiple turn natural conversation, respectively. 


\begin{table}[t]
  \caption{Data summary (R:random, C:context, F:female, M:male).}
  \label{data_stat}
  \centering
  \begin{tabular}{p{0.4\textwidth} p{0.4\textwidth}} 
    \toprule
    \multicolumn{2}{c}{\textbf{Monologue Subset}}\\
    \cmidrule(r){1-2}
    Mean no. of utterances/monologue     & $9.69\pm2.55$    \\
    Mean duration of utterances       & $12.44 \pm 6.72$ seconds   \\
    Total no. of utterances        & 2,648    \\
    Selected no. of utterances     & 2,574    \\
    Gender distribution        & 19 (M) and 9 (F)   \\
    Total annotated speech duration & $\sim10$ hours\\
    \midrule
    \multicolumn{2}{c}{\textbf{Crowdsourced Data}}\\
    \cmidrule(r){1-2}
    Num of workers                 & 160 (R) and 72 (C)    \\
    Blocked workers     & 8   \\
    \multirow{2}{*}{Mean activation}     & 3.62$\pm$0.91 (R)   \\
                                  &   3.69$\pm$0.81 (C)\\
    \multirow{2}{*}{Mean valence}       & 5.26$\pm$0.95 (R)   \\
     &    5.37$\pm$1.00 (C)\\
    \bottomrule
  \end{tabular}
\end{table}

\begin{figure}
    \hspace{-3.0ex}
    \begin{tabularx}{\textwidth}{XX}
        \includegraphics[scale=0.45]{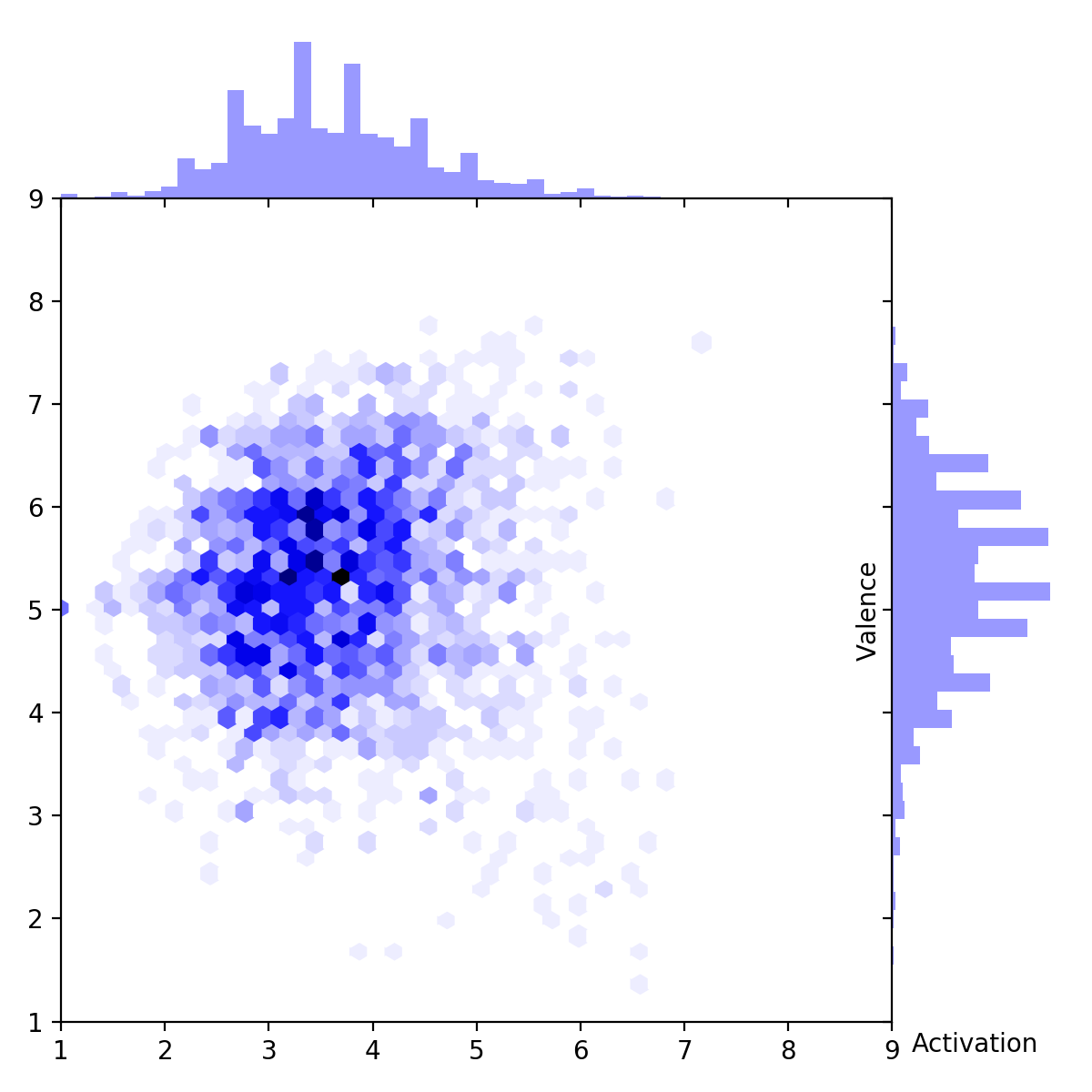}
    &
    \hspace{-3.0ex}
        \includegraphics[scale=0.45]{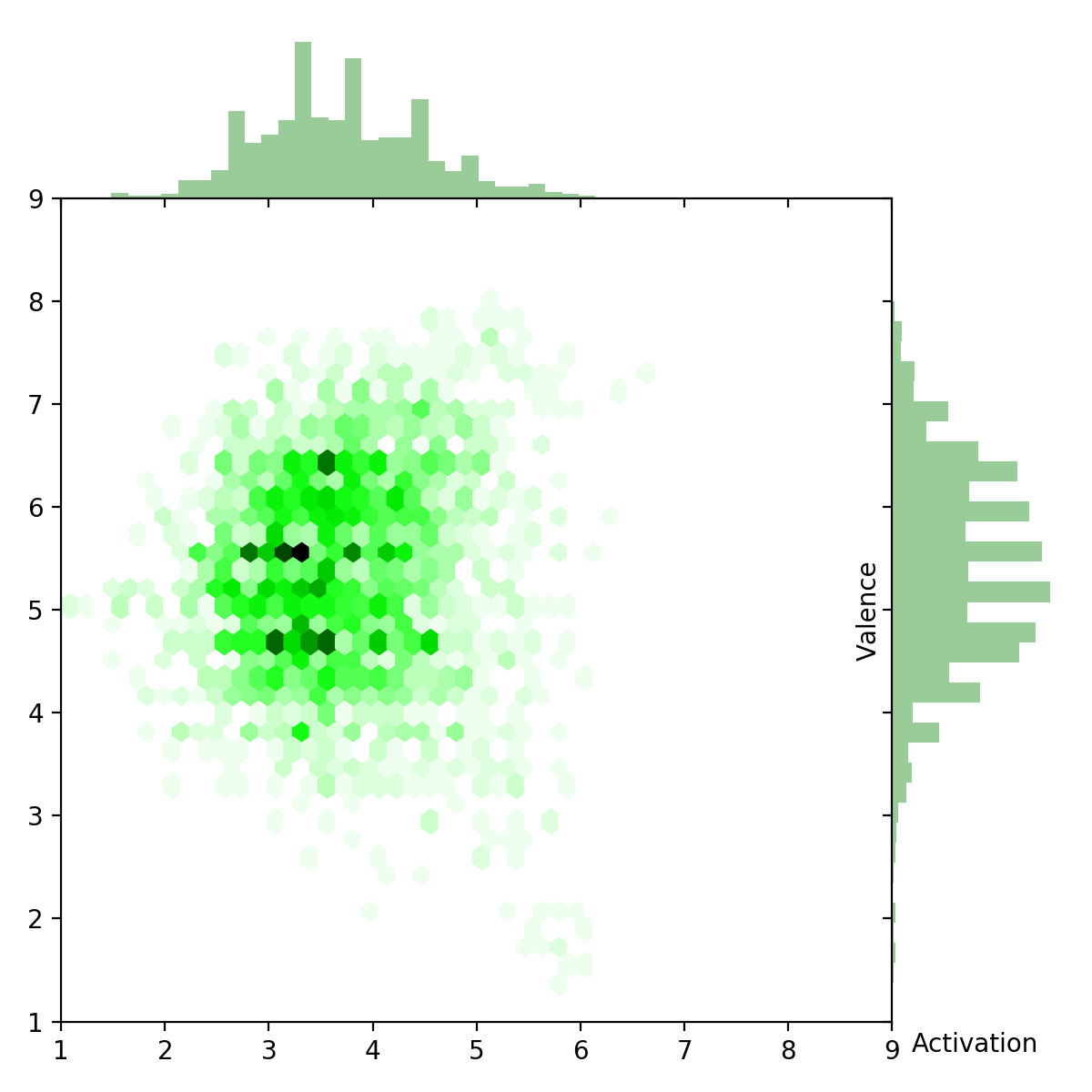}
    
\end{tabularx}
\caption{Distribution of the activation and valence
ratings in random labeling scheme (on left) and contextual labeling scheme (on right).}
\label{fig:emotion-density}
\end{figure}

\subsection{Emotion Content Analysis}
We show the distribution of the annotations received in both the random and contextual setting in Table~\ref{data_stat} and Figure~\ref{fig:emotion-density}. The labels obtained for our dataset form a distribution that mostly covers negative and neutral levels of activation, and all but extremities for valence. This can also be seen in the data summary in Table~\ref{data_stat}. We performed a paired t-test between the labels obtained from random vs contextual presentation and found that these labels are significantly different (using paired t-test at $p<0.05$ for both activation and valence for utterances in the non-stressed situation). Although the obtained labels are significantly different for valence in the stressed category using the same method as above, 
the same does not hold true for the activation annotations in this category.

\begin{figure}[tb]
    \centering
    \includegraphics[scale=0.4]{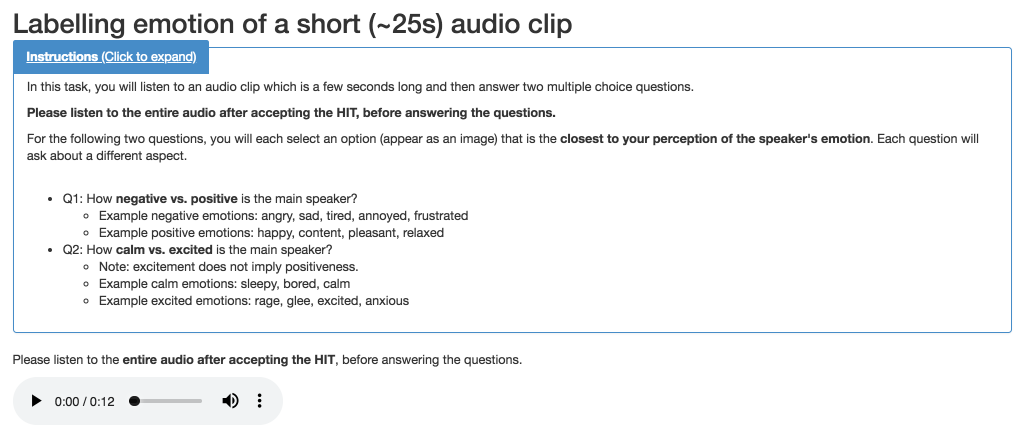}
    \caption{An overview of the instructions provided to the annotators for annotating an utterance.\\\\}
    \label{fig:label-ins}
\end{figure}

\begin{figure}[tb]
    \centering
    \includegraphics[scale=0.4]{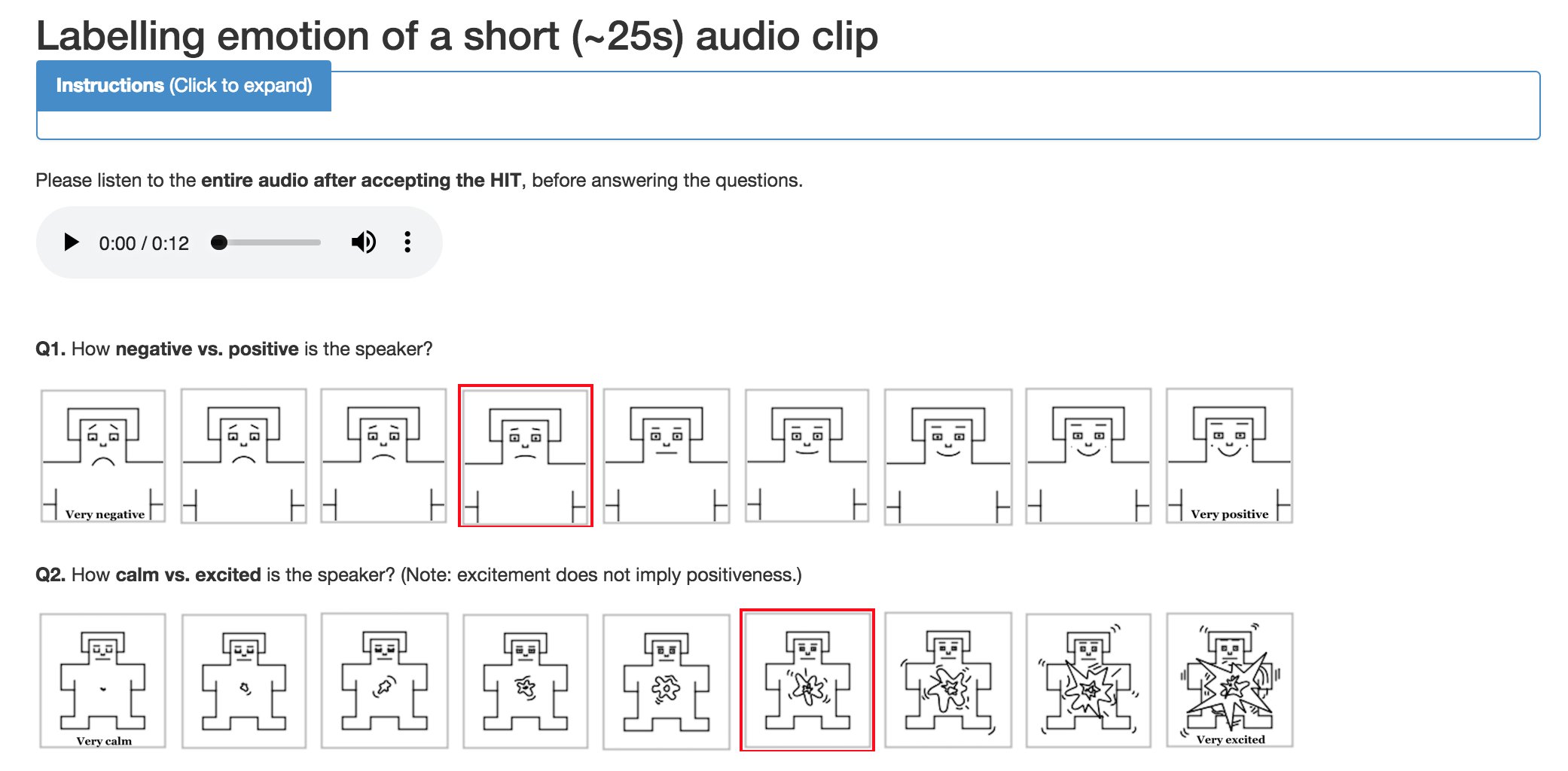}
    \caption{Annotation scale used by MTurk workers to annotate the emotional content of the corpus. They annotate valence and activation for each utterance.}
    \label{fig:label-sam}
\end{figure}

\section{Experiments}
In this section, we describe our baseline experiments for predicting emotion and stress in the recorded modalities. We have a more granular marked annotation of emotion, i.e., over each utterance, as compared to stress over the complete monologue. Hence, we extract features for each modality over continuous one second frame intervals for predicting stress, and over the complete utterance for emotion. Audio and lexical features are still extracted over a complete utterance for stress due to higher interval of variation over time.

\subsection{Evaluation of Emotion Recognition}
We use the following set of features for our baseline models:

\begin{enumerate}
\itemsep 0pt
\item \textbf{Acoustic Features.} We extract acoustic features using OpenSmile~\cite{eyben2010opensmile} with the eGeMAPS configuration~\cite{eyben2016geneva}. The eGeMAPS feature set consists of $88$ utterance-level statistics over the low-level descriptors of  frequency, energy, spectral, and cepstral parameters. We perform speaker-level $z$-normalization on all features.

\item \textbf{Lexical Features.} We extract lexical features using Linguistic Inquiry and Word Count (LIWC)~\cite{pennebaker2001linguistic}. These features have been shown to be indicative of stress, emotion, veracity and satisfaction~\cite{golbeck2011predicting,monin2012linguistic,newman2003lying}. We normalize all the frequency counts by the total number of words in the sentence accounting for the variations due to utterance length.

\item \textbf{Thermal Features.} For each subject a set of four regions were selected in the thermal image: the forehead area, the eyes, the nose and the upper lip as previously used in \cite{pavlidis2002thermal,garbey2007contact,abouelenien2016analyzing}. These regions were tracked for the whole recording and a 150-bin histogram of temperatures was extracted from the four regions per frame, i.e., 30 frames a second for thermal recordings. We further reduced the histograms to the first four measures of central tendency, e.g. Mean, Standard Deviation, Skewness and Kurtosis. We combined these features over the utterance using first delta measures (min, max, mean, SD) of all the sixteen extracted measures per frame, resulting in 48 measures in total.

\item \textbf{Close-up Video Features.} 
We use OpenFace~\cite{baltruvsaitis2016openface} to extract the subject's facial action units. The AUs used in OpenFace for this purpose are AU1, AU2,
AU4, AU5, AU6, AU7, AU9, AU10, AU12, AU14, AU15,
AU17, AU20, AU23, AU25, AU26, AU28 and AU25 comprising of eyebrows, eyes and mouth. These features have been previously shown to be indicative of emotion~\cite{wegrzyn2017mapping,du2014compound} and have been shown to be useful for predicting deception~\cite{jaiswal2016truth}. We summarize all frames into a feature using summary statistics (maximum, minimum, mean, variance, quantiles) across the frames and across delta between the frames resulting in a total of 144 dimensions.
\end{enumerate}

\textbf{Network Setup.} We train and evaluate multiple unimodal Deep Neural Networks (DNN) models for predicting valence and activation using Keras~\cite{gulli2017deep}. 
~\cite{muse-ing-icassp} have shown that a match between the context provided to the classifier and the annotator leads to better classification performance. Because we are performing single utterance classification, for all further experiments, we use the annotations obtained in a random manner as mentioned above. In all cases, we predict the continuous annotation using regression.  

We also use an ensemble of these four networks (audio, lexical, visual and thermal) to measure multimodal performance. For each network setup, we follow a five-fold subject independent evaluation scheme and report the average RMSE across the folds. For each test-fold, we use the previous fold for hyper-parameter selection and early stopping. The hyper-parameters include: number of layers $\{2, 3, 4\}$ and layer width $\{64, 128, 256\}$. We use ReLU activation and train the networks with MSE loss using the Adam optimizer. 

We train our networks for a maximum of 50 epochs and monitor the validation loss after each epoch. We perform early stopping if the loss doesn't decrease for 15 consecutive epochs. We save the weights that achieved the lowest validation performance during training. We train each network five times with different seeds and average the predictions to account for variations due to random initialization.


\begin{table}[t]
  \caption{RMSE for emotion classification models using multiple modalities. Significance established at $p<0.05$.\\}

  \label{exp-emotion}
  \centering
  \begin{tabular}{lcc}
    \toprule
    & \textbf{Activation} & \textbf{Valence}  \\
   
    \midrule
    \textbf{Unimodal Models} \\
    \midrule
   Acoustic (A) & \textbf{1.004}$^{*}$   & 1.122 \\
   Lexical (L) & 1.343 & 0.980 \\
   Close Video (V) & 1.111 & \textbf{0.879}$^{**}$\\
   Thermal (T) &2.012 & 1.565\\
    \midrule
    \textbf{Ensemble}          \\
    \midrule
    A+L &  0.987 & 0.981\\
    A+V & 0.970 & 0.899 \\ 
    L+V & 0.981  & 0.901 \\
    A+L+V & 0.972 & \textbf{0.856}$^{*}$ \\
    A+L+V+T (All) & \textbf{0.961}$^{*}$ & 0.868 \\
    \bottomrule
  \end{tabular}
\end{table}

\textbf{Results.} We show our results in Table~\ref{exp-emotion}. We find that between acoustic and lexical modalities, the acoustic modality carries more information about activation and the lexical for valence. This is in line with previous research~\cite{yang2011prediction,cambria2017benchmarking}. We also note that the visual modality significantly outperforms both the speech and lexical modalities for valence prediction. 

When we merge these networks using late voting on each modality (decision fusion), we find that the combination of all modalities performs the best for predicting activation. But for predicting valence, the best performance is shown by the combination of acoustic, lexical, visual and thermal modalities. We believe this is true because previous work has shown that thermal features are mostly indicative of intensity and discomfort~\cite{herborn2015skin} and hence improves performance on activation prediction, while the visual expressions are most informative about valence~\cite{rubo2018social}.

\subsection{Evaluation of Presence of Stress}
We use the following set of features for our baseline models. Given that stress vs non-stressed state is classified for the complete section (monologue or neutral recording), we extract visual features differently to use the the sequential information over the whole segment, i.e., a monologue. We also use physiological features for our network, since we found that even though they are highly variable over shorter segments (utterances), they are informative for recognizing physiological state on a whole section.
\begin{enumerate}
\itemsep 0pt
\item \textbf{Acoustic, Lexical, and Thermal Features.} We use the same features as extracted for predicting emotion.

\item \textbf{Wide-angle Video Features.} We extract the subject's pose using OpenPose \cite{cao2017realtime,simon2017hand,wei2016cpm} at 25 frames per second. For each frame, we extract 14 three-dimensional points representing anchor points for the upper body. For classification of each 3D point is interpolated over one second using a $5^{th}$ order spline \cite{oikonomopoulos2008human,3-D-motion-estimation-and-object-tracking-using-B-spline-curve-modeling}. The parameters of the splines are then used as features for classification. 

\item \textbf{Close-up Video Features.} 
We use OpenFace to extract the subject's action units~\cite{baltruvsaitis2016openface}. The features are extracted for every frame. In each frame, features include the gaze direction vectors, gaze angles, 2D eye region landmarks, head locations, rotation angles of the head, landmark locations, and facial action units. Landmarks locations offset by the nose location. We window the data into segments of one-second windows with 0.5 second overlap and calculate summary statistics (maximum, minimum, mean, variance). We retain the top 300 features based on the F values between the training features and corresponding labels (stressed vs non-stressed).

\item \textbf{Physiological Features.} While the physiological features varied greatly per second to be informative for emotion, they are informative for recognizing presence or absence of stress. We consider the raw measurements for heart rate, breathing rate, skin conductance and skin temperature and compute the first four measures of central tendency, e.g. mean, standard deviation, skewness, and kurtosis.

\end{enumerate}


\begin{table}[t]
\caption{Baseline results for classifying stressed and non-stressed situations per time unit, unless specified otherwise. A - Accuracy, P - Precision, R - Recall.\\}
    \centering
    \begin{tabular}{lrrrr}
    \toprule
     Recording Parts & $A$ & $P$ & $R$ & $F_{1}$\\
        \hline
        & \multicolumn{4}{c}{Thermal} \\
        \hline
        \textbf{Neutral} & \textbf{0.61} & \textbf{0.67} & \textbf{0.62} &\textbf{0.64} \\
        Questions & 0.50 & 0.64 & 0.52 & 0.57 \\
        \hline
        & \multicolumn{4}{c}{Wide-angle Video} \\
       
        \hline
        Neutral & 0.66 & 0.41 & 0.96 & 0.58 \\
        Questions & 0.69 & 0.45 & 0.82 & 0.58 \\
        \hline
        & \multicolumn{4}{c}{Close-up Video} \\
       
        \hline
        Neutral & 0.61 & 0.78 & 0.33 & 0.46 \\
        Questions & 0.65 & 0.65 & 0.69 & 0.67  \\
        \hline
        & \multicolumn{4}{c}{Physiological} \\
       
        \hline
        Neutral & 0.66 & 0.47 & 0.89 & 0.64  \\
        Questions & 0.70 & 0.55 & 0.88 & 0.67  \\
        \hline
        & \multicolumn{4}{c}{Audio - Per utterance} \\
        
        \hline
        Questions & 0.67 & 0.70 & 0.69 & 0.69  \\
        \hline
        & \multicolumn{4}{c}{Text - Per utterance} \\
        
        \hline
        Questions & 0.60 & 0.74 & 0.61 & 0.67  \\
        \hline
        & \multicolumn{4}{c}{Late Fusion - Voting} \\
        
        \hline
        Questions & 0.60 & 0.74 & 0.61 & 0.67  \\
        \hline
    \end{tabular}
    
    \label{tab:Results}
\end{table}






\textbf{Network.} We train a DNN to perform binary classification, i.e., to recognize stressed vs. non-stressed situation using ReLU as activation, with softmax as the classification method.The final layer uses a soft-max activation. We train six different networks for thermal, wide-angle video, close-up video, physiological, audio, and lexical modalities. Each network is trained in a subject-independent manner.
We train network to recognize stress vs non-stress situation in both neutral recording,i.e., when the subject isn't speaking at the beginning of the recording, and during emotional monologue questions. To do so, we decide the final prediction by a majority vote over one-second predictions for the complete section of the recording. For the lexical and acoustic modality, we train the network for the question monologues, and decide the final prediction based on a majority vote over prediction for each utterance.


         


\textbf{Results.} We report our results for prediction of stress vs non-stress situation using various modalities in Table~\ref{tab:Results}. We see that the captured modalities are indeed informative for recognizing stress vs non-stressed situations.
We find that for recognizing this distinction when the subjects are speaking, audio and physiological features perform the best. This is in agreement with previous related work \cite{lazarus1977environmental,yaribeygi2017impact,horvath1978experimental}.
Interestingly, we also find that the thermal and physiological modality is apt at recognizing differences in stress, even in the neutral recording, i.e., when the subject is not speaking. This advantage of thermal modality has been previously documented by researchers~\cite{Abouelenien:2014:DDU:2663204.2663229,pavlidis2000thermal,pavlidis2002thermal,garbey2007contact}. We find that answering emotional monologue questions interferes with the recorded thermal modality, leading to a poorer performance at stress recognition.
\section{Conclusions and Future Work}
In this chapter, we introduced a dataset that aims to capture the interplay between psychological factors such as stress and emotion.
While various other datasets have explored the relationship between gender or personality measures and emotion production and perception, the relationship between psychological factors and emotion is understudied from a data collection point of view, and hence an automated modeling perspective.

We verified that the presence of emotion and stress can be detected in our dataset. Our baseline results for emotion classification using DNNs with acoustic, linguistic and visual features on our dataset are similar to reported results on other datasets such as IEMOCAP~\cite{busso2008iemocap} and MSP-Improv~\cite{busso2017msp}. 

%% file: Chapters/chap5.tex
\section{Motivation and Contributions}
Speech emotion recognition is an important component of any human centered system. But speech characteristics produced and perceived by a person can be influenced by a multitude of reasons, both desirable such as emotion, and undesirable such as noise. To train robust emotion recognition models, we need a large, yet realistic data distribution, but emotion datasets are often small and hence are augmented with noise. Often noise augmentation makes one important assumption, that the prediction label should remain the same in presence or absence of noise, which is true for automatic speech recognition  but not necessarily true for perception based tasks. In this chapter we make three novel contributions. We validate through crowdsourcing that the presence of noise does change the annotation label and hence may alter the original ground truth label. We then show how disregarding this knowledge and assuming consistency in ground truth labels propagates to downstream evaluation of ML models, both for performance evaluation and robustness testing. We end the chapter with a set of recommendations for noise augmentations in speech emotion recognition datasets.

\section{Introduction}
\label{sec:introduction}
Speech emotion recognition is increasingly included as a component in many real-world human-centered machine learning models. 
Modulations in speech can be produced for a multitude of reasons, both desirable and undesirable. In our case desirable modulations encode information that we want our model to learn and be informed by, such as  speaker characteristics or emotion. Undesirable modulations encode information that are extrinsic factors change with the environment, such as noise. In order to handle these modulations, we need large datasets that capture the range of possible speech variations and their relationship to emotion expression.  But, such datasets are generally not available for emotion tasks. To bridge this gap, researchers have proposed various methods to generate larger datasets.  One of the most common is noise augmentation. The baseline assumption of noise augmentation is that the labels of the emotion examples do not change once noise has been added~\cite{pappagari2021copypaste}. While this assumption can be confidently made for tasks such as automatic speech recognition (ASR), the same cannot be said for perception-based tasks, such as emotion recognition.

In this chapter, we question the assumption that the annotation label remains the same in the presence of noise.  We first create a noise augmented dataset and conduct a perception study to label the emotion of these augmented samples, focused on the type of noise in samples whose perception has changed or remained the same given the agumentation. 
We use the results from this study to classify the complete set of augmentation noises into two categories, \textit{perception-altering} (i.e., noises that may change the perception of emotion) and  \textit{perception-retaining} (i.e., noises that do not change the perception of emotion). 
We propose that the perception-altering noises should not be used in supervised learning or evaluation frameworks because we cannot confidently maintain that the original annotation holds for a given sample. 
We evaluate the effects of disregarding emotion perception changes by examining how the performance of emotion recognition models and analyses of their robustness change in unpredictable manners when we include samples that alter human perception in the training of these models. Lastly, we provide a set of recommendations for noise based augmentation of speech emotion recognition datasets based on our results.

Researchers have considered the impact of noise on emotion perception and thereby the annotation of emotions. [X] looked at how pink and white noises in varying intensities change the perception of emotion. Another set of research has concentrated on training and validating noise robust models with the assumption that intent label prediction remains consistent in the presence of noise. For example, [X] have looked at training student teacher models that aim to ignore the effect of noise introduced to the model. On the other hand [X] have proposed copy pasting various emotion segments together along with neutral noise to balance the classes in an emotion dataset, thus improving performance.

In this chapter, we claim that the standard assumption about perception and hence, label retention of emotion in the presence of noise may not hold true in a multiple noise categories. To understand which noises impact emotion perception, we use a common emotion dataset, IEMOCAP and introduce various kinds of noises to it, at varying signal to noise ratio (SNR) levels as well as at different positions in the sample. We then perform a crowdsourcing experiment that asks workers to annotate their perception of emotion for both the clean and the corresponding noise-augmented sample. This enables us to divide noise augmentation options into groups characterized by their potential to either influence or not influence human perception.

The results of the crowdsourcing experiments inform a series of empircal analyses focused on model performance and model robustness.  We first present an empirical evaluation of the effects of including perception-altering noises in training.  It will allow us to observe how the inclusion of perception-altering noises creates an impression of performance improvement.  We will discuss how this improvement is a myth, this new model will have learned to predict labels that are not truly associated with a given sample due to the perceptual effects of these noises.  
We consider both a general recurrent neural network (RNN) model and an end-to-end model for this purpose. We evaluate conditions in which novel augmentation noises are either introduced during training (matched) or seen for the first time during testing (mismatched).
The second empirical evaluation analyzes whether the gap in performance between the matched and mismatched conditions can be bridged using noise robust modeling techniques. 
The third and final evaluation is focused on the robustness of the model.  It will allow us to observe how the inclusion of these perception altering noises ultimately leads to a model that is more susceptible to attack compared to a model that does not include these noises.
We train an attack model for robustness testing. It considers a pool of noises and picks the best noise with a minimal SNR degradation that is able to change a model's prediction. We consider a condition in which the attack model has black-box access to the trained model.  The attack has a fixed number of allowed queries to the trained model, but not the internal gradients or structure (i.e., the attack model can only provide input and can only access the trained model's prediction). We test and  monitor the difference in the observed robustness of these aforementioned models.

We find that the crowdsourced labels do change in the presence of some kinds of noise. We then verify that the models perform worse on noisy samples when trained only on clean datasets. But, we show that this decrease in performance is different when using the complete set of noises for augmenting the test set vs. when only using the perception-retaining noises for augmentation. We show similar patterns for noise-robust models, specifically showing how there is an increased drop in performance for the end-to-end noise-robust model when excluding performance-altering noises during augmentation. We then discuss how our conventional metrics, those that look only at model performance, may be incorrectly asserting improvements as the model is learning to predict an emotion measure that is not in line with human perception. Troublingly, we find that the attack model is generally more effective when it has access to the set of all noises as compared to when excluding perception-altering noises for allowed augmentations. We also specifically find that given just a pool of carefully crafted reverberation modulations, the attack model can be successful in almost 65\% of the cases with minimal degradation in SNR and in less than ten queries to the trained model.
We end the chapter with a general set of recommendations for noise augmentations in speech emotion recognition datasets.

\section{Research Questions}
In this chapter, we investigate five research questions:

\textbf{Purpose 1: }Premise Validation through Crowdsourcing
\\
\textbf{RQ1: } Does the presence of noise affect emotion perception as evaluated by \emph{human raters}? Is this effect dependent on the utterance length, loudness, the type of the added noise, and the original emotion?
\\
\textbf{Reason: }Noise has been known to have masking effect on humans in specific situations. Hence, humans can often understand verbalized content even in presence of noise. Our goal is to understand whether the same masking effect extends to paralinguistic cues such as emotion, and to what extent. Our continuing claim from hereon remains that only noises that do not change human perception should be used for the training and evaluation of machine learning models. Not doing so, can lead to gains or drops in performance measurement that may not actually extend to real world settings. We call these changes "unverified" because we cannot, with certainity, be sure that the model should have predicted the original label (i.e., the label of the sample before noise was added) because the human did not neccessarily label the noisy instance with that same label.

\textbf{Purpose 2: }Noise Impact Quantification
\\
\textbf{RQ2: }  Can we verify previous findings that the presence of noise affects the performance of \emph{emotion recognition models}? Does this effect vary based on the type of the added noise? 
\\
\textbf{Reason: }We have known that presence of noise in data shifts the data distribution~\cite{chenchah2016speech}. This shift often leads to poor performance by machine learning models. We aim to quantify the amount of performance drop based on the type of noise in these systems, both, for any kind of noise, and then, specifically for noises that do not change human perception (perception-retaining).

\textbf{Purpose 3: }Denoising and Augmentation Benefits Evaluation
\\
\textbf{RQ3: } Does dataset augmentation (Q3a) and/or sample denoising (Q3b) help improve the robustness of emotion recognition models to unseen noise?
\\
\textbf{Reason: }We test whether the commonly-used methods for improving the performance of these models under distribution shifts is helpful. We focus on two main methods, augmentation and denoising. We specifically look at how performance changes when we augment with noises that include those that are perception-altering vs. when we exclude such noises.

\textbf{Purpose 4: }Model Robustness Testing Conditions
\\
\textbf{RQ4: } How does the robustness of a model to attacks compare when we are using test samples that with are augmented with perception-retaining noise vs. samples that are augmented with all types of noise, regardless of their effect on perception?
\\
\textbf{Reason: } Another major metric for any deployable machine learning algorithm is its performance on "unseen situations" or handling incoming data shifts (i.e., robustness testing). We test robustness using a noise augmentation algorithm that aims to forcefully and efficiently change a model's output by augmenting test samples with noise. We look at how often this algorithm is unsuccessful in being able to "fool" a model with its augmented samples. We look at the changes in frequency with which a model is successfully able to defend itself when the attack algorithm uses a set that includes all types of noises vs. when it only uses  perception-retaining noises.

\textbf{Purpose 5: }Recommendations
\\
\textbf{RQ5: } What are the recommended practices for speech emotion dataset augmentation and model deployment?
\\
\textbf{Reason: }We then provide a set of recommendations based on our empirical studies for deploying emotion recognition models in real world situations.

\section{Noise}
\label {subsec:noise}

We investigate the effects of two types of noise, environmental and signal distortion. Environmental noises are additive, while signal distortion noise involves other types of signal manipulation.

\label{sec:noise}
\subsection{Environmental Noise}
We define environmental noises (\textit{ENV}) as additive background noise, obtained from the ESC-50 dataset\cite{piczak2015dataset}\footnote{https://github.com/karoldvl/ESC-50}.  ESC-50 is generally used for noise contamination and environmental sound classification ~\cite{xu2021head}. 
These environmental sounds are representative of many types of noise seen in real world deployments, especially in the context of virtual and smart home conversational agents. We use the following categories:
 
\begin{itemize}
    \item Natural soundscapes  (\textit{Nat}), e.g., rain, wind.
    \item Human, non-speech sounds (\textit{Hum}), e.g., sneezing, coughing, laughing or crying in the background etc.
    \item Interior/domestic sounds (\textit{Int}), e.g., door creaks, clock ticks etc.
\end{itemize}

We manipulate three factors when adding the noise sources: 
\begin{itemize}
    \item 
    \textit{Position}: The position of the introduction of sound that: (i) starts and then fades out in loudness or (ii) occurs during the entirety of the duration of the utterance. In the second case, this complete additive background would represent a consistent noise source in real world (e.g., fan rotation).
    \item \textit{Quality Degradation}: The decrease in the signal to noise ratio (SNR) caused by the addition of the additive background noise at levels of 20dB, 10dB and 0dB.  This is used only when noise is added to the entirety of the utterance.
\end{itemize}

\subsection{Signal Distortion}
We define signal distortion noise as modulations that aren't additive in the background. These kinds of noise in the audio signal can occur from linguistic/paralinguistic factors, room environment, internet lags, or the physical locomotion of the speaker. 

We use the nine following categories: 
\begin{itemize}
    \item \textit{SpeedUtt}: The utterance is sped up by either 1.25$\times$ or 0.75$\times$. 
    \item \textit{SpeedSeg}: A random segment within an utterance is sped up by 1.25$\times$.  The package pyAudio~\footnote{https://people.csail.mit.edu/hubert/pyaudio/}
    that we used to speed up a segment did not permit slowing a segment down.  Thus, the 0.75$\times$ was not used here. 
    \item \textit{Fade}: The loudness of the utterance is faded by 2\% every second, which emulates the scenario of a user moving away from the speaker. The loudness is increased for fade in, and decreased for fade out. 
    \item \textit{Filler}: Non-verbal short fillers such as `uh', `umm' (from the same speaker) are inserted in the middle of a sentence. The insertion is either just the filler 
    or succeeded and preceded by a long pause 
    \footnote{Fillers are obtained by parsing audio files for a given speaker and finding occurrences of any of the options from the above mentioned set. We will release the extracted fillers per speaker for IEMOCAP}.  
    \item \textit{DropWord}: A randomly selected set of non-essential words belonging to the set: \{a, the, an, so, like, and\} are dropped from an utterance using word-aligned boundaries and stiching the audio segments together.
    \item \textit{DropLetters}: Following the same approach as drop word, letters are dropped in accordance with various linguistic styles chosen from the set: \{/\textbf{h}/+vowel, vowel+/n\textbf{d}/+consonant(next word), consonant+/\textbf{t}/+consonant(next word), \\ vowel+/\textbf{r}/+consonant, /ihn\textbf{g}/\}. This is supported by research that has studied phonological deletion or dropping of letters in the native US-English dialect~\cite{phono, yuan2011automatic}. 
    \item \textit{Laugh/Cry}: ``Sob'' and ``short-laughter'' sounds are added to the utterance.  They are obtained from AudioSet~\cite{gemmeke2017audio}.  
    \item \textit{Pitch}: The pitch is changed by $\pm$ 3 half octaves using the pyAudio library. 
\item \textit{Rev}: Room reverberation is added to the utterance using py-audio-effects (pysndfx\footnote{https://github.com/carlthome/python-audio-effects
}). We vary metrics such as reverberation ratio or room size to vary the type and intensity of reverberation added.
\end{itemize}

\subsection{Sampling and Noise-Perturbations}
\label{sec:datasample}
We randomly select 900 samples from the IEMOCAP dataset, which is far larger than the ones used for previous perception studies~\cite{parada-cabaleiro2017the, scharenborg2018the}. We select 100 samples from each activation and valence pair bin, i.e., 100 samples from the bin with activation: \emph{low}, valence: \emph{low}; 100 samples from the bin with activation: \emph{low}, and valence: \emph{mid}, and so on. This ensures that the chosen 900 samples cover the range of emotions expressed.  We impose another constraint on these 100 samples from each bin, 30 of them are shorter than the first quartile or greater than fourth quartile of utterance length in seconds to cover both extremities of the spectrum, and the remaining 70 belong in the middle. We also ensure that the selected samples had a 50-50 even split amongst gender.
We introduce noise to the 900 samples (Section~\ref{sec:data}).  Each sample is modulated in ten ways: four randomly chosen types of environmental noise and six randomly chosen signal distortion noise modulations, giving us a total of 9,000 noisy samples\footnote{We will release the script to create these files.}.  

\section{User study}
\label{sec:userstudy}
We first analyze the effects of noise on human perception by relabeling the noise-enhanced data using the Amazon Mechanical Turk (AMT) platform.  We use insights from this experiment to guide the machine learning analyses that follow.

\subsection{Crowdsourcing Setup} 
We recruited 147 workers using Amazon Mechanical Turk who self-identify as being from the United States and as native English speakers, to reduce the impact of cultural variability.  We ensured that each worker had $>98$\% approval rating and more than 500 approved Human Intelligence Tasks (HITs). We ensured that all workers understood the meaning of activation and valence using a qualification task that asked workers to rank emotion content similar to~\cite{jaiswal2019muse}.  The qualification task has two parts: (i) we explain the difference between valence and activation and how to identify those, and, (ii) we ask them to identify which of the two samples has a higher/lower valence and a higher/lower activation, to ensure that they have understood the concept of activation and valence annotations. All HIT workers were paid a minimum wage ($\$9.45/$hr), pro-rated to the minute. Each HIT was annotated by three workers.

For our main task, we created pairs that contained one original and one modulated sample.  We then asked each worker to annotate whether or not they perceived the pair to have the same emotion. If they said \textit{yes} for both activation and valence, the noisy sample was labeled \textit{same} and they could directly move to the next HIT. If they said \textit{no}, the noisy sample was labeled \textit{different}. In this case, they were asked to assess the activation and valence of the noisy sample using Self Assessment Manikins \cite{bradley1994measuring} on a scale of [1, 5] (similar to the original IEMOCAP annotation). 

We also include three kinds of attention checks:
	\begin{enumerate}
		\item We show two samples that have not been modified and ask them to decide if the emotion represented was different. If the person says yes, then the experiment ends.
		\item We observe the time spent on the task. If the time spent on the task is less than the combined length of both samples, then the user's qualification to annotate the HITs is rescinded and their responses are discarded.
		\item We show two samples, one which has a gold standard label, and another, which has been contaminated with significant noise (performance degradation $>$30dB), such that the resulting sample is incomprehensible. If people do not mark this set of samples as being different, the experiment ends.
	\end{enumerate}
\noindent The failure rate based on the above  criteria was 8\%.
We ensured the quality of the annotations by paying bonuses based on time spent, not just number of HITs, and by disqualifying annotators
if they annotated any sample (including those outside of the attention checks) more quickly than the combined length of the audio samples.

We then created two sets of labels for each noise-augmented clip.  The \emph{first type} of label compared a noise-augmented clip to its original.  The noise-augmented clip was labeled the \emph{same} if the modified and original clip were perceived to have the same valence or activation, otherwise it was labeled \emph{different}.  We created this label by taking the majority vote over all evaluations.  The \emph{second type} of label included valence and activation.  A noise-augmented clip was given the average valence and activation over all evaluations.  

The inter-annotator agreement was measured using Cohen's kappa. Conventionally, when estimating Cohen's kappa, annotators are not considered as individuals, instead reducing annotators to the generic 1, 2, and 3. The challenge is that this often leads to artificially inflated inter-annotator agreement because individual characteristics and behavior of a particular worker are not taken under consideration~\cite{hoek2017evaluating}. We take a different approach, creating a table for the calculation of the statistic that considers annotators as individuals with separate entries for each clip, following the approach of~\cite{hoek2017evaluating}.  If an annotator didn't evaluate a given clip, the cell has a null (missing data) value.
We found that the Cohen's kappa was 79\% for activation and 76\% for valence~\footnote{The sample name, code to create the paired noisy examples, and the resulting annotations will be made available for further research}.



\begin{algorithm}
\scriptsize
\caption{Pseudo-code for testing model robustness. \textit{Exit Code} is \textit{\textcolor{blue}{Success}} when the algorithm finds a noise-augmented version of the sample that the model changes prediction for. \textit{Exit Code} is \textit{\textcolor{red}{Failure}} when the model maintains its predictions over the any of the noise-augmented versions tried.\label{algo-attack}}
Randomly sample 1 noise variation from each category mentioned in Section~\ref{sec:noise}.\;
$numAttempts = 0$\;
\For{each noise in selected random noises:}
{Add noise to the sample such that the decrease in SNR is 1.\;
Get the classifier output with this new sample variation.\;
$numAttempts += 1$\;
\If{$numAttempts>k$}{\Return{Exit Code = \textcolor{red}{Failure}}}
\If {classifier output changes}{\Return{Exit Code = \textcolor{blue}{Success}}}
}
 
\For{each noise in selected random noises:}
{Add noise to the sample such that the decrease in SNR is 5.\;
Get the classifier output with this new sample variation.\;
$numAttempts += 1$\;
\If{$numAttempts>k$}{\Return{Exit Code = \textcolor{red}{Failure}}}
\If {classifier output changes}{\Return{Exit Code = \textcolor{blue}{Success}}}
\If {classifier output changes}
{\While{classifier output does not change}{
Iterate over all SNR decreases from 2-5\;
Get classifier output for the modified sample\;
$numAttempts += 1$\;
\If{$numAttempts>k$}{\Return{Exit Code = \textcolor{red}{Failure}}}
\If{classifier output changes}{\Return{Exit Code = \textcolor{blue}{Success}}}
}
}
}
 
\For{each noise in selected random noises:}
{Add noise to the sample such that the decrease in SNR is 10.\;
Get the classifier output with this new sample variation.\;
\If {classifier output changes}{\Return{Exit Code = \textcolor{blue}{Success}}}
\If {classifier output changes}
{\While{classifier output does not change}{
Iterate over all SNR decreases from 6-10
Get classifier output for the modified sample\;
$numAttempts += 1$\;
\If{$numAttempts>k$}{\Return{Exit Code = \textcolor{red}{Failure}}}
\If{classifier output changes}{\Return{Exit Code = \textcolor{blue}{Success}}}
}
}
}
 
\Return{Exit Code = \textcolor{red}{Failure}}

\end{algorithm}

\begin{table}
\centering
\caption{The table shows the ratio of samples marked by human evaluators as imperceptible to difference in emotion perception. \textit{V: Valence, A: Activation, $\delta$V:Average change in Valence on addition of noise, $\delta$A:Average change in Activation on addition of noise.}}
\label{tab:model_res}
\begin{tabular}{llllll}
\toprule
\multicolumn{2}{l}{}                    & V    & A  & $\delta$V & $\delta$A    \\ 
\midrule
\multicolumn{6}{l}{\textbf{Environmental Noise}}       \\ 
\midrule
NatSt      &                            & 0.01 & 0.00  \\
NatdB (Co) & -10dB                      & 0.01 & 0.00  \\
           & Same                       & 0.02 & 0.00  \\
           & +10dB                      & 0.03 & 0.01  \\ 
\hline
HumSt      &                            & 0.01 & 0.00  \\
HumdB (Co) & -10dB                      & 0.03 & 0.00  \\
           & Same                       & 0.02 & 0.00  \\
           & +10dB                      & 0.04 & 0.01  \\ 
\hline
IntSt      &                            & 0.05 & 0.01  \\
IntdB (Co) & -10dB                      & 0.02 & 0.01  \\
           & Same                       & 0.02 & 0.00  \\
           & +10dB                      & 0.04 & 0.01  \\ 
\midrule
\multicolumn{6}{l}{\textbf{Signal Distortion}}           \\ 
\midrule
SpeedSeg   &                            & 0.01 & 0.0   \\ 
\hline
Fade       & In                         & 0.04 & 0.01  \\
           & Out                        & 0.04 & 0.00  \\ 
\hline
DropWord      &                            & 0.01 & 0.00  \\ 
\hline
DropLetters     &                            & 0.01 & 0.00  \\ 
\hline
Reverb     &                            & 0.04 & 0.01  \\ 
\midrule
Filler     & L                          & 0.10 & 0.06  \\
           & S                          & 0.06 & 0.03  \\ 
\hline
Laugh      &                            & 0.16 & 0.17  & + .11 & + .26\\ 
\hline
Cry        &                            & 0.20 & 0.22  & - .20 & - .43\\ 
\hline
SpeedUtt   & 1.25x                      & 0.13 & 0.03 & - .10 & - .13 \\
           & 0.75x                      & 0.28 & 0.06 & - .18 & - .23 \\ 
\hline
Pitch      & 1.25x                      & 0.22 & 0.07 & - .11 & + .19 \\
           & 0.75x                      & 0.29 & 0.10 & - .07 & - .15 \\
\bottomrule
\end{tabular}
\end{table}

\begin{table*}
	\caption{State of the art model performance in terms of UAR when using the general versions of traditional deep learning and end-to-end deep learning models. No noise refers to clean speech, all noises refers to the combined set of perception-retaining and perception-altering noise.  The environmental and signal distortion categories shown include only the perception-retaining noises.  As a reminder, samples in the all noises category have an uncertain ground truth, the row is marked with two stars ($\ast\ast$). \textit{V: Valence, A: Activation, Clean: Training on clean dataset, Clean$+$Noise | Mismatch: Cleaning on noisy dataset and testing on mismatched noisy partition, Clean$+$Noise | Match: Cleaning on noisy dataset and testing on matched noisy partition.} Random chance UAR is 0.33. 
    }
	\scriptsize
	\renewcommand{\arraystretch}{1.3}
	\setlength{\tabcolsep}{0.5em}
	\centering
	\label{tab:sota}
	\begin{tabular}{c|ccccccccccccccccccc} \toprule
	\multicolumn{1}{c}{} &  &  &  &  & \multicolumn{7}{c}{\textbf{Traditional Deep Neural Network}} &  & \multicolumn{7}{c}{\textbf{End-To-End Deep Neural Network}} \\ \cmidrule[\heavyrulewidth]{1-20}
	\multicolumn{1}{c}{} &  &  &  &  & \multicolumn{2}{c}{\multirow{2}{*}{\textbf{Clean}}} &  & \multicolumn{4}{c}{\textbf{Clean+Noise}} &  & \multicolumn{2}{c}{\multirow{2}{*}{\textbf{Clean}}} &  & \multicolumn{4}{c}{\textbf{Clean+Noise}} \\ \cline{9-12}\cline{17-20}
	\multicolumn{1}{c}{} &  &  &  &  & \multicolumn{2}{c}{} &  & \multicolumn{2}{c}{\textbf{\textit{Mismatch}}} & \multicolumn{2}{c}{\textbf{\textit{Match}}} &  & \multicolumn{2}{c}{} &  & \multicolumn{2}{c}{\textbf{\textit{Mismatch}}} & \multicolumn{2}{c}{\textbf{\textit{Match}}} \\ 
	\cmidrule(lr){6-7}\cmidrule(lr){9-10}\cmidrule(lr){11-12}\cline{14-15}\cmidrule(lr){17-18}\cmidrule(lr){19-20}
	\multicolumn{1}{c}{} &  &  &  &  & A & V &  & A & V & A & V &  & A & V &  & A & V & A & V \\ \midrule
	
	\multicolumn{4}{c|}{\textbf{No Noise}} &  &  
	
	0.67 &  0.59 &  & - & - & - & - &  & 
	
	0.70 & 0.63 &  & - &-  & - & - \\ \midrule

 \multicolumn{4}{c|}{\textbf{All Noises**}} &  &  
	
	0.40 &  0.38 &  & 0.55 & 0.42 & 0.66 & 0.59 &  & 
	
	0.70 & 0.63 &  & 0.44 & 0.38  & 0.67 & 0.60 \\ 

	\multicolumn{4}{c|}{\textbf{Perception Retaining Noises}} &  &  
	
	0.50 &  0.42 &  & 0.57 & 0.48 & 0.60 & 0.52 &  & 
	
	0.53 & 0.45 &  & 0.60 & 0.50  & 0.62 & 0.54 \\ \midrule
	
	\multirow{12}{*}{\rotatebox[origin=c]{90}{\textbf{ Environmental Category}}} & \textbf{Nature} & At Start &\multicolumn{1}{c|}{} &  & 
	 0.50 & 0.45 &  & 0.61 & 0.53 & 0.63 & 0.55 &  & 
	
	0.56 & 0.48 &  & 0.64 & 0.55 & 0.66 & 0.58 \\
	
	 &  & \multirow{3}{*}{\rotatebox[origin=c]{90}{Cont.}} & \multicolumn{1}{||c|}{-5dB} &  & 
	 0.45 & 0.39 &  & 0.55 & 0.44 & 0.59 & 0.50 &  & 
	 
	 0.49 & 0.42 &  & 0.58 & 0.47 & 0.62 & 0.53 \\
	 
	 &  &  & \multicolumn{1}{||c|}{-10dB} &  & 
	 
	 0.42 & 0.35 &  & 0.56 & 0.46 & 0.59 & 0.49 &  & 
	 
	 0.47 & 0.38 &  & 0.57 & 0.48 & 0.61 & 0.51 \\
	 
	 &  &  & \multicolumn{1}{||c|}{-20dB} &  & 
	 
	 0.40 & 0.35 &  & 0.51 & 0.44 & 0.55 & 0.47 &  & 
	 
	 0.47 & 0.39 &  & 0.52 & 0.46 & 0.56 & 0.50 \\
	   \cmidrule{2-20}
	
	 & \textbf{Interior} & At Start & \multicolumn{1}{c|}{}&  & 
	 
	 0.53 & 0.44 &  & 0.61 & 0.52 & 0.64 & 0.57 &  & 
	 
	 0.57 & 0.47 &  & 0.64 & 0.56 & 0.66 & 0.59 \\
	 
	 &  & \multirow{3}{*}{\rotatebox[origin=c]{90}{Cont}} & \multicolumn{1}{||c|}{-5dB} &  & 
	 0.46 & 0.36 &  & 0.55 & 0.44 & 0.58 & 0.49 & &
	 0.49 & 0.39 &  & 0.59 & 0.49 & 0.62 & 0.51 \\

	 &  &  & \multicolumn{1}{||c|}{-10dB} &  & 
	 
	 0.44 & 0.36 &  & 0.54 & 0.43 & 0.57 & 0.48 &  & 
	 
	 0.49 & 0.39 &  & 0.56 & 0.45 & 0.59 & 0.52 \\
	 
	 &  &  & \multicolumn{1}{||c|}{-20dB} &  & 
	 
	 0.40 & 0.35 &  & 0.52 & 0.44 & 0.55 & 0.49 &  & 
	 
	 0.46 & 0.37 &  & 0.56 & 0.45 & 0.58 & 0.51 \\
	   \cmidrule{2-20}
	
	 & \textbf{Human} & At Start & \multicolumn{1}{c|}{} &  & 
	 
	 0.52 & 0.45 &  & 0.60 & 0.51 & 0.63 & 0.55 &  & 
	 
	 0.58 & 0.47 &  & 0.62 & 0.52 & 0.66 & 0.57 \\
	 
	 &  & \multirow{3}{*}{\rotatebox[origin=c]{90}{Cont}} & \multicolumn{1}{||c|}{-5dB} &  & 
	 
	 0.45 & 0.37 &  & 0.52 & 0.43 & 0.55 & 0.48 &  & 
	 
	 0.49 & 0.40 &  & 0.56 & 0.44 & 0.57 & 0.50 \\
	 
	 &  &  & \multicolumn{1}{||c|}{-10dB} &  & 
	 
	 0.42 & 0.34 &  & 0.51 & 0.43 & 0.53 & 0.47 &  & 
	 
	 0.49 & 0.38 &  & 0.53 & 0.45 & 0.55 & 0.49 \\
	 
	 &  &  & \multicolumn{1}{||c|}{-20dB} &  & 
	 
	 0.40 & 0.34 &  & 0.50 & 0.41 & 0.53 & 0.46 &  & 
	 
	 0.46 & 0.38 &  & 0.54 & 0.43 & 0.56 & 0.48 
	 
	\rule{0pt}{3ex}    \\
	 \cmidrule{1-20}
	
	\multirow{6}{*}{\rotatebox[origin=c]{90}{\textbf{ Signal Distortion }}} & 
	
	\multicolumn{3}{l|}{Speed Segment} &  & 
	
	0.61 & 0.52 &  & 0.63 & 0.53 & 0.64 & 0.55 &  & 
	
	0.63 & 0.55 &  & 0.64 & 0.55 & 0.67 & 0.58 \\
	
	 & \multicolumn{2}{l}{Fade} & \multicolumn{1}{||c|}{In} &  & 
	 
	 0.62 & 0.53 &  & 0.65 & 0.55 & 0.67 & 0.58 &  & 
	 
	 0.64 & 0.55 &  & 0.67 & 0.58 & 0.68 & 0.59 \\
	 
	 &  & & \multicolumn{1}{||c|}{Out} &  & 
	 
	 0.61 & 0.51 &  & 0.62 & 0.54 & 0.64 & 0.57 &  & 
	 
	 0.63 & 0.54 &  & 0.64 & 0.56 & 0.66 & 0.59 \\
	 
	 & \multicolumn{3}{l|}{DropWord} &  & 
	 
	 0.64 & 0.56 &  & 0.65 & 0.56 & 0.67 & 0.59 &  & 
	 
	 0.65 & 0.58 &  & 0.67 & 0.58 & 0.69 & 0.61 \\
	 
	 & \multicolumn{3}{l|}{DropLetters} &  & 
	 
	 0.65 & 0.58 &  & 0.69 & 0.60 & 0.71 & 0.62 &  & 
	 
	 0.66 & 0.59 &  & 0.72 & 0.60 & 0.74 & 0.63 \\
	 
	 & \multicolumn{3}{l|}{Reverb} &  & 
	 
	 0.43 & 0.37 &  & 0.50 & 0.43 & 0.53 & 0.45 &  & 
	 
	 0.35 & 0.34 &  & 0.51 & 0.42 & 0.55 & 0.46 \rule{0pt}{2ex} \\ 
    
	
	 \bottomrule
	
	\end{tabular}
\end{table*}

\begin{table*}
	\caption{Noise-Robust (NR) state of the art model performance in terms of UAR when using the noise-robust versions of traditional deep learning and end-to-end deep learning models. No noise refers to clean speech, all noises refers to the combined set of perception-retaining and perception-altering noise.  The environmental and signal distortion categories shown include only the perception-retaining noises.  As a reminder, samples in the all noises category have an uncertain ground truth, the row is marked with two stars ($\ast\ast$). \textit{V: Valence, A: Activation, Clean: Training on clean dataset, Clean$+$Noise | Mismatch: Cleaning on noisy dataset and testing on mismatched noisy partition, Clean$+$Noise | Match: Cleaning on noisy dataset and testing on matched noisy partition, NR: Noise Robust versions of the corresponding models} Random chance UAR is 0.33. 
    }
	\scriptsize
	\renewcommand{\arraystretch}{1.3}
	\setlength{\tabcolsep}{0.5em}
	\centering
	\label{tab:noise-robust}
	\begin{tabular}{c|ccccccccccccccccccc} \toprule
	\multicolumn{1}{c}{} &  &  &  &  & \multicolumn{7}{c}{\textbf{NR-Traditional Deep Neural Network}} &  & \multicolumn{7}{c}{\textbf{NR-End-To-End Deep Neural Network}} \\ \cmidrule[\heavyrulewidth]{1-20}
	\multicolumn{1}{c}{} &  &  &  &  & \multicolumn{2}{c}{\multirow{2}{*}{\textbf{Clean}}} &  & \multicolumn{4}{c}{\textbf{Clean+Noise}} &  & \multicolumn{2}{c}{\multirow{2}{*}{\textbf{Clean}}} &  & \multicolumn{4}{c}{\textbf{Clean+Noise}} \\ \cline{9-12}\cline{17-20}
	\multicolumn{1}{c}{} &  &  &  &  & \multicolumn{2}{c}{} &  & \multicolumn{2}{c}{\textbf{\textit{Mismatch}}} & \multicolumn{2}{c}{\textbf{\textit{Match}}} &  & \multicolumn{2}{c}{} &  & \multicolumn{2}{c}{\textbf{\textit{Mismatch}}} & \multicolumn{2}{c}{\textbf{\textit{Match}}} \\ 
	\cmidrule(lr){6-7}\cmidrule(lr){9-10}\cmidrule(lr){11-12}\cline{14-15}\cmidrule(lr){17-18}\cmidrule(lr){19-20}
	\multicolumn{1}{c}{} &  &  &  &  & A & V &  & A & V & A & V &  & A & V &  & A & V & A & V \\ \midrule
	
	\multicolumn{4}{c|}{\textbf{No Noise}} &  &  
	
	0.67 &  0.59 &  & - & - & - & - &  & 
	
	0.70 & 0.63 &  & - &-  & - & - \\ \midrule

 \multicolumn{4}{c|}{\textbf{All Noises**}} &  &  
	
	0.44 &  0.40 &  & 0.58 & 0.44 & 0.68 & 0.60 &  & 
	
	0.50 & 0.40 &  & 0.50 & 0.40  & 0.72 & 0.61 \\ \midrule

	\multicolumn{4}{c|}{\textbf{Perception Retaining Noises}} &  &  
	
	0.52 &  0.44 &  & 0.59 & 0.50 & 0.61 & 0.51 &  & 
	
	0.55 & 0.48 &  & 0.61 & 0.52  & 0.63 & 0.54 \\ \midrule
	
	\multirow{12}{*}{\rotatebox[origin=c]{90}{\textbf{ Environmental Category}}} & \textbf{Nature} & At Start &\multicolumn{1}{c|}{} &  & 
	 0.54 & 0.49 &  & 0.63 & 0.55 & 0.63 & 0.55 &  & 
	
	0.57 & 0.50 &  & 0.66 & 0.55 & 0.67 & 0.56 \\
	
	 &  & \multirow{3}{*}{\rotatebox[origin=c]{90}{Cont.}} & \multicolumn{1}{||c|}{-5dB} &  & 
	 0.50 & 0.42 &  & 0.58 & 0.50 & 0.60 & 0.52 &  & 
	 
	 0.49 & 0.42 &  & 0.61 & 0.51 & 0.63 & 0.53 \\
	 
	 &  &  & \multicolumn{1}{||c|}{-10dB} &  & 
	 
	 0.48 & 0.38 &  & 0.59 & 0.48 & 0.61 & 0.51 &  & 
	 
	 0.50 & 0.46 &  & 0.63 & 0.52 & 0.65 & 0.53 \\
	 
	 &  &  & \multicolumn{1}{||c|}{-20dB} &  & 
	 
	 0.44 & 0.38 &  & 0.55 & 0.49 & 0.59 & 0.51 &  & 
	 
	 0.50 & 0.43 &  & 0.53 & 0.46 & 0.58 & 0.47 \\
	   \cmidrule{2-20}
	
	 & \textbf{Interior} & At Start & \multicolumn{1}{c|}{}&  & 
	 
	 0.53 & 0.44 &  & 0.61 & 0.52 & 0.65 & 0.54 &  & 
	 
	 0.58 & 0.52 &  & 0.63 & 0.56 & 0.66 & 0.58 \\
	 
	 &  & \multirow{3}{*}{\rotatebox[origin=c]{90}{Cont}} & \multicolumn{1}{||c|}{-5dB} &  & 
	 
	 0.46 & 0.36 &  & 0.55 & 0.44 & 0.59 & 0.47 &  & 
	 
	 0.51 & 0.42 &  & 0.58 & 0.48 & 0.62 & 0.52 \\
	 
	 &  &  & \multicolumn{1}{||c|}{-10dB} &  & 
	 
	 0.44 & 0.36 &  & 0.54 & 0.43 & 0.57 & 0.43 &  & 
	 
	 0.49 & 0.44 &  & 0.57 & 0.48 & 0.61 & 0.50 \\
	 
	 &  &  & \multicolumn{1}{||c|}{-20dB} &  & 
	 
	 0.40 & 0.35 &  & 0.52 & 0.44 & 0.55 & 0.46 &  & 
	 
	 0.46 & 0.40 &  & 0.55 & 0.48 & 0.58 & 0.49 \\
	   \cmidrule{2-20}
	
	 & \textbf{Human} & At Start & \multicolumn{1}{c|}{} &  & 
	 
	 0.52 & 0.45 &  & 0.60 & 0.51 & 0.63 & 0.52 &  & 
	 
	 0.59 & 0.49 &  & 0.63 & 0.53 & 0.66 & 0.56 \\
	 
	 &  & \multirow{3}{*}{\rotatebox[origin=c]{90}{Cont}} & \multicolumn{1}{||c|}{-5dB} &  & 
	 
	 0.45 & 0.37 &  & 0.52 & 0.43 & 0.55 & 0.46 &  & 
	 
	 0.49 & 0.42 &  & 0.56 & 0.48 & 0.58 & 0.50 \\
	 
	 &  &  & \multicolumn{1}{||c|}{-10dB} &  & 
	 
	 0.42 & 0.34 &  & 0.51 & 0.43 & 0.54 & 0.44 &  & 
	 
	 0.47 & 0.38 &  & 0.54 & 0.49 & 0.55 & 0.45 \\
	 
	 &  &  & \multicolumn{1}{||c|}{-20dB} &  & 
	 
	 0.40 & 0.34 &  & 0.50 & 0.41 & 0.52 & 0.43 &  & 
	 
	 0.43 & 0.38 &  & 0.54 & 0.45 & 0.58 & 0.49 \\ 
	 \cmidrule{1-20}
	
	\multirow{6}{*}{\rotatebox[origin=c]{90}{\textbf{ Signal Distortion }}} & 
	
	\multicolumn{3}{l|}{Speed Segment} &  & 
	
	0.63 & 0.55 &  & 0.65 & 0.57 & 0.67 & 0.58 &  & 
	
	0.66 & 0.58 &  & 0.67 & 0.59 & 0.67 & 0.60 \\
	
	 & \multicolumn{2}{l}{Fade} & \multicolumn{1}{||c|}{In} &  & 
	 
	 0.64 & 0.55 &  & 0.66 & 0.57 & 0.68 & 0.59 &  & 
	 
	 0.65 & 0.58 &  & 0.67 & 0.59 & 0.69 & 0.60 \\
	 
	 &  & & \multicolumn{1}{||c|}{Out} &  & 
	 
	 0.64 & 0.56 &  & 0.65 & 0.57 & 0.67 & 0.58 &  & 
	 
	 0.66 & 0.57 &  & 0.68 & 0.59 & 0.69 & 0.63 \\
	 
	 & \multicolumn{3}{l|}{DropWord} &  & 
	 
	 0.67 & 0.60 &  & 0.66 & 0.58 & 0.66 & 0.59 &  & 
	 
	 0.68 & 0.60 &  & 0.69 & 0.60 & 0.69 & 0.60 \\
	 
	 & \multicolumn{3}{l|}{DropLetters} &  & 
	 
	 0.67 & 0.60 &  & 0.65 & 0.62 & 0.66 & 0.60 &  & 
	 
	 0.68 & 0.64 &  & 0.69 & 0.60 & 0.69 & 0.60 \\
	 
	 & \multicolumn{3}{l|}{Reverb} &  & 
	 
	 0.48 & 0.41 &  & 0.56 & 0.47 & 0.58 & 0.48 &  & 
	 
	 0.52 & 0.40 &  & 0.55 & 0.45 & 0.60 & 0.45 \\ 

  \bottomrule
	
	\end{tabular}
\end{table*}

\begin{table*}
	\centering
	\scriptsize
	\renewcommand{\arraystretch}{1.3}
	\caption{Success of misclassification attempts on different models with varying number of allowed attempts (lower is better). As a reminder, samples in the all noises category have an uncertain ground truth, the row is marked with two stars ($\ast\ast$). Reverberation (reverb) is a perception-retaining noise that is also analyzed separately.  \textit{Trad: Traditional Deep Learning Model, E2E: End to End deep learning model, NR: Noise Robust version of the deep learning model.}}
	\begin{tabular}{@{}ccc|cccc|cccc} \toprule
	
	& \multirow{2}{*}{ \myboxsm{\textbf{Noise Set}} } & \multirow{2}{*}{\myboxsm{\textbf{No. of Attempts}}  } & 
	\multicolumn{4}{c}{\textbf{Activation}} & \multicolumn{4}{c}{\textbf{Valence}} 
	\\ \cmidrule{4-7} \cmidrule{8-11}
	 & 
	 & 
	 & \textbf{Trad} & \textbf{E2E} & \textbf{NR-Trad} & \textbf{NR-E2E}  & 
	 \textbf{Trad} & \textbf{E2E} & \textbf{NR-Trad} & \textbf{NR-E2E} \\ 
	  \cmidrule[\heavyrulewidth]{1-3}
	 \cmidrule[\heavyrulewidth]{4-7}  \cmidrule[\heavyrulewidth]{8-11}
	
	\multicolumn{1}{c|}{\multirow{10}{*}{
	
	\mybox{\textbf{Performance Impact Per Noise Is Unknown}}}} & 
	
	\multirow{4}{*}{
	
	\rotatebox[origin=c]{90}{All Noises**}} & 5 &
	
	0.29 & 0.22 & 0.15 & 0.10 &  
	
	0.11 & 0.15 & 0.13 & 0.05 \\
	
	 \multicolumn{1}{c|}{}&  & 15 &   
	 
	 0.31 & 0.33 & 0.31 & 0.32 &   
	 
	 0.23 & 0.24 & 0.22 & 0.22 \\

	 \multicolumn{1}{c|}{}&  & 25 &   
	 
	 0.40 & 0.28 & 0.40 & 0.33 &   
	 
	 0.22 & 0.19 & 0.21 & 0.20 \\
	 
	 \multicolumn{1}{c|}{}&  & inf &   
	 
	 0.43 & 0.41 & 0.44 & 0.35 &   
	 
	 0.25 & 0.20 & 0.26 & 0.18 \\ 
	 
	 \cmidrule{2-3}\cmidrule(r){4-11}
	 \multicolumn{1}{c|}{}& \multirow{4}{*}{
	 
	 \myboxmed{Perception-Retaining}} & 5 &   
	 
	 0.18 & 0.11 & 0.07 & 0.05 &   
	 
	 0.11 & 0.05 & 0.02 & 0.02 \\

	 \multicolumn{1}{c|}{}&  & 15 &  
	 
	 0.25 & 0.12 & 0.25 & 0.15 &   
	 
	 0.14 & 0.08 & 0.18 & 0.10 \\
	 
	 \multicolumn{1}{c|}{}&  & 25 &   
	 
	 0.32 & 0.24 & 0.32 & 0.20 &   
	 
	 0.13 & 0.10 & 0.11 & 0.09 \\
	 
	 \multicolumn{1}{c|}{}&  & inf &   
	 
	 0.40 & 0.26 & 0.36 & 0.23 &   
	 
	 0.19 & 0.14 & 0.17 & 0.14 \\ 
	 
  \cmidrule(r){4-11}
	 
	 \multicolumn{1}{c|}{}& \multirow{2}{*}{
	 
	 \myboxsm{Reverb}} & 5 &   
	 
	 0.33 & 0.15 & 0.22 & 0.18 &   
	 
	 0.20 & 0.14 & 0.22 & 0.09 \\
	 
	 \multicolumn{1}{c|}{}&  & 15 &   
	 
	 0.40 & 0.23 & 0.30 & 0.21 &   
	 
	 0.30 & 0.13 & 0.34 & 0.16  
	 
	  \rule{0pt}{3ex}  \\ \midrule
	\multicolumn{1}{c|}{\multirow{10}{*}{
	
	\mybox{\textbf{Performance Impact Per Noise Is Known}}}} 
	
	& \multirow{4}{*}{
	
	\rotatebox[origin=c]{90}{All Noises**}} & 5 &   
	
	0.33 & 0.28 & 0.24 & 0.22 &   
	
	0.15 & 0.14 & 0.12 & 0.12 \\
	
	 \multicolumn{1}{c|}{}&  & 15 &   
	 
	 0.38 & 0.38 & 0.32 & 0.33 &   
	 
	 0.20 & 0.21 & 0.18 & 0.16 \\
	 
	 \multicolumn{1}{c|}{}&  & 25 &   
	 
	 0.52 & 0.32 & 0.44 & 0.37 &   
	 
	 0.25 & 0.16 & 0.18 & 0.16 \\
	 
	 \multicolumn{1}{c|}{}&  & inf &   
	 
	 0.54 & 0.42 & 0.46 & 0.41 &   
	 
	 0.24 & 0.20 & 0.22 & 0.22 \\ 
	 
	 \cmidrule{2-3}
   \cmidrule(r){4-11}
	 
	 & \multicolumn{1}{|c}{\multirow{4}{*}{
	 
	 \myboxmed{Perception-Retaining}}} & 5 &   
	 
	 0.29 & 0.16 & 0.22 & 0.13 &   
	 
	 0.14 & 0.10 & 0.15 & 0.08 \\
	 
	 \multicolumn{1}{c|}{}&  & 15 &   
	 
	 0.32 & 0.32 & 0.32 & 0.28 &   
	 
	 0.14 & 0.15 & 0.16 & 0.12 \\
	 
	 \multicolumn{1}{c|}{}&  & 25 &   
	 
	 0.47 & 0.30 & 0.47 & 0.29 &   
	 
	 0.22 & 0.18 & 0.23 & 0.16 \\
	 
	 \multicolumn{1}{c|}{}&  & inf &   
	 
	 0.51 & 0.36 & 0.50 & 0.32 &   
	 
	 0.22 & 0.19 & 0.25 & 0.17 \\ 
	 
  \cmidrule(r){4-11}
	 
	 \multicolumn{1}{c|}{}& \multirow{2}{*}{
	 
	 \myboxsm {Reverb}} & 5 &   
	 
	 0.38 & 0.22 & 0.33 & 0.22 &   
	 
	 0.28 & 0.20 & 0.31 & 0.21 \\
	 
	 \multicolumn{1}{c|}{}&  & 15 &   
	 
	 0.47 & 0.29 & 0.41 & 0.28 &   
	 
	 0.30 & 0.23 & 0.35 & 0.26 
	 \rule{0pt}{3ex}    \\
	 \bottomrule
	\end{tabular}
 \label{tab:robustnessperf}
\end{table*}

\section{Methods}
\begin{table}
\small
\centering
\caption{Hyper-parameters used to select the best performing model on validation subset whilst training the traditional deep learning model.}
\label{tab:model_hyp1}
\begin{tabular}{ll}
\toprule
Hyper-parameter & Values\\
\midrule
\textbf {Traditional} & \\
\midrule
No. of Convolution Kernels & \{64, 128\} \\
Convolution Kernels Width & \{2\} \\
Number of Convolution Layers & \{5\} \\
Number of GRU layers & \{1, 2, 3\} \\
Pooling Kernel Width & \{2, 4\}\\
GRU Layers Width & \{32, 64\}\\
Number of Dense Layers & \{1, 2, 3\}\\
\midrule
\textbf{End to End} & \\
\midrule
No. of Dense Layers & \{1, 2\}\\
\bottomrule
\end{tabular}
\end{table}
We now describe the emotion recognition approaches, presenting two separate pipelines, one that relies upon direct feature extraction (Section~\ref{sec:dnn_features}) and the other that is end-to-end (Section~\ref{sec:dnn_endtoend}).  This allows us to investigate whether noise has a consistent effect.  We discuss approaches to improve noise robustness by training models with noise-augmented data or denoised data (Section~\ref{sec:noise_aug}).  Finally, we describe the setup and evaluation of the model robustness using an untargeted model misclassification test, 
which measures a model's fragility in terms of how likely it is that the model's decisions will change when specific types of noise are observed at test time (Section~\ref{sec:attack}).

\subsection{Creation of Data Partitions} 
We use a subject-independent five-fold cross validation scheme to select our train, test and validation sets.  In the first iteration, sessions 1-3 are used for training, session 4 is used as validation, and session 5 is used for testing.  This is repeated in a round-robin fashion, resulting in each session serving as a validation and a test fold. We also divide possible noises in two different categories based on results of crowdsourcing study (see Section~\ref{sec:rq1}).  The first category is \emph{perception-altering}, those that changed perception of humans and hence cannot be used for model training or evaluation with the old annotations.  The second category is \emph{perception-retraining}, those that did not change human perception, and hence, the model should produce no change in predictions when using those noise categories for sample augmentation. 

We use the noise categories (seeSection~\ref{subsec:noise}) in two varying circumstances.  The first category is \textit{matched}, where both the training and testing sets are augmented with same kinds of noise (e.g., both have nature-based sounds in them).  The second category is \textit{mismatched},
where the testing set is augmented with a noise \emph{category} not used for augmenting the training set (e.g., only the test set is augmented with nature-based noise while the train set is augmented with human or interior noises).

\subsection{Traditional Deep Learning Network}
\label{sec:dnn_features}

We first explore a common ``traditional'' deep learning network that is used in speech emotion recognition.  In this method we extract Mel Filterbank (MFB) features as input to a model composed of convolutional and gated recurrent unit (GRU) layers.

\subsubsection{Features}
We extract 40-dimensional Mel Filterbank (MFB) features using a 25-millisecond Hamming window with a step-size of 10-milliseconds using python-speech-features~\footnote{https://github.com/jameslyons/python\_speech\_features}. Each utterance is represented as a sequence of 40-dimensional feature vectors. We $z$-normalize the acoustic features using parameters extracted from the training dataset.  During each cross-validation fold, the parameters are chosen from the training data and are applied to both the validation and testing data. 

\subsubsection{Network}
Our baseline network is a state-of-art single utterance emotion classification model which has been used in previous research~\cite{aldeneh2017pooling, khorram2017capturing, krishna2018study}. The extracted MFBs are processed using a set of convolution layers and GRUs (see Table~\ref{tab:model_hyp1} for the hyperparameters used for these layers). The output of these layers is then fed through a mean pooling layer to produce an acoustic representation which is then fed into a set of dense layers to classify activation or valence. 


\subsubsection{Training.} 
We implement the models using the Keras library~\cite{chollet2015}. We use a cross-entropy loss function for each task (e.g., valence or activation).  We learn the model parameters using the RMSProp optimizer. We train our networks for a maximum of 50 epochs and use early stopping if the validation loss does not improve after five consecutive epochs. Once the training process ends, we revert the network's weights to those that achieved the lowest validation loss. We repeat the experiment five times.  We report the results in terms of  Unweighted Average Recall (UAR, chance is 0.33), averaged over all test samples and five repetitions.  We compare the performance of different models or the same model in different noisy conditions/partitions using a paired t-test using the Bonferroni correction, asserting significance when $p\leq0.05$.

\subsection{End-to-End Deep Learning Networks}
\label{sec:dnn_endtoend}

Next, we explore a transformer-based model.  In this method the raw audio signal is used as input to a pre-trained and fine-tuned network and the emotion prediction is directly obtained as an output. These models do not require us to perform manual or domain knowledge-based extraction of features. They instead have a feature encoder component inside the model, which is dynamic in nature, and hence, can change its output for the same signal based on the dataset and nature of the task.

\subsubsection{Features}
For the end-to-end deep learning models, we do not need to extract audio features. Instead we rely on the network itself to both normalize and extract features, that are later passed onto the deeper layers of the network.
The feature set here is the original wav files that are not modified in any capacity. The eventual representations are of size 512, reproducing the setup in the state-of-the-art implementation~\cite{pepino2021emotion}.

\subsubsection{Network}
Our baseline network is the state-of-the-art wav2vec2.0 emotion recognition model~\cite{pepino2021emotion}. The wav2vec model is comprised of three parts: (i) a convolutional neural network (CNN) that acts as feature encoder, (ii) a quantizier module, and (iii) a transformer module.  The input to the model is raw audio data (16kHz) that is passed to a multi-block 1-d CNN to generate audio representations (25ms).  The quantizer is similar to a variational autoencoder that encodes and extracts features using a contrastive loss.  The transformer is used for masked sequence prediction and encodes the bi-directional temporal context of the features. 

We use the base model, which has not been fine-tuned for ASR (wav2vec2.0-PT). We then fine-tune the base model to predict the binned emotion labels.  We use the final representation of the output as an input to dense layers 
to produce the final output.

\subsubsection{Training}
We implement the model provided in the speech brain library~\footnote{https://speechbrain.readthedocs.io/en/latest/API/\\speechbrain.lobes.models.fairseq\_wav2vec.html}. 
As in the other pipeline (Section~\ref{sec:dnn_features}), we use cross-entropy loss for each task and learn the dense layer parameters. Reproducing the state of the art model
~\cite{pepino2021emotion}
We run the model for a maximum of eight epochs.  We revert the network's state to the one that achieved the lowest validation loss. We repeat this experiment five times.  Again, we use UAR and report the results averaged over both subjects and repetitions.

\subsection{Noise Augmentation Overview}
\label{sec:noise_aug}
We will be assessing a model's ability to classify emotion given either environmental or signal distortion noise.  
We perform two kinds of analysis, one when using the set of noises that includes those that do alter human perception, and another when only using noises that are perception-retaining. We report overall model performances for both of these categories. 

For a more thorough analysis, we then
specifically focus on the categories of noise that do not significantly affect human perception.  This allows us to evaluate a model's robustness, or its fragility, with respect to variations that wouldn't alter a human's perception of emotion.  This is important because the overwhelming majority of the noise-augmented utterances in the IEMOCAP dataset were not included in the user study and, therefore, do not have perceptual labels (Section~\ref{sec:userstudy}).  We consider three types of environmental noise \{Human (Hum), Interior (Int), Natural (Nat)\} and three types of signal distortion noise \{Speeding a segment (SpeedSeg), Fade,  Reverberation (Reverb)\}.

We use two separate testing paradigms: (i) matched testing, in which all noise types are introduced to the training, testing, and validation data and (ii) mismatched testing, in which $n$-1 types of noise are introduced to the training and validation sets and the heldout type of noise is introduced to the test set. In all cases, we analyze the test data in terms of specific noise categories.  Therefore, the test sets are the same between the two paradigms.

We run both the matched and mismatched experiments twice, first with the noise-augmented data and second with a noise-robust/denoising pipeline.  The first iteration will allow us to quantify the effect of the noise on the traditional and end-to-end classification pipelines.  We then repeat the experiment with either denoised data for the traditional classifier  (Section~\ref{sec:denoise}) or using the noise-robust implementation of wav2vec2.0 for the end-to-end classifier (Section~\ref{sec:noiserobust}).  This allows us to investigate how, or if, noise-robust implementations can offset the effects of background noise.

\subsubsection{Denoising}
\label{sec:denoise}
We implement denoising using the well-known Recurrent Neural Network Noise Suppression (RNNNoise, denoising feature space) approach, proposed in 2017 for noise suppression~\cite{valin2018hybrid}.  RNNNoise is trained on environmental noise, and these noises overlap considerably with those in our dataset. We use the algorithm's default parameters  
and use it on an `as-is' basis for our experiments. We assume that the system does not have the knowledge of which noise, from the set of available noise categories, is introduced and, therefore, we do not compare with other denoising algorithms that assume a priori knowledge of noise category. 
The result is a set of `noise-suppressed' samples in the training, validation and testing sets. 

We pass all the data, including both the original and noise-augmented data, through a denoising algorithm.  This allows us to ensure that acoustic artifacts, if any, are introduced to both the original and noise-augmented data.  We then train the traditional deep learning model as described in Section~\ref{sec:dnn_features}. 

\subsubsection{Using a Noise-Robust Model}
\label{sec:noiserobust}
In the end-to-end model, we need to use a different denoising approach because the approach described in the previous section does not return a wav file, but instead is applied to the feature-space directly. 
Here, we enforce robustness to noise using a model trained to be noise-robust in an end-to-end fashion. We use the noise-robust version (Wav2Vec2-Large-Robust) of the aforementioned wav2vec2.0 model~\cite{hsu2021robust}. 
The noise-robust large model was pretrained on 16kHz sampled speech audio. Noisy speech datasets from multiple domains were used to pretrain the model: Libri-Light, CommonVoice, Switchboard, and, Fisher~\cite{hsu2021robust}. 
We then train the end-to-end model as described in Section~\ref{sec:dnn_endtoend}. 


\subsection{Model Robustness Testing}
\label{sec:attack}
Deployed emotion recognition models must be robust. One of the major scenarios that we robustness test  any speech-based model for is the presence of noise. But the set of noises we choose to test robustness on can lead  to different conclusions about the robustness of the models. In our case, we consider two different scenarios:
\begin{enumerate}
    \item Robustness evaluation when using perception-retaining samples, noise samples that do not change human perception
    \item Robustness evaluation when using any kind of noise (i.e., both perception-retaining and perception-altering)
\end{enumerate}

We perform robustness evaluation of a model by using the model's output predictions to create new noise-enhanced samples that change the model's output, compared to the original clean sample.  We do this using an untargeted model misclassification test, in which we add noise to the samples. The intentional misclassification algorithm assumes black-box model access. For our purposes, it needs to have access to: (i) a subset of the dataset, (ii) noises to add to create perturbed samples, and (iii) model input and output.

As in any other perturbation-based robustness testing, the goal is to introduce perturbations to the samples such that the resulting samples are as close to the original sample as possible. 
The minimally perturbed sample should be the one that causes a classifier to change its original classification. We measure the amount of perturbation using SNR, calculated using the logarithmic value of the ratio between the original signal and the noise-augmented signal's power. We note that the lower the decrease in SNR, the more minimally perturbed a sample is. The maximal decrease in SNR that we use in the algorithm is a difference of 10 dB. This condition ensures that the sample is not audibly judged as contaminated by humans~\cite{kidd2016determining}.

The algorithm to choose this minimally perturbed sample has four major components:
\begin{enumerate}
	\item Requirements: some labelled samples, noise files, model input and output access, unlabelled samples for testing, and, optionally, correlation between noise type and performance degradation for a given model.
	\item Looping: The algorithm then loops over each noise category to figure out whether it can successfully force the model to misclassify. The noise category order is random if we do not have access to the optional performance degradation correlations.
	\item Minimizing: The algorithm then aims to find the lowest decrease in dB, such that the model still misclassifies. This ensures that the resultant noisy sample is as imperceptible to humans as possible.
	\item End Condition: The algorithm ends if a noise addition has been found, or if it runs out of number of tries allowed for model access.
\end{enumerate}

Please see Algorithm~\ref{algo-attack} for more details. In the algorithm, \textit{numAttempts} is the number of times the algorithm is allowed to access the model's input-output pairs. \textit{Classifier output} refers to  the prediction made by the model when the attack algorithm sends an input to the model to be classified. \textit{Classifier output changes} is true when the model predicted the emotion label differently after noise was added to the sample, compared to the original clean sample. \textit{\textcolor{blue}{Success}} implies that the algorithm was successfully able to force the model to misclassify a particular sample in the allowed number of attempts. \textit{\textcolor{red}{Failure}} implies that the algorithm could not force the model to misclassify in the allowed number of attempts and that the model can be considered robust for that sample.

We use the above algorithm in two different settings, under two different pre-known assumptions, with four levels of allowed queries, and four models (64 categories):
\begin{enumerate}
    \item Settings $\{$All vs. Not-Altering Human Perception$\}$
    \item Pre-Known Assumptions $\{$No Knowledge vs.  Knowledge About Noise Category Degradation Level$\}$
    \item Allowed Queries $\{$5, 15, 20, inf$\}$
    \item Models $\{$Traditional Deep Neural Network (T-DNN), End to End Deep Neural Network (E-DNN), Noise-Robust T-DNN (T-DNN-NR), Noise-Robust E-DNN (E-DNN-NR)$\}$
\end{enumerate}

For all the test samples, we execute five runs of the above algorithm to account for randomization in noise choices. These five runs are then averaged to obtain the average success of misclassification or average robustness for a given sample (1- average success of misclassification). We then average the robustness value over all the test samples. We report our obtained results for the above mentioned scenarios.

\section{Analysis}
\subsection{Research Question 1 (Q1): Does the presence of noise affect emotion perception as evaluated by \emph{human raters}? Is this effect dependent on the utterance length, loudness, the type of the added noise, and the original emotion?}
\label{sec:rq1}
We find that the presence of environmental noise, even when loud, rarely affects annotator perception, suggesting that annotators are able to psycho-acoustically mask the background noise in various cases, as also shown in prior work (e.g.,~\cite{stenback2016speech}).


We find that the addition of signal distortion noise alters human perception.  The reported change in valence and activation values is on a scale of -1 to 1 (normalized). The addition of laughter changes the activation perception of 16\% of the utterances, with an average change of +22\% (+.26). The valence perception is altered in 17\% of the utterances, with an average change of +14\% (+.11).  Similarly for crying, valence is altered in 20\% of the cases, with an average change of -21\% (-.20). Crying changes activation perception in 22\% of the cases, with an average change of -32\% (-.43).
Raises in pitch also alter the perception of emotion.  In 22\% of utterances, the perception of activation is changed.  This contrasts with the perception of valence, which was altered only in 7\% of utterances. In this scenario, activation increases by an average of 26\% (+.19), and valence decreases by 12\% (-.11). On the other hand, decreases in pitch change the perception of activation in 10\% of the cases and of valence in 29\% of the cases. In this scenario, activation decreases by an average of 16\% (-.15), and valence decreases by 7\% (-.07). This ties into previous work~\cite{busso2009shrikanth}, which looked into how changes and fluctuations in pitch levels influenced the perception of emotions.  Changes in the speed of an utterance affect human perception of valence in 13\% (average of -.13) of the cases when speed is increased, and 28\% (average of -.23) when speed is decreased. On the other hand, changes in the speed of an utterance do not affect activation as often, specifically, 3\% in case of increase and 6\% in case of decrease.

We ensured that our crowsdourcing samples had an even distribution over gender of the speaker and the length of the sample (see Section~\ref{sec:datasample}). We performed paired t-test to evaluate whether these variables influenced the outcome of emotion perception change in presence of noise. We found that the changes in perception were not tied to characteristics of the speakers. For example, there was no correlation between changes in perception and variables such as, the original emotion of the utterance, the gender of the speaker, and the length of the utterance.

The human perception study provides insight into how emotion perception changes given noise.  This also provides information about the potential effects of noise addition on model behavior.  In the sections that follow, we will evaluate how machine perception changes given these sources of noise. 

\subsection{RQ2: Can we verify previous findings that the presence of noise affects the performance of \emph{emotion recognition models}? Does this effect vary based on the type of the added noise?}  
We first assess the performance of the model on the original IEMOCAP data and  find that the traditional model obtains a performance of 0.67 UAR on the activation and 0.59 UAR on the valence task. On the other hand, the end-to-end model obtains a performance of 0.73 on activation and 0.64 on the valence task. We hypothesize that the wav2vec2 model has an added advantage of being trained to recognize word structures that can incorporate some paralinguistic/langauge information in the fine-tuned model.

Next, we augment the test samples of each fold with each of the noise types (Section~\ref{sec:noise}) and investigate how the performance of the model changes.
We include two cases: (i) only perception-retaining noises and,(ii) all noises.

In the first scenario, we do not include noise types that were found to  affect human perception (e.g., \textit{Pitch, SpeedUtt, Laugh}) because once these noises are added, the ground truth is no longer reliable.  This lack of reliable ground-truth data hinders the evaluation of the model's performance on these samples because the majority of the utterances were not part of the original crowdsourcing experiment and are thus unlabeled.  The remainder of this section focuses on the second scenario only.

We find that for matched train and test noise conditions, the traditional machine learning model's performance decreases by an average of 28\% for environmental noise while it drops by 32\% for signal manipulation. On the other hand, for end-to-end deep learning model, the model's performance decreases by an average of 22\% and 26\% for environmental and manipulated noises, respectively.  In mismatched noise conditions, the models' performance decreases by an average of 33\% for environmental noise, fading, and reverberation. There is also a smaller drop in performance for speeding up parts of the utterance and dropping words, showing the brittleness of these models. Table~\ref{tab:sota} reports the percentage change in performance when testing on noisy test data, compared to clean test data.

We see that the end-to-end deep learning model is less affected by environmental noise, but has a larger drop due to fading and reverberation. We observe a larger drop on performance when dropping words, which possibly can be attributed to the change in audio-structure and non-controllable feature extraction for this model.

In the second scenario, we observe a large drop in performance for both the traditional and the end to end machine learning model. For example, in the case of a traditional deep learning model, the valence prediction performance drops to a near-chance performance when including all kinds of noises (see Table~\ref{tab:sota}).

We specifically want to point out how the inclusion of all noises in the test conditions changes the observed model performance. Primarily, the models on an average seem to do 20\% worse than they would if we only consider noises that do not alter human perception. We note the discrepancy between the results of the two noise addition scenarios and that results should be described with respect to the perceptual effects of noise, if noise augmentation is used.
\subsection{RQ3a: Does dataset augmentation help improve the robustness of emotion recognition models to unseen noise?}
\label{sec:modelaug}
We first report results for only perception-retaining noises.
When the training datasets are augmented with noise, we observe an average performance drop of 26\% and 10\% for matched noise conditions when using the traditional and the end-to-end deep learning model, respectively. For the mismatched noise conditions, we observe an average performance drop of 31\% and 16\% for the traditional and end-to-end deep learning models, respectively.

Both models see improved performance when the training dataset is augmented with continuous background noise in the matched noise setting.
We find that data augmentation improves performance on mismatched noisy test data over a baseline system trained only on the clean IEMOCAP data.  For example, the end-to-end model tested on environmental noise-augmented dataset (as compared to traditional deep learning model), reduces the performance drop to nearly zero.  This improvement is particularly pronounced (an increase of 22\% as compared to when trained on the clean partition) when the environmental noise is introduced at the start of the utterance (e.g., when the test set is introduced with nature-based noises at the start of the utterance and the train set is introduced with human and interior noises at the start of the utterance). We speculate that the network learns to assign different weights to the start and end of the utterance to account for the initial noise. 

However, we find that in both matched and mismatched conditions, it is hard to handle utterances contaminated with reverberation, a common use case, even when the training set is augmented with other types of noise. We find that this improvement in performance is even more reduced when using the wav2vec model, alluding to the model's fragility towards data integrity/structural changes. This can be because reverberation adds a continuous human speech signal in the background delayed by a small period of time. None of the other kind of noise types have speech in them, and hence augmentation doesn't aid the model to learn robustness to this kind of noise.


Finally, we investigate the differences in model performance when we use all types of noise vs. those that are perception-retaining.  Specifically, we focus on the perception-altering noises because samples augmented with noises in this category no longer have a known ground truth.  We inquire as to whether the use of samples that alter perception may lead to the appearance of model performance improvement (note: appearance because the samples now have uncertain ground truth).  
To maintain equivalence, we ensure that the training and validation dataset sizes are equal even when they are augmented with more noise conditions.
We observe that many cases of performance improvement occur when the noises include those that are perception-altering (see ``Al noises'' in Table~\ref{tab:sota}).
We observe a difference of 12\% to 25\% between the numbers that we obtain for the perception-retaining noises vs. when not distinguishing between the two noise categories. This supports our claim that the choice between types of noises used for data augmentation during model training and performance evaluation affects the empirical observations and should be carefully considered.
We hypothesize that this improvement in performance may be due to the inherent nature of noises that change emotion perception, if they are perceptible enough to change emotion perception, then they may stand out enough that the model can adequately learn to separate them out and improve its prediction towards the original ground truth annotation.  However, if the noise alteration truly does change perception, then the model is learning to ignore this natural human perceptual phenomenon.  This may have negative consequences during model deployment.

\subsection{RQ3b: Does sample denoising help improve the robustness of emotion recognition models to unseen noise?}
In the \emph{matched training testing condition}, we find that the traditional deep learning model has an average performance of 0.57 across all the datasets and testing setups, while the end-to-end models do substantially better at 0.61 UAR. See Table~\ref{tab:noise-robust} for details.

In the \emph{mismatched training testing condition}, we find that for the traditional deep learning model, adding a denoising component to the models leads to a significant improvement when the original SNR is high (e.g., after continuous noise introduction the SNR decreases only by 10dB).  In this case, we see an average improvement of $23\%\pm3\%$ across all environmental noise categories, compared to when there is no denoising or augmentation performed. 

However, when the SNR decreases by 20dB, we observe a decline in performance when using the noise suppression algorithms.  We believe that this decline in performance is reflective of the mismatch in goals: the goal of noise suppression is to maintain, or improve, the comprehensibility of the speech itself, not necessarily highlight emotion.  As a result, it may end up masking emotional information, as discussed in~\cite{ma2015human}.  

We further show that the addition of a denoising component does not significantly improve performance in the presence of signal distortion noise (other than reverberation) as compared to the presence of environmental noise (noise addition).  For example, when samples were faded in or out or segments were sped up, the performance is significantly lower ($-28\%$) than when tested on a clean test set. However, we did see an improvement in the performance  for \emph{unseen} reverberation contaminated samples as compared to data augmentation (an average of $+36\%$). Finally, we observe a general trend of increase in emotion recognition performance for the combined dataset (noisy and non-noisy samples), as compared to when the model is trained on the clean training set, which supports the findings from previous dataset augmentation research~\cite{aldeneh2017using}.

For the end-to-end deep learning model, we use the noise-robust version. We find that the model is effective at countering environmental noise when trained on a dataset augmented with environmental noise, even in the mismatched condition.  The performance is equivalent to the model evaluated on the clean data.
We further delve into the amount of noise augmentation needed to achieve this equivalency.  We consider all of the original training data.  We augment a percentage of the training data, starting by augmenting a random sample of 10\% with perception-retaining noise and increasing by 10\% each time.  We find that we obtain equivalency after augmenting with only 30\% of the training data.
We compare this compares to the traditional model, in which all of the training data are noise-augmented and we still do not see equivalency. 

We separately consider the signal distortion noise samples.  These were not part of the training of the wav2vec2-Large-robust model.  However, this model only sees a 6\% loss in performance, where the traditional robust model saw a 
20\%
loss in performance.

However, as discussed in the original traditional model, the end-to-end noise-robust model also fails on reverberation-based contamination even when trained on a similarly augmented dataset (note that the denoised traditional model could effectively handle reverberation). We believe that this may be because the wav2vec model is trained on continuous speech and relies on the underlying linguistic structure of speech.  However, in reverberation, there is an implicit doubling of the underlying information, which is at odds with how this model was trained.  This may explain why it is not able to compensate for this type of signal manipulation.

Next, we analyze whether the perception category of noise used for data augmentation of the samples, in both, the train and test dataset influences the reported results for noise-robust model improvement. We find that there is a significant difference in performance when the testing dataset is augmented with any kind of noise vs. when augmented with perception-retaining noise. 
Specifically, we observe that the maximal gains in performance when testing on matched noisy conditions are for samples for which we do not know whether or not the ground truth holds (i.e., both noise categories). For example, when using the noise robust traditional deep learning model, where the test and train dataset is augmented with any type of noises, we observe a performance improvement, as compared to that using a clean train dataset, of 12\%. Similarly for noise robust end to end models, the performance improvement difference when using all noises vs. only perception retaining ones is 15\% for activation and 13\% for valence.  Again, this is a problem when we think about deploying models in the real world because although the perception of these emotion expressions may change, we are assuming that the system should think of these perception labels as rigid and unchanging.
\subsection{RQ4: How does the robustness of a model to attacks compare when we are using test samples that with are augmented with perception-retaining noise vs. samples that are augmented with all types of noise, regardless of their effect on perception?}

In this section, we aim to show the effects of noise augmentation in general and specifically highlight noise categories that do not alter human perception.  We will show that if we are not careful with the selection of our noise types, moving from noise that we know not to alter perception to noise that may, the resulting noise sources can not only impact the brittleness of models, but also lead to inaccurate evaluation metrics. We also specifically report robustness performance when using reverberation-based contamination, as we observed that it is the most likely noise category to degrade the robustness of the model.

We allow the decision boundary attack algorithm a maximum of five queries to create a noise augmented sample that will change the model output.  We find that if the attacker is also given perception-altering noises, compared to perception-retaining, it can more effectively corrupt the sample.  It achieves an increase in success rate from 35\% (only perception-retaining) to 48.5\% (all noise categories).
This increase in the success rate when perception-altering noises are included implies that the model does not remain robust when the effects of noise on human perception are not considered.

We next consider the type of noise (environmental vs. signal manipulation).  We find that the success rate of flipping a model's output is 18\% for noises belonging to the environmental category, which is generally a category of perception-retaining noise. The success rate of flipping a model's output is 37\% for all noises belonging to the signal manipulation category.
When we constrain our possible noise choices to perception-retaining signal manipulations, we see that the success rate of the intentional misclassification algorithm drops to 24\%.
On the other hand, we observe that when we also consider the signal manipulations that are perception-altering, the success rate of flipping a model output is 39\%.  See Table~\ref{tab:robustnessperf} for more details.

We previously discussed the fragility of end-to-end models towards reverberation-based noise contamination, noise that is perception-retaining for human evaluators.
Here, we specifically run an experiment to use only that particular noise category for the model fragility testing. 
If the attacker knows that the model is susceptible to reverberation-based prediction changes, the intentional misclassification algorithm can land on an optimal set of room and reverberation parameters in a maximum of five queries to be able to produce a flipped output for that particular sample. It achieves a success rate of 24\%, compared to 12\% for other perception-retaining noises.
The traditional noise-robust deep learning model is even more challenged, compared to the end-to-end model.  The number of queries required to flip the output is three, vs. five for the end-to-end model, suggesting that it is less robust.
This empirical evaluation is performed primarily to demonstrate how such noise inclusions can not only invalidate the ground truth but also lead to inaccurate and fragile benchmarking and evaluation of adversarial efficiency and robustness.
\subsection{RQ5: What are the recommended practices for speech emotion dataset augmentation and model deployment?}
We propose a set of recommendations, for both augmentation and deployment of emotion recognition models in the wild, that are grounded in human perception. For augmentation, we suggest that:

\begin{enumerate}
    \item  Environmental noise should be added to datasets to improve generalizability to varied noise conditions, whether using denoising, augmentation, or a combination of both.
    \item  It is good to augment datasets by fading the loudness of the segments, dropping letters or words, and speeding up small (no more than 25\% of the total sample length) segments of the complete sound samples in the dataset. But it is important to note that these augmented samples should not be passed through the denoising component as the denoised version loses emotion information.
    \item One should not change the speed of the entire utterance more than 5\% and should not add intentional pauses or any background noises that elicit emotion behavior, e.g., sobs or laughter.
\end{enumerate}
Regarding deployment, we suggest that:
\begin{enumerate}
    \item Noisy starts and ends of utterances can be handled by augmentation, hence, if the training set included these augmentations, there should be no issue for deployed emotion recognition systems.
    \item Reverberation is hard to handle for even augmented emotion recognition models. Hence, the samples must either be cleaned to remove the reverberation effect, or must be identified as low confidence for classification.
    \item Deploy complementary models that identify the presence of noise that would change a human's perception.
\end{enumerate}

\section{Conclusion}
In this work, we study how the presence of real world noise, environmental or signal distortion, affects human emotion perception. We identify noise sources that do not affect human perception, such that they can be confidently used for data augmentation. We look at the change in performance of the models that are trained on the original IEMOCAP dataset, but tested on noisy samples and if augmentation of the training set leads to an improvement in performance. We conclude that, unlike humans, machine learning models are extremely brittle to the introduction of many kinds of noise. While the performance of the machine learning model on noisy samples is aided from augmentation, the performance is still significantly lower when the noise in the train and test environments does not match. In this chapter, we demonstrate fragility of the emotion recognition systems and valid methods to augment the datasets, which is a critical concern
in real world deployment.

%% file: Chapters/chap6.tex






\section{Motivation and Contributions}
Emotion recognition algorithms rely on data annotated with high quality labels. However, emotion expression and perception are inherently subjective.  There is generally not a single annotation that can be unambiguously declared ``correct''.  As a result, annotations are colored by the manner in which they were collected.  In this chapter, we conduct crowdsourcing experiments to investigate this impact on both the annotations themselves and on the performance of these algorithms.  We focus on one critical question: the effect of context.  We present a new emotion dataset, Multimodal Stressed Emotion (MuSE), and annotate the dataset using two conditions: randomized, in which annotators are presented with clips in random order, and contextualized, in which annotators are presented with clips in order.
We find that contextual labeling schemes result in annotations that are more similar to a speaker's own self-reported labels and that labels generated from randomized schemes are most easily predictable by automated systems.



\section{Introduction}
Emotion technologies, both recognition and synthesis, are heavily dependent on having reliably annotated emotional data, annotations that describe the observed emotional display.  The hope is often that these annotations capture the speaker's true underlying state.  Yet, in practice, this true \emph{felt sense} emotion is unknown, and researchers must resort to manual labeling of data.  The hope is that these manual labels are sufficiently ``correct'' to enable the training and evaluation of emotion technologies.  One method of ensuring quality labels has been to require the participation of expert raters. However, it can be both expensive and time consuming to hire expert raters. More recently, researchers have embraced crowdsourcing services (e.g., \textit{Amazon Mechanical Turk}) to efficiently collect annotations from non-expert workers in a cost-effective and timely manner \cite{soleymani2010crowdsourcing}. Once collected, annotations from non-expert workers are aggregated to form ground-truth labels that are used for training and evaluating automated systems. However, the method through which these annotations are collected can profoundly impact the behavior of the annotators.  In this chapter, we study how the setup of a crowdsourcing task can influence both the collected emotion labels as well as the performance of classifiers trained using these labels.

The effective use of crowdsourcing for collecting reliable emotion labels has been an active research topic. Burmania et al. investigated the trade-off between the number of annotators and underlying reliability of the annotations~\cite{burmania2016tradeoff}. 
Other work has looked at quality-control techniques to improve the reliability of annotations.  For example, Soleymani et al. used qualification tests to filter out spammers and retain high-quality annotators~\cite{soleymani2010crowdsourcing}. Burmania et al. investigated the use of gold-standard samples to monitor annotators' reliability and fatigue~\cite{burmania2016increasing}.

However, variability also results from context, relevant past information that provides cues as to how to interpret an emotional display. Context, such as tone, words, expressions can affect how individuals perceive emotion \cite{laplante2003things}.
Context is also implicitly included in the labeling schemes of many of the most common emotion datasets (e.g., IEMOCAP~\cite{busso2008iemocap} and MSP-Improv~\cite{busso2017msp}) because annotators rate each utterance (or time period) in order.  That means that annotators are influenced by information that they recently observed~\cite{yannakakis2017ordinal}.  However, emotion recognition systems are often trained over single utterances~\cite{aldeneh2017using,abdelwahab2017incremental,mirsamadi2017automatic, sarma2018emotion}, leading to a mismatch in the information available to annotators and to classification systems. 

In this work, we study the difference between annotations obtained for audio clips when emotional displays are presented to annotators with context and when presented randomly. In both cases, annotators are affected by the emotion displays that they have recently observed \cite{qiao2017transient, russell2017emotion}.  However, only in the contextual presentation there is also a cohesive story. 
We investigate the following research questions:

\begin{itemize}
\itemsep0em
\item Q1: Is there a significant difference between annotations obtained from random and contextual presentations?

\item Q2: Are annotations obtained from contextual presentations more similar to a speaker's own self-reported labels than those from random presentations?

\item Q3: Is there a significant difference between the inter-rater agreements obtained from random and contextual presentations?

\item Q4: How does the performance of an emotion recognition system, operating on single utterances, vary given annotations obtained from random and contextual presentations?

\item Q5: How does the performance gain of an emotion recognition system operating across multiple utterances vary given different amounts of context (defined as number of prior utterances) and labels obtained from random and contextual presentations?

\end{itemize}

The findings from this work will provide insight into performance implications of emotion recognition system given mismatches between the amount of context provided to the annotators generating the labels and the ultimate classification system.\looseness=-1

The main aim of the work was to annotate the dataset described in the previous chapter. For all further experiments, we use the utterances created from the MuSE dataset.

\subsection{Crowdsourcing}
We posted our experiments as Human Intelligence Tasks (HITs) on \textit{Amazon Mechanical Turk}. HITs were defined as sets of utterances in either the contextual or random presentation condition.  In each condition, workers were presented with a single utterances and were asked to annotate the activation and valence values of that utterance using Self Assessment Manikins \cite{bradley1994measuring}. Once completed, the worker was presented with a new HIT and could not go back to revise a previous estimate of emotion. This annotation strategy is different than the one deployed in \cite{chen2018emotionlines},where the workers could go back and re-evaluate utterances. 


In the randomized experiment, each HIT is an utterance from any section, by any speaker, from any session and all HITs appear in random order. So, a worker might see the first HIT as \textit{Utterance 10 from Section 3 of Subject 4's stressed recording} and see the second HIT as \textit{Utterance 1 from Section 5 of Subject 10's non-stressed recording}. This setup ensured that the workers couldn't condition to any speaker's specific style or contextual information.

In the contextual experiment, we posted each HIT as a collection of ordered utterances from a section of a particular subject's recording. Because each section's question was designed to elicit a particular emotion, we still posted the HITs in a random order over sections from all subjects.  This prevented workers from conditioning to the speaking style of an individual participant. For example, a worker might see the first HIT as \textit{Utterance 1...N from Section 3 of Subject 4's stressed recording} and see the second HIT as \textit{Utterance 1...M from Section 5 of Subject 10's non-stressed recording} where {\it N, M} are the number of utterances in those sections respectively.

We recruited from a population of workers in the United States who are native English speakers, to reduce the impact of cultural variability.  We ensured that each worker had $>98$\% approval rating and number of HITs approved as $>500$. We ensured that all workers understood the meaning of activation and valence using a qualification task that asked workers to rank emotion content. The workers were asked to select, given two clips, which clip had the higher valence and which had the higher activation.  The options were chosen from a set including: (1) a speaker in low activation, high valence state and (2) a speaker in high activation, low valence state.  

We assigned each HIT to eight workers.  All HIT workers were paid a minimum wage ($\$9.25/$hr), pro-rated to the minute.  We removed and re-posted assignments where the worker completed the assignment in time shorter than the audio length.  The ground-truth for each utterance was formed by taking the average of the eight annotations. Table~\ref{data_stat} shows the data summary for the collected annotations and corresponding data points.

    


\section{Experimental Setup}
\label{exp_setup}

\textbf{Acoustic Features.} We extract acoustic features using OpenSmile~\cite{eyben2010opensmile} with the eGeMAPS configuration~\cite{eyben2016geneva}. The eGeMAPS feature set consists of $88$ utterance-level statistics over the low-level descriptors of frequency, energy, spectral, and cepstral parameters. We perform speaker-level $z$-normalization on all features.\looseness=-1

\textbf{Static Network Setup (Hypothesis 4).} We train and evaluate four Deep Neural Networks (DNN) models: \{random, contextual\}$\times$ \{valence, activation\}. In all cases, we predict the continuous annotation using regression.  For each network setup, we follow a five-fold evaluation scheme and report the average RMSE across the folds. For each test-fold, we use the previous fold for hyper-parameter selection and early stopping. The hyper-parameters include: number of layers $\{2, 3, 4\}$ and layer width $\{64, 128, 256\}$. We use ReLU activation and train the networks with MSE loss using Adam optimizer. 

\textbf{Dynamic Network Setup (Hypothesis 5).} We use Gated Recurrent Unit networks (GRU). The hyper-parameters are: number of layers $\{1, 2\}$ and layer width $\{64, 128, 256\}$. We pass the GRU output of the last time step through a regression layer to get the final outputs. We train the networks with MSE loss using Adam optimizer.

\textbf{Network Training.} We train our networks for a maximum of 100 epochs and monitor the validation loss after each epoch. We stop the training if the validation loss does not improve for 15 consecutive epochs. We revert the network's weights to those that achieved the lowest validation loss during training. Finally we train each network five times and average the predictions to reduce variance due to random initialization.

\section{Results and Analysis}
\subsection{RQ1: Differences in Obtained Annotations}
\uline{Hypothesis}: \textit{Human annotations collected through randomized labeling are significantly different from those collected through contextualized labeling.} Prior work has shown context effects emotion perception \cite{yannakakis2017ordinal}, even when observers are explicitly asked not to take it under consideration \cite{ngo2015use,cauldwell2000did}. Hence, we believe that context provided by previous utterances would lead to a change in perception of a particular utterance.
Tables~\ref{exp1-stress} and~\ref{exp1-section} (sets of significantly different means are bolded ($t$-test, $p<0.01$)) show the mean activation and valence, for the random and contextualized labeling schemes, grouped by condition and question, respectively. 
Table~\ref{exp1-stress} shows that, for non-stress conditions, the mean of the activation ratings obtained through contextual labeling is significantly higher than that obtained through random labeling. The table also shows that, for both stress and non-stress conditions, the valence means obtained through contextual labeling are significantly higher than those obtained through random labeling.
Table~\ref{exp1-section} shows that, although the mean valence and activation values were consistently different for the labelling schemes across all emotion elicitation techniques, the differences were significant in some elicitation techniques and not in others.


\begin{table}[t]
  \caption{Mean activation and valence values obtained from the two crowdsourcing labeling schemes (random and context) grouped by speaker condition (stress and non-stress).  
  }
  \label{exp1-stress}
  \centering
  \begin{tabular}{lcccc}
    \toprule
    & \multicolumn{2}{c}{Activation} & \multicolumn{2}{c}{Valence}  \\
    \cmidrule(r){2-3} \cmidrule(r){4-5}
         & Random     &  Context  & Random     &  Context\\
    \midrule
    Stress         & 3.63        & 3.59       & {\bf 5.27}  & \textbf{5.36}   \\
    Non-Stress     & {\bf 3.61}  & {\bf 3.79} & {\bf 5.26}  & {\bf 5.39}      \\
    \bottomrule
  \end{tabular}
\end{table}
\begin{table}[t]
  \caption{Mean activation and valence values obtained from the two crowdsourcing labeling schemes (random and context) grouped by emotion elicitation question.
  }
  \label{exp1-section}
  \centering
  \begin{tabular}{lcccc}
    \toprule
    & \multicolumn{2}{c}{Activation} & \multicolumn{2}{c}{Valence}  \\
    \cmidrule(r){2-3} \cmidrule(r){4-5}
         & Random     &  Context  & Random     &  Context\\
    \midrule
    Icebreaker & 3.55  & 3.60 & {\bf 5.41}  & \textbf{5.61}    \\
    Positive   & 3.64  & 3.71 & 5.11  & 5.13      \\
    Negative   & {\bf 3.57}  & {\bf 3.67} & {\bf 5.40}  & {\bf 5.55}      \\
    Intensity  & {\bf 3.64}  & {\bf 3.74} & {\bf 5.17}  & {\bf 5.31}      \\
    Ending     & 3.69  & 3.71 & {\bf 5.23}  & {\bf 5.29}      \\
    \bottomrule
  \end{tabular}
\end{table}
\subsection{RQ2: Self-Annotations and Crowdsourced Annotations}
\begin{figure}[t]
  \centering
  \caption{Mean difference between the self-reported activation and valence ratings and the random and contextual presentations.}
  \includegraphics[scale=0.75]{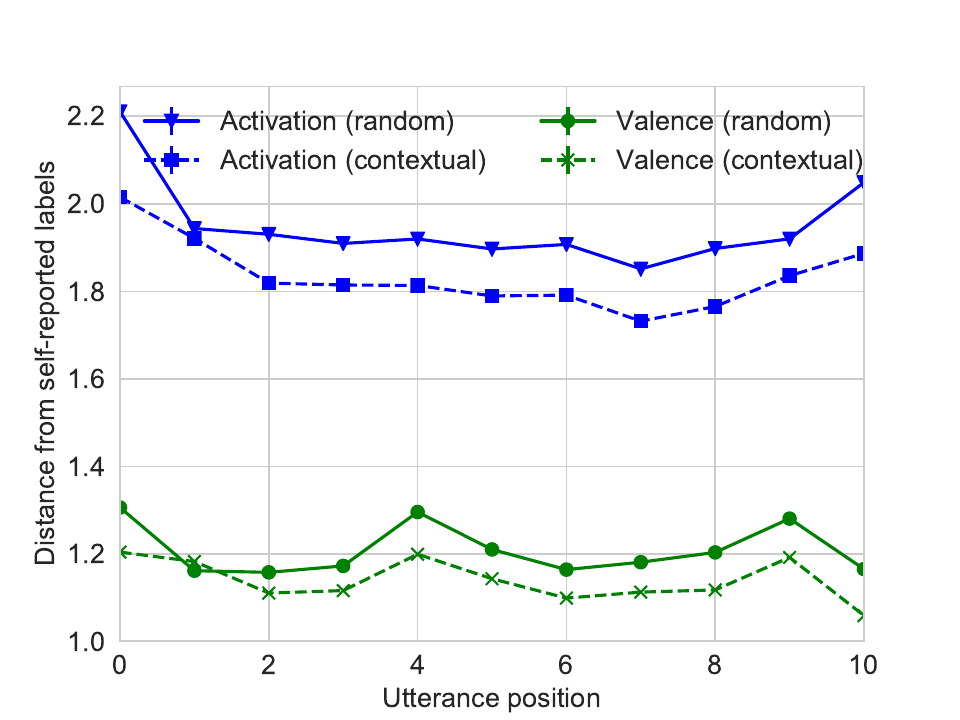}
  \label{fig:exp2}
\end{figure}

\uline{Hypothesis}: \textit{Annotations of outside observers are more similar to self-annotations in the contextual case, compared to the randomized case.} Path models \cite{banziger2015path} suggest that subjective voice variation, from the established mental baseline accounts for much of the variance in emotion inference. Hence, emotion inference is aided with more cues about the speech patterns that are more readily provided through context.
Figure~\ref{fig:exp2} shows the absolute differences between the mean crowdsourced labels (valence and activation, each for random and contextual schemes) and self-reported scores as a function of utterance position. The figure shows that contextual labels have consistently lower absolute differences, compared to self-reported labels, than the random labels. A paired $t$-test shows that these differences between the contextual and random labels are significant ($p<0.01$) for both valence and activation.

Our results suggest that crowdsourced emotion labels collected with access to contextual information are closer to self-reported emotion labels. Our results further suggest that these differences are consistent across recording conditions (Table~\ref{exp2-stress}) and emotion elicitation questions ( 
Table~\ref{exp2-section}, sets of significantly different means are bolded, $t$-test, $p<0.01$). 

\begin{table}[t]
    \caption{Mean difference between the self-reported activation and valence ratings from the two labeling schemes (random and context) grouped by speaker condition (stress and non-stress).
    }  
  \label{exp2-stress}
  
  \centering
  \begin{tabular}{lcccc}
    \toprule
    & \multicolumn{2}{c}{Activation} & \multicolumn{2}{c}{Valence}  \\
    \cmidrule(r){2-3} \cmidrule(r){4-5}
         & Random     &  Context  & Random     &  Context\\
    \midrule
    Stress         & {\bf 2.03}        & {\bf 1.96}       & {\bf 1.20}  & \textbf{1.14}   \\
    Non-Stress     & {\bf 1.82}  & {\bf 1.67} & {\bf 1.20}  & {\bf 1.12}      \\
    \bottomrule
  \end{tabular}
\end{table}

\begin{table}[t]
  \caption{Mean difference between the self-reported activation and valence ratings from the two labeling schemes (random and context) grouped by emotion elicitation question.  
  }
  \label{exp2-section}
  \centering
  \begin{tabular}{lcccc}
    \toprule
    & \multicolumn{2}{c}{Activation} & \multicolumn{2}{c}{Valence}  \\
    \cmidrule(r){2-3} \cmidrule(r){4-5}
         & Random     &  Context  & Random     &  Context\\
    \midrule
    Icebreaker & 1.81  & 1.80 & {\bf 0.97}  & \textbf{0.85}    \\
    Positive   & {\bf 1.89}  & {\bf 1.74} & 1.14  & 1.11      \\
    Negative   & {\bf 1.96}  & {\bf 1.76} & {\bf 1.18}  & {\bf 1.07}      \\
    Intensity  & {\bf 2.19}  & {\bf 2.08} & {\bf 1.49}  & {\bf 1.44}      \\
    Ending     & {\bf 1.81}  & {\bf 1.73} & 1.23  & 1.28      \\
    \bottomrule
  \end{tabular}
\end{table}



\subsection{RQ3: Inter-Annotator Agreement}
\uline{Hypothesis}: \textit{Individual annotators differ in annotation similarity in the contextual presentations, compared to the randomized presentation.} Joseph et al. in \cite{joseph2017constance} show that while insufficient context results in noisy and uncertain annotations, an overabundance of context may cause the context to outweigh other signals and lead to lower agreement. Further, contextual information biases different people differently on both temporal and intensity metrics \cite{van2009immediacy,walker2009fading}.  Our results highlight the impact of context: the agreement is significantly higher in the case of labels obtained from the randomized presentations, compared to the contextualized presentations: ($1.55$ vs. $1.62$) for activation and ($1.07$ vs. $1.14$) for valence. This trend holds true for all experimental design setups i.e. \{random, contextual\}$\times$ \{valence, activation\} and \{random, contextual\}$\times$ \{icebreaker, positive, negative, intensity and ending\}. As shown in Tables~\ref{exp1-stress} and \ref{exp1-section}, the labels obtained in both cases are significantly different due to context-based conditioning. However, the conditioning may not impact the labels consistently across all workers, which may lead to lower inter-annotator agreement values. This suggests that it may be beneficial to consider the distribution of annotations as ground-truth, rather than averaging labels, which presumes that the impact of conditioning is consistent across all workers \cite{zhang2017predicting}.

\subsection{RQ4: Non-Contextual Annotations and Static Classifiers}
\uline{Hypothesis}: \textit{A static classifier will perform better when trained and evaluated using labels annotated with a randomized presentation, compared to a  contextualized presentation.} Prior studies have shown that it is easier to classify data with less target variation \cite{liu2016classification} and matched classifier input, which in our case is labels obtained from the random labelling presentation (the classifier processes single utterances at a time, no context).

We test this hypothesis by training and evaluating classifiers for the four possible setups: $\{random,$ $ contextual\}$ x $\{valence, $ $activation\}$. The classifier is described in Section~\ref{exp_setup}.  We find that the RMSEs are lower for the contextual labels in the case of activation ($0.91$ vs. $1.00$) while the errors are lower for the random labels in the case of valence ($1.13$ vs. $1.20$). Using a paired $t$-test, we find that the differences in errors are significant in the case of valence but not activation. 
These findings suggest that classification performance is impacted by the labelling methodology, but that this effect may depend on emotion dimension.  

Prior work has demonstrated the importance of considering long-term context when predicting valence (the same effect has not been shown in activation)~\cite{khorram2017capturing}.  The contextual annotations provided the annotators with this information, but the classifier could not take advantage of this effect.  This mismatch may have contributed to the relatively lowered performance of the valence classifier, compared to the activation classifier.


\subsection{RQ5: Contextualized Annotations and Long-Context Classifiers}
\label{q5}
\uline{Hypothesis}: \textit{We anticipate that systems trained on contextualized labels will see greater increases in performance as the amount of provided context increases.}  This finding would support results in the literature regarding the ordinal nature of emotion perception \cite{yannakakis2017ordinal} and previous works in emotion recognition that have demonstrated that context can influence the performance of emotion classifiers~\cite{khorram2017capturing}. 

The classifier is described in Section~\ref{exp_setup}.  We test this hypothesis by using the contextual annotations in one classifier and the non-contextual (random) annotations for the other classifier.
We select a subset of utterances in each section that have at least five consecutive utterances before them ($59\%$ of the original data). The initial classifier is trained without temporal context (but with the contextualized labels).  We incrementally increase the number of past utterances (from zero to five). We run this for every task combination and report the results in Table~\ref{exp5-errors}.

Table~\ref{exp5-errors} shows the performance gains after incrementally adding the past utterance, relative to the baseline performance. The addition of past utterances improves the performance over baseline for all setups. Where using contextual labels, however, the performance gains are generally higher than the gains obtained after using random labels. Our results suggest that it is necessary to consider the mismatch the amount of context provided to the annotators generating the labels and the ultimate classification system.

\begin{table}[t]
  \caption{Relative improvement in RMSE (\%) obtained for each additional previous utterance, comparing random and contextual labels.}
  \label{exp5-errors}
  \centering
  \begin{tabular}{ccccc}
    \toprule
     & \multicolumn{2}{c}{Activation} & \multicolumn{2}{c}{Valence}  \\
    \cmidrule(r){2-3} \cmidrule(r){4-5}
         Past steps & Random     &  Context  & Random     &  Context\\
    \midrule
    0 & -    & -    & -    & -\\
    1 & $+1.96\%$ & $+1.24\%$ & $+0.85\%$ & $+3.32\%$\\
    2 & $+2.28\%$ & $+2.93\%$ & $+5.23\%$ & $+7.63\%$\\
    3 & $+3.36\%$ & $+8.72\%$ & $+6.08\%$ & $+8.43\%$\\
    4 & $+4.41\%$ & $+10.5\%$ & $+8.23\%$ & $+8.36\%$\\
    \bottomrule
  \end{tabular}
\end{table}
\section{Conclusion}
In this work we showed that the amount of context provided to annotators when assigning emotion labels affects both the annotations themselves and the performance of classifiers using these annotations. We also studied the implications of a mismatch between annotation context and classifier context on classifier performance. 




%% file: Chapters/chap7.tex
\section{Motivation and Contributions}
Various psychological factors affect how individuals express emotions. Yet, when we collect data intended for use in building emotion recognition systems, we often try to do so by creating paradigms that are designed just with a focus on eliciting emotional behavior. 
Algorithms trained with these types of data are unlikely to function outside of controlled environments because our emotions naturally change as a function of these other factors. 
In this work, we study how the multimodal expressions of emotion change when an individual is under varying levels of stress. We hypothesize that stress produces modulations that can hide the true underlying emotions of individuals and that we can make emotion recognition algorithms more generalizable by controlling for variations in stress. To this end, we use adversarial networks to decorrelate stress modulations from emotion representations.  We study how stress alters acoustic and lexical emotional predictions,  paying special attention to how modulations due to stress affect the transferability of learned emotion recognition models across domains. 
Our results show that stress is indeed encoded in trained emotion classifiers and that this encoding varies across levels of emotions and across the lexical and acoustic modalities. 
Our results also show that emotion recognition models that control for stress during training have better generalizability when applied to new domains, compared to models that do not control for stress during training. We conclude that is is necessary to consider the effect of extraneous psychological factors when building and testing emotion recognition models.

\section{Introduction}
Many extraneous psychological factors influence how individuals express and perceive emotions~\cite{paulmann2016psychological}.
However, most emotion recognition algorithms, rely on data collected in controlled laboratory environments (e.g.,~\cite{busso2008iemocap, busso2017msp}) where influences from such factors are either not present, or kept constant.
The performance of emotion recognition algorithms is likely to vary when applied to data where these external psychological factors are present.
In this work, we study how an extraneous psychological factor, stress, affects multimodal (acoustic+lexical) emotion classifiers.
Stress 
can affect how individuals produce and perceive emotion~\cite{paulmann2016psychological}. 
Yet, the effect of stress levels on the performance of state-of-the-art emotion recognition systems has not been explored.

Extraneous psychological factors can act as confounding factors, 
variables that influence both the output (e.g., emotion) and the input (e.g., acoustic and lexical features).
Not controlling for confounding variables when training emotion classifiers
can cause the classifiers to learn unintentional associations
between the variables, associations that might not replicate in real world scenarios.
For instance, consider a dataset where all the ``sad'' samples were unintentionally recorded from individuals who were experiencing stress at the time of recording.
Not taking special care when building the models could cause a trained classifier to erroneously associate experiencing stress with being sad. 
In this work, we study how stress alters the performance of trained emotional classifiers in the context of neural networks. We then see how performance is affected when tested on samples out of domain, when we explicitly impede the network from learning such associations.

Previous research showed that controlling for confounding variables when training emotion recognition classifiers results in more robust models when compared to models trained without controlling for the same confounding variables. For instance, Abdelwahab et al.~\cite{abdelwahab2018domain} and Gideon et al.~\cite{gideon2019barking} showed that controlling for domain (i.e., data source), as a confounding factor, when training emotion recognition models results in improved cross-corpus generalization performance when compared to performance of models that were trained without controlling for domain as a confounding factor. 
Most of the above mentioned methods rely on samples obtained from the target domain to extract representations that retain information only about emotion and not domains. Our goal is to go beyond studying the effects of variations due to domain and background noise on the robustness of trained emotion recognition models, and instead
focus on how stress affects the learned acoustic and lexical emotional representations. Unlike the commonly used methods for learning domain invariance, we aim to accomplish generalizing person specific behavior by proactively ``unlearning'' the modulations due to the presence of stress while still retaining emotion information in representations.

In particular, we seek to answer the following questions:
\begin{enumerate}
\item Can we recognize stress given representations trained solely for recognizing emotion? Is the stress recognition performance similar across the lexical and acoustic modalities?
\item Can we completely decorrelate emotion representations from stress representations? If so, how does this decorrelation impact the performance of emotion classifiers? 
\item Does the impact of decorrelation on the performance of emotion classifiers vary given different levels of stress?
\item Does decorrelating these representations (i.e., emotion and stress) aid in model generalizability?
\item Can we proactively remove other types of confounders (e.g., spontaneity) to improve cross-dataset performance?
\item Are there identifiable lexical patterns in samples that are especially successfully classified by the adversarially trained model for emotion classification?
\end{enumerate}
To the best of our knowledge, this is the first work that studies the interplay between emotion and stress in the context of automatic emotion recognition and representation learning.

\section{Experimental Setup}

We use three datasets to study the effect of stress on emotion recognition:
(1) Multimodal Stressed Emotion (MuSE) dataset~\cite{jaiswal2019muse};
(2) Interactive Emotional Dyadic MOtion Capture (IEMOCAP) dataset~\cite{busso2008iemocap}; 
and (3) MSP-Improv dataset~\cite{busso2017msp}. We use the acoustic and lexical features, MFBs and word2vec respectively, as defined in Section~\ref{sec:features}. We then describe the network architecture and the training recipe of the two emotion recognition models, one that controls for 
stress as a confound and one that does not.

\begin{figure}[t]
  \centering
  \includegraphics[width=0.8\linewidth]{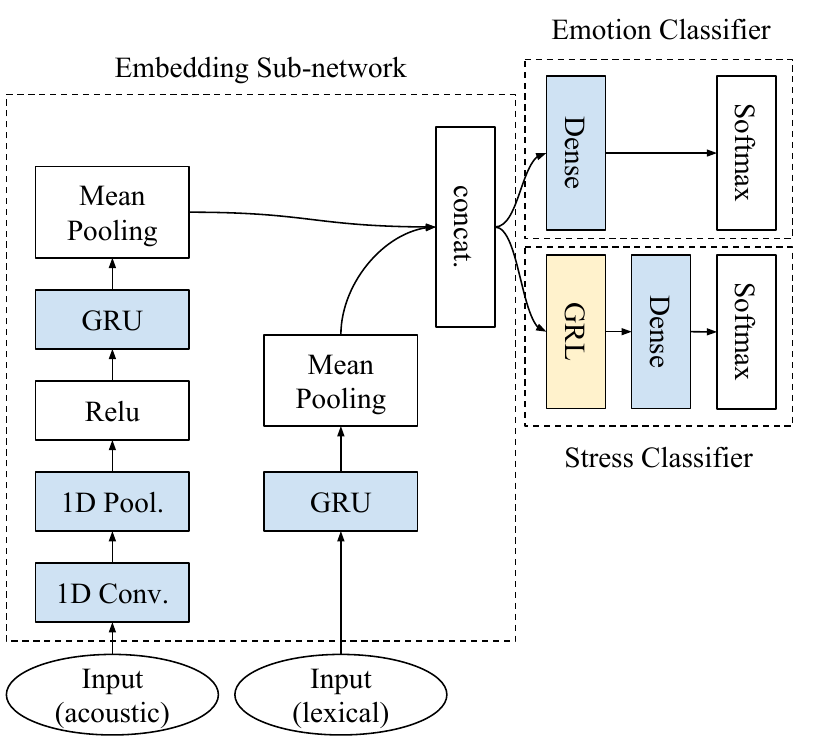}
  \caption{Adversarial multi-task network architecture.}
  \label{fig:model}
\end{figure}
\subsection{Architecture}
\label{sec:setup_arch}

The network consists of three components (Figure~\ref{fig:model}): (1) embedding sub-network; 
(2) emotion classifier; and (3) stress classifier.
The embedding sub-network induces fixed-size representations given the 
acoustic and lexical input streams. In Figure~\ref{fig:model}, the concatenation layer of acoustic and lexical stream shows the induced fixed-size representations. The emotion and stress classifiers
perform their respective classification tasks given the fixed-size representations from
the embedding sub-network. 
We use two variants of the embedding sub-network in this work: a unimodal
and a multimodal variant.
The unimodal embedding sub-network takes a single stream (acoustic or lexical) input while the multimodal embedding sub-network takes a two-stream (acoustic and lexical)
input.
The objective of the adversarial multi-task system is to maximize the performance of the emotion classifier while minimizing the performance of the stress classifier.  

\textbf{Stress-Invariance.}
The network is trained to unlearn stress.  We achieve this goal using a Gradient Reversal Layer (GRL)~\cite{ganin2014unsupervised}.
The use of GRLs is a common approach that can be used to train networks that are invariant to specific properties
\cite{meng2018speaker, shinohara2016adversarial, abdelwahab2018domain, ganin2016domain, mchardy2019adversarial, elazar2018adversarial}.
During the backward pass of the training phase, the GRL
multiplies the backpropagated gradients by $-\lambda$.
During the forward pass, the GRL acts as an identity function.
To make the network invariant to stress, we place the GRL between 
the embedding sub-network and the stress classifier as shown in 
Figure~\ref{fig:model}.

\textbf{Model Variations.} We use 12 variants of the network shown in Figure~\ref{fig:model} with the following combinations:
\textit{\{normal-classification, adversarial-classification\}} $\times$ \textit{\{activation, valence\}} $\times$
\textit{\{uni-lexical, uni-acoustic, multimodal\}}.
The normal classification setup consists of the embedding sub-network (lexical, acoustic, or multimodal) and the emotion classifier (activation or valence) parts of the model.
The adversarial classification setup adds the adversarial stress classifier.
\subsection{Training}
We implement models using the Keras library~\cite{keras2015}.
We use a weighted cross-entropy loss function for each task and learn the model parameters using the RMSProp optimizer~\cite{rmsprop}.
We train our networks for a maximum of 50 epochs and monitor the validation loss from the emotion classifier after each epoch, stopping the training if the validation loss does not improve after five consecutive epochs.
Once the training process ends, we revert the network's weights to those that
achieved the lowest validation loss on the emotion classification task. 
For the adversarial classification model, we ensure that the chosen model yields a validation unweighted average recall (UAR) that is random (0.33) for the stress classification task. 
Finally, we train each setup three times with different random seeds and average the predictions over these runs to reduce variations due to random initialization.

We use validation samples for hyper-parameter selection and early stopping. 
The hyper-parameters that we use in our search include:
number of convolutional layers \{3, 4\}, 
number of convolutional kernels \{2, 3\}, 
conv. layers width  \{32, 64, 128\}, 
1D max-pooling kernel width  \{2\}, 
number of GRU layers  \{2, 3\}, 
GRU layers width  \{32\}, 
number of dense layers  \{1, 2\}, 
dense layers width  \{32, 64\}, 
GRL $\lambda$  \{0.3, 0.6, 0.8\}.
For the adversarial emotion classification setups, we use the hyper-parameters that maximize the validation emotion classification performance while minimizing the validation stress classification performance.
We assess performance using UAR, given the imbalanced nature of our data~\cite{rosenberg2012classifying}.

\section{Analysis}

\subsection{RQ1: Implicit Stress Encoding}
\label{sec:analysis_1}

\begin{table*}
          \centering
          \captionsetup[subtable]{position = below}
          \captionsetup[table]{position=top}
          
\caption{
UAR (chance = 0.333) for predicting activation (top) and valence (bottom) in adversarial and non-adversarial (normal) setups. Bold signifies significantly different performance (paired $t$-test, $\alpha<0.05$).}
     \label{exp-perf}
  
\begin{subtable}{\columnwidth}\centering
\setlength{\tabcolsep}{0.65em}
                     \begin{tabular}{lcccc}
     \toprule
     & \multicolumn{2}{c}{Normal} & \multicolumn{2}{c}{Adversarial}  \\
     \cmidrule(r){1-1} \cmidrule(r){2-3} \cmidrule(r){4-5}
          Setup & Act.     &  Stress  & Act.     &  Stress\\
     \midrule
     Unimodal (A) & \textbf{0.611} & 0.412 & \textbf{0.572}  & 0.305    \\
     Unimodal (L) & 0.550  & 0.394  & 0.527 & 0.332       \\
     Multimodal (A+L)   & \textbf{0.659}  & 0.425 & \textbf{0.613}  & 0.322      \\
     \bottomrule
  \end{tabular}
              \caption{
              Activation}
              \label{exp-performance-act}
          \end{subtable}%
\quad
\hfill
\begin{subtable}{\columnwidth}\centering
\setlength{\tabcolsep}{0.6em}
              \centering
                  \begin{tabular}{lcccc}
                    \toprule
                    & \multicolumn{2}{c}{Normal} & \multicolumn{2}{c}{Adversarial}  \\
                    \cmidrule(r){1-1} \cmidrule(r){2-3} \cmidrule(r){4-5}
                        Setup & Val.     &  Stress  & Val.     &  Stress\\
                    \midrule
                    Unimodal (A) & 0.460 & 0.396 & 0.431  & 0.332    \\
                    Unimodal (L)   & 0.685  & 0.353 & 0.674  & 0.323      \\
                    Multimodal (A+L)   & \textbf{0.666}  & 0.397 & \textbf{0.641}  & 0.328      \\
                    \bottomrule
                  \end{tabular}    
                \caption{
                Valence}
                 \label{exp-performance-val}
          \end{subtable}
      \end{table*}
%

\begin{table*}[t]
\caption{
Confusion matrices for predicting activation (top) and valence (bottom) showing percentage change in classification performance of the multimodal setup for each emotion class after controlling for stress.}
\begin{subtable}{\columnwidth}\centering
\setlength{\tabcolsep}{0.65em}
\begin{tabular}{l|ccc|ccc}
    \toprule
    & \multicolumn{3}{c}{Low Stress} & \multicolumn{3}{c}{High Stress}  \\
    \cmidrule(r){2-4} \cmidrule(r){5-7}
    Act.     & (0) & (1) & (2) & (0) & (1) & (2) \\
    \midrule
    (0) & $-1.21$ & $-0.22$ & $+1.44$ & $+1.30$ & $-2.11$ & $+7.35$ \\
    (1)  & $+4.22$ & $-2.73$ & $+1.31$ & $+18.40$ & $-22.03$ & $+14.80$\\
    (2)  & $-2.01$ & $+6.66$ & $-6.38$ & $-8.81$ & $+6.21$ & $-3.14$\\
    
    \bottomrule
  \end{tabular}
  \caption{
  Activation}
\end{subtable}
\quad
\hfill
\begin{subtable}{\columnwidth}\centering
\setlength{\tabcolsep}{0.6em}
\begin{tabular}{l|ccc|ccc}
    \toprule
    & \multicolumn{3}{c}{Low Stress} & \multicolumn{3}{c}{High Stress}  \\
    \cmidrule(r){2-4} \cmidrule(r){5-7}
    Val.     & (0) & (1) & (2) & (0) & (1) & (2) \\
    \midrule
    (0) & $-1.66$ & $+1.12$ & $+1.11$ & $-8.22$ & $+7.88$ & $+1.33$ \\
    (1)  & $-0.76$ & $-2.22$ & $+0.87$ & $-2.31$ & $-6.11$ & $+4.26$ \\
    (2)  & $-1.01$ & $+0.31$ & $+1.45$ & $-1.15$ & $-2.10$ & $+0.79$\\
    
    \bottomrule
  \end{tabular}
  \caption{
  Valence}
\end{subtable}
\label{emotion-confusion-matrix}
\end{table*}

\noindent\textsc{Question}: \textit{Can we recognize stress given representations trained solely for recognizing emotion?}\\
\textsc{Hypothesis}: \textit{We expect the performance of detecting stress from representations obtained by training emotion classifiers to vary based on the modality, and the emotion being modeled.}\\
Stress has been shown to have varying effects on both the linguistic~\cite{buchanan2014acute} and para-linguistic~\cite{paulmann2016psychological, steeneken1999speech} components of
communication. Previous work has also demonstrated that the lexical part of speech carries more information about valence while the para-linguistic part carries more information about activation~\cite{khorram2017capturing}.
As a result, we expect the performance of stress 
classification to vary based on modality, and emotion dimension being modeled. 

To test our hypothesis, we train the 12 model variants described in section~\ref{sec:setup_arch} with five-fold speaker-independent cross-validation.
We report the average across the five folds for the normal classification and the adversarial classification setups in 
Tables~\ref{exp-performance-act} and~\ref{exp-performance-val} for predicting activation and valence, respectively. 
Our results show that a network trained to only recognize emotion is generally discriminative for stress. For instance, we obtain a maximum UAR of 0.425 when using a multimodal network that was trained to only detect activation; and a maximum UAR of 0.397 when using multimodal network that was trained to only detect valence.

Our results in Table~\ref{exp-performance-act} suggest that the acoustic modality encodes information that is relevant for recognizing stress and activation. 
In contrast, the results show that the representations trained on lexical modality encode information that is relevant for detecting valence but not for stress.
Our findings are consistent with previous research that demonstrated that stress is encoded in acoustic features \cite{buchanan2014acute, freeman2009comprehension, dedora2011acute}. 

\subsection{RQ2: Decorrelate Emotion and Stress Representations}
\label{sec:analysis_2}

\noindent\textsc{Question}: \textit{Can we decorrelate emotion representations from stress representations? How does it impact performance of emotion classifiers?}\\
\textsc{Hypothesis}: \textit{Decorrelating the stress and emotion representations will cause a decrease in emotion classification performance on the source domain.}\\
Previous research demonstrated that controlling for confounders during the training process can cause the performance of the main task on the same dataset to decrease~\cite{satire-adv,subbaswamy2018learning}.
For instance, Zhang et al.~\cite{satire-adv} showed that the performance of detecting sarcasm decreases when controlling for publication as a confounding variable in the training process, but the prediction accuracy increases on an unseen publication set. Similarly, Ganin et al.~\cite{ganin2016domain} showed that controlling for domain while training a network for detecting sentiment can result in a performance reduction on the main task.
The reduction in performance on the main task, after controlling for an extraneous confounding variable, can be attributed to the removal of information that the model can use as a ``shortcut'' for achieving the main task.

Our results show that (Tables~\ref{exp-performance-act} and~\ref{exp-performance-val}):
\begin{itemize}
  \item Activation classification performance decreases given adversarial training.  This decrease is statistically significant for the acoustic (6.382\% drop in UAR) and multimodal (6.980\% drop in UAR) setups.
  \item Valence classification performance decreases given adversarial training. This decrease is statistically significant for the multimodal setup (3.754\% drop).
\end{itemize}
The reduction in performance in the main task after controlling for a confounding variable can also be caused by the removal of information that is equally beneficial for both detecting stress and detecting emotion.
Our results in 
further sections
, however, show that models that control for stress are better able to recognize emotion in new domains, compared to models that do not control for stress. This suggests that the process of ``unlearning'' stress does not come at the expense of the primary task of emotion detection.

\subsection{RQ3: Impact of Stress Levels on Emotion Recognition}
\label{sec:analysis_3}
\noindent\textsc{Question}: \textit{Does the impact on the performance of emotion classifiers vary given different levels of stress}\\
\textsc{Hypothesis}: \textit{The valence and activation emotion classes (low, medium, and high) are impacted differently by stress.}\\
Prior research demonstrated that emotions produced by stressed individuals are not recognized in the same way as 
those
by non-stressed individuals~\cite{paulmann2016psychological}. In particular, researchers  found that speech patterns of negative emotions produced by stressed individuals are more difficult to recognize than negative emotions produced by non-stressed individuals \cite{paulmann2016psychological}. 
We expect similar patterns to hold in automatic emotion recognition systems.
That is, we expect the presence of stress to have a varying effect on the performance of the classifier depending on the emotion class (for valence and activation), and the amount of stress induced.

To test our hypothesis, we study how the performance of the classifier varies for each emotion class when we control for
stress. We report the changes in performance, after controlling for stress, for each emotion class, grouped by stress level
(low, high), in Table~\ref{emotion-confusion-matrix}. The results in the table show:

\begin{itemize}
\item High levels of stress impact classification more strongly (3.89\% and 2.41\% drop in UAR for activation and valence, respectively) than low levels of stress do (1.44\% and 0.31\% drop in UAR for activation and valence, respectively). This is generally true for all emotion classes (valence and activation).
\item High levels of stress have the biggest impact on mid level of activation predictions (22.03\% drop in accuracy for detecting neutral activation).
\item High levels of stress have the biggest impact on low valence predictions (8.22\% drop for low valence).

\end{itemize}
The results show that stress level effects emotion recognition, for both activation and valence. 
This drop in performance can be attributed to changes in the perceived emotions by the annotators. 
Researchers 
have
demonstrated that stressed sentences are usually perceived by annotators to be more neutral than they were originally intended to be \cite{paulmann2016psychological}. 


\subsection{RQ4: Decorrelation and Model Generalizability}
\label{sec:analysis_4}
\noindent\textsc{Question}: \textit{Does the process of decorrelating these representations (i.e., emotion and stress) aid in model generalizability?}\\
\textsc{Hypothesis}: \textit{Removing the confounding factor stress would aid in creating models that are more generalizable across datasets.}\\ Previous research has shown that laboratory collected datasets are too small and often fail to capture the complete distribution of the domain~\cite{hu2017frankenstein, locatello2018challenging} present in the real world. These datasets are often plagued with unintentional correlational factors~\cite{locatello2018challenging, landeiro2016robust}. Hence we believe that removing modulations due to stress should aid the generalizability of the model to datasets, where this psychological factor is either unmeasured or the distribution is non-uniform between training and testing.

To answer if the models trained on MuSE dataset generalize better, we perform two sets of experiments: (a) self generalizability in artificially partitioned datasets with different stress distributions for evidence of concept and (b) cross-dataset generalizability.

\textbf{Artificially Segmented Within-Dataset Performance.} We run the first set of experiments by creating partitions of data by stress level. We do this to create artificially mismatched environments between training and testing. 
We reserve one set to be test set (target), while keeping the other two for training and validation combined (source). This is in similar vein to partitioning created across confounding factor for UCI Bike Rentals Dataset in \cite{subbaswamy2018learning}. To ensure speaker independent sets, we divide the training set using an 80:20  split (train and validation), ensuring no speakers overlap. We run these models $n$ times where $n$ is the number of speakers in test data, that are also present in train/validation data. For each run, we remove one speaker from the train/validation data and test on that speaker. We calculate the average test performance of all these runs as the performance of the model for that setup.

\begin{table}[]
\caption{
Performance (UAR)  predicting activation (left) and valence (right) in non-adversarial and adversarial (for best \textit{lambda} value) for self-partition on MuSE. Bold signifies significantly different performance (paired $t$-test, $\alpha<0.05$).
  }
  \label{exp-tri-self-partition}
\centering
\begin{tabular}{lllll}
\toprule
           & \multicolumn{2}{c}{Normal} & \multicolumn{2}{c}{Adversarial} \\ \cmidrule(r){2-3} \cmidrule(r){4-5}
                 & Act            & Val            & Act              & Val             \\ \midrule
\multicolumn{5}{c}{\textbf{Train: Stress (Medium + High) Test: Stress (Low)}}                                                      \\ \midrule
Unimodal (A)     &       \textbf{0.623}        &       0.451         &        0.582          &     0.453            \\
Unimodal (L)     &      0.561          &      0.654         &       0.548           &       \textbf{0.691}          \\
Multimodal (A+L) &       0.650         &        0.673        &        \textbf{0.662}          &       \textbf{0.703}          \\ \midrule
\multicolumn{5}{c}{\textbf{Train: Stress (Low + High) Test: Stress (Medium)}}                                                      \\ \midrule
Unimodal (A)     &      0.610          &        0.420        &       \textbf{0.628}           &    \textbf{0.432}             \\
Unimodal (L)     &        0.520        &      0.672         &      \textbf{0.545}            &      0.669           \\
Multimodal (A+L) &      0.602          &        0.621        &        \textbf{0.638}          &   \textbf{0.649}              \\ \midrule
\multicolumn{5}{c}{\textbf{Train: Stress (Medium + High) Test: Stress (Low)}}                                                      \\ \midrule
Unimodal (A)     &      0.582          &     0.384           &        \textbf{0.605}          &   \textbf{0.411}              \\
Unimodal (L)     &    0.540            &     0.630          &       0.513           &      \textbf{0.652}           \\
Multimodal (A+L) &        0.642        &     0.621           &      0.647            &     \textbf{0.637}            \\ 
 \bottomrule
\end{tabular}
\end{table}

We report our results in Table~\ref{exp-tri-self-partition}.
When we consider low levels of stress as our target, we see that adversarial classification significantly improves performance over normal classification for multimodal setup for activation and for both, lexical and multimodal setup for valence.
Considering mid levels of stress as our target, adversarial classification significantly improves performance over normal classification for all setups for activation and for both, acoustic and multimodal setup for valence.
Subsequently considering high levels of stress as our target, adversarial classification significantly improves performance over normal classification for acoustic setups for activation and for all setups for valence.

\textbf{Cross-dataset Performance.} Now that we have evidence for concept for artificially mismatched distributions that removing stress as a confounding factor can aid generalizability, we ask if adversarially removing encoded stress from emotion representation improves cross-dataset performance. We assume that we previously do not have any samples from the target dataset to train our model, to test generalizability at deployment. We train a dataset on complete MuSE data, keeping 20\% of speaker independent data for validation of hyper-parameters. Then we use the trained model to test on IEMOCAP for a combination of acoustic, lexical and multimodal inputs and on MSP-Improv for acoustic inputs.

We report our results of comparing the performance of the adversarial and normal models (Table~\ref{exp-tri-other}):
\begin{itemize}
    \item Activation: There is a significant increase in performance in all setups when the adversarial classification model is tested on IEMOCAP.
    We observe a significant increase in performance in acoustic setup of adversarial classification model (0.404 vs 0.421) when tested on MSP-Improv.
    \item Valence: There is a significant increase in performance using acoustic setup (0.376 vs 0.401) and multimodal setup (0.431 vs 0.472) of adversarial model when tested on IEMOCAP.
    We see no significant difference in performance when testing on MSP-Improv using adversarial classification model .
\end{itemize}
Based on these results, we understand that removal of a psychological confounding factor, stress, generally aids in the generalizability of the model on completely unseen data, where the distribution of this confounding factor is unknown. 

\begin{table}[]
\caption{
Performance (in UAR)  predicting activation (left) and valence (right) in  non-adversarial and adversarial (for best \textit{lambda} value) setups across datasets when trained on MuSE. Bold signifies significantly different performance (paired $t$-test, $\alpha<0.05$).
  }
  \label{exp-tri-other}
\centering
\begin{tabular}{lllll}
\toprule
           & \multicolumn{2}{c}{Normal} & \multicolumn{2}{c}{Adversarial} \\ \cmidrule(r){2-3} \cmidrule(r){4-5}
                 & Act            & Val            & Act              & Val             \\ \midrule
\multicolumn{5}{c}{\textbf{MuSE to IEMOCAP}}                                                      \\ \midrule
Unimodal (A)     &      0.419          &      0.376          &   \textbf{0.448}               &       \textbf{0.401}          \\
Unimodal (L)     &    0.401            &    0.433       &   \textbf{0.436}               &           0.447      \\
Multimodal (A+L) &      0.422          &      0.431          &      \textbf{0.459}            &   \textbf{0.472}              \\ \midrule
\multicolumn{5}{c}{\textbf{MuSE to MSP-Improv}}                                                   \\ \midrule
Unimodal (A)     &      0.404          &      0.368          &          \textbf{0.431}        &   0.372              \\ \bottomrule
\end{tabular}
\end{table}

\subsection{RQ5: Spontaneity as Confounding Factors}
\label{sec:analysis_5}

\noindent\textsc{Question}: \textit{Can we proactively remove other types of confounders to improve cross-dataset performance?}\\
\textsc{Hypothesis}: \textit{Removing the confounding factor of spontaneity in IEMOCAP will improve cross-dataset performance.}\\
We observed in 
the last question
that ``unlearning" the confounder stress can aid generalizability. Now, we want to see if the same method can be used to make models trained using other datasets more reliable to change in target data distribution. 
We hypothesize that decorrelating the effect of spontaneity on emotion representation will lead to models that generalize better. This is because, as shown in \cite{mangalam2017learning}, the emotional content expression is different in scripted vs spontaneous speech, and hence should modulate the emotion representation in trained model. To this end, we use the IEMOCAP dataset which has utterances that come from sessions that are both scripted and improvised. We do not use MSP-Improv for similar analysis here, because the scripted sessions, by corpus design, have limited lexical content and hence wouldn't cover enough input representation space for generalizability. We train the same 12 model variants described in section~\ref{sec:setup_arch} replacing the adversarial stress classifier sub-component with adversarial spontaneity classifier for this analysis.

We report our results in Table~\ref{exp-iemocap-other}. We compare the performance of the adversarial and normal models:
\begin{itemize}
    \item Activation: There is a significant increase in performance in lexical (0.401 vs 0.425) and multimodal setup (0.433 vs 0.467) when the adversarial model is tested on MuSE dataset.
    We see no significant difference in performance when the adversarially trained model is tested on MSP-Improv.
    \item Valence: There is a significant increase in the performance using all setups of the adversarial classification model when tested on MuSE.
    We observe a significant increase in the performance in the acoustic setup of the adversarial model (0.410 vs 0.438) when tested on MSP-Improv.
\end{itemize}
We see that the removal of modulations due to the data collection methodology improves generalizability for many cross-dataset cases. This suggests that this method can be extended to train stabler models by explicitly accounting for confounding variables in limited amounts of training data.

\begin{table}[t]
\caption{
Performance (in UAR)  predicting activation (left) and valence (right) in non-adversarial and adversarial (for best \textit{lambda} value) setups across datasets when trained on IEMOCAP. Bold signifies significantly different performance (paired $t$-test, $\alpha<0.05$).
  }
 \label{exp-iemocap-other}
\centering
\begin{tabular}{lllll}
\toprule
           & \multicolumn{2}{c}{Normal} & \multicolumn{2}{c}{Adversarial} \\ \cmidrule(r){2-3} \cmidrule(r){4-5}
                 & Act            & Val            & Act              & Val             \\ \midrule
\multicolumn{5}{c}{\textbf{IEMOCAP to MuSE}}                                                      \\ \midrule
Unimodal (A)     &       0.428         &      0.401          &   0.427               &      \textbf{0.431}           \\
Unimodal (L)     &      0.401          &    0.423            &     \textbf{0.425}             &         \textbf{0.455}        \\
Multimodal (A+L) &     0.430           &       0.429         &        \textbf{0.463}          &          \textbf{0.468}       \\ \midrule
\multicolumn{5}{c}{\textbf{IEMOCAP to MSP-Improv}}                                                   \\ \midrule
Unimodal (A)     &       \textbf{0.414}         &      0.410          &       0.402           & \textbf{0.439}           \\ \bottomrule
\end{tabular}
\end{table}

\subsection{RQ6: Lexical Patterns in Samples That Benefit from Adversarial Training}
\label{sec:analysis_6}
\noindent\textsc{Question}: \textit{Are there identifiable lexical patterns in samples that are especially successfully classified by the adversarially trained model for emotion classification?}\\
\textsc{Hypothesis}: \textit{Certain properties of input features correlate with the increase in probability of successful classification in adversarially trained emotion recognition models.}\\
Our results in questions 4 and 5 of section~\ref{sec:analysis_4} 
demonstrate that decorrelating the representations from modulations due to confounding factors can positively affect the classification performance of our trained emotion recognition models when applied to datasets whose properties differ from the data on which the models were trained.

In this section, we aim to understand what properties of input features in a given sample
correlate with the probability of successful classification in trained emotion recognition models due to decorrelation of such modulations. 
Understanding the relationship between the properties of the input features and the likelihood of success 
can help us identify data points that are more likely to be correctly classified using adversarially trained models. This can help us identify samples in an unseen dataset for which we can trust the classification label obtained from the adversarial model as compared to the normal classification model.
We analyze this relationship using word tokens, which are low-dimensional and human-interpretable.  
\subsubsection{\textbf{Adjusted Probability of Success}}
We study the correlation between the lexical patterns of data samples and the probability that those samples are correctly classified.  We focus on improvements in classifaction, moving from the normal model to the adversarial model.  This allows us to focus on improvement and mitigates the challenge that certain samples may just be particularly easy or hard to classify. 
We define probability of success for a sample as the $P_{A, s}(Success)$ where A can either be a normal (\textit{normal}) classification model or an adversarial classification model (\textit{adv}) and $s$ refers to the index of a particular sample.  We calculate $P_A(Success)$ as the ratio of the number of times a model correctly classifies a given sample to the total number of fifteen runs. 
If a sample is correctly classified across all runs by adversarial model, the $P_{adv}(Success)$ for that sample would be $1$. But we want to concentrate on the gain in performance of using adversarial over normal classification. It might be the case that this sample is correctly classified across all runs by normal classification model as well, the $P_{normal}(Success)$ for that sample would be $1$. In this case, we do not see any betterment as a result of using adversarial training paradigm.
To mitigate the above limitation, we define adjusted probability of success in the following manner:
We define adjusted probability of success (APS) for sample $s$ as: $P_{adv, s}(Success)-P_{norma, sl}(Success)$.  When the APS is greater than 0, the sample is more accurately classified using the adversarial model.  When the APS is less than zero, it is more accurately classifed using the normal model.

\subsubsection{\textbf{Features}}
Our goal is to correlate APS with interpretable lexical features. We 
use the Linguistic Inquiry and Word Count (LIWC)~\cite{pennebaker2001linguistic} tool. 
LIWC assigns a predefined category to a word based on its association with social, affective and cognitive process.
These categories have been shown to be highly predictive of both emotion~\cite{kahn2007measuring}, spontatenity~\cite{clark2002using} and stress~\cite{wang2016twitter}.

We form a twelve length feature vector for each utterance by counting the number of words that fall into each of the nine LIWC categories (adverb, pronoun, social process, negation, positive and negative emotion, insight, tentative and certainty).  We normalize the feature vector by how many words in the utterance.
We augment this feature vector to include: (1) fillers (e.g., ``uhh''),  hesitation (e.g., ``like''), and discourse markers (e.g., ``so'') and (2) content rate, defined as the number of words per unit length of time. The final feature vector comprises of all the above mentioned categories.
\subsubsection{\textbf{Discussion}}
\textbf{Decorrelating Stress.} We report the Pearson correlation coefficient and the resulting Benjamini-Hochberg adjusted~\cite{benjamini1995controlling} $p$-values that we obtain between each feature in the vector, and the APS for each sample.  We perform this assessment for both the activation and valence normal and adversarial lexical models.  We focus on the cross-dataset case in which the models were trained on MuSE and tested on IEMOCAP (in Table~\ref{exp-linguistic-correlation-stress}). A large positive correlation between a category and APS implies that samples with larger numbers of words in a given category are likely to be classified correctly more often given the adversarial model versus the normal model. 

\begin{table}[t]
  \caption{
  Correlation between LIWC features and APS due to stress decorrelation, for activation and valence.
  $p$-values are Benjamini-Hochberg adjusted for multiple comparisons ($\alpha = 0.05$). p-value codes: `**'$<$0.01; `*'$<$0.05; `-'$<$0.1}
  \label{exp-linguistic-correlation-stress}
  \centering
  \begin{tabular}{lcccc}
    \toprule
    & \multicolumn{2}{c}{Act.} & \multicolumn{2}{c}{Val.}  \\
    \cmidrule(r){2-3} \cmidrule(r){4-5}
         & r     &  p  & r     &  p\\
    \midrule
    \textbf{LIWC} \\
    \midrule
    Adverb                      & \textbf{0.217}   & ** & \textbf{0.177}  &  *  \\
    Pronoun                     & \textbf{0.165}   & -  & 0.082 & * \\
    Social Process (social)     & 0.084  & -  &   0.001    & -      \\
    Negations (negate)          & -0.018   & -  & 0.005  & -\\
    Positive emotion (posemo)   & \textbf{0.154}    & *  &  0.093  & - \\
    Negative emotion (negemo)   & 0.086   & -  & \textbf{0.143} & * \\
    Insight                     & -0.021   & -  & -0.012   &   -   \\
    Tentative (tentat)          & 0.074   & -  &    0.101    &  -    \\
    Certainty (certain)         & \textbf{0.138} & *  &  0.116     &  -\\
    \midrule
    \textbf{Hesitation}          \\
    \midrule
    Fillers      &  \textbf{0.154} & *  & \textbf{0.182} & * \\
    Discourse marker &  \textbf{0.141}     & *  & 0.111 & - \\
    Content Rate    &   \textbf{0.196}    & ** & \textbf{0.178} & ** \\
    \bottomrule
  \end{tabular}
\end{table}

\begin{itemize}
\item Activation recognition: APS is significantly positively correlated with the presence of words that relate to: Adverb (0.217), pronoun (0.165), positive emotion (0.154), certainity (0.138), fillers (0.154), discourse markers (0.141), and content rate (0.196).
\item Valence recognition: APS is significantly positively correlated with the presence of words that relate to: Adverb (0.177), negative emotion (0.143), fillers (0.182), content rate (0.178).

\end{itemize}
This finding is consistent with previous research~\cite{mehl2017natural}, where the authors have shown that there is often an  increase in usage of function words and intensifiers in stressed conditions. So, for example a sentence "I am really really sad about losing my pen" would have more likelihood of being correctly classified by the adversarial model compared to the normal emotion classification model. Hence, we can hypothesise that an increase in the likelihood of correct classification of samples containing these intensifiers occurs due to reduced weightage of adverbs in adversarial training paradigm.

There are fewer significant categories for valence than for activation. This is consistent with the results in Table~\ref{exp-tri-other} for the lexical modality.
Although we see a significant correlation between filler words and APS for activation classification~\cite{Duvall2014ExploringFW}, we do not observe the same for discourse markers and presence of social process words.  
The absence of significance in these cases implies that though these values are markers of stress, the normal classifier is still able to learn reliable representations invariant of stress for predicting the correct target label, resulting in negligible impact on classification performance when decorrelating the representations.



\textbf{Decorrelating Spontatenity.} We do a similar analysis for analyzing lexical properties of samples that were aided by decorrelating spontaneity. We report the Pearson correlation coefficient and the resulting Benjamini-Hochberg adjusted~\cite{benjamini1995controlling} $p$-values that we obtain from the LIWC features and APS for both emotion axes lexical-based classification models (trained on IEMOCAP; tested on MuSE) in Table~\ref{exp-linguistic-correlation-spont}.

\begin{table}[t]
  \caption{
  Correlation between LIWC features and APS due to spontaenity decorrelation, for activation and valence.
  $p$-values are Benjamini-Hochberg adjusted for multiple comparisons ($\alpha = 0.05$). p-value codes: `**'$<$0.01; `*'$<$0.05; `-'$<$0.1}
  \label{exp-linguistic-correlation-spont}
  \centering
  \begin{tabular}{lcccc}
    \toprule
    & \multicolumn{2}{c}{Act.} & \multicolumn{2}{c}{Val.}  \\
    \cmidrule(r){2-3} \cmidrule(r){4-5}
         & r     &  p  & r     &  p\\
    \midrule
    \textbf{LIWC} \\
    \midrule
    Adverb                      & 0.121   & ** & 0.088  &  -  \\
    Pronoun                     & \textbf{0.138}   & -  & 0.016 & * \\
    Social Process (social)     & 0.132  & -  &   \textbf{0.166}     & *      \\
    Negations (negate)          & -0.003   & -  & -0.011  & -\\
    Positive emotion (posemo)   & 0.122   & -  &  \textbf{0.161}  & * \\
    Negative emotion (negemo)   & \textbf{0.137}   & *  & \textbf{0.148} & * \\
    Insight                     & 0.017   & -  & 0.099   &   -   \\
    Tentative (tentat)          & \textbf{0.155}   & *  &    0.112    &  -    \\
    Certainty (certain)         & \textbf{0.191}  & *  &  \textbf{0.172}     &  *\\
    \midrule
    \textbf{Hesitation}          \\
    \midrule
    Fillers      &  \textbf{0.221} & **  & 0.119 & * \\
    Discourse marker &  \textbf{0.189}  & *  & 0.122 & - \\
    Content Rate                &   \textbf{0.165}   & * & \textbf{0.144} & - \\
    \bottomrule
  \end{tabular}
\end{table}

\begin{itemize}

\item Activation recognition: APS is significantly positively correlated with the presence of words that relate to: Pronoun (0.138), negative emotion (0.137), tentativeness (0.155), certainty (0.191), fillers (0.221), discourse markers (0.189) and content rate (0.165).

\item Valence recognition: APS is significantly positively correlated with the presence of words that relate to: Social Process (0.166), positive (0.161) and negative (0.148) emotion, certainty (0.172), and content rate (0.144).
\end{itemize}

The results suggest that there are identifiable linguistic properties of samples whose likelihood of correct classification benefits from the model trained adversarially to decorrelate spontaneity and emotion representation as compared to normal classification model. This is especially true for the use of words in the certainty category for both emotion dimensions and all hesitation categories for activation. Spontaneous speech has been shown to have more of these words in~\cite{clark2002using}. 
Scripted content has been shown to have more exaggerated displays of emotion through words and facial expressions~\cite{jurgens2015effect}. Controlling for the weights assigned to words in positive and negative emotion categories using the adversarial model, leads to better classification of samples that are comprised of these word tokens.


\section{Conclusions}
This work focused on the interplay between stress and emotion in the context of automatic emotion recognition.
We first showed that the presence of stress affects the performance of emotion recognition models.
We then observed that these effects vary depending on modality (acoustic or lexical) and task (activation or valence classification). We then showed how decorrelating stress modulations from emotion representations aids the generalizability of the model. Next, we showed how a similar method could be used to control for variations due to spontaneity; facilitating the generalizability of the model.
Finally, we identified human interpretable lexical markers that correlate with successful generalization of the model; especially concentrating on samples that are aided by decorrelation of stress and emotion representation.

Our results suggest that an extraneous psychological factor, such as stress, can significantly impact the performance of emotion recognition systems both within and across datasets. As a result, extraneous psychological factors should be accounted for when collecting data for training emotion recognition systems, especially when being used to predict labels of data that may or may not be modulated by those same factors. We then show how proactive decorrelation of this confounder can improve generalization of the model to other dataset at time of deployment. 





%% file: Chapters/chap8.tex
\section{Motivation and Contributions}
The rise of mobile applications and conversational agents harnessing emotion recognition capabilities has brought a crucial challenge to the forefront: the safeguarding of sensitive user information. This data, encapsulating detailed demographic aspects, is often extracted from user devices and stored on centralized servers. In worst-case scenarios, these details could be exploited without user consent or manipulated by harmful entities. Storing abstracted representations rather than raw data, while a potential solution, doesn't fully counter the problem. Indeed, these representations, intended for tasks like emotion recognition, are susceptible to inadvertent demographic leakage undermining user-defined settings (for example, to not receive gender targetted ads) and exposing sensitive variables from unimodal (textual or acoustic) or multimodal data.

The consequences of unintentional demographic leakage are significant. Continual exposure of sensitive attributes threatens the integrity of user data. Moreover, the inherent tie-in of demographics such as age, gender, and race with emotion recognition models makes these models potential reservoirs of sensitive user information—even when storing only representational data. As a result, measures to counter demographic leakage have become a pressing need.

Addressing these issues, our research focuses on developing robust methods to mitigate demographic leakage. We propose to use an adversarial learning model, which is trained with the aim to remove the learnt sensitive information encoded in the generated representation from any emotion recognition system. Varying the intensity of this "removal" offers us insight into the effects on the primary task and the ability of an attacker to accurately predict demographic information from just the input to a model and the generated representations. Additionally, we introduce a metric aimed at quantifying the defense against sensitive information exposure and the probability of membership identification — an unauthorized person's ability to determine if a user was part of the training data set. Our work across multiple datasets yields encouraging results; enhancing protection of sensitive data without significantly hampering the performance of primary emotion-recognition tasks. Ultimately, our work represents a pivotal stride in bolstering user data security within emotion recognition systems, setting in place a strong foundation for future research and improvements in this critical field.

\section{Introduction}
Virtual conversational agents strive to emulate human-like interaction to have more naturally flowing conversation~\cite{metcalf2019mirroring}.
These agents often employ models that classify aspects of communication, including the classification of the emotional content of speech.
~\cite{huang2018automatic}. 
The resulting predictions can then be used to bias 
response generation. Emotion classification is also used in mobile and web applications to identify heightened risk of suicidal ideation or mood fluctuations~\cite{Khorram2018,matton2019into,gideon2019emotion}, for the purpose of tracking or intervention. Data are sent from users' devices, including mobile applications~\cite{Khorram2018} and Alexa or Google home devices~\cite{piersol2019pre}, and are stored on central servers for analysis.

However, data transmitted from users' devices are vulnerable to data hacking and re-identification~\cite{barbaro2006face}. Eavesdroppers can use these data for identification of an individual and to gain access to sensitive information.
A way to counter this issue in data collected by mobile or smart home applications is to generate a data representation on the device and then to transfer that representation to the server for additional processing. 
The benefit is that these representations can decrease leakage of content and sensitive information by partially obfuscating the actual content of the conversation~\cite{bengio2013representation}. However, they still contain sensitive demographic information.

The implications of sensitive information leakage is profound: research has shown that discrimination occurs across variables of age, race, and gender in hiring, policing and credit ratings~\cite{hajian2012study}.~\cite{abadi2016deep} showed how adding random noise to aggregated dataset or individual samples can ensure defense against attacks that aim to classify sensitive information from the representation.
But, previous research has shown that using additional noise can often be exploited if the adversary has access to the network used to generate anonymity~\cite{kifer2011no}.
Therefore to ensure robustness, we consider a scenario of the attacker having access to the same embedding sub-network to generate the representations that will be used to train its attack network.

In this work, we focus on sensitive information encoding in the context of emotion recognition.  Emotion recognition provides an important test case because emotion production varies significantly across gender and race.  As a result, the outputs of emotion recognition models are often highly correlated with these secondary demographic signals~\cite{chaplin2015gender,soto2009emotion}.  We design approaches to first measure leakage and then to counteract this leakage.
We measure sensitive information encoding in two ways: 1) using a sensitive information reduction metric, which we define as the incapability of an attacker to recover demographic information from representations, and 2) by determining an adversary's ability to perform membership identification~\cite{li2013membership}, defined as the ability to determine if a given user was in a dataset from which the emotion recognition models were trained (this can be harmful if the training data are collected in a sensitive context, such as counselling or therapy).  We ask the following seven questions:
\begin{enumerate}
\itemsep-0.2em 
    \item Does demographic leakage differ in umimodal and multimodal emotion recognition models?
    \item How does the sensitive information reduction metric change when a network is trained to not encode sensitive information?
    \item How does emotion recognition performance change when networks are trained to not encode sensitive information?
    \item How does the  adversarial component's strength impact emotion recognition performance and the sensitive information reduction metric?
    \item Focusing on gender, how does the performance of emotion recognition change when a network is trained to not encode sensitive information?
    \item Does the location of the adversarial component within a network  affect the sensitive information reduction metric and emotion recognition performance?
    \item Does the sensitive information reduction paradigm that we used to reduce encoding of gender information also help defend against other attacks such as membership identification?
    
\end{enumerate}

Our results show that representations obtained for emotion recognition can be exploited by an adversary to predict sensitive variables given unimodal information (either audio or lexical). We further show that multimodal models contain even more sensitive information, as lexical and audio each encode different aspects of demographic information. We show that we can increase the defense against this attack by adversarially training representations to be invariant to gender. The novelty of this work is two fold: (1) we analyze how the demographic variable encoded in a representation differs across modalities and how it can be increased using adversarial paradigm; and, (2) we obtain enhanced representations that defend against multiple sensitive information leakage or prediction attacks while still maintaining performance on emotion recognition.

Given most of the previous work on representations that aim to reduce encoded sensitive information concentrates on just lexical information, we tackle the questions that arise from desiring such preservation in multimodal representations for emotion recognition. While the primary goal of most previous works has been to avoid unintentional inference by the application itself, we concentrate on minimizing the potency of an attacker to deliberately recover sensitive attributes from an invariant representation.


\section{Experimental Setup}

We use four common emotion recognition datasets:
MSP-Improv~\cite{busso2017msp}, MSP-Podcast~\cite{lotfian2017building}, Interactive Emotional Dyadic MOtion Capture (IEMOCAP) dataset~\cite{busso2008iemocap}, and
Multimodal Stressed Emotion (MuSE) dataset~\cite{jaiswal2019muse}. We use the acoustic and lexical features, MFBs and word2vec respectively, as defined in Section~\ref{sec:features}. Next, we describe the network architecture, the training recipe, and the metrics in consideration.

\subsection{Architecture}
\label{sec:setup_arch_c8}

\begin{figure}[t]
  \centering
  \includegraphics[width=\linewidth]{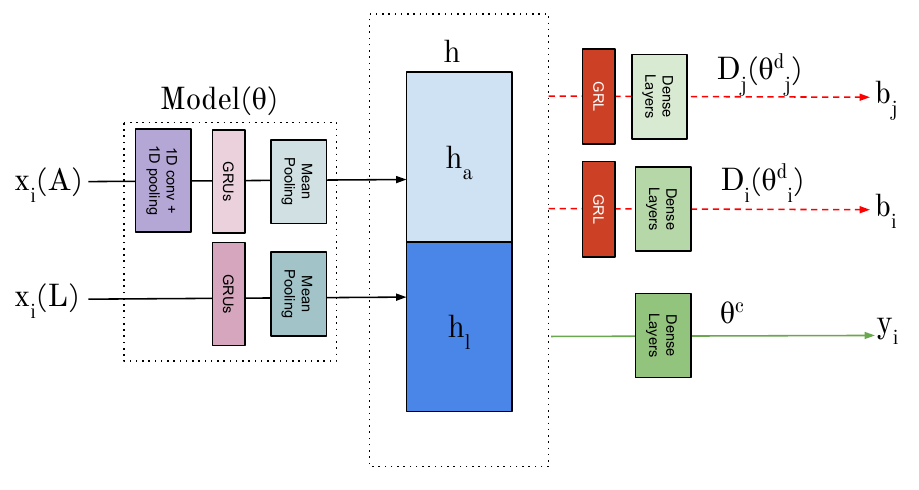}
  \caption{Privacy preserving network architecture.}
  \label{fig:model_chap8}
\end{figure}

The objective of this system is to maximize the performance of the emotion classifier while minimizing the performance of the gender classifier (see Figure~\ref{fig:model_chap8}).
The main network consists of three components: (1) embedding sub-network, $Model(\theta)$; (2) emotion classifier, $\theta^c$ and output $y_i$; and (3) gender classifier, $D_i(\theta^d_i)$, with output, $b_i$. We then disucss how an attacker network could maliciously use this information to obtain sensitive demographic information.

\textbf{Main Network. }The \textit{embedding sub-network} uses a state-of-the-art multimodal approach in emotion recognition~\cite{aldeneh2017pooling} in which the acoustic and lexical information are processed separately and then joined after the application of modality-specific global mean pooling.  The acoustic input stream $(x_i(a))$, where $i$ represents an input frame (40-dimensional) and $a$ represents acoustic, is processed using a set of convolution layers and Gated Recurrent Units (GRU), which are fed through a mean pooling layer to produce an acoustic representation $(h_a)$. The lexical input $(x_i(l))$,  where $i$ represents an input word (300-dimensional) and $l$ represents lexical, is passed through GRUs and pooled to obtain a lexical representation $(h_l)$. For the multimodal setup, these two representations, $(h_a)$ and $(h_l)$, are concatenated $(h)$. The representations ($h, h_a, h_l$) are of fixed-length given acoustic and lexical inputs. 
The \textit{emotion classifier} takes in the representation ($h, h_a,$ or $h_l$) as input and estimates valence or activation using a set of dense layers. The \textit{gender classifier} estimates the gender label (i.e., male or female) using a set of dense layers.

\textbf{Gender-Leakage.}
The main network is trained to unlearn gender. To achieve this goal, we use a Gradient Reversal Layer (GRL)~\cite{ganin2014unsupervised}. GRLs are a common multi-task approach to train networks such that they are invariant to specific properties~\cite{meng2018speaker,mimansa2019controlling}.
During the backward pass of the training phase, the GRL multiplies the backpropogated gradients by $-\lambda$ (i.e., the strength of the adversarial component). During the forward pass, the GRL acts as an identity function. To make the network invairant to gender, we place the GRL function between the embedding sub-network and the gender classifier. We obtain gender-invariant representations using the following loss function:
\[\widehat{\theta}= \underset{\theta_{M}}{min}\underset{\{\theta_{D^i}\}^N_{i=1}}{max}\chi(\widehat{y}(x;\theta_M),y)-\sum_{i=1}^{N}(\lambda_i.\chi(\widehat{b}(x;\theta_{D^i}),b_{i}))\] 
\noindent where $N$ is the number of targeted sensitive variables (here $N$=$1$). The loss function ensures that while the output components are trained to be good predictors, the  representation is trained to be maximally good for the primary task (emotion) and maximally poor for the secondary task (gender).

\textbf{Attacker Network.}
We assume that the attacker has access to a held-out dataset (either a different dataset or a section of the original dataset) with known gender labels. The attacker generates representations for this dataset using the previously described embedding sub-network. 
The network then learns to predict gender labels from the generated representations using a set of dense layers.  Since the parameters used to construct the representation are fixed, the attacker only acts upon its own parameters to optimize the gender prediction linear loss. 
The purpose of the attacker's network is to recover gender information from representations whose labels are unknown. Though testing using a singular network isn't a guarantee of robustness of the representation to attacks that aim to predict or extract sensitive information from representations, for the scope of this chapter, we use a feed forward network, one of the powerful learning methods on a fixed size static representation.

\textbf{Model Variations.} We use 12 variants of the network shown in Figure~\ref{fig:model_chap8}.
We train combinations of the following setups:
\textit{\{general-classification-model (Gen), sensitive-information-reduction-classification-model (SIR)\}} $\times$ \textit{\{activation, valence\}} $\times$
\textit{\{uni-lexical, uni-acoustic, multimodal\}}.

The general classification setup makes use of the embedding sub-network with text, speech, or both as input streams and the emotion classifier.  The sensitive information reducing classification setup adds the gender classifier to the general setup.

\textbf{Training.}
We implement models using the Keras library.
~\cite{keras2015}.
We use a weighted cross-entropy loss function for each task and learn the model parameters using the RMSProp optimizer.
~\cite{rmsprop}.
We train our networks for a maximum of 50 epochs and monitor the validation loss for the emotion classifier after each epoch, stopping the training if the validation loss does not improve for five consecutive epochs.
Once the training ends, we revert the network's weights to those that
achieved the lowest validation loss on the emotion classification task. 
For the classification model trained to not encode sensitive information, we ensure that the chosen model yields a chance unweighted average recall (UAR) for the gender classification task on the validation set. 
Finally, we train each setup three times with different random seeds and average the predictions over these runs to reduce variations due to random initialization.

We use validation samples for hyper-parameter selection and early stopping. 
The hyper-parameters that we use for the main network include:
number of convolutional layers \{3, 4\}, 
width of the convolutional layers \{2, 3\}, 
number of convolutional kernels \{32, 64, 128\}, 
number of GRU layers  \{2, 3\}, 
GRU layers width  \{32\}, 
number of dense layers  \{1, 2\}, 
dense layers width  \{32, 64\}, 
GRL $\lambda$  \{0.3, 0.5, 0.75, 1\}.
For the adversarial emotion classification setups, we use the hyper-parameters that maximize the validation emotion classification performance while minimizing the validation gender classification performance.
For the attacker's model, we use the following hyper-parameters:
number of dense layers  \{2, 3, 4\}, 
dense layers width  \{32, 64\}.
We report the UAR performance of our models, given the imbalanced nature of our data.
~\cite{rosenberg2012classifying}.
\subsection{Metrics}
\label{subsec:metrics}
  \textbf{Performance.} We define performance for emotion recognition as the ability of the model to correctly classify either activation or valence into 3 categories: low, medium, and high.  We measure performance using UAR (chance is 0.33).
  \\\textbf{Demographic Leakage.} Leakage is defined as the ability of a trained gender classifier to predict gender from the representations which are obtained when the network is simultaneously trained to perform the primary task.
  \\\textbf{Demographic Sensitive Information Reduction Metric.} We define the sensitive information reduction metric as the inability of an attacker to be able to recover gender from the representations trained on a primary task. To test this, we use four phases of training.
  \begin{enumerate}
  \itemsep-0.2em 
    \item 
    We train the main network on a dataset (D1), represented by the pair $(x_{D1},y_{D1})$, where $x$ is the data input while $y$ is the gender label. We obtain representations for this dataset ($h(x_{D1})$).
    \item 
    We consider that the attacker has access to another dataset or unused subset of the same dataset (D2) represented by the pair $(x_{D2},y_{D2})$. We generate representations $h(x_{D2})$ for the pairs in this dataset using the embedding sub-network of the main network.
    \item 
    We train a model $(M_{att})$ to predict gender labels using the representations obtained in step 2, represented as $M_{att}((h(x_{D2}), y_{D2}))$.
    \item 
    Using the model obtained previously $(M_{att})$, we choose $h(x_{D1})$ as inputs, and measure the gender prediction capability of the attacker $UAR(M_{att}((h(x_{D1}), y_{D1}))$. The Demographic Sensitive Information Reduction metric of an attacker is then quantitatively defined as $1-UAR(M_{att})$.
  \end{enumerate}
The range of the sensitive information reduction metric goes from $0$ (the attacker is always correct) to $0.5$ (the attacker has a chance UAR). 
  \\\textbf{Membership Identification.} Membership identification is the possibility of an attacker being able to recognize if a speaker belongs to the training set. 
  We assume that the adversary can obtain samples from a speaker from the same distribution as that for the training set.
  Consider that the adversary knows some speakers for whom representations definitely exist in the training set and some for whom they definitely don't. We test the possibility of membership identification using four steps:
  
  \begin{enumerate}
  \itemsep-0.2em 
    \item We simulate the above using cross-validation. Given five speaker independent folds, we use three for the training set. From the remaining two folds, we add some samples of the selected speakers to the training set. 
    \item We consider that the attacker knows both, the speakers selected and not selected for training from set four $(s4)$, but has no information about this split for set five $(s5)$. The objective of the attacker is to predict whether speakers were selected for inclusion in the training set from $(s5)$.
    \item The attacker trains a binary classification model comprised of dense layers ($M_{att-mi}$) using the representations obtained from dataset D1 as $M_{att-mi}(h(x_{D1}),`Yes')$. It obtains representations of the samples not used in training for the the selected speakers included in the training set and trains its model as $M_{att-mi}(h(x_{s4_selected}),`Yes')$ and for the speakers not included in training as $M_{att-mi}(h(x_{s4_selected}),`No')$. A speaker is saved from each label for validation.
    \item We then define the UAR of the performance of $M_{att-mi}(h(x_{s5}))$ as membership identification.
  \end{enumerate}

\begin{landscape}
  
\begin{table*}[t]
\small
\caption{\centering Results using general (left) and models trained to not encode sensitive information (right) for activation and valence prediction. U-UAR, U(M/F)-UAR for male/female, L-leakage, SIR-sensitive information reduction metric, MI-membership identification.
\textit{\textbf{Bold-Italic}} shows significant improvement in metrics as compared to general classification model and \textit{Italic} shows significant difference in metrics as compared to the model trained to not encode sensitive information. Significance is established using paired t-test at adjusted p-value$<0.05$.}
\centering
\begin{subtable}[t]{\textwidth}
\centering
\subcaption{Prediction of activation using general (left) and models trained to not encode sensitive information (right)}
\begin{tabular}{ll|llllll|llllllll|}
\hline
                            &&  \multicolumn{6}{c|}{General Classification} & \multicolumn{5}{c}{Privacy Preserving Classification} \\ \hline
\multicolumn{2}{l|}{}                  & L     & U(M)     & U(F)        & U        & SIR & MI & U(M)     & U(F)        & U        & SIR & MI \\ \hline
\multirow{4}{*}{Audio}         & Imp    &   0.69 &   0.65   &  \textit{0.62}    & \textit{0.63}       & 0.35  &  0.71 & 0.64   &   0.57   &    0.60   &    \textbf{\textit{0.44}}   &   0.68 \\
                               & Pod  &  0.71  &   0.69   &   0.70  &  0.70       &  0.32 &  0.73  &     0.68   &   0.69   &    0.69   &  \textbf{\textit{0.44}} &   \textbf{\textit{0.68}} \\
                               & Iem  &   0.73 &   0.66   &  0.69    &  0.67       &  0.30 &  0.72  &    \textbf{\textit{0.68}}     &     0.70     &   \textbf{\textit{0.69}} &     \textbf{\textit{0.43}} &    \textbf{\textit{0.67}}\\
                               & MuS  &  0.72  &    \textit{0.61}  &  \textit{0.64}    &  \textit{0.63}        &  0.33 &   0.75 &     0.58   &   0.61   &    0.60   &   \textbf{\textit{0.45}}&     \textbf{\textit{0.69}} \\ \hline
\multirow{2}{*}{Lexical}          & Iem   &  0.62 &  0.51    &  0.52    &  0.52        &  0.39 & 0.59 &  \textbf{\textit{0.55}} & \textbf{\textit{0.56}} &\textbf{\textit{0.56}}&\textbf{\textit{0.48}}&\textbf{\textit{0.55}}   \\
                               & MuS  &    0.64  &   0.54   & 0.56     &  0.55      &  0.38 &  0.60&  \textbf{\textit{0.58}} &0.57   &   \textbf{\textit{0.58}} &  \textbf{\textit{0.47}}   &   0.58  \\
                               \hline
\multirow{2}{*}{Multimodal}    & Iem   &    0.74 &   0.66   & 0.70     &  0.68      &  0.30 &  0.74 & 0.66 & 0.69 & 0.68 & \textbf{\textit{0.41}} & \textbf{\textit{0.67}}  \\
                               & MuS   &   0.73 &  0.65    &   0.66   &  0.66       &  0.31 &  0.76  & 0.65 & 0.64  & 0.65 & \textbf{\textit{0.43}} & \textbf{\textit{0.69}} \\ \hline
\end{tabular}
\label{normal-results-act}
\label{adv-results-act}
\end{subtable}

\begin{subtable}[t]{\textwidth}
\centering

\subcaption{Prediction of valence using general (left) and models trained to not encode sensitive information (right)}
\begin{tabular}{ll|llllll|llllllll|}
\hline
                            &&  \multicolumn{6}{c|}{General Classification} & \multicolumn{5}{c}{Privacy Preserving Classification} \\ \hline
\multicolumn{2}{l|}{}                  & L     & U(M)     & U(F)        & U        & SIR & MI & U(M)     & U(F)        & U        & SIR & MI \\ \hline
\multirow{4}{*}{Audio}         & Imp   &   0.56 &  0.53    &  0.49    &   \textit{0.51}      &  0.44 &  0.70  & 0.51 & 0.49 & 0.48 & \textbf{\textit{0.48}} & 0.68\\
                               & Pod   &   0.60 &  0.56    &  0.57    &  0.56       &  0.42 &  0.71 & 0.55 & 0.56 & 0.56 & \textbf{\textit{0.47}} & 0.70 \\
                               & Iem  & 0.62 &   0.60   &  0.61    &   0.60         &  0.39 &  0.70 & 0.60 & 0.62 & 0.61 &\textbf{\textit{0.45}} & 0.68 \\
                               & MuS   &  0.58 &   0.50   &  0.47    &  0.48        & 0.42  &  0.72 & 0.48 & 0.47 & \textbf{\textit{0.47}} & 0.46 &0.71 \\ \hline
\multirow{2}{*}{Lexical}          & Iem    &   0.61 &  0.64    &  0.65    &  0.65      & 0.41  &  0.62 & \textbf{\textit{0.67}} & \textbf{\textit{0.68}} &\textbf{\textit{0.67}} & \textbf{\textit{0.46}} & 0.62 \\
                               & MuS  &    0.57  &   0.68   &  0.69    &  0.68      &  0.45 &  0.63 & 0.70 & 0.71 & 0.70 & 0.47 &0.62 \\ \hline
\multirow{2}{*}{Multimodal}    & Iem   &   0.68 &  0.67    &  0.71    &  0.69       &  0.32 &  0.70 & 0.68 & 0.70 & 0.69 & \textbf{\textit{0.45}} & \textbf{\textit{0.68}} \\
                               & MuS   &   0.64  &  \textit{0.67}    &   0.66   &  \textit{0.67}      & 0.38  &   0.71 & 0.64 & 0.65 & 0.65 & \textbf{\textit{0.46}} & 0.71\\ \hline
\end{tabular}
\label{normal-results-val}
\label{adv-results-val}
\end{subtable}
\end{table*}
\end{landscape}

\section{Analysis}
In all the tables, \textbf{U} is the unweighted average recall (UAR), and \textbf{U(M)} and \textbf{U(F)} represent the performance of the model for emotion recognition when gender is male and female respectively. Leakage in the model is represented by \textbf{L}, the lower the better, where chance leakage is 0.5. sensitive information reduction metric, represented by \textbf{P}, ranges from $[0,0.5]$, and is the incapability of an attacker to obtain demographic information from the representation, the higher the better.
Membership identification represented by \textbf{MI}, ranges from $[0.5$ (chance UAR),$1]$, and is the capability of an attacker to identify if the subject belongs in the training set, for which the lower the value, the better. We code identify the datasets as follows: \textbf{Imp}-MSP-Improv; \textbf{Pod}-MSP-Podcast; \textbf{Iem}-IEMOCAP; and \textbf{MuS}-MuSE. All significance tests are paired t-test, with significance established (shown in bold) when \textbf{Benjamini-Hochberg adjusted} (FDR $= 5\%$) $p$-value$<0.05$.

\subsection{RQ1: Unimodal vs Multimodal Leakage}
\label{subsec:q1}
\noindent\textsc{Q:} \textit{Does demographic leakage differ in umimodal and multimodal emotion recognition models?}\\
\textsc{Hypothesis}: \textit{Multimodal representations leak more gender information than unimodal representations.}\\
Previous research has shown that different modalities have varying capabilities of capturing demographic information, such as age or gender~\cite{levitan2016automatic} information.The authors showed that audio, as compared to lexical, is used more successfully to predict gender. Hence, we hypothesize that a combination of these modalities leads to an increase in the leakage of the sensitive variable. 

We train the six setups separately for each dataset described in Section~\ref{sec:setup_arch_c8} for activation and valence. We report the average across five-fold speaker-independent cross-validation in Table~\ref{normal-results-act} and Table~\ref{normal-results-val}. We find that:
\begin{itemize}
\itemsep-0.2em 
    \item A network trained to only recognize emotion is generally discriminative for gender as well. For instance we obtain a leakage of 0.73 when training a multi-modal network for activation and of 0.64 when trained for valence on MuSE.
    \item In unimodal systems, leakage is higher when systems are trained using only audio streams compared to lexical.
    \item Leakage of gender in learned representation is higher for multimodal systems than that for the unimodal systems for both, MuSE and IEMOCAP (the two datasets with both audio and lexical information).
\end{itemize}

Our results suggest that models that aren't explicitly trained for gender recognition, or, that don't use gender as an input feature, still learn representations that are discriminative to identify gender. This leakage is more prominent when the input stream is audio as compared to lexical, but the leakage compounds
in multimodal systems.

\subsection{RQ2: Privacy Preservation Performance}
\label{subsec:q2}
\noindent\textsc{Q:} \textit{How does the sensitive information reduction metric change when a network is trained to not encode sensitive information?}\\
\textsc{Hypothesis}: \textit{Representations that are gender-invariant are less prone to leakage when attacked by an adversary, leading to reduced leakage of sensistive information.}\\
Previous research has shown that obtaining a representation from a model trained to be invariant to gender, age, or location leads to better protection from an attacker who tries to recover this information~\cite{coavoux2018privacy}. Previous research~\cite{elazar2018adversarial} has also shown that while the representations might be trained such that leakage of sensitive variable is reduced to chance, the attacker might still be able to recover this information. Hence, we concentrate on using this incapability as our primary metric.
To test our hypothesis, we train the adversarial variants of the six models as mentioned above, while making sure that the leakage in the models is reduced to chance performance and compare our results to those in Table~\ref{normal-results-act} and Table~\ref{normal-results-val}. We train the multimodal models only for MuSE and IEMOCAP. 
Our results in Table~\ref{adv-results-act} and Table~\ref{adv-results-val} show that:
\begin{itemize}
\itemsep-0.2em 
    \item The sensitive information reduction metric is always higher when the representations are trained adversarially, compared to generally.
    \item Even when leakage is adversarially reduced to chance, the attacker is still able to recover information about gender.
    \item The sensitive information reduction metric is in general always lower for audio than for lexical based unimodal systems.
    \item Multimodal systems often have the lowest sensitive information reduction metric.
\end{itemize}
Our results suggest that, though the sensitive information encoded in the learned representation is reduced using the proposed method, the attacker can still recover that information. This effect is especially compounded for multimodal systems. While previous work has concentrated on text (Section ~\ref{chap:relworkmodel}), our work shows how audio is the major culprit and that models involving audio as input are easier to exploit, even when trained adversarially for not encoding sensitive information.

\subsection{RQ3: Privacy Preserving Emotion Recognition Performance}
\label{subsec:q3}
\noindent\textsc{Q:} \textit{How does emotion recognition performance change when networks are trained to not encode sensitive information?} \\
\textsc{Hypothesis}: \textit{There is a minor drop in emotion recognition performance when models are trained to not encode sensitive information.}\\
Previous research has shown that training a model invariant to a dataset variable might lead to drop in performance on the primary task, especially when there exists known correlations or biases in the datasets between the target label for the primary task and the secondary task~\cite{meng2018speaker} 

We compare the performance for predicting activation and valence of the models trained just to predict emotion (Act: Table~\ref{normal-results-act}, Val: Table~\ref{normal-results-val}) versus the model trained to not encode sensitive information while still predicting emotion in (Act: Table~\ref{adv-results-act}, Val: Table~\ref{adv-results-val}).
Our results suggest that, in general there is no significant effect on the performance on the primary task when we train networks to not encode sensitive information. We find that the performance is either maintained, e.g., Act: multimodal-MuSE; Val: multimodal-IEMOCAP, or there is a slight decrease in performance for some setups, e.g., Act: unimodal-acoustic-MuSE; Val: multimodal-MuSE. In multiple cases, such as Act/Val:unimodal-lexical-MuSE/IEMOCAP, contrary to some previous work, we also see a significant increase in performance, implying that making the model invariant to gender increases its robustness by not learning replicable associations between gender and emotion label. 

\subsection{RQ4: Privacy Preservation Strength}
\noindent\textsc{Q}: \textit{How does the  adversarial component's strength impact emotion recognition performance and the sensitive information reduction metric?}\\
\textsc{Hypothesis}: \textit{As the strength of the adversarial component increase, the sensitive information reduction metric increases and the performance on the pimary task is unchanged.}\\
Our results in Section~\ref{subsec:q2} suggest that while the leakage of the model was reduced to chance performance, the attacker is still capable of recovering this information. We analyze the effect of the strength of the adversarial component on the performance of the primary task and the sensitive information reduction metric. 

We find that the emotion recognition performance is generally unaffected with a change in $\lambda$, as expected from the results in Section~\ref{subsec:q3}. 
We observe that the the attacker is usually less capable of inferring gender from learned representations when $\lambda=0.50$ as compared to when $\lambda=0.75$. For example, the sensitive information reduction metric for the unimodal audio system trained on MuSE increases from $0.39$ to $0.45$. But contrary to our expectation, we often see a decrease in the sensitive information reduction metric when we move from $\lambda=0.75$ to $\lambda=1.00$ for both activation and valence. For example, the sensitive information reduction metric for the unimodal audio system trained on MuSE decreases from $0.45$ to $0.41$. 
 The decrease in the sensitive information reduction metric as $\lambda\rightarrow1$ could be attributed to overfitting of data~\cite{schmidt2018adversarially} when being trained for invariance to the sensitive variable which the attacker network is able exploit.
This suggests that an increase in the strength of the adversarial component doesn't necessarily correlate to an increase in the sensitive information reduction metric.

\subsection{RQ5: Privacy Preserving Emotion Recognition Performance Change \& Gender}
\noindent\textsc{Q:} \textit{Focusing on gender, how does the performance of emotion recognition change when a network is trained to not encode sensitive information?}\\
\textsc{Hypothesis:} \textit{Learning representations invariant to gender will affect performance on the primary task in an imbalanced manner across subgroups.}\\
Previous research~\cite{bagdasaryan2019differential} has shown that training models invariant to race or gender can harm performance for one group more than others. This may be worrying when the prediction is used for sensitive application such as intervention or policing. Hence, we analyze if the performance on emotion recognition is affected in an imbalanced way for the models trained to not encode sensitive information. 

We compare the performance for predicting activation and valence of the models trained just to predict emotion (Act: Table~\ref{normal-results-act}, Val: Table~\ref{normal-results-val}) versus the model trained to not encode sensitive information while still predicting emotion in (Act: Table~\ref{adv-results-act}, Val: Table~\ref{adv-results-val}). We find that while the performance is affected differently for the subgroups, the effect is not consistent across multiple setups and datasets. For example, the unimodal-acoustic system trained on MSP-Improv for activation classification decreases in performance for both the male and female groups, but the effect on the female group is greater. But the pattern isn't consistent across other datasets for the same model setup. Our takeaway from this analysis is cautionary, that though the sensitive information reduction metric increases when a model is adversarially trained to not encode sensitive information, we need to ensure that the performance of the model on that dataset doesn't harm one subgroup more than the other.

\subsection{RQ6: Location of Adversarial Component}
\noindent\textsc{Q:} \textit{Does the location of the adversarial component within a network  affect the sensitive information reduction metric and emotion recognition performance?}\\
\textsc{Hypothesis:} \textit{Unlearning the demographic variable in separate pooled streams will improve the sensitive information reduction metric.}\\
Previous work has shown that curtailing a variable on intermediate layers often leads to a difference in the performance of the classifier~\cite{chabanne2017privacy}. As seen in Section~\ref{subsec:q1}, audio is more prone to leakage than lexical information, hence, a multimodal system's sensitive information reduction metric might benefit from curtailing audio separately. Our initial multimodal model (Fig~\ref{fig:model_chap8}) only allows for the same strength and parameters of the adversarial component to be applied for both audio and lexical streams. 
To test our hypothesis, we place the same adversarial component after the mean pooling layer of both input streams, allowing us separate control of gender invariance for both modalities, before concatenation of representation.

We show our results in Table~\ref{adv-double-act-val}.
We find that, using adversarial component separately for each input stream improves sensitive information reduction metric for emotion recognition models trained on both datasets, as compared to using one adversarial component. This suggests that a granular control of invariance over modalities leads to better defense of representations against gender identification.

\begin{table}[t]
\centering
\caption{Results for activation (Act) and valence (Val) prediction using multimodal input, when adversarially unlearning gender in each input (SIR-E) [left] stream separately. U-UAR, U(M/F)-UAR for male/female, P-sensitive information reduction metric, MI-membership identification.
\textbf{\textit{ Bold-Italic}} shows significant improvement in the sensitive information reduction metric as compared to model trained to not encode sensitive information by maximizing loss on the concatenated representation (SIR-C) [right]. Significance is established using paired t-test, adjusted p-value$<0.05$.
}
\begin{tabular}{llllllll}
\hline
                           &    & \multicolumn{3}{c}{SIR-E} & \multicolumn{3}{c}{SIR-C} \\ \cline{3-8} 
                            &    & U          & P         & MI     & U       & P       & MI       \\ \hline
                            
\multirow{2}{*}{Act} & Iem &    0.66    &   \textit{\textbf{0.43}}   &    0.73   &    0.68     &     0.41    &   0.67      \\
                            & MuS &    0.65    &    0.44   &   0.74    &    0.65     &     0.43    &   0.69      \\ \hline
\multirow{2}{*}{Val} & Iem &    0.67   &    \textit{\textbf{0.46}}   &    0.70   &    0.69     &     0.45    &   0.68      \\
                            & MuS &    0.66    &    0.47   &   0.74    &    0.65     &     0.46    &   0.71      \\ \hline
\end{tabular}
\label{adv-double-act-val}
\end{table}

\subsection{RQ7: Defending Set-Based Membership Identification}

\begin{table*}[t]
\centering
\caption{\centering Results for activation and valence prediction, for general classification (General), and, when adversarially unlearning subject identity (SIR-SubjectID) and both subject identity and gender (SIR-Multiple). U-UAR, U(M/F)-UAR for male/female, SIR-sensitive information reduction metric, MI-membership identification.
\textit{\textbf{Bold-Italic}} shows significant improvement in metrics as compared to general classification model and \textit{Italic} shows significant difference in metrics as compared to the models trained to not encode sensitive information. Significance is established using paired t-test at adjusted p-value$<0.05$.
}
\begin{tabular}{lllllllllll}
\hline
                           &    & \multicolumn{3}{c}{General} &\multicolumn{3}{c}{SIR-SpeakerID} & \multicolumn{3}{c}{SIR-Multiple}  \\ \cline{3-11}  
\hline

                            &    & U          & SIR         & MI     & U       & SIR       & MI  & U       & SIR       & MI     \\ \hline
                            &    & \multicolumn{6}{c}{Activation}\\ \hline
\multirow{4}{*}{Audio}         & Imp   &\textit{0.63} &0.35 &0.71 & 0.59 & \textbf{\textit{0.40}} & \textbf{\textit{0.58}} &    0.59   &  \textbf{\textit{0.45}}   &  \textbf{\textit{0.58}} 
 \\
                               & Pod   & 0.70 & 0.32 & 0.73 & 0.67 & \textbf{\textit{0.37}} & \textbf{\textit{0.60}} &   0.69    & \textit{\textbf{0.46}}     &  \textbf{\textit{0.59}}      \\
                               & Iem  & 0.67 & 0.30 & 0.72 & 0.66 & \textbf{\textit{0.35}} & \textbf{\textit{0.58}} &  0.67    &  \textbf{\textit{0.43}}    &  \textbf{\textit{0.57}}     \\
                               & MuS   & \textit{0.63} & 0.33 & 0.75 & 0.61 & \textbf{\textit{0.36}} & \textbf{\textit{0.62}} &   0.59      &   \textbf{\textit{0.44}}   &   \textbf{\textit{0.60}}     \\ \hline
\multirow{2}{*}{Lexical}          & Iem   & 0.52 & 0.39 & 0.59 & 0.51 & 0.40 & 0.52 &    0.53  &   \textbf{\textit{0.48}}   &   \textbf{\textit{0.52}}    \\
                               & MuS & 0.55 & 0.38 & 0.60 & 0.52 & 0.39 & 0.53 &   0.54      &    \textbf{\textit{0.47}}  &  \textbf{\textit{0.52}}     \\ \hline
\multirow{2}{*}{Multimodal}    & Iem & 0.68 & 0.30 & 0.74 & 0.67 & 0.33 & \textbf{\textit{0.58}} &   0.66    & \textbf{\textit{0.40}} &   \textbf{\textit{0.57}}    \\
                               & MuS & 0.66  & 0.31 & 0.76 & 0.65 & 0.33 &\textbf{\textit{0.60}} &   0.65   &   \textbf{\textit{0.40}}   &  \textbf{\textit{0.58}}    \\ \hline
                                &    & \multicolumn{6}{c}{Valence}\\ \hline
\multirow{4}{*}{Audio}         & Imp & \textit{0.51}  & 0.44 & 0.70 & 0.47 & 0.45 & \textbf{\textit{0.54}} &
0.47    & 0.48  &  \textbf{\textit{0.53}}  \\
                               & Pod & 0.56  & 0.42 & 0.71 &  0.55 & 0.43 & \textbf{\textit{0.56}} &    0.54   & \textbf{\textit{0.48}}  & \textbf{\textit{0.56}} \\
                               & Iem & 0.60 & 0.39 & 0.70 & 0.61 & 0.41 & \textbf{\textit{0.59}} &  0.60    &  \textbf{\textit{0.47}} &   \textbf{\textit{0.57}} \\
                               & MuS & \textit{0.48} & 0.42 & 0.72 & 0.45 & 0.42 & \textbf{\textit{0.60}} & 0.46    & \textbf{\textit{0.46}}   &  \textbf{\textit{0.58}}  \\ \hline
\multirow{2}{*}{Lexical}          & Iem  & 0.65 & 0.41  & 0.62 & 0.67 & 0.41 & \textbf{\textit{0.52}} &  0.66    & \textbf{\textit{0.47}}  &  \textbf{\textit{0.53}}  \\
                               & MuS  & 0.68 & 0.45 & 0.63 &  0.68 & 0.44 & \textbf{\textit{0.53}} &   0.68   & 0.46   &  \textbf{\textit{0.53}}  \\ \hline
\multirow{2}{*}{Multimodal}    & Iem   &  0.69 & 0.32 & 0.70 & 0.69 & 0.34 & \textbf{\textit{0.57}} &   0.65   & \textbf{\textit{0.43}}  &  \textbf{\textit{0.56}}  \\
                               & MuS    & \textit{0.67} & 0.38 & 0.71 & 0.64  & 0.37 & \textbf{\textit{0.58}} & 0.62    & \textbf{\textit{0.44}}  &   \textbf{\textit{0.58}} \\ \hline                               
\end{tabular}
\label{adv-id-act-val}
\label{adv-multi-act-val}
\end{table*}
\noindent\textsc{Q:} \textit{Does the sensitive information reduction paradigm that we used to reduce encoding of gender information also help defend against other attacks such as membership identification?}\\
\textsc{Hypothesis:} \textit{Membership identification will decrease when models are trained to be invariant to speaker.}\\
Membership identification is defined as an attack that tries to identify if samples from a speaker `x' were present in the training set~\cite{li2013membership}. ~\cite{papernot2018marauder} showed that removing identifying factors from learned representations reduces the probability of membership leakage. 
For this analysis, we ask two questions: 
(a) can we defend against membership identification using a proxy task and,
(b) can we defend against both, gender and membership identification?

We train an attack model for membership identification as specified in Section~\ref{subsec:metrics}. We find that while adversarial removal of gender in the learned representation (Act: Table~\ref{adv-results-act} and Val: Table~\ref{adv-results-val}) does lead to reduced membership identification, as compared to a model trained solely for emotion recognition (Act: Table~\ref{normal-results-act} and Val: Table~\ref{normal-results-val}), the membership identification is still far higher than chance.

Our goal is to be unable to identify whether samples from speaker `x' exist in the training set.
This is different from the usual membership defense that prevents prediction of presence of a data-point pair $(input_x,output_x)$ is in the training set. As a result, we require a proxy task, 
because our model cannot use samples from the speakers not in the training set even to induce invariance. We hypothesize that given randomly chosen speakers from the population, speaker-invariant training leads to representations that are less likely to encode speaker-specific information. This will make it harder for the attacker to identify membership of a particular speaker in the training set. 
We train the emotion recognition models specified in Section~\ref{sec:setup_arch_c8} and replace the gender invariance sub-network with speaker invariance and use the same membership attack network. 


We show our results in Table~\ref{adv-id-act-val}.
We find that models trained to be invariant to speaker identity have significantly lower UAR for membership identification than those trained solely to recognize emotion, or trained invariant to gender, which matches our hypothesis.

\textbf{Extension towards multi-attribute invariance.}
We train our emotion recognition model using both the adversarial components (speaker id and gender) and the primary classification task i.e., emotion recognition. This ensures that the model can defend against both, gender and membership identification attacks. We report our results in Table~\ref{adv-multi-act-val}. We find that we can successfully train models that are safer against both, gender and membership identification attacks, while still maintaining similar performance on the primary task, as an evidence towards multi-attribute invariance.


\section{Conclusion}
In this work, we show how sensitive information preserving networks trained for emotion recognition can be used to protect against gender and membership identification. 
This provides a compelling case for separating the process of data processing on user devices and of task-specific training on central servers. 
While in this chapter we concentrate on a single primary task i.e., emotion recognition, this method can be extended to maximize utility on multiple primary tasks that are loosely related to each other and are benefited from a multi-task setup as shown for dialogue act and turn detection, and sentiment and topic classification~\cite{ruder2017overview}.



%% file: Chapters/chap9.tex
\section{Motivation and Contribution}
Large language models have limitations in subjective tasks like emotion recognition, partly due to insufficient annotation diversity and inadequate data coverage. A diverse range of annotations is crucial for capturing the complex expressions of human emotions, while extensive data coverage renders it more likely that the many possible variations of these expressions are adequately represented. However, obtaining a comprehensive and diverse set of annotations in emotion datasets is both time-consuming and expensive, involving the recruitment of numerous human annotators to provide their unique perspectives on emotional expressions.

These annotation costs are further compounded by the need to evaluate models against diverse emotions, which often require not only human annotation for the collected data but also for the generated emotion label on unseen data for validation purposes. Current strategies rely heavily on costly, time-consuming human-based feedback for model evaluation, which, while effective in their own right, may not always be economically viable or efficient.

To address these challenges, we propose using metrics that are based off sociological literature to capture the nuances of human behavior and emotions in a more cost-effective manner. We propose model-agnostic evaluation templates can be instantiated with sociological word lists, and these templates can be generalized to any task with existing word lists. We introduce two metrics: one for emotion generalization and another for intentional reduction of learnt sensitive information. These metrics aim to comprehensively assess model performance, while reducing dependency on expensive human-based feedback. By effectively evaluating the robustness and sensitive variable leakage of models, the proposed metrics provide valuable insights into model performance, resulting in more efficient utilization of resources.

Evaluating their effectiveness in enhancing model performance and reducing leakage of sensitive demographic, we find that these metrics are significantly correlated with model performance and can improve cross-corpus results. The proposed method empowers us to leverage the benefits of diverse and comprehensive emotion recognition models, while mitigating the costs associated with the traditional, labor-intensive feedback processes. Ultimately, this approach leads to the development of more accurate and relevant emotion recognition models in a cost-effective manner.

\section{Introduction}
Recent advances in natural language processing have led to the development of large and very large language models that can be trained on massive amounts of labeled data. However, despite their impressive performance on various natural language processing tasks, these models do not necessarily learn behavior that is similar to that of humans.
Data diversity is particularly important in subjective, paralinguistic tasks, such as those seen in emotion recognition and other behavior modeling areas.
But paralinguistic datasets are often relatively small, compared to the size of datasets seen in fields such as speech recognition. Further, it can be difficult to identify appropriate metrics for these subjective tasks and, as a result, those common in emotion recognition tend to be metrics such as cross-entropy loss between the predicted and original labels.

The challenge with the common metrics is that they do not capture the full complexity of the problem, leading to models that have limitations at their core.  Models trained to optimize for cross-entropy are often suffused with biases.
Reducing learnt sensitive information or generalizing these models can be extremely difficult because sensitive information reduction processes generally rely on these same data and labels.  Therefore, limitations in the original dataset are still present when the model attempts to unlearn these problematic associations~\cite{Schick2021SelfDiagnosisAS, GarridoMuoz2021ASO}.

Researchers have looked at countering these issues in several different ways: from data generation, to data augmentation, and, from annotation diversification to comprehensive data collection~\cite{chatziagapi2019data}. Each of these methods comes with its own pros and cons. For example, generating data samples that are not perceived differently by humans is not only challenging, but also, requires extensive post-hoc evaluations~\cite{Jaiswal2021BestPF}. Researchers have also looked at diverse data collection methods, focusing on active learning optimizing for data diversity~\cite{ren2021survey}. While this method generally improves performance by a significant measure, we can still end up with datasets that reinforce unwanted stereotypes and prejudices~\cite{chen2021understanding}.

This has led to a growing need for feedback that is based on human evaluation to ensure that the models produce more accurate and relevant results.
One of the main challenges of using human-based feedback though is that it can be expensive to obtain, particularly for tasks that involve emotions. This is because emotions are subjective and can vary widely depending on demographic and situational factors. As a result, there is a need for alternative methods that can provide metrics for evaluating the performance of language models that are centered around human behavior.

To address this issue, we propose leveraging metrics that are based on sociological literature, which can capture the nuances of human behavior and emotions. By drawing on insights from sociology, we can develop more nuanced and context-sensitive metrics that can better capture the complexities of human behavior. This approach has the potential to provide a more comprehensive optimization of language models, while also reducing the dependence on expensive and time-consuming human-based feedback.

\begin{figure*}[t]
\centering
\includegraphics[width=\textwidth]{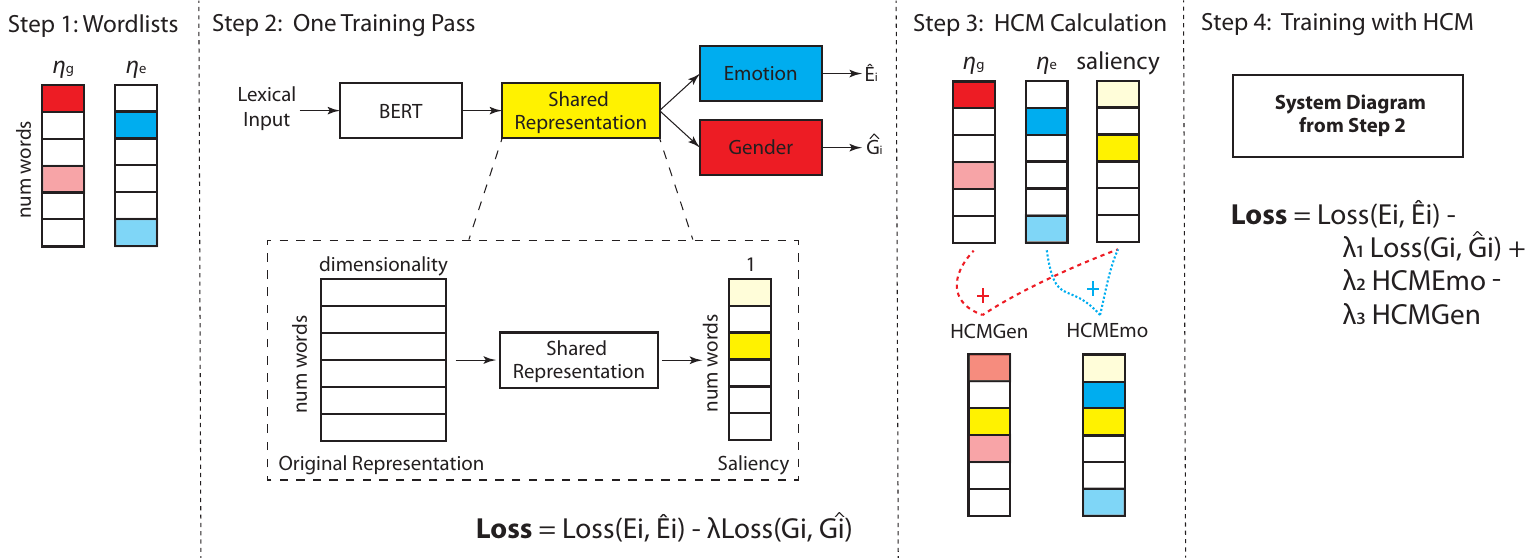}
\caption{Diagram of the approach. In step 1, the wordlists are created (e.g., for gender, $w_g$, and for emotion, $w_e$.  The length is the number of words.  The brightness corresponds to the strength of the relationship between the word and the category (gender or emotion).  In step 2, the model learns a shared representation with the goal of learning emotion, $\hat{E}_i$, and unlearning gender, $\hat{G}_i$.  The shared representation is used to create a word-level saliency vector.  In step 3, the HCM metrics are calculated by combining the salience with either $w_g$ or $w_e$.  In step 4, the model is trained with a loss that includes the original target (maximizing emotion accuracy and minimizing gender accuracy) in addition to the two HCM metrics. }
\label{img:model}
\end{figure*}

\section{Intuition Behind Human-Centered Metric Design}
In this chapter, we propose a novel approach to developing task-invariant metric templates that are centered around human perception. To begin with, we define metric blueprints, called as metric templates. These templated metrics that have generic components that can be adapted for specific tasks. Our aim is to leverage these templates to evaluate a model, and to add a training objective that relies on sociological literature in addition to human-obtained labels. This is a significant departure from the traditional approach of using labeled data, which can be costly and time-consuming to obtain.

Ideally, important words identified by models should align well with human's interpretations. To achieve this, we use model interpretability; any method that provides us some intuition as to the words that the model pays attention to for its prediction to define the metric templates. This allows us to have a training objective that optimizes for importance assigned to certain words based on sociolinguistic literature. This approach is different from using labels because it enables us to incorporate sociolinguistic knowledge into the metric templates.

The metric templates we propose consist of three components: directionality (sign), input interpretation saliency weights, and word lists. The combination of the three components results in a task-invariant metric template.  

The sign component specifies whether the metric is meant to be maximized or minimized. By setting the sign, we can control the direction of the correlational relationship between the input and output variables, and thus, improve the accuracy of our predictions.

The input interpretation saliency weights component specifies how the input data should be interpreted.  They are saliency values obtained from the model for each word token in the input. These values are highly correlated with the predicted label and can be positive, negative, or zero. A positive value indicates that the presence of a word correlates with the prediction of that label. A negative value indicates that the presence of that word correlates with “not predicting the given label.” A zero value means that the word has no influence on the prediction.

The wordlists component provides a list of words that should be given special consideration when computing the metric. 
These lists can be grouped by class labels and may have associated weights with each word, but both of these features are optional. Wordlists are an essential tool for training machine learning models, as they allow us to focus on specific words or concepts that are relevant to our task. By using wordlists, we can improve the accuracy and efficiency of our models, as well as gain valuable insights into the relationships between words and concepts in our data.

In summary, our metric template allows us to focus on three separate aspects, each aligned with one of the components.  Sign allows us to encourage the model to learn the correlation between a particular word and the output (sign of loss function).  The input interpretation saliency weights allow us to measure how well the model captures the correlation between all words in a sentence and the output.  The wordlists allow us to impact the learned correlations to align with our expectations about the relationshp between a given category and word choice.

We now ground these aspects, focusing on a given task-related wordlist focusing on the task of emotion recognition. For our purposes, we use the NRC-VAD~\cite{mohammad2018obtaining} and a self curated, Emotion-Gender wordlists.  We first calculate the overlap between the words in the input sentence and the task-related wordlist(s). This overlap allows us to identify words whose correlations (learned by a model) may need to change. For class-based wordlists, we use the signed correlation that corresponds to the predicted class label, which is controlled by the sign component of our metric template. By setting the sign, we can encourage the model to learn the correlation between a particular word and the output. In cases where we want to unlearn a task completely, we aim to move the correlation of those words to zero to remove any influence on the model output. The input interpretation saliency weights component of our metric template specifies how well the model captures the correlation between all words in a sentence and the output. By leveraging the overlap between task-related words and the input sentence, we can effectively optimize the correlations and improve the model’s performance. Our proposed metric is designed to help models learn the correlations that are relevant to the task at hand while minimizing the impact of irrelevant words. By using the metric template, we can focus on three separate aspects, each aligned with one of the components: sign, iiw, and wordlists, allowing us to impact the learned correlations to align with our expectations about the relationship between a given category and word choice.

The focus of this chapter is on the extensive testing of a metric template designed for the task of emotion recognition. We had the option to choose between task coverage or usability. Task coverage involves testing multiple tasks on a single state-of-the-art (SoTA) dataset and model, while usability involves extensively evaluating the proposed metric for one subjective task of emotion, but testing it on multiple datasets and different model combinations. We chose the latter approach to validate the usability of the metric rather than the task coverage.
Our approach involves selecting one task for learning, which is emotion, and one subtask for unlearning, which is demographics. Through this chapter, we iterate over various combinations of this setup, thoroughly validating the utility of the metric for both cases, including emotion classification improvement and reducing encoded sensitive information in emotion classification models, while varying the datasets and other factors. Our results demonstrate the effectiveness of the proposed metric in improving emotion classification and producing models with reduced sensitive information leakage.

To ensure that our proposed metric aligns with annotator choices, we conduct a crowdsourcing survey. The survey results indicate that our metric aligns well with annotator choices, thereby validating the effectiveness of our approach. The metric can be similarly applied to any other human-centered subjective tasks, such as, deception detection, opinion analysis etc.


\subsection{Requirements}
The proposed templates has three inputs: input interpretation saliency weights, direction of learning (sign) and relevant wordlists. 

\subsubsection{Input Interpretation Saliency Weights}
\label{sec:integratedgradients}
The goal of input interpretation saliency weights is to understand what a trained black-box neural network is doing. This is achieved by examining the learned correlations from the model, which are obtained through a post-hoc method. By analyzing these correlations, we can gain insights into how the model is making its predictions and identify any potential leakage of sensitive information or limitations.

We calculate the reliance of a model on any given word in the input using the Captum interpretability library for PyTorch~\cite{kokhlikyan2020captum}. Captum provides state-of-the-art algorithms to identify how input features contribute to a model’s output. 

We use the attribution algorithms implemented via integrated gradients~\cite{sundararajan2017axiomatic}. Integrated gradients represent the integral of gradients with respect to inputs along the path from a given baseline (absence of the cause) to input sample. The output is a set of words for each model instance that contribute towards the prediction along with their attribution weights.

The sign of the attribution weight shows the direction of the correlation, and the value is the strength of the correlation. Words that do not influence prediction should have zero correlation. 
	
\subsubsection{Wordlists}
\label{sec:wordlists}
A wordlist here refers to linguistic indicators extracted from the literature for a given task.  The lists can either have singular words or phrases. A wordlist has three main properties: the variable it is created for, existence of label associations, and associated weights or importances.  For example, if we want to measure `emotion' then words such as `excited', or `lonely' are strong linguistic indicators, the former for positive emotion, the latter for negative emotion.
	
Similarly, if our measurement variable is `age', then phrases such as `cancel culture' or `groovy' have been found influencing perception of age.  If the aim is to evaluate `unlearning' of information in a given model, we do not necessarily need to have a pairwise list (e.g., words related to females or males).  Instead, a combined list will still allow us to mitigate the possibility of encoding sensitive variables in the learned representations. A wordlist can also have weights associated with a particular entry that correlate to strength of distinguishing ability of perception. For example, a wordlist associated with `emotion' can have a weight of 4 for `amazing' vs. 2 for `good', the former being perceived as a stronger positive emotion as compared to the latter.

\subsection{Terminology}
\label{exp_setuo_eg}
First, let's consider a model trained for for a binary classification task ($[T]$), using dataset ($D$) of size $|D|$ with two output labels ($L$), A and B ($\{A, B\}$), i.e., ($[T] \rightarrow L_{\{A, B\}}$).  We refer to samples in the dataset as $s$ and individual words as $x$, and if a word occurs in both as, $x \in s$.

We create a wordlist ($\eta$) that has human-centered linguistic indicators corresponding to $[T]$.  Depending on the nature of the task, the wordlist can either be combined, for both output labels, ($\eta_{L_{A+B}}$), or can have pairwise association with only one output label  (e.g., $\eta_{L_A}$ or $\eta_{L_B}$). While this does not necessarily need to be the case, combined wordlists can be used for reduced sensitive information leakage tasks ($\mathbf{HCM^{SIR}}$), i.e., where the aim is to minimize the presence of information relevant to the sensitive variable. On the other hand, pairwise associative wordlists can be used for generalizability ($\mathbf{HCM^{gz}}$) tasks, i.e., we need to learn and investigate associations between learned model correlations and the task outputs.   The wordlist can also have weights, $\omega$, associated with the individual entries (e.g., individual words, $\omega_x$). If the associated weights are absent, all entries are assigned a weight of 1.

During prediction, the model outputs a saliency value ($\lambda_{x}$) for each word.  We group words that have a positive ($\lambda_x > 0.05$), negative ($\lambda_x < -0.05$), or negligible ($|\lambda_x|\leq 0.05$) saliency attribution value.  The set of words in a sample with positive saliency will be noted $+_x = \{x | \lambda_x > 0.05, for x \in s\}$ and negative saliency $-_x = \{x | \lambda_x < -0.05, for x \in s\}$ \footnote{We do  not specifically call out words with negligible saliency.}.  

We describe how this information can be used at a sample-level ($\mathbf{HCM:s}$) or of the dataset ($\mathbf{HCM}$) as a whole to either promote generalizability or reduced sensitive variable information encoding:
\begin{enumerate}
    \item $\mathbf{HCM^{gz}Emo}$ is a metric calculated over a dataset that is focused on generalizability ($gz$) in the task of emotion recognition ($Emo$).  It can be evaluated in either a \textbf{standalone fashion} ($stn, \mathbf{HCM_{stn}^{gz}Emo}$), in terms of performance on a dataset.  It can also be calculated as the difference in performance between two models (relative improvement), ($ri, \mathbf{HCM_{ri}^{gz}Emo}$). 

    \item $\mathbf{HCM^{sir}Dem}$ is a metric calculated over a dataset that is focused on sensitive information reduction ($sir$) with respect to a demographic variable ($Dem$, e.g., gender).  It can also be evaluated in terms of either standalone performance or relative improvement.
\end{enumerate}

\subsection{Metric Types and Calculation}
The template enables two different types of model evaluations: (a) standalone (\emph{stn}), and, (b) relative improvement (\emph{ri}).

\subsubsection{Standalone ($\mathbf{HCM_{stn}}$)}  
We refer to standalone evaluation as benchmarking a model in isolation. We will first consider the case where we have combined wordlists (e.g., $\eta_{L_{A+B}}$).  This condition would be used when we \textbf{do not have the class labels for a specific task in our setup}, as in the case for the sensitive variable of age.  It allows us to identify words that are broadly associated with age in general, rather than a specific age grouping (e.g., older adults).

We define $\{ \lambda_x \}_{x\in s}$ as the set of saliency values for all words in a given sample $s$.  We then define an intersection set, which identifies words that occur both in the sample (${x\in s}$) and in a wordlist ($\eta_{L_{A+B}}$).  
The intersection set is defined as:

\begin{equation}
 int = \{x|x\in s\} \cap \eta_{L_{A+B}}
 \label{eq:standaloneintersectionset}
\end{equation}

We can then derive the standalone metric ($\mathbf{HCM_{stn}}$) from this intersection set for a given sample.  We consider each word ($x$) in the intersection set defined by Equation~\ref{eq:standaloneintersectionset}. We weigh the saliency attribution for that word ($\lambda_x$) by the wordlist weight ($\omega_x$) and sum this value over all words in intersection set for the sample.  We normalize by the number of words in the intersection set ($n_s$).  This gives us the sample-wise generalizability, $\mathbf{HCM_{stn}^{gz}[T]:s}$, for sample, $s$ (Equation~\ref{eq:standalonegen}).  

\begin{equation}
      \mathbf{HCM_{stn}^{gz}[T]:s} = \frac{1}{n}\sum_{x\in int}(|\omega_x| \times |\lambda_x|) 
      \label{eq:standalonegen}
\end{equation}

We calculate the generalizability metric for a dataset by averaging over the number of samples in the dataset  (Equation~\ref{eq:standalonegendataset}).

\begin{equation}
    \mathbf{HCM_{stn}^{gz}[T]} = \frac{1}{|D|} \sum_{s \in D} \mathbf{HCM_{stn}^{gz}[T]:s}
    \label{eq:standalonegendataset}
\end{equation}

We use $\mathbf{HCM_{stn}^{gz}[T]}$ to anticipate the generalizability of the model for a given task $[T]$.  Higher values of $\lambda_x$ for words with higher wordlist weights $\omega_x$ suggest that the model is relying on words that are known to be related to a given task (e.g., the word ``thrilled'' and positive valence).  

We can use the same metric to anticipate the degree to which a model reduces encoded sensitive information by replacing $|\lambda_x|$ with $1-|\lambda_x|$ (Equation~\ref{eq:standalonedebias}) for sample $s$.  Lower values of $\lambda_x$ (and therefore high weights of $1-|\lambda_x|$) for words with higher wordlist weights $\omega_x$ suggest that the model is not relying upon words that are known to be related to a demographic variable (e.g., the word ``she'' and the task of gender classification).

\begin{equation}
    \mathbf{HCM_{stn}^{sir}[T]:s} = \sum_x |\omega_x| \times (1-|\lambda_x|) \label{eq:standalonedebias}
\end{equation}

We again calculate the sensitive information reduction metric for a dataset by averaging over the number of samples in the dataset  (Equation~\ref{eq:standalonedebiasdataset}).

\begin{equation}
    \mathbf{HCM_{stn}^{sir}[T]} = \frac{1}{|D|} \sum_{s \in D} \mathbf{HCM_{stn}^{sir}[T]:s}
    \label{eq:standalonedebiasdataset}
\end{equation}

Next, we consider the case where we \textbf{do have labels for a given sensitive variable} (e.g., gender).  In this case, we calculate the value of the metric with respect to the pairwise association between the wordlist for a given label (e.g., $\eta_{L_A}$) and the model's output saliency attribution value for words within the sample ($\{ \lambda_x \}_{x\in s}$).  

We must consider four separate intersections, each of which evaluates the alignment between the directionality of the saliency attribution for a given word (i.e., positive vs. negative, $\lambda_{x,x\in s}$) and whether or not the word is in a given class $A$.  For simplicity, we will discuss the metric applied to a single class\footnote{The metric can be extended to multiclass problems.}, class $A$.

We first define four intersections, using the notation: ($int_i, i\in\{1,2,3,4\}$).  We will discuss sets of words with certain properties:  

\begin{enumerate}
    \item Words with positive saliency and in wordlist, $\eta_{L_A}$: \\
    $int_1 = +_x \cap \{x | x\in \eta_{L_A} \}$

    \item Words with negative saliency and \textit{not} in wordlist, $\eta_{L_A}$: \\
    $int_2 = -_x\cap \{x | x \notin \eta_{L_A} \}$
    
    \item Words with negative saliency and in wordlist, $\eta_{L_A}$:\\
    $int_3 = -_x \cap \{x | x\in \eta_{L_A} \}$
    
    \item Words with positive saliency and \textit{not} in wordlist, $\eta_{L_A}$:\\
    $int_4 = +_x \cap \{x | x\notin \eta_{L_A} \}$
\end{enumerate}

Intuitively, we would like to encourage the model to maximize the number of words in the first and second intersection sets ($int_1, int_2$) and minimize the number of words in the third and fourth intersection sets ($int_3, int_4$).  When this occurs, the model's behavior is aligned with the sociolinguistic wordlists.  

We can then calculate the generalizability and sensitive information reduction metrics.  For a given sample, we calculate either $\mathbf{HCM_{stn}^{gz}[T]:s}$ (Equation~\ref{eq:standalonegen}) or $\mathbf{HCM_{stn}^{sir}[T]:s}$ (Equation~\ref{eq:standalonedebias}) for each of the four interaction sets.  We calculate the metric by summing over weighting saliency values for matching sets and subtracting out the sum of weighted saliency values for the mistmatched sets.  We show the calculation of the generalizability metric here:

\begin{dmath}
    \mathbf{HCM_{stn}^{gz}[T]:s} = {\sum_{i=1}^2 \sum_{x\in int_i} \omega_x \times \lambda_x} - \sum_{i=3}^4 \sum_{x\in int_i} \omega_x \times \lambda_x
    \label{eq:pair}
\end{dmath}


We average over the samples within the dataset to obtain the dataset-level metric for generalizability metric as in Equation~\ref{eq:standalonegendataset} and for the sensitive information reduction metric as in Equation~\ref{eq:standalonedebiasdataset}.

\subsubsection{Relative Improvement ($\mathbf{HCM_{ri}}$)}
We refer to relative improvement evaluation as measuring how a model improves in a given task after adding an additional component (e.g., to reduce sensitive information encoded in a model). In aggregated level metrics such as accuracy, the relative improvement is measured by difference in performance aggregated over all sample's predictions. In our proposed method, we instead of focus on relative improvement per sample, as measured by the weights assigned to words/phrases. This is to measure the shift in the learnt associations of the model as compared to just the output predictions. As in the standalone case, we first calculate the relative improvement at the sample-level (e.g., $\mathbf{HCM_{ri}^{gz}[T]:s}$ or $\mathbf{HCM_{ri}^{sir}[T]:s}$).  For example, we can capture the relative improvement in the generalizability of the new model, by calculating the standalone values of both the new and original models (each using Equation~\ref{eq:standalonegen}) and dividing by the standalone of the original:

\begin{equation}
    \mathbf{HCM_{ri}^{sir}[T]:s} = \frac{\mathbf{HCM_{stn}^{sir}[T]_{new}:s} - \mathbf{HCM_{stn}^{sir}[T]_{orig}:s}}{\mathbf{HCM_{stn}^{sir}[T]_{orig}:s}}
\end{equation}

We can then average this value over all samples within the dataset, to calculate the dataset-level relative improvement metric for generalizability as in Equation~\ref{eq:standalonegendataset} and for the sensitive information reduction metric as in Equation~\ref{eq:standalonedebiasdataset}.  The final sensitive information reduction metric for relative improvement, moving from model $orig$ (the original) to model $new$ can be written as:

\begin{equation}
    \mathbf{HCM_{ri}^{sir}[T]} = \frac{1}{|D|} \sum_{s\in D} \mathbf{HCM_{ri}^{sir}[T]:s}
    \label{eq:ridebdataset}
\end{equation}

\subsection{Metrics for Measuring and Improving Emotion Recognition Model Generalizability}
We use  the sample-wise metric ($\mathbf{HCM_{stn}^{gz}Emo:s}$, Equation~\ref{eq:standalonegen}) for model training.  We use the dataset-level metric ($\mathbf{HCM_{stn}^{gz}Emo}$, Equation~\ref{eq:standalonegendataset}) to quantify emotion recognition performance.  We empirically evaluate both the correlation and impact of including this metric on the performance of the model, both in within and cross-dataset model analyses. 

We create wordlists using the valence annotations from National Research Council Canada - Valence, Arousal, and Dominance (NRC-VAD) Lexicon~\cite{mohammad2018obtaining} consisting of ~20,000 annotated unigrams. We normalize the annotated valence values between \emph{{-1,1}} and bin those into 3 quantile categories \emph{{low, med, high}} to match our model training label schema. This wordlist has information about the class label and the associated word.  Therefore, we use the metric that leverages this information (Eq.~\ref{eq:pair}).

\subsection{Metrics for Relative Measurement and Improvement of Reducing Sensitive Information Encoded in Emotion Recognition Models}
We use  the sample-wise metric ($\mathbf{HCM_{stn}^{sir}Dem:s}$, Equation~\ref{eq:standalonedebiasdataset}) for model training.  We use the dataset-level metric ($\mathbf{HCM_{ri}^{sir}Dem}$, Equation~\ref{eq:ridebdataset}) to quantify the sensitive information reduction efficacy.  

We use various sociological studies to create wordlists (Section~\ref{sec:wordlists}) and derive sensitive information reduction metrics corresponding to three demographic variables: gender ($\mathbf{HCM_{stn}^{sir}Gen:s}$), age ($\mathbf{HCM_{stn}^{sir}Age:s}$), and race ($\mathbf{HCM_{stn}^{sir}Race:s}$).  For gender and race, we use (a) word lists that have been tested using the Implicit Association Test (IAT)~\cite{nosek2005understanding} to measure population-level bias between the genders and (b) sociological studies~\cite{newman2008gender} and the corresponding category based words from LIWC~\cite{Tausczik2010ThePM}. For age, we consider sociolinguistic literature  for both, biological and social age~\cite{Eckert2017AgeAA}, using word trends dataset~\cite{Shparberg2021GoogleBN}, dictionary updates~\cite{LoVecchio2021UpdatingTO} and vocabulary grade levels~\cite{Yovanoff2005GradeLevelIO}.

\section{Research Questions}
\label{sec:rq}
\noindent \textbf{RQ0-emo}: Can we accurately recognize emotion, here labeled in terms of valence (positive vs. negative)?  This baseline question demonstrates that our system is capable of recognizing the information of interest. \\ 

\noindent \textbf{RQ0-SIR}: Can we use representations trained in the context of emotion recognition to recognize a demographic variable, here gender\footnote{In this chapter, we rely on binary markers of gender because these are the labels available within existing emotion corpora.  Binary labels of gender to not align with how many individuals identify.}?  We use this second research question to demonstrate that emotion representations encode information beyond the categories of interest, as shown in our prior work~\cite{jaiswal2020privacy}.\\
        
\noindent \textbf{RQ1}: How do metrics that capture sensitive information reduction ($\mathbf{HCM_{stn}^{sir}Dem}$) and generalizability ($\mathbf{HCM_{stn}^{gz}Emo}$) relating to reliance on emotionally-unrelated words vary across different sensitive information reduction methods? This research question provides a validation of the metrics and a way of measuring the effects of sensitive information reduction by itself.  \\
    
\noindent \textbf{RQ2}: How can the $\mathbf{HCM_{stn}^{sir}Dem}$ metric be used in model training to decrease the leakage of gender information? This research question provides evidence that the metric can be effectively leveraged as a sensitive information reduction technique in model training.\\
    
\noindent \textbf{RQ3}: How does the inclusion of $\mathbf{HCM_{stn}^{gz}Emo}$ increase cross-dataset performance? This research question provides evidence that increasing reliance on emotional words, and decreasing reliance on non-emotional words, promotes better performance in unknown environments.\\ 
    
\noindent \textbf{RQ4}: How does cross-dataset performance change when we include both $\mathbf{HCM_{stn}^{gz}Emo}$ and $\mathbf{HCM_{stn}^{sir}Dem}$?  Here, we investigate generalizable metrics that are based on knowledge of emotion expression and on spurious correlations between the perception of emotion and demographic categories (i.e., gender, age, race).\\
    
\noindent \textbf{RQ5}: For each of these tasks, emotion recognition and sensitive information reduction, do humans' model preferences correlate with $\mathbf{HCM_{stn}^{gz}Emo}$ and $\mathbf{HCM_{ri}^{sir}Dem}$, respectively, when considering the sample, model's prediction and salient explanations?\\

\section{Methods}
\label{sec:methods}
In this section, we provide an overview for the all the baseline models that we will evaluate with respect to $\mathbf{HCM}$. 

\subsection{Emotion Recognition}
\label{sec:emo}
\subsubsection{Baseline Emotion Recognition Model: \textbf{Base}}
\label{sec:genmodel}
We use the base version of Bidirectional Encoder Representations from Transformers (BERT) model due to the prevalence of this approach~\cite{devlin2018bert}. We use a pre-trained BeRT tokenizer for the model. We implement and fine-tune the model using the HuggingFace library~\cite{wolf2019huggingface}. 

\subsubsection{Multi-Dataset Training: \textbf{Multi-D}}
Our goal is to create an emotion recognition model that is not biased by dataset.  To do this, we start with the same model described above (\textbf{Base}).  We introduce an additional adversarial task, recognizing the dataset (e.g., IEMOCAP, Section~\ref{sec:data}), following the domain adversarial networks method suggested in~\cite{Ganin2016DomainAdversarialTO}.  We train our model on combinations of two datasets and test on the third.

\subsection{Sensitive Information Reduction}
\label{sec:deb}
Our goal is to reduce a given model's ability to detect a sensitive attribute (e.g., gender) from embeddings learned for the task of emotion recognition.  In all cases, the models are initialized with the \textbf{Base} model.

We measure sensitive information reduction success using Sensitive Information Leakage Measurement~\cite{mendelson2021debiasing}.  It assumes that an adversary has black-box access to the representations learned from the emotion recognition algorithm and access to a gender labeled subset of the dataset. The adversary uses these representation to train an auxiliary model to predict gender for any representation from that model, as described in our prior work~\cite{jaiswal2020privacy}. 

\subsubsection{Sensitive Information Reduction by Adversarial Training: \textbf{SIRAdv}}
\label{sec:privadv}
\textbf{SIRAdv} uses an adversarial training paradigm to reduce the sensitive information encoded in the generated embeddings with respect to gender. The main network is trained to unlearn gender using a Gradient Reversal Layer (GRL)~\cite{ganin2014unsupervised}, a multi-task approach to train models invariant to specific properties~\cite{meng2018speaker}. We place the GRL  between the embedding sub-network and gender classifier to obtain gender-invariant representations. 

\subsubsection{Sensitive Information Reduction by Data Augmentation: \textbf{SIRAug}}
\textbf{SIRAug} is is trained using an augmented data set using gender-swapping to compare our proposed method and resultant metric to other successful approaches~\cite{iosifidis2018dealing}. We use a pronoun-based word list and create a gender-swapped equivalent for each sentence, e.g., replacing ``he'' with ``she'', ``his'' with ``hers'', and so on. Data augmentation has its own issues: it doubles the training data size with no added label information, has expensive list creation of gender-based words in the dataset to be replaced, and nonsensical sentence creation~\cite{belinkov2019analysis}.

\subsubsection{Sensitive Information Reduction by Bias Fine-Tuning: \textbf{SIRBias}}
\textbf{SIRBias} leverages an additional outside dataset that is has reduced correlations with respect to the sensitive variable, which is used to train an initial model. We use the improvised turns subset of MSP-Improv (see Section~\ref{sec:data}).  We chose this subset because both actors (the male and female actor) perform improvisations using the same initial set of scenarios, controlling the word choice due to topic variations.  Therefore, the manner and speaking and word choice is more likely to be related to gender differences, rather than topic differences. We then fine tune a BERT model to be unable to distinguish between genders for a particular improvisational target.  This clusters samples together by prompt, accounting for topic-based word variations. The result of this fine-tuning should ideally be a model with reduced leakage of gender information. This resulting model is then finetuned for emotion recognition on IEMOCAP (see Section~\ref{sec:data}).

\subsection{Helper Model}
\subsubsection{Gender Control: \textbf{GenControl}}
Gender Control is a sanity check for verifying the reliability of our template. We hypothesize that if our proposed method indeed captures relevant information, then a model trained specifically to recognize gender should have the lowest $\mathbf{HCM_{ri}^{sir}Gen}$ value. We train a multi-task model for predicting both gender and emotion.

\subsubsection{Artificially Noisy Model: \textbf{ArtNoise}}
We create a parallel corpus of IEMOCAP to use as an attention check for the crowdsourced task. We add six artificial ``noisy" features, \{'zq0', 'zq1', 'zq2', 'zx0', 'zx1', 'zx2'\}, such that they correlate specifically with both emotion and gender and train a classification model to predict emotion on this dataset. The added signals are 100\% correlated to the sub-classes, ensuring that the model always learns these added nonsensical tokens as salient features.

\section{Experimental Setup}

\subsection{Models for Metric Correlation To Performance}
We perform all training experiments three times, using one dataset for training at a time, and testing on the same dataset (within-corpus testing) as well as the other two (cross-corpus testing). We train six models as described in Section~\ref{sec:emo} and Section~\ref{sec:deb}: two solely for emotion recognition $\{\mathbf{Base}, \mathbf{Multi-D}\}$, three for reduced sensitive information encoded emotion recognition $\{\mathbf{SIRAdv}, \mathbf{SIR, Aug}, \mathbf{SIRBias}\}$, and one control $\{\mathbf{GenControl}\}$.  

\subsection{Using the Metrics to Train Generalizable and Sensitive Information Reduced Models}
\label{sec:retrain}
We evaluate whether including the proposed metric in the training process improves emotion recognition and sensitive information reduction performance. We initialize the model with the baseline emotion recognition model~(\textbf{Base}, Section~\ref{sec:genmodel}). We retrain the model with the approaches discussed in Section~\ref{sec:deb}.  Please see Figure~\ref{img:model} for a pictorial example for how metrics are integrated within the training process.  We consider five different training conditions: 
\begin{enumerate}
    \item Only $\mathbf{HCM_{stn}^{gz}Emo:s}$ 
    \item Cross-entropy + $\mathbf{HCM_{stn}^{gz}Emo:s}$
    \item Only $\mathbf{HCM_{stn}^{sir}Dem:s}$ 
    \item Adversarial component + $\mathbf{HCM_{stn}^{sir}Dem:s}$ 
    \item Cross-entropy + adversarial component + $\mathbf{HCM_{stn}^{gz}Emo:s}$ + $\mathbf{HCM_{stn}^{sir}Dem:s}$
\end{enumerate}

We also perform experiments with each demographic metric on its own and other permutations of the setup.

\begin{figure}[t]
\centering
\includegraphics[width=8cm]{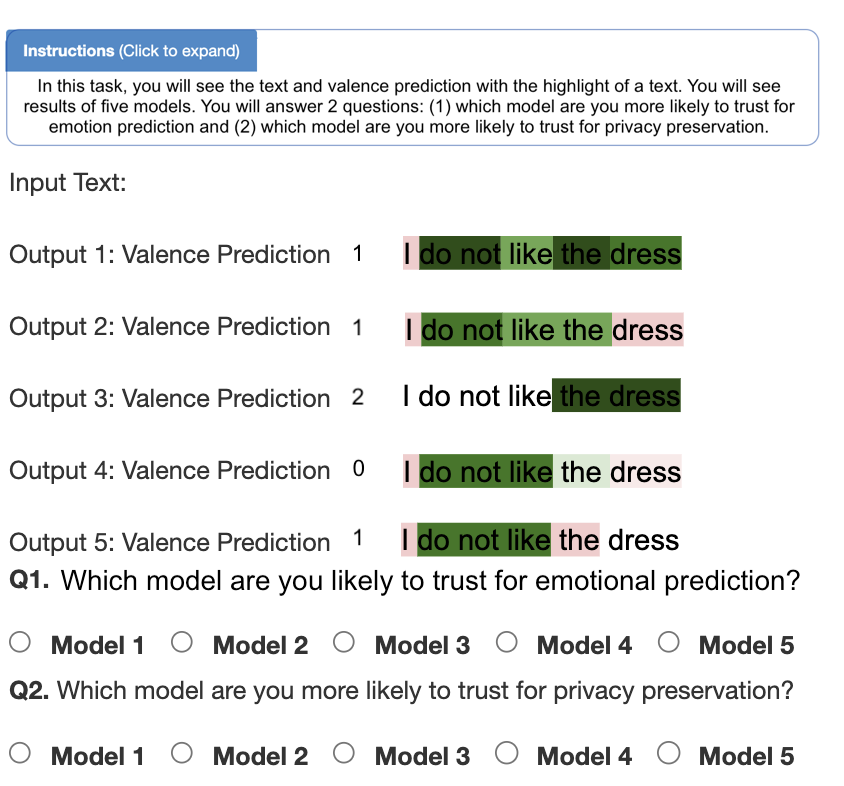}
\caption{The crowdsourcing interface.  Users are asked to indicate their model preference given visualizations of the relative importance of specific words in the prediction of emotion.  The evaluators were not presented with the names of the models and the order of model outputs was randomized for each viewing.  Note that the sentence was chosen to clearly convey gender, and potentially, gendered information.  In the figure, green indicates that the model is using the word in the prediction, while red indicates that it is not.  The outputs correspond to: 1) GenControl - note the heavy focus on the word ``dress'', 2) SIRBias, 3) ArtNoise, 4) SIRAug, 5) SIRAdv (for model details, see Section~\ref{sec:deb}).  Observe that the three sensitive information reduction methods are no longer focusing on the word ``dress''. }
\label{img:crowdsourcing}
\end{figure}

\subsection{Crowdsourcing}
\label{sec:questions}
Finally, we explore whether user preference aligns with our proposed metrics.  We select a representative sample population by considering three levels of: 1) valence (low, medium, and high, see Section~\ref{sec:labels}) and 2) the metrics of $HCM_{stn}^{sir}Gnd:s$ (low, medium, and high, measured via quantiles).  This results in nine bins. 

We select 60 random samples from each bin for a total 540 samples.  We classify each sample using our five models (\textbf{SIRAdv}, \textbf{SIRAug}, \textbf{SIRBias}, \textbf{GenControl}, \textbf{ArtNoise}, described above).  We extract sample-level predictions.  In addition, we extract sample-level explanations using Captum~\cite{sundararajan2017axiomatic},  Section~\ref{sec:integratedgradients}).  

\textbf{ArtNoise} is used as a attention check baseline for human evaluation, ensuring that the evaluators are paying attention to the task at hand. We consider this option as an attention check and discard any response where the crowdsourced worker preferred the result and/or the explanation from this model (7.31\% samples were discarded in total).

We presented the predictions from the models and heat map-based explanations.  We asked the workers to choose between the five models, focusing on the following questions: (1) which model are you more likely to trust for emotion prediction and (2) which model are you more likely to trust for not being able to predict sensitive information. 

We recruited annotators using Prolific from a population of workers with the following characteristics: 1) in the United States, and 2) native English speakers. Each task was annotated by three workers. We used qualification tests to check that all workers understood valence.
Each task took an average of one-minute. The compensation was $\$9.45/$hr.

\section{Results and Discussion}

\subsection{RQ0-emo: Can we accurately recognize emotion, here labeled dimensionally for valence?}

The baseline emotion classification model (\textbf{Base}) for within-corpus valence classification has 0.65 UAR for IEMOCAP, 0.68 for MuSE, and, 0.54 for MSP-Podcast (chance is 0.33 UAR), which is comparable to existing approaches on this domain~\cite{aldeneh2017pooling}.

In the cross-corpus setting, we see an average drop of at least 15\% on all the models and datasets permutations. For example, a model trained on IEMOCAP and tested on MuSE has a UAR of 0.58 for \textbf{Base}, and, 0.53 for the \textbf{SIRAdv} model, which is significantly lower (paired t-test adjusted for multiple comparisons, p=0.0017) than the performance of these models when trained and tested on MuSE (0.68 and 0.71 respectively).

The multi-dataset emotion classification model has an overall higher performance in the cross-corpus setting. For example, the \textbf{Multi-D}~ has 0.64 UAR for IEMOCAP as compared to 0.61 and 0.57 obtained using \textbf{Base}.

We find that while reducing sensitive information leakage reduces within-corpus emotion recognition performance in many cases, these sensitive information leakage reduced models, especially \textbf{SIRAdv} usually have equivalent or better performance as compared to \textbf{Base} in cross-corpus setting, e.g., \textbf{Base} and \textbf{SIRAdv} trained on MuSE and tested on IEMOCAP has a UAR of 0.61 and 0.63, respectively.

\subsection{RQ0-SIR: Can we use representations trained in the context of emotion recognition to recognize a demographic variable, here gender?}

Gender classification within dataset is highest for models using the emotion representations learned from the baseline model and the control gender-addition model, supporting prior work ~\cite{jaiswal2020privacy}.  The multi-dataset model usually has a lower or equivalent gender UAR as compared to the baseline model. Gender classification operating on the representations learned from all three sensitive information reduction models have lower UAR, compared to the multi-dataset model.

The leakage of gender information is higher for all models in the cross-corpus condition, compared to within corpus testing, excepting the multitask emotion-gender model.  This 
suggests some loss of efficacy of the sensitive information reduction methods in cross-corpus settings. In almost all the cases, \textbf{SIRAdv} has the lowest cross-corpus gender leakage while almost retaining emotion recognition performance. Hence, we use \textbf{SIRAdv} as our main model for retraining purposes (please see results tables in appendix for full details).

\subsection{RQ1: How do metrics that capture sensitive information reduction, $\mathbf{HCM_{ri}^{sir}Dem}$, and reliance on words unrelated to emotion, $\mathbf{HCM_{stn}^{gz}Emo}$, correlate to a model's emotion recognition and sensitive information reduction performance?}
 
We expect that models that effectively reduce leakage of sensitive information the model have high $\mathbf{HCM_{ri}^{sir}Dem}$ values, while models that do not will have low $\mathbf{HCM_{stn}^{gz}Emo}$ values.  

We initially validate the assumption that the $\mathbf{HCM_{ri}^{sir}Gen}$ metric is very low for the GenControl model (-0.21).  This aligns with our expectation that the relative sensitive information reduction should be low.  The goal of the GenControl model is to \textit{learn}, rather than ignore, gender identity.  

For within corpus settings, our expectation is that the SIRAug~ and SIRAug models will have the highest $\mathbf{HCM_{ri}^{sir}Gen}$ values. We find that across all the datasets, and models, gender leakage is significantly negatively correlated (paired t-test adjusted for multiple comparisons) with $\mathbf{HCM_{ri}^{sir}Gen}$ for both within corpus (${-0.95}^*$) and cross-corpus (${-0.89}^*$) evaluation. ($*$ refers to significant value when using paired t-test adjusted for multiple comparisons)

We expect that models that generalize well will have high $\mathbf{HCM_{stn}^{gz}Emo}$ values, while models that generalize poorly will have low $\mathbf{HCM_{stn}^{gz}Emo}$ values.  We find that in two out of three cases, \textbf{SIRAdv} has a higher $\mathbf{HCM_{stn}^{gz}Emo}$ than \textbf{Base}. We find that across all the datasets, and models, emotion recognition performance is significantly positively correlated (paired t-test adjusted for multiple comparisons) with $\mathbf{HCM_{stn}^{gz}Emo}$ for both within corpus (${0.72}^*$), and cross-corpus (${0.80}^*$) evaluation.

We analyze the intersections between word-lists and saliency attribution. We consider instances in which $\mathbf{HCM_{stn}^{gz}Emo}$ positively impacts model performance, when a sample moves from being incorrectly to correctly classified after $\mathbf{HCM_{stn}^{gz}Emo}$ is included in model training.  We observe that in 87.3\% of the words in the intersections  (Equation~\ref{eq:standaloneintersectionset}), in the better performing model have medium valence label (from the set: low, medium, and high).  This suggests that the metric enables the model improve in the recognition of subtle displays of emotion.

\subsection{RQ2: How does additionally optimizing for $\mathbf{HCM_{stn}^{sir}Gen}$ affect gender information leakage?}

We observe that when using the $\mathbf{HCM_{stn}^{sir}Gen}$ metric as the only optimizer, we fall short of the sensitive information reduction performance of \textbf{SIRAdv}, indicating need for composite metrics. We find that augmenting \textbf{SIRAdv} training with $\mathbf{HCM_{stn}^{gz}Gen:s}$ improves valence and decreases gender classification performance (0.64 in \textbf{SIRAdv} vs 0.61 with \textbf{SIRAug}) in cross-corpus settings (Table~\ref{tab:resultsonlyemo}). In within corpus settings, we find that the addition of metric still benefits sensitive information reduction but can also lead to a non-negligible drop in emotion recognition performance. We also find that while age ($\mathbf{HCM_{stn}^{sir}Age}$) and race ($\mathbf{HCM_{stn}^{sir}Race}$) metrics on their own do not signify any positive change in cross-corpus dataset performance.

We find that the metrics can be jointly considered in a simple multi-task setup between the primary task (i.e., emotion) and an adversarial component ($\mathbf{HCM_{stn}^{sir}Gen}$).  The multi-task setup results in the best performing gender information reduction (we do not have labels to measure age/race) of emotion recognition model (Table~\ref{tab:resultscomposite}).

\subsection{RQ3: How does additionally optimizing for $\mathbf{HCM_{stn}^{gz}Emo}$ affect emotion recognition?}

We find that optimizing for alignment with the human-centered metric leads to better cross-corpus performance across every setup (Table~\ref{tab:resultsonlyemo}).  For example, when we augment the \textbf{Base} model trained on IEMOCAP with $\mathbf{HCM_{stn}^{gz}Emo}$.  The UAR increases to 0.6 and 0.45 for Muse and MSP-Podcast respectively, as compared to 0.58 and 0.43. Additionally, we find that, training the model with these metrics give us better or equivalent performance to when using the combined dataset method, which was the best solely emotion recognition model. This shows the effectiveness of the metric, and lowers the required computation time and power.

\begin{table}
\begin{center}
\footnotesize
 \caption{Training models with only $\mathbf{HCM_{stn}^{gz}Gen:s}$ \label{tab:resultsonlyemo}}
	\begin{tabular}{cccc}
 \multirow{2}{*}{Training} & \multicolumn{3}{c}{Testing} \\
 & IEMOCAP & MuSE & MSP-Improv \\\\
 \hline\hline\\
 IEMOCAP &  0.64 & 0.60 & 0.45 \\
 MuSE & 0.63 & 0.67 & 0.41  \\
 MSP-Podcast & 0.60 & 0.62 & 0.52 \\


\end{tabular}
\end{center}
\end{table}

We also see that while we see a performance increase for cross-dataset testing, some within dataset performances are lower than the generally trained model (e.g., \textbf{Base} has UAR 0.54 for MSP-Podcast, but \textbf{Base}+$\mathbf{HCM_{stn}^{gz}Emo}$ has a UAR 0.52). We hypothesize that this occurs because the models were relying on spurious correlations (sprurious, with respect to emotion recognition), which, when removed, decreased the model's performance. 

\subsection{RQ4: Does emotion recognition and sensitive information reduction performance change when including both $\mathbf{HCM_{stn}^{gz}Emo}$ and $\mathbf{HCM_{stn}^{sir}Dem}$ during training?}  

We include all sensitive information reduction components ($\mathbf{HCM_{stn}^{sir}Gen}$, $\mathbf{HCM_{stn}^{sir}Age}$, $\mathbf{HCM_{stn}^{sir}Race}$) and the emotion generalizability component ($\mathbf{HCM_{stn}^{gz}Emo}$) in multi-task setup method in model training.  We find that this composite metric (combination of all metrics: $\mathbf{HCM_{stn}^{sir}Gen}:s$ + $\mathbf{HCM_{stn}^{sir}Age}:s$ + $\mathbf{HCM_{stn}^{sir}Race}:s$) leads to better model performance across the board.  For example, models trained with the composite metric improves the UAR to 0.7 on MuSE within-corpus, as compared to 0.68 from the \textbf{Base} model).

\begin{table}
\begin{center}
\footnotesize
 \caption{Training models with composite metrics: $\mathbf{HCM_{stn}^{gz}Gen:s}$ + \textbf{SIRAdv}(Gender) + $\mathbf{HCM_{stn}^{sir}Dem}:s$ ($\mathbf{HCM_{stn}^{sir}Gen}:s$ + $\mathbf{HCM_{stn}^{sir}Age}:s$ + $\mathbf{HCM_{stn}^{sir}Race}:s$)\label{tab:resultscomposite}}
	\begin{tabular}{cccc}
 \multirow{2}{*}{Training} & \multicolumn{3}{c}{Testing} \\
 & IEMOCAP & MuSE & MSP-Improv \\
 \hline\hline\\
 IEMOCAP & 0.71 & 0.59 & 0.45 \\
 MuSE & 0.67 & 0.70 & 0.46  \\
 MSP-Podcast & 0.57 & 0.56 & 0.50 \\


\end{tabular}
\end{center}
\end{table}

We find that the two components interact to give better performance on both cross-corpus emotion recognition and sensitive information reduction performance.  For example, we see an improved UAR of 0.67 on IEMOCAP (cross-corpus), compared to 0.62 from the \textbf{Base} method). Further, we also see that the gender leakage universally reduces when training the models on this combination (cross-corpus testing on MSP-Podcast when using composite metric training on Muse improves performance by 6\%). We hypothesize that encouraging certain generalizable correlations while discouraging spurious correlations leads to models that are more robust and have reduced sensitive information leakage across the spectrum, but the enforcement of specific `unlearning' often leads to negligible drops in within-corpus sensitive information reduction performance.

Finally, we found that inclusion of age and race, to gender, as sensitive information reduction components improves emotion recognition performance in cross corpus evaluation (Please see the full results table in the appendix).

\subsection{RQ5: For each of these tasks, emotion recognition and sensitive information reduction, do humans' model preferences correlate with $\mathbf{HCM_{stn}^{gz}Emo}$ and $\mathbf{HCM_{ri}^{sir}Dem}$ respectively, when considering the sample, model's prediction and salient explanations?}

We show that user preference depends on what the user was asked to consider the model's behavior for, with respect to emotion or to sensitive information leakage~\cite{iachello2007end}. The user preference for emotion recognition models is significantly correlated with within-corpus valence recognition performance (0.78) and with the $\mathbf{HCM_{stn}^{gz}Emo}$ metric (0.71).  

The correlation between user preference for emotion recognition and the $\mathbf{HCM_{ri}^{sir}Gen}$ metric is negligible (0.09).  This suggests that when asked to focus on emotion recognition performance with models that have the same prediction but different highlighted words, both performance and the potential for generalizability (when grasped using the highlighted words/phrases for attributed by the model for the prediction) dominate user preference.  

When choosing sensitive information leakage reducing model, we find that user's preference is moderately correlated with emotion recognition performance (0.46) and strongly negatively correlated with the ability to predict gender (-0.86).  This is also supported by the correlation between user preference and $\mathbf{HCM_{ri}^{sir}Gen}$ (0.64) and $\mathbf{HCM_{stn}^{gz}Emo}$ (0.81).  This support the hypothesis that, human perception can be encoded into models through the proposed metric template ($\mathbf{HCM[T]}$).

\section{Conclusion}
In this chapter, we present a novel set of emotion generalizability metrics focused on reducing sensitive information encoding, and reliance on emotionally-unrelated words.  We evaluate these metrics across the tasks of emotion recognition and classification of demographic variables (here, gender).  We demonstrate that these metrics are correlated with the performance of algorithms, both within and across corpus, and that these metrics can be used during model training to improve classification performance and leakage of gender information.  In future work, we 
aim to look at finer grained controls over reducing learnt sensitive information when evaluated by the end user, and methods to incorporate user-specific personalization of the baseline metric.


%


Please see the table on the following page that provides a detailed accounting for the results discussed in the rest of the chapter.

\begin{table*}
\begin{center}

\footnotesize
\caption{Within and cross dataset performance of models across various datasets, correlation with these metrics and using metrics for training.  Note that for space, the column headings refer to H, rather than HCM.  We use the notation "Dem" to refer to the combined demographic categories of Gender, Age, and Race. SIR refers to sensitive information reduction.}
\scriptsize
	\begin{tabular}{cp{30mm}ccccccccc}
\multirow{2}{*}{Train} & \multirow{2}{*}{ Model } & \multicolumn{3}{c}{Valence} & \multirow{2}{*}{ $\mathbf{H_{stn}^{gz}E}$} & \multicolumn{3}{c}{Gender} & \multirow{2}{*}{ $\mathbf{H_{ri}^{sir}G}$}  & \multirow{2}{*}{ 1- $\mathbf{H_{ri}^{sir}G}$}  \\
& & IEM & MuSE & MSP-P & & IEM & MuSE & MSP-P & & \\
\hline
\hline\\
\multirow{11}{*}{ \myboxsm{IEMOCAP} } & Baseline & 0.65 & 0.58 & 0.43 & 0.64 & 0.71 & 0.74 & 0.68 & - & \\
& SIRAdv & 0.67 & 0.53 & 0.38 & 0.62 & 0.56 & 0.62 & 0.59 & 0.65 & 0.35 \\
& SIRAug & 0.64 & 0.59 & 0.44 & 0.68 & 0.55 & 0.64 & 0.55 & 0.68 & 0.32 \\
& SIRBias & 0.62 & 0.53 & 0.38 & 0.6 & 0.62 & 0.69 & 0.7 & 0.56 & 0.44 \\
& GenControl & 0.63 & 0.52 & 0.37 & 0.59 & 0.82 & 0.76 & 0.76 & -0.21 & 1.21 \\
& $\mathbf{HCM_{stn}^{sir}Gen}$ & 0.64 & 0.56 & 0.4 & 0.61 & 0.58 & 0.61 & 0.6 & 0.6 & 0.4 \\
& $\mathbf{HCM_{stn}^{sir}Age}$ & 0.65 & 0.51 & 0.38 & 0.63 & - & - & - & 0.32 & 0.68 \\
& $\mathbf{HCM_{stn}^{sir}Race}$ & 0.63 & 0.52 & 0.39 & 0.65 & - & - & - & 0.51 & 0.49 \\
& $\mathbf{HCM_{stn}^{sir}Dem}$ & 0.65 & 0.54 & 0.41 & 0.64 & - & - & - & 0.63 & 0.37 \\
& SIRAdv+$\mathbf{HCM_{stn}^{sir}Gen}$ & 0.69 & 0.56 & 0.43 & 0.67 & - & - & - & 0.74 & 0.26 \\
& SIRAdv+$\mathbf{HCM_{stn}^{sir}Dem}$ & 0.67 & 0.58 & 0.44 & 0.66 & 0.54 & 0.6 & 0.52 & 0.72 & 0.28 \\\\
\hline\\
\multirow{11}{*}{ \myboxsm{MuSE} } & Baseline & 0.61 & 0.68 & 0.38 & 0.65 & 0.67 & 0.57 & 0.72 & - & \\
& SIRAdv & 0.63 & 0.71 & 0.36 & 0.67 & 0.58 & 0.53 & 0.56 & 0.67 & 0.33 \\
& SIRAug & 0.57 & 0.62 & 0.34 & 0.6 & 0.6 & 0.51 & 0.59 & 0.61 & 0.39 \\
& SIRBias & 0.56 & 0.6 & 0.34 & 0.59 & 0.62 & 0.55 & 0.63 & 0.59 & 0.41 \\
& GenControl & 0.4 & 0.53 & 0.3 & 0.52 & 0.8 & 0.88 & 0.72 & -0.17 & 1.17 \\
& $\mathbf{HCM_{stn}^{sir}Gen}$ & 0.58 & 0.62 & 0.34 & 0.6 & 0.65 & 0.54 & 0.58 & 0.58 & 0.42 \\
& $\mathbf{HCM_{stn}^{sir}Age}$ & 0.6 & 0.62 & 0.35 & 0.58 & - & - & - & 0.21 & 0.79 \\
& $\mathbf{HCM_{stn}^{sir}Race}$ & 0.57 & 0.61 & 0.34 & 0.59 & - & - & - & 0.33 & 0.67 \\
& $\mathbf{HCM_{stn}^{sir}Dem}$ (all) & 0.6 & 0.64 & 0.38 & 0.62 & - & - & - & 0.55 & 0.45 \\
& SIRAdv+$\mathbf{HCM_{stn}^{sir}Gen}$ & 0.66 & 0.7 & 0.44 & 0.73 & - & - & - & 0.7 & 0.3 \\
& SIRAdv+$\mathbf{HCM_{stn}^{sir}Dem}$ & 0.64 & 0.69 & 0.4 & 0.7 & 0.55 & 0.52 & 0.55 & 0.69 & 0.31 \\\\
\hline\\
\multirow{11}{*}{ \myboxsm{MSP Podcast} } & Baseline & 0.57 & 0.58 & 0.54 & 0.53 & 0.65 & 0.55 & 0.72 & - & \\
& SIRAdv & 0.54 & 0.48 & 0.5 & 0.42 & 0.54 & 0.51 & 0.57 & 0.68 & 0.32 \\
& SIRAug & 0.52 & 0.43 & 0.48 & 0.38 & 0.59 & 0.52 & 0.59 & 0.64 & 0.36 \\
& SIRBias & 0.44 & 0.42 & 0.48 & 0.38 & 0.61 & 0.54 & 0.61 & 0.53 & 0.47 \\
& GenControl & 0.37 & 0.38 & 0.4 & 0.32 & 0.78 & 0.7 & 0.79 & -0.19 & 1.19 \\
& $\mathbf{HCM_{stn}^{sir}Gen}$ & 0.44 & 0.46 & 0.45 & 0.37 & 0.59 & 0.6 & 0.58 & 0.5 & 0.5 \\
& $\mathbf{HCM_{stn}^{sir}Age}$ & 0.46 & 0.48 & 0.47 & 0.38 & - & - & - & 0.45 & 0.55 \\
& $\mathbf{HCM_{stn}^{sir}Race}$ & 0.47 & 0.5 & 0.47 & 0.39 & - & - & - & 0.5 & 0.5 \\
& $\mathbf{HCM_{stn}^{sir}Dem}$ & 0.47 & 0.51 & 0.48 & 0.4 & - & - & - & 0.55 & 0.45 \\
& DemAdv+$\mathbf{HCM_{stn}^{sir}Gen}$ & 0.54 & 0.56 & 0.57 & 0.55 & - & - & - & 0.69 & 0.31 \\
& DemAdv+$\mathbf{HCM_{stn}^{sir}Dem}$ & 0.52 & 0.55 & 0.53 & 0.51 & 0.53 & 0.52 & 0.52 & 0.63 & 0.37 \\\\
 \hline
\end{tabular}
\end{center}
\end{table*}

%% file: Chapters/chap10.tex
The objective of this dissertation is to investigate the effect of implicit decisions made during the machine learning (ML) process on the performance of ML models trained for emotion recognition. Implicit decisions refer to the choices made by practitioners during the various stages of the ML pipeline, such as data preprocessing, feature selection, and model selection. These decisions are often made based on intuition, experience, or convenience, rather than a rigorous analysis of their impact on the final model. Chapters~\ref{chap:muse} and Chapter~\ref{chap:emotion_annotation} of this dissertation focus on identifying key areas for dataset creation and annotation that can significantly impact the evaluation and performance of trained machine learning (ML) models. Chapters~\ref{chap:emotion_annotation} and~\ref{chap:conf} of this dissertation are dedicated to investigating the impact of confounding factors in data on the performance and reliability of machine learning (ML) models. Chapters~\ref{chap:leakage} and~\ref{chap:metric} of this dissertation are dedicated to addressing two critical challenges in the development and deployment of machine learning (ML) models: unintentional encoding of demographic and memership information in model representations and generalizability of these models. 

\section{Summary of Contributions}

\begin{itemize}

\item \textbf{Chapter~\ref{chap:muse}} presented the following contributions:
    \begin{itemize}
        \item We unveiled a unique dataset, Multimodal Stressed Emotion (MuSE), purposed for studying the interaction between stress and emotion in speech.
        \item We elucidated the procedure for data collection, the potential applications, and the methodology for annotating emotional content.
        \item We showcased the dataset's potential in the development and evaluation of models focused on emotion recognition and stress detection.
    \end{itemize}

\item \textbf{Chapter~\ref{chap:noise_aug}} brought forth:
    \begin{itemize}
        \item Augmenting IEMOCAP with realistic noisy samples through the use of various types of environmental and synthetic noise.
        \item An evaluation of how noise influences both the ground truth and predicted labels with associated guidance for noise-based augmentation within speech emotion datasets.
        \item A strong emphasis on the need to consider the influence of noise on human emotion perception and the necessity of robustly testing models.
    \end{itemize}

\item \textbf{Chapter~\ref{chap:emotion_annotation}} provided insights on:
    \begin{itemize}
        \item How context influences the annotation of emotional content within the MuSE dataset.
        \item The comparison between two labeling methods, randomized and contextualized, and the discovery that contextualized labeling produces annotations closer to self-reported labels.
        \item The finding that labels generated using the randomized method can be more efficiently predicted by automated systems.
    \end{itemize}

\item \textbf{Chapter~\ref{chap:conf}} offered:
    \begin{itemize}
        \item A new method for separating stress modulations from emotion representations utilizing adversarial networks.
        \item Evidence showing that controlling for stress during training boosts the generalizability of emotion recognition models across new domains.
        \item An effective demonstration of our unique approach's usability using the MuSE dataset and its potent applicability to other confounding variables present in emotion datasets.
    \end{itemize}

\item \textbf{Chapter~\ref{chap:leakage}} achieved:
    \begin{itemize}
        \item A comprehensive analysis of demographic leakage in representations obtained from textual, acoustic, and multimodal data when trained for emotion recognition.
        \item The introduction of an adversarial learning paradigm to reduce demographic leakage from generated representations.
        \item A valid demonstration of our novel approach's efficacy on numerous datasets and its potential use to defend against set-based membership identification.
    \end{itemize}

\item \textbf{Chapter~\ref{chap:metric}} proposed:
    \begin{itemize}
        \item Automatic and quantifiable metrics for measuring generalizability and sensitive information encoding in the representation of machine learning models within the context of speech emotion recognition.
        \item A verification method of the utility of the proposed metrics using crowdsourcing, showcasing their adaptability for evaluating cross-corpus generalization.
        \item A cost-efficient and dependable way of appraising the effectiveness of machine learning models within speech emotion recognition.
    \end{itemize}
\end{itemize}

These points, in sum, illustrate the specific contributions this dissertation makes to improve understanding, methodologies, and tools in the area of emotion recognition research.

\section{Future Work}

Building upon the contributions of this dissertation, there are two primary areas where future research could focus. 

Firstly, when using interpretation and explanation, whether for emotion recognition or other human centered machine learning tasks, future research should focus on developing a standardized evaluation checklist for deep learning model explanations. Currently, the field is marked by inconsistent evaluation methodologies in various research papers, making it challenging to compare results. It is essential to address these variations by designing a checklist that encapsulates evaluation criteria applicable across different studies. The introduction of such a tool will help simplify the evaluation process and foster a clearer understanding and comparison of diverse research outcomes. The current environment lacks a standardized approach, and individual studies often adopt unique methods, which detracts from the comparability of research. Creating a universal evaluation framework is thus a crucial step for future advancements in the field.

Secondly, in terms of evaluation metrics for any human centered tasks, the use of closed language models (LLMs) poses certain challenges when it comes to testing their performance. If we do not know what went into the training dataset of a closed LLM, it is very likely that the test dataset we use or generate will be contaminated. This can result in inaccurate performance metrics and difficulty in verifying the correctness of the model's responses. Prompt-based benchmarking can exacerbate these challenges further by introducing another layer of complexity. If we are not directly testing for an ``input sample" to ``output" correspondence, the way we write the prompt and what we ask for as an output will significantly change how we benchmark the ability of any model. This can make evaluating the performance of a closed LLM more difficult and error-prone, and increase the risk of producing biased results. Future work can look into Prompt Based Experimental Design Schemas. These schemas can be thought of as templates, but with a key distinction. They not only focus on varying the input sample, but also on modifying the prompt specifications. By doing so, they allow us to closely monitor and analyze the changes that occur from one prompt to another. This approach provides a valuable way to study and understand the impact of different prompts on the overall outcome of an experiment.

In essence, these proposed trajectories, in tandem with this dissertation's contributions, provide a roadmap for further research in the field of models trained for human centered tasks such as emotion recognition. The continuance of these research undertakings will augment our understanding and enhance the effectiveness of the techniques used in the recognition and evaluation of human centered predictive models.

%% file: thesis.bbl
\begin{thebibliography}{100}

\bibitem{phono}
Phonological history of english consonant clusters=
  \url{https://bit.ly/3l08zbu}.

\bibitem{abadi2016deep}
Martin Abadi, Andy Chu, Ian Goodfellow, H~Brendan McMahan, Ilya Mironov, Kunal
  Talwar, and Li~Zhang.
\newblock Deep learning with differential privacy.
\newblock In {\em Proc. 2016 ACM Conference CCS}.

\bibitem{abdelwahab2017incremental}
Mohammed Abdelwahab and Carlos Busso.
\newblock Incremental adaptation using active learning for acoustic emotion
  recognition.
\newblock In {\em Acoustics, Speech and Signal Processing (ICASSP), 2017 IEEE
  International Conference on}, pages 5160--5164. Ieee, 2017.

\bibitem{abdelwahab2018domain}
Mohammed Abdelwahab and Carlos Busso.
\newblock Domain adversarial for acoustic emotion recognition.
\newblock {\em IEEE/ACM Transactions on Audio, Speech, and Language
  Processing}, 26(12):2423--2435, 2018.

\bibitem{abdullah2019practical}
Hadi Abdullah, Washington Garcia, Christian Peeters, Patrick Traynor, Kevin~RB
  Butler, and Joseph Wilson.
\newblock Practical hidden voice attacks against speech and speaker recognition
  systems.
\newblock {\em arXiv preprint arXiv:1904.05734}, 2019.

\bibitem{abouelenien2016analyzing}
Mohamed Abouelenien, Rada Mihalcea, and Mihai Burzo.
\newblock Analyzing thermal and visual clues of deception for a non-contact
  deception detection approach.
\newblock In {\em Proceedings of the 9th ACM International Conference on
  PErvasive Technologies Related to Assistive Environments}, page~35. Acm,
  2016.

\bibitem{Abouelenien:2014:DDU:2663204.2663229}
Mohamed Abouelenien, Veronica P{\'e}rez-Rosas, Rada Mihalcea, and Mihai Burzo.
\newblock Deception detection using a multimodal approach.
\newblock In {\em Proceedings of the 16th International Conference on
  Multimodal Interaction}, Icmi '14, pages 58--65, New York, NY, USA, 2014.
  Acm.

\bibitem{aguiar2014voce}
Ana Aguiar, Mariana Kaiseler, Mariana Cunha, Hugo Meinedo, Pedro~R Almeida, and
  Jorge Silva.
\newblock Voce corpus: Ecologically collected speech annotated with
  physiological and psychological stress assessments.
\newblock In {\em LREC 2014, Ninth International Conference on Language
  Resources and Evaluation}, 2014.

\bibitem{aldeneh2017pooling}
Zakaria Aldeneh, Soheil Khorram, Dimitrios Dimitriadis, and Emily~Mower
  Provost.
\newblock Pooling acoustic and lexical features for the prediction of valence.
\newblock In {\em Proceedings of the 19th ACM International Conference on
  Multimodal Interaction}, pages 68--72. Acm, 2017.

\bibitem{aldeneh2017using}
Zakaria Aldeneh and Emily~Mower Provost.
\newblock Using regional saliency for speech emotion recognition.
\newblock In {\em Acoustics, Speech and Signal Processing (ICASSP), 2017 IEEE
  International Conference on}, pages 2741--2745. Ieee, 2017.

\bibitem{aron1997experimental}
Arthur Aron, Edward Melinat, Elaine~N Aron, Robert~Darrin Vallone, and Renee~J
  Bator.
\newblock The experimental generation of interpersonal closeness: A procedure
  and some preliminary findings.
\newblock {\em Personality and Social Psychology Bulletin}, 23(4):363--377,
  1997.

\bibitem{audibert2010prosodic}
Nicolas Audibert, V{\'e}ronique Auberg{\'e}, and Albert Rilliard.
\newblock Prosodic correlates of acted vs. spontaneous discrimination of
  expressive speech: a pilot study.
\newblock In {\em Speech Prosody 2010-Fifth International Conference}, 2010.

\bibitem{averill1980constructivist}
James~R Averill.
\newblock A constructivist view of emotion.
\newblock In {\em Theories of emotion}, pages 305--339. Elsevier, 1980.

\bibitem{bagdasaryan2019differential}
Eugene Bagdasaryan and Vitaly Shmatikov.
\newblock Differential privacy has disparate impact on model accuracy.
\newblock {\em arXiv preprint arXiv:1905.12101}, 2019.

\bibitem{baltruvsaitis2016openface}
Tadas Baltru{\v{s}}aitis, Peter Robinson, and Louis-Philippe Morency.
\newblock Openface: an open source facial behavior analysis toolkit.
\newblock In {\em 2016 IEEE Winter Conference on Applications of Computer
  Vision (WACV)}, pages 1--10. Ieee, 2016.

\bibitem{banda2011noise}
Ntombikayise Banda and Peter Robinson.
\newblock Noise analysis in audio-visual emotion recognition.
\newblock In {\em Proceedings of the 11th International Conference on
  Multimodal Interaction (ICMI)}, 2011.

\bibitem{banziger2015path}
Tanja B{\"a}nziger, Georg Hosoya, and Klaus~R Scherer.
\newblock Path models of vocal emotion communication.
\newblock {\em PloS one}, 10(9):e0136675, 2015.

\bibitem{barbaro2006face}
Michael Barbaro, Tom Zeller, and Saul Hansell.
\newblock A face is exposed for aol searcher no. 4417749.
\newblock {\em New York Times}, 2006.

\bibitem{barrick1991big}
Murray~R Barrick and Michael~K Mount.
\newblock The big five personality dimensions and job performance: a
  meta-analysis.
\newblock {\em Personnel psychology}, 44(1):1--26, 1991.

\bibitem{bartolini2011eliciting}
Ellen~Elizabeth Bartolini.
\newblock Eliciting emotion with film: Development of a stimulus set.
\newblock 2011.

\bibitem{batliner1995can}
Anton Batliner, Ralf Kompe, A~Kie{\ss}ling, E~N{\"o}th, and H~Niemann.
\newblock Can you tell apart spontaneous and read speech if you just look at
  prosody?
\newblock In {\em Speech Recognition and Coding}, pages 321--324. Springer,
  1995.

\bibitem{belinkov2019analysis}
Yonatan Belinkov and James Glass.
\newblock Analysis methods in neural language processing: A survey.
\newblock {\em Transactions of the Association for Computational Linguistics},
  7:49--72, 2019.

\bibitem{ben2010theory}
Shai Ben-David, John Blitzer, Koby Crammer, Alex Kulesza, Fernando Pereira, and
  Jennifer~Wortman Vaughan.
\newblock A theory of learning from different domains.
\newblock {\em Machine learning}, 79(1-2):151--175, 2010.

\bibitem{bengio2013representation}
Yoshua Bengio, Aaron Courville, and Pascal Vincent.
\newblock Representation learning: A review and new perspectives.
\newblock {\em IEEE transactions on pattern analysis and machine intelligence},
  2013.

\bibitem{benjamini1995controlling}
Yoav Benjamini and Yosef Hochberg.
\newblock Controlling the false discovery rate: a practical and powerful
  approach to multiple testing.
\newblock {\em Journal of the Royal statistical society: series B
  (Methodological)}, 57(1):289--300, 1995.

\bibitem{berger2012acute}
Rebecca~H Berger, Alison~L Miller, Ronald Seifer, Stephanie~R Cares, and
  Monique~K LeBourgeois.
\newblock Acute sleep restriction effects on emotion responses in 30-to
  36-month-old children.
\newblock {\em Journal of sleep research}, 21(3):235--246, 2012.

\bibitem{bertero2016real}
Dario Bertero, Farhad~Bin Siddique, Chien-Sheng Wu, Yan Wan, Ricky Ho~Yin Chan,
  and Pascale Fung.
\newblock Real-time speech emotion and sentiment recognition for interactive
  dialogue systems.
\newblock In {\em Proceedings of the 2016 Conference on Empirical Methods in
  Natural Language Processing}, pages 1042--1047, 2016.

\bibitem{bianchi2002modeling}
Nadia Bianchi-Berthouze and Christine~L Lisetti.
\newblock Modeling multimodal expression of user’s affective subjective
  experience.
\newblock {\em User modeling and user-adapted interaction}, 12:49--84, 2002.

\bibitem{bolukbasi2016man}
Tolga Bolukbasi, Kai-Wei Chang, James~Y Zou, Venkatesh Saligrama, and Adam~T
  Kalai.
\newblock Man is to computer programmer as woman is to homemaker? debiasing
  word embeddings.
\newblock In {\em Advances in neural information processing systems}, 2016.

\bibitem{bradley1994measuring}
Margaret~M Bradley and Peter~J Lang.
\newblock Measuring emotion: the self-assessment manikin and the semantic
  differential.
\newblock {\em Journal of behavior therapy and experimental psychiatry},
  25(1):49--59, 1994.

\bibitem{bruno2017temperature}
Pascal Bruno, Valentyna Melnyk, and Franziska V{\"o}lckner.
\newblock Temperature and emotions: Effects of physical temperature on
  responses to emotional advertising.
\newblock {\em International Journal of Research in Marketing}, 34(1):302--320,
  2017.

\bibitem{buchanan2014acute}
Tony~W Buchanan, Jacqueline~S Laures-Gore, and Melissa~C Duff.
\newblock Acute stress reduces speech fluency.
\newblock {\em Biological psychology}, 97:60--66, 2014.

\bibitem{burmania2016tradeoff}
Alec Burmania, Mohammed Abdelwahab, and Carlos Busso.
\newblock Tradeoff between quality and quantity of emotional annotations to
  characterize expressive behaviors.
\newblock In {\em Acoustics, Speech and Signal Processing (ICASSP), 2016 IEEE
  International Conference on}, pages 5190--5194. Ieee, 2016.

\bibitem{burmania2017stepwise}
Alec Burmania and Carlos Busso.
\newblock A stepwise analysis of aggregated crowdsourced labels describing
  multimodal emotional behaviors.
\newblock In {\em Interspeech}, pages 152--156, 2017.

\bibitem{burmania2016increasing}
Alec Burmania, Srinivas Parthasarathy, and Carlos Busso.
\newblock Increasing the reliability of crowdsourcing evaluations using online
  quality assessment.
\newblock {\em IEEE Transactions on Affective Computing}, 7(4):374--388, 2016.

\bibitem{busso2008iemocap}
Carlos Busso, Murtaza Bulut, Chi-Chun Lee, Abe Kazemzadeh, Emily Mower, Samuel
  Kim, Jeannette~N Chang, Sungbok Lee, and Shrikanth~S Narayanan.
\newblock Iemocap: Interactive emotional dyadic motion capture database.
\newblock {\em Language resources and evaluation}, 42(4):335, 2008.

\bibitem{busso2009shrikanth}
Carlos Busso, Murtaza Bulut, and Sungbok Lee.
\newblock Shrikanth narayanan fundamental frequency analysis for speech emotion
  processing.
\newblock {\em The role of prosody in affective speech}, 97:309, 2009.

\bibitem{busso2017msp}
Carlos Busso, Srinivas Parthasarathy, Alec Burmania, Mohammed AbdelWahab,
  Najmeh Sadoughi, and Emily~Mower Provost.
\newblock Msp-improv: An acted corpus of dyadic interactions to study emotion
  perception.
\newblock {\em IEEE Transactions on Affective Computing}, 8(1):67--80, 2017.

\bibitem{cambria2017benchmarking}
Erik Cambria, Devamanyu Hazarika, Soujanya Poria, Amir Hussain, and RBV
  Subramanyam.
\newblock Benchmarking multimodal sentiment analysis.
\newblock In {\em International Conference on Computational Linguistics and
  Intelligent Text Processing}, pages 166--179. Springer, 2017.

\bibitem{cao2017realtime}
Zhe Cao, Tomas Simon, Shih-En Wei, and Yaser Sheikh.
\newblock Realtime multi-person 2d pose estimation using part affinity fields.
\newblock In {\em Cvpr}, 2017.

\bibitem{carlini2019secret}
Nicholas Carlini, Chang Liu, {\'U}lfar Erlingsson, Jernej Kos, and Dawn Song.
\newblock The secret sharer: Evaluating and testing unintended memorization in
  neural networks.
\newblock In {\em 28th $\{$USENIX\$\}\$}, 2019.

\bibitem{carlini2018audio}
Nicholas Carlini and David Wagner.
\newblock Audio adversarial examples: Targeted attacks on speech-to-text.
\newblock In {\em 2018 IEEE Security and Privacy Workshops (SPW)}.

\bibitem{cauldwell2000did}
Richard~T Cauldwell.
\newblock Where did the anger go? the role of context in interpreting emotion
  in speech.
\newblock In {\em ISCA Tutorial and Research Workshop (ITRW) on Speech and
  Emotion}, 2000.

\bibitem{chabanne2017privacy}
Herv{\'e} Chabanne, Amaury de~Wargny, Jonathan Milgram, Constance Morel, and
  Emmanuel Prouff.
\newblock Privacy-preserving classification on deep neural network.
\newblock {\em IACR Cryptology ePrint Archive}, 2017.

\bibitem{chaplin2015gender}
Tara~M Chaplin.
\newblock Gender and emotion expression: A developmental contextual
  perspective.
\newblock {\em Emotion Review}, 2015.

\bibitem{chatziagapi2019data}
Aggelina Chatziagapi, Georgios Paraskevopoulos, Dimitris Sgouropoulos, Georgios
  Pantazopoulos, Malvina Nikandrou, Theodoros Giannakopoulos, Athanasios
  Katsamanis, Alexandros Potamianos, and Shrikanth Narayanan.
\newblock Data augmentation using gans for speech emotion recognition.
\newblock In {\em Interspeech}, pages 171--175, 2019.

\bibitem{friends_dataset}
Sheng{-}Yeh Chen, Chao{-}Chun Hsu, Chuan{-}Chun Kuo, Ting{-}Hao~K. Huang, and
  Lun{-}Wei Ku.
\newblock Emotionlines: An emotion corpus of multi-party conversations.
\newblock {\em CoRR}, abs/1802.08379, 2018.

\bibitem{chen2018emotionlines}
Sheng-Yeh Chen, Chao-Chun Hsu, Chuan-Chun Kuo, Lun-Wei Ku, et~al.
\newblock Emotionlines: An emotion corpus of multi-party conversations.
\newblock {\em arXiv preprint arXiv:1802.08379}, 2018.

\bibitem{chen2021understanding}
Yunliang Chen and Jungseock Joo.
\newblock Understanding and mitigating annotation bias in facial expression
  recognition.
\newblock In {\em Proceedings of the IEEE/CVF International Conference on
  Computer Vision}, pages 14980--14991, 2021.

\bibitem{chenchah2016speech}
Farah Chenchah and Zied Lachiri.
\newblock Speech emotion recognition in noisy environment.
\newblock In {\em 2016 2nd International Conference on Advanced Technologies
  for Signal and Image Processing (ATSIP)}, pages 788--792. Ieee, 2016.

\bibitem{chollet2015}
Fran\c{c}ois Chollet.
\newblock keras.
\newblock \url{https://github.com/fchollet/keras}, 2015.

\bibitem{keras2015}
Fran\c{c}ois Chollet.
\newblock keras.
\newblock \url{https://github.com/fchollet/keras}, 2015.

\bibitem{clark2002using}
Herbert~H Clark and Jean E~Fox Tree.
\newblock Using uh and um in spontaneous speaking.
\newblock {\em Cognition}, 84(1):73--111, 2002.

\bibitem{clark2019makes}
Leigh Clark, Nadia Pantidi, Orla Cooney, Philip Doyle, Diego Garaialde, Justin
  Edwards, Brendan Spillane, Emer Gilmartin, Christine Murad, Cosmin Munteanu,
  et~al.
\newblock What makes a good conversation?: Challenges in designing truly
  conversational agents.
\newblock In {\em Proceedings of the 2019 CHI Conference on Human Factors in
  Computing Systems}, page 475. Acm, 2019.

\bibitem{coavoux2018privacy}
Maximin Coavoux, Shashi Narayan, and Shay~B Cohen.
\newblock Privacy-preserving neural representations of text.
\newblock {\em arXiv preprint arXiv:1808.09408}, 2018.

\bibitem{cohen1988perceived}
Sheldon Cohen.
\newblock Perceived stress in a probability sample of the united states.
\newblock 1988.

\bibitem{cohen1994perceived}
Sheldon Cohen, T~Kamarck, R~Mermelstein, et~al.
\newblock Perceived stress scale.
\newblock {\em Measuring stress: A guide for health and social scientists},
  pages 235--283, 1994.

\bibitem{cohen1983global}
Sheldon Cohen, Tom Kamarck, and Robin Mermelstein.
\newblock A global measure of perceived stress.
\newblock {\em Journal of health and social behavior}, pages 385--396, 1983.

\bibitem{corbett2018measure}
Sam Corbett-Davies and Sharad Goel.
\newblock The measure and mismeasure of fairness: A critical review of fair
  machine learning.
\newblock {\em arXiv preprint arXiv:1808.00023}, 2018.

\bibitem{dalpe2019personality}
Julien Dalp{\'e}, Martin Demers, J{\'e}r{\'e}mie Verner-Filion, and Robert~J
  Vallerand.
\newblock From personality to passion: The role of the big five factors.
\newblock {\em Personality and Individual Differences}, 138:280--285, 2019.

\bibitem{davidson2019racial}
Thomas Davidson, Debasmita Bhattacharya, and Ingmar Weber.
\newblock Racial bias in hate speech and abusive language detection datasets.
\newblock {\em arXiv preprint arXiv:1905.12516}, 2019.

\bibitem{dedora2011acute}
Daniel~J DeDora, Joshua~M Carlson, and Lilianne~R Mujica-Parodi.
\newblock Acute stress eliminates female advantage in detection of ambiguous
  negative affect.
\newblock {\em Evolutionary Psychology}, 9(4):147470491100900406, 2011.

\bibitem{devlin2018bert}
Jacob Devlin, Ming-Wei Chang, Kenton Lee, and Kristina Toutanova.
\newblock Bert: Pre-training of deep bidirectional transformers for language
  understanding.
\newblock {\em arXiv preprint arXiv:1810.04805}, 2018.

\bibitem{du2014compound}
Shichuan Du, Yong Tao, and Aleix~M Martinez.
\newblock Compound facial expressions of emotion.
\newblock {\em Proceedings of the National Academy of Sciences},
  111(15):E1454--e1462, 2014.

\bibitem{Duvall2014ExploringFW}
Emily~D. Duvall, Alan~Stuart Robbins, Thomas~R Graham, and Scott Divett.
\newblock Exploring filler words and their impact.
\newblock 2014.

\bibitem{Eckert2017AgeAA}
Penelope Eckert.
\newblock Age as a sociolinguistic variable.
\newblock 2017.

\bibitem{elazar2018adversarial}
Yanai Elazar and Yoav Goldberg.
\newblock Adversarial removal of demographic attributes from text data.
\newblock {\em arXiv preprint arXiv:1808.06640}, 2018.

\bibitem{engler2021auditing}
Alex Engler.
\newblock Auditing employment algorithms for discrimination.
\newblock {\em Brookings Institute, Center for Technology Innovation}, 2021.

\bibitem{evfimievski2002randomization}
Alexandre Evfimievski.
\newblock Randomization in privacy preserving data mining.
\newblock {\em ACM Sigkdd Explorations Newsletter}, 2002.

\bibitem{eyben2016geneva}
Florian Eyben, Klaus~R Scherer, Bj{\"o}rn~W Schuller, Johan Sundberg, Elisabeth
  Andr{\'e}, Carlos Busso, Laurence~Y Devillers, Julien Epps, Petri Laukka,
  Shrikanth~S Narayanan, et~al.
\newblock The geneva minimalistic acoustic parameter set (gemaps) for voice
  research and affective computing.
\newblock {\em IEEE Transactions on Affective Computing}, 7(2):190--202, 2016.

\bibitem{eyben2010opensmile}
Florian Eyben, Martin W{\"o}llmer, and Bj{\"o}rn Schuller.
\newblock Opensmile: the munich versatile and fast open-source audio feature
  extractor.
\newblock In {\em Proceedings of the 18th ACM international conference on
  Multimedia}, pages 1459--1462. Acm, 2010.

\bibitem{fernandez2013emotion}
Jos{\'e}-Miguel Fern{\'a}ndez-Dols and Carlos Crivelli.
\newblock Emotion and expression: Naturalistic studies.
\newblock {\em Emotion Review}, 5(1):24--29, 2013.

\bibitem{ferrario2019ai}
Andrea Ferrario, Michele Loi, and Eleonora Vigan{\`o}.
\newblock In ai we trust incrementally: a multi-layer model of trust to analyze
  human-artificial intelligence interactions.
\newblock {\em Philosophy \& Technology}, pages 1--17, 2019.

\bibitem{fitzpatrick2017delivering}
Kathleen~Kara Fitzpatrick, Alison Darcy, and Molly Vierhile.
\newblock Delivering cognitive behavior therapy to young adults with symptoms
  of depression and anxiety using a fully automated conversational agent
  (woebot): a randomized controlled trial.
\newblock {\em JMIR mental health}, 2017.

\bibitem{freeman2009comprehension}
Thomas~W Freeman, John Hart, Tim Kimbrell, and Elliott~D Ross.
\newblock Comprehension of affective prosody in veterans with chronic
  posttraumatic stress disorder.
\newblock {\em The Journal of neuropsychiatry and clinical neurosciences},
  21(1):52--58, 2009.

\bibitem{fussell2002verbal}
Susan~R Fussell.
\newblock The verbal communication of emotion: Introduction and overview.
\newblock In {\em The verbal communication of emotions}, pages 9--24.
  Psychology Press, 2002.

\bibitem{ganin2014unsupervised}
Yaroslav Ganin and Victor Lempitsky.
\newblock Unsupervised domain adaptation by backpropagation.
\newblock {\em arXiv preprint arXiv:1409.7495}, 2014.

\bibitem{Ganin2016DomainAdversarialTO}
Yaroslav Ganin, E.~Ustinova, Hana Ajakan, Pascal Germain, H.~Larochelle,
  Fran\c{c}ois Laviolette, Mario Marchand, and Victor~S. Lempitsky.
\newblock Domain-adversarial training of neural networks.
\newblock In {\em J. Mach. Learn. Res.}, 2016.

\bibitem{ganin2016domain}
Yaroslav Ganin, Evgeniya Ustinova, Hana Ajakan, Pascal Germain, Hugo
  Larochelle, Fran{\c{c}}ois Laviolette, Mario Marchand, and Victor Lempitsky.
\newblock Domain-adversarial training of neural networks.
\newblock {\em The Journal of Machine Learning Research}, 17(1):2096--2030,
  2016.

\bibitem{garbey2007contact}
Marc Garbey, Nanfei Sun, Arcangelo Merla, and Ioannis Pavlidis.
\newblock Contact-free measurement of cardiac pulse based on the analysis of
  thermal imagery.
\newblock {\em IEEE transactions on Biomedical Engineering}, 54(8):1418--1426,
  2007.

\bibitem{GarridoMuoz2021ASO}
Ismael Garrido-Mu{\~n}oz, A.~Montejo-R{\'a}ez, Fernando Mart{\'i}nez-Santiago,
  and L.~Alfonso Ure{\~n}a-L{\'o}pez.
\newblock A survey on bias in deep nlp.
\newblock volume~11, page 3184, 2021.

\bibitem{gemmeke2017audio}
Jort~F. Gemmeke, Daniel P.~W. Ellis, Dylan Freedman, Aren Jansen, et~al.
\newblock Audio set: An ontology and human-labeled dataset for audio events.
\newblock In {\em 2017 {IEEE} International Conference on Acoustics, Speech and
  Signal Processing, {ICASSP} 2017, New Orleans, LA, USA, March 5-9, 2017}.
  Ieee, 2017.

\bibitem{gideon2019barking}
John Gideon, Melvin~G McInnis, and Emily Mower~Provost.
\newblock Barking up the right tree: Improving cross-corpus speech emotion
  recognition with adversarial discriminative domain generalization (addog).
\newblock {\em arXiv preprint arXiv:1903.12094}, 2019.

\bibitem{gideon2019emotion}
John Gideon, Heather~T Schatten, Melvin~G McInnis, and Emily~Mower Provost.
\newblock Emotion recognition from natural phone conversations in individuals
  with and without recent suicidal ideation.
\newblock In {\em The 20th Annual Conference of the International Speech
  Communication Association INTERSPEECH 2019}, 2019.

\bibitem{giraud2013multimodal}
Tom Giraud, Mariette Soury, Jiewen Hua, Agnes Delaborde, Marie Tahon, David
  Antonio~Gomez Jauregui, Victoria Eyharabide, Edith Filaire, Christine
  Le~Scanff, Laurence Devillers, et~al.
\newblock Multimodal expressions of stress during a public speaking task:
  Collection, annotation and global analyses.
\newblock In {\em Humaine Association Conference on Affective Computing and
  Intelligent Interaction}. Ieee, 2013.

\bibitem{golbeck2011predicting}
Jennifer Golbeck, Cristina Robles, Michon Edmondson, and Karen Turner.
\newblock Predicting personality from twitter.
\newblock In {\em 2011 IEEE third international conference on privacy,
  security, risk and trust and 2011 IEEE third international conference on
  social computing}, pages 149--156. Ieee, 2011.

\bibitem{goldberg1992development}
Lewis~R Goldberg.
\newblock The development of markers for the big-five factor structure.
\newblock {\em Psychological assessment}, 4(1):26, 1992.

\bibitem{gomez2010data}
Jose~Maria Gomez-Hidalgo, Jose~Miguel Martin-Abreu, Javier Nieves, Igor Santos,
  Felix Brezo, and Pablo~G Bringas.
\newblock Data leak prevention through named entity recognition.
\newblock In {\em IEEE Second International Conference on Social Computing},
  2010.

\bibitem{gong2017crafting}
Yuan Gong and Christian Poellabauer.
\newblock Crafting adversarial examples for speech paralinguistics
  applications.
\newblock {\em arXiv preprint arXiv:1711.03280}, 2017.

\bibitem{griffiths2003iii}
Paul~E Griffiths.
\newblock Iii. basic emotions, complex emotions, machiavellian emotions 1.
\newblock {\em Royal Institute of Philosophy Supplements}, 52:39--67, 2003.

\bibitem{gulli2017deep}
Antonio Gulli and Sujit Pal.
\newblock {\em Deep Learning with Keras}.
\newblock Packt Publishing Ltd, 2017.

\bibitem{hajian2012study}
Sara Hajian and Josep Domingo-Ferrer.
\newblock A study on the impact of data anonymization on anti-discrimination.
\newblock In {\em 2012 IEEE 12th ICDM}.

\bibitem{Hansen2021AGS}
Lasse Hansen, Yan-Ping Zhang, Detlef Wolf, Konstantinos Sechidis, Nicolai
  Ladegaard, and Riccardo Fusaroli.
\newblock A generalizable speech emotion recognition model reveals depression
  and remission.
\newblock 2021.

\bibitem{herborn2015skin}
Katherine~A Herborn, James~L Graves, Paul Jerem, Neil~P Evans, Ruedi Nager,
  Dominic~J McCafferty, and Dorothy~EF McKeegan.
\newblock Skin temperature reveals the intensity of acute stress.
\newblock {\em Physiology \& behavior}, 152:225--230, 2015.

\bibitem{hoek2017evaluating}
Jet Hoek and Merel Scholman.
\newblock Evaluating discourse annotation: Some recent insights and new
  approaches.
\newblock In {\em Proceedings of the 13th Joint ISO-ACL Workshop on
  Interoperable Semantic Annotation (isa-13)}, 2017.

\bibitem{horvath1978experimental}
Frank Horvath.
\newblock An experimental comparison of the psychological stress evaluator and
  the galvanic skin response in detection of deception.
\newblock {\em Journal of Applied Psychology}, 63(3):338, 1978.

\bibitem{horvath1982detecting}
Frank Horvath.
\newblock Detecting deception: the promise and the reality of voice stress
  analysis.
\newblock {\em Journal of Forensic Science}, 27(2):340--351, 1982.

\bibitem{hsu2021robust}
Wei-Ning Hsu, Anuroop Sriram, Alexei Baevski, Tatiana Likhomanenko, Qiantong
  Xu, Vineel Pratap, Jacob Kahn, Ann Lee, Ronan Collobert, Gabriel Synnaeve,
  et~al.
\newblock Robust wav2vec 2.0: Analyzing domain shift in self-supervised
  pre-training.
\newblock {\em arXiv preprint arXiv:2104.01027}, 2021.

\bibitem{hsueh2009data}
Pei-Yun Hsueh, Prem Melville, and Vikas Sindhwani.
\newblock Data quality from crowdsourcing: a study of annotation selection
  criteria.
\newblock In {\em Proceedings of the NAACL HLT 2009 workshop on active learning
  for natural language processing}, pages 27--35. Association for Computational
  Linguistics, 2009.

\bibitem{hu2017frankenstein}
Guosheng Hu, Xiaojiang Peng, Yongxin Yang, Timothy~M Hospedales, and Jakob
  Verbeek.
\newblock Frankenstein: Learning deep face representations using small data.
\newblock {\em IEEE Transactions on Image Processing}, 27(1):293--303, 2017.

\bibitem{huang2018automatic}
Chenyang Huang, Osmar Zaiane, Amine Trabelsi, and Nouha Dziri.
\newblock Automatic dialogue generation with expressed emotions.
\newblock In {\em Proceedings of NAACL}, 2018.

\bibitem{3-D-motion-estimation-and-object-tracking-using-B-spline-curve-modeling}
Z.~{Huang} and F.~S. {Cohen}.
\newblock 3-d motion estimation and object tracking using b-spline curve
  modeling.
\newblock In {\em Proceedings of IEEE Conference on Computer Vision and Pattern
  Recognition}, pages 748--749, June 1993.

\bibitem{iachello2007end}
Giovanni Iachello and Jason Hong.
\newblock {\em End-user privacy in human-computer interaction}, volume~1.
\newblock Now Publishers Inc, 2007.

\bibitem{iosifidis2018dealing}
Vasileios Iosifidis and Eirini Ntoutsi.
\newblock Dealing with bias via data augmentation in supervised learning
  scenarios.
\newblock {\em Jo Bates Paul D. Clough Robert J{\"a}schke}, 24, 2018.

\bibitem{jaiswal2019muse}
Mimansa Jaiswal, Zakaria Aldeneh, Cristian-Paul Bara, Yuanhang Luo, Mihai
  Burzo, Rada Mihalcea, and Emily Mower~Provost.
\newblock Muse-ing on the impact of utterance ordering on crowdsourced emotion
  annotations.
\newblock In {\em 2019 IEEE International Conference on Acoustics, Speech and
  Signal Processing (ICASSP)}. Ieee, 2019.

\bibitem{muse-ing-icassp}
Mimansa Jaiswal, Zakaria Aldeneh, Cristian{-}Paul Bara, Yuanhang Luo, Mihai
  Burzo, Rada Mihalcea, and Emily~Mower Provost.
\newblock Muse-ing on the impact of utterance ordering on crowdsourced emotion
  annotations.
\newblock {\em CoRR}, abs/1903.11672, 2019.

\bibitem{mimansa2019controlling}
Mimansa Jaiswal, Zakaria Aldeneh, and Emily Mower~Provost.
\newblock Controlling for confounders in multimodal emotion classification via
  adversarial learning.
\newblock {\em arXiv preprint arXiv:1908.08979}, 2019.

\bibitem{jaiswal2020privacy}
Mimansa Jaiswal and Emily~Mower Provost.
\newblock Privacy enhanced multimodal neural representations for emotion
  recognition.
\newblock In {\em Aaai}, pages 7985--7993, 2020.

\bibitem{Jaiswal2021BestPF}
Mimansa Jaiswal and Emily~Mower Provost.
\newblock Best practices for noise-based augmentation to improve the
  performance of emotion recognition "in the wild".
\newblock volume abs/2104.08806, 2021.

\bibitem{jaiswal2016truth}
Mimansa Jaiswal, Sairam Tabibu, and Rajiv Bajpai.
\newblock The truth and nothing but the truth: Multimodal analysis for
  deception detection.
\newblock In {\em 2016 IEEE 16th International Conference on Data Mining
  Workshops (ICDMW)}, pages 938--943. Ieee, 2016.

\bibitem{jiang2014predicting}
Yu-Gang Jiang, Baohan Xu, and Xiangyang Xue.
\newblock Predicting emotions in user-generated videos.
\newblock In {\em Twenty-Eighth AAAI Conference on Artificial Intelligence},
  2014.

\bibitem{joseph2017constance}
Kenneth Joseph, Lisa Friedland, William Hobbs, Oren Tsur, and David Lazer.
\newblock Constance: Modeling annotation contexts to improve stance
  classification.
\newblock {\em arXiv preprint arXiv:1708.06309}, 2017.

\bibitem{jurgens2015effect}
Rebecca J{\"u}rgens, Annika Grass, Matthis Drolet, and Julia Fischer.
\newblock Effect of acting experience on emotion expression and recognition in
  voice: Non-actors provide better stimuli than expected.
\newblock {\em Journal of nonverbal behavior}, 39(3):195--214, 2015.

\bibitem{kahn2007measuring}
Jeffrey~H Kahn, Renee~M Tobin, Audra~E Massey, and Jennifer~A Anderson.
\newblock Measuring emotional expression with the linguistic inquiry and word
  count.
\newblock {\em The American journal of psychology}, pages 263--286, 2007.

\bibitem{Kaushik2021OnTE}
Divyansh Kaushik, Douwe Kiela, Zachary~Chase Lipton, and Wen tau Yih.
\newblock On the efficacy of adversarial data collection for question
  answering: Results from a large-scale randomized study.
\newblock volume abs/2106.00872, 2021.

\bibitem{khorram2017capturing}
Soheil Khorram, Zakaria Aldeneh, Dimitrios Dimitriadis, Melvin McInnis, and
  Emily~Mower Provost.
\newblock Capturing long-term temporal dependencies with convolutional networks
  for continuous emotion recognition.
\newblock {\em Proc. Interspeech}, 2017.

\bibitem{Khorram2018}
Soheil Khorram, Mimansa Jaiswal, John Gideon, Melvin McInnis, and Emily {Mower
  Provost}.
\newblock The priori emotion dataset: Linking mood to emotion detected
  in-the-wild.
\newblock In {\em Interspeech 2018}.

\bibitem{khorram2018priori}
Soheil Khorram, Mimansa Jaiswal, John Gideon, Melvin McInnis, and Emily~Mower
  Provost.
\newblock The priori emotion dataset: Linking mood to emotion detected
  in-the-wild.
\newblock {\em arXiv preprint arXiv:1806.10658}, 2018.

\bibitem{kidd2016determining}
Gerald Kidd~Jr, Christine~R Mason, Jayaganesh Swaminathan, Elin Roverud,
  Kameron~K Clayton, and Virginia Best.
\newblock Determining the energetic and informational components of
  speech-on-speech masking.
\newblock {\em The Journal of the Acoustical Society of America}, 2016.

\bibitem{kifer2011no}
Daniel Kifer and Ashwin Machanavajjhala.
\newblock No free lunch in data privacy.
\newblock In {\em Proceedings of the ACM SIGMOD International Conference on
  Management of data}, 2011.

\bibitem{kim2014convolutional}
Yoon Kim.
\newblock Convolutional neural networks for sentence classification.
\newblock {\em arXiv preprint arXiv:1408.5882}, 2014.

\bibitem{kingston2016life}
Cara Kingston and James Schuurmans-Stekhoven.
\newblock Life hassles and delusional ideation: Scoping the potential role of
  cognitive and affective mediators.
\newblock {\em Psychology and Psychotherapy: Theory, Research and Practice},
  89(4):445--463, 2016.

\bibitem{kirschbaum1993trier}
Clemens Kirschbaum, Karl-Martin Pirke, and Dirk~H Hellhammer.
\newblock The `trier social stress test'--a tool for investigating
  psychobiological stress responses in a laboratory setting.
\newblock {\em Neuropsychobiology}, 28(1-2):76--81, 1993.

\bibitem{kokhlikyan2020captum}
Narine Kokhlikyan, Vivek Miglani, Miguel Martin, Edward Wang, Bilal Alsallakh,
  Jonathan Reynolds, Alexander Melnikov, Natalia Kliushkina, Carlos Araya, Siqi
  Yan, et~al.
\newblock Captum: A unified and generic model interpretability library for
  pytorch.
\newblock {\em arXiv preprint arXiv:2009.07896}, 2020.

\bibitem{kolavr2008automatic}
J{\'a}chym Kol{\'a}{\v{r}}.
\newblock {\em Automatic Segmentation of Speech into Sentence-like Units}.
\newblock PhD thesis, University of West Bohemia in Pilsen, 2008.

\bibitem{krishna2018study}
Kalpesh Krishna, Liang Lu, Kevin Gimpel, and Karen Livescu.
\newblock A study of all-convolutional encoders for connectionist temporal
  classification.
\newblock In {\em 2018 IEEE International Conference on Acoustics, Speech and
  Signal Processing (ICASSP)}, pages 5814--5818. Ieee, 2018.

\bibitem{kurniawan2013stress}
Hindra Kurniawan, Alexandr~V Maslov, and Mykola Pechenizkiy.
\newblock Stress detection from speech and galvanic skin response signals.
\newblock In {\em Proceedings of the 26th IEEE International Symposium on
  Computer-Based Medical Systems}, pages 209--214. Ieee, 2013.

\bibitem{landeiro2016robust}
Virgile Landeiro and Aron Culotta.
\newblock Robust text classification in the presence of confounding bias.
\newblock In {\em Thirtieth AAAI Conference on Artificial Intelligence}, 2016.

\bibitem{laplante2003things}
Debi Laplante and Nalini Ambady.
\newblock On how things are said: Voice tone, voice intensity, verbal content,
  and perceptions of politeness.
\newblock {\em Journal of Language and Social Psychology}, 22(4):434--441,
  2003.

\bibitem{Lassalle2019}
Amandine Lassalle, Delia Pigat, Helen O'Reilly, Steve Berggen, Shimrit
  Fridenson-Hayo, Shahar Tal, Sigrid Elfstr{\"o}m, Anna R{\aa}de, Ofer Golan,
  Sven B{\"o}lte, Simon Baron-Cohen, and Daniel Lundqvist.
\newblock The eu-emotion voice database.
\newblock {\em Behavior Research Methods}, 51(2):493--506, Apr 2019.

\bibitem{lazarus1977environmental}
Richard~S Lazarus and Judith~Blackfield Cohen.
\newblock Environmental stress.
\newblock In {\em Human behavior and environment}, pages 89--127. Springer,
  1977.

\bibitem{lech2014stress}
Margaret Lech and Ling He.
\newblock Stress and emotion recognition using acoustic speech analysis.
\newblock In {\em Mental Health Informatics}, pages 163--184. Springer, 2014.

\bibitem{lee2011apparatus}
Mihee Lee, Seokwon Bang, and Gyunghye Yang.
\newblock Apparatus and method for inducing emotions, June~7 2011.
\newblock US Patent 7,955,259.

\bibitem{levitan2016automatic}
Sarah~Ita Levitan, Taniya Mishra, and Srinivas Bangalore.
\newblock Automatic identification of gender from speech.
\newblock In {\em Proceeding of Speech Prosody}, pages 84--88, 2016.

\bibitem{li2014overview}
Jinyu Li, Li~Deng, Yifan Gong, and Reinhold Haeb-Umbach.
\newblock An overview of noise-robust automatic speech recognition.
\newblock {\em IEEE/ACM Transactions on Audio, Speech, and Language
  Processing}, 2014.

\bibitem{li2013membership}
Ninghui Li, Wahbeh Qardaji, Dong Su, Yi~Wu, and Weining Yang.
\newblock Membership privacy: a unifying framework for privacy definitions.
\newblock In {\em Proceedings of the 2013 ACM SIGSAC conference on Computer \&
  communications security}. Acm.

\bibitem{Li2021ControllableET}
Tao Li, Shan Yang, Liumeng Xue, and Lei Xie.
\newblock Controllable emotion transfer for end-to-end speech synthesis.
\newblock pages 1--5, 2021.

\bibitem{li2016towards}
Wei Li, Farnaz Abtahi, Christina Tsangouri, and Zhigang Zhu.
\newblock Towards an ``in-the-wild'' emotion dataset using a game-based
  framework.
\newblock In {\em 2016 IEEE Conference on Computer Vision and Pattern
  Recognition Workshops (CVPRW)}, pages 1526--1534. Ieee, 2016.

\bibitem{liao2015laboratory}
Li-Mei Liao and Mary~G Carey.
\newblock Laboratory-induced mental stress, cardiovascular response, and
  psychological characteristics.
\newblock {\em Reviews in cardiovascular medicine}, 16(1):28--35, 2015.

\bibitem{lichtenauer2011mahnob}
JEROEN Lichtenauer and MOHAMMAD Soleymani.
\newblock Mahnob-hci-tagging database, 2011.

\bibitem{liu2016classification}
Tongliang Liu and Dacheng Tao.
\newblock Classification with noisy labels by importance reweighting.
\newblock {\em IEEE Transactions on pattern analysis and machine intelligence},
  38(3):447--461, 2016.

\bibitem{locatello2018challenging}
Francesco Locatello, Stefan Bauer, Mario Lucic, Sylvain Gelly, Bernhard
  Sch{\"o}lkopf, and Olivier Bachem.
\newblock Challenging common assumptions in the unsupervised learning of
  disentangled representations.
\newblock {\em arXiv preprint arXiv:1811.12359}, 2018.

\bibitem{lotfian2017building}
Reza Lotfian and Carlos Busso.
\newblock Building naturalistic emotionally balanced speech corpus by
  retrieving emotional speech from existing podcast recordings.
\newblock {\em IEEE Transactions on Affective Computing}, 2017.

\bibitem{lucas2014s}
Gale~M Lucas, Jonathan Gratch, Aisha King, and Louis-Philippe Morency.
\newblock It's only a computer: Virtual humans increase willingness to
  disclose.
\newblock {\em Computers in Human Behavior}, 37:94--100, 2014.

\bibitem{ma2015human}
Weiyi Ma and William~Forde Thompson.
\newblock Human emotions track changes in the acoustic environment.
\newblock {\em Proceedings of the National Academy of Sciences},
  112(47):14563--14568, 2015.

\bibitem{mangalam2017learning}
Karttikeya Mangalam and Tanaya Guha.
\newblock Learning spontaneity to improve emotion recognition in speech.
\newblock {\em arXiv preprint arXiv:1712.04753}, 2017.

\bibitem{matsumoto1989cultural}
David Matsumoto.
\newblock Cultural influences on the perception of emotion.
\newblock {\em Journal of Cross-Cultural Psychology}, 20(1):92--105, 1989.

\bibitem{matton2019into}
Katie Matton, Melvin~G McInnis, and Emily~Mower Provost.
\newblock Into the wild: Transitioning from recognizing mood in clinical
  interactions to personal conversations for individuals with bipolar disorder.
\newblock {\em Proc. Interspeech 2019}, pages 1438--1442, 2019.

\bibitem{satire-adv}
Robert McHardy, Heike Adel, and Roman Klinger.
\newblock Adversarial training for satire detection: Controlling for
  confounding variables.
\newblock {\em CoRR}, abs/1902.11145, 2019.

\bibitem{mchardy2019adversarial}
Robert McHardy, Heike Adel, and Roman Klinger.
\newblock Adversarial training for satire detection: Controlling for
  confounding variables.
\newblock {\em arXiv preprint arXiv:1902.11145}, 2019.

\bibitem{mehl2017natural}
Matthias~R Mehl, Charles~L Raison, Thaddeus~WW Pace, Jesusa~MG Arevalo, and
  Steve~W Cole.
\newblock Natural language indicators of differential gene regulation in the
  human immune system.
\newblock {\em Proceedings of the National Academy of Sciences},
  114(47):12554--12559, 2017.

\bibitem{mendelson2021debiasing}
Michael Mendelson and Yonatan Belinkov.
\newblock Debiasing methods in natural language understanding make bias more
  accessible.
\newblock {\em arXiv preprint arXiv:2109.04095}, 2021.

\bibitem{meng2018speaker}
Zhong Meng, Jinyu Li, Zhuo Chen, Yang Zhao, Vadim Mazalov, Yifan Gang, and
  Biing-Hwang Juang.
\newblock Speaker-invariant training via adversarial learning.
\newblock In {\em 2018 IEEE International Conference on Acoustics, Speech and
  Signal Processing (ICASSP)}, pages 5969--5973. Ieee, 2018.

\bibitem{metcalf2019mirroring}
Katherine Metcalf, Barry-John Theobald, Garrett Weinberg, Robert Lee, Ing-Marie
  Jonsson, Russ Webb, and Nicholas Apostoloff.
\newblock Mirroring to build trust in digital assistants.
\newblock {\em arXiv preprint arXiv:1904.01664}, 2019.

\bibitem{mikolov2013distributed}
Tomas Mikolov, Ilya Sutskever, Kai Chen, Greg~S Corrado, and Jeff Dean.
\newblock Distributed representations of words and phrases and their
  compositionality.
\newblock In {\em Advances in neural information processing systems}, pages
  3111--3119, 2013.

\bibitem{mills2014validity}
Caitlin Mills and Sidney D'Mello.
\newblock On the validity of the autobiographical emotional memory task for
  emotion induction.
\newblock {\em PloS one}, 9(4):e95837, 2014.

\bibitem{mirsamadi2017automatic}
Seyedmahdad Mirsamadi, Emad Barsoum, and Cha Zhang.
\newblock Automatic speech emotion recognition using recurrent neural networks
  with local attention.
\newblock In {\em Acoustics, Speech and Signal Processing (ICASSP), 2017 IEEE
  International Conference on}, pages 2227--2231. Ieee, 2017.

\bibitem{mitchell2006relationship}
Lyndon~A Mitchell.
\newblock The relationship between emotional recognition and personality
  traits.
\newblock 2006.

\bibitem{mohammad2018obtaining}
Saif Mohammad.
\newblock Obtaining reliable human ratings of valence, arousal, and dominance
  for 20,000 english words.
\newblock In {\em Proceedings of the 56th Annual Meeting of the Association for
  Computational Linguistics (Volume 1: Long Papers)}, pages 174--184, 2018.

\bibitem{mohammad2022ethics}
Saif~M Mohammad.
\newblock Ethics sheet for automatic emotion recognition and sentiment
  analysis.
\newblock {\em Computational Linguistics}, 48(2):239--278, 2022.

\bibitem{monin2012linguistic}
Joan~K Monin, Richard Schulz, Edward~P Lemay~Jr, and Thomas~B Cook.
\newblock Linguistic markers of emotion regulation and cardiovascular
  reactivity among older caregiving spouses.
\newblock {\em Psychology and aging}, 27(4):903, 2012.

\bibitem{nasir2017predicting}
Md~Nasir, Brian~Robert Baucom, Panayiotis Georgiou, and Shrikanth Narayanan.
\newblock Predicting couple therapy outcomes based on speech acoustic features.
\newblock {\em PloS one}, 12(9):e0185123, 2017.

\bibitem{newman2008gender}
Matthew~L Newman, Carla~J Groom, Lori~D Handelman, and James~W Pennebaker.
\newblock Gender differences in language use: An analysis of 14,000 text
  samples.
\newblock {\em Discourse Processes}, 45(3):211--236, 2008.

\bibitem{newman2003lying}
Matthew~L Newman, James~W Pennebaker, Diane~S Berry, and Jane~M Richards.
\newblock Lying words: Predicting deception from linguistic styles.
\newblock {\em Personality and social psychology bulletin}, 29(5):665--675,
  2003.

\bibitem{ngo2015use}
Nhi Ngo and Derek~M Isaacowitz.
\newblock Use of context in emotion perception: The role of top-down control,
  cue type, and perceiver's age.
\newblock {\em Emotion}, 15(3):292, 2015.

\bibitem{nosek2005understanding}
Brian~A Nosek, Anthony~G Greenwald, and Mahzarin~R Banaji.
\newblock Understanding and using the implicit association test: Ii. method
  variables and construct validity.
\newblock {\em Personality and Social Psychology Bulletin}, 31(2):166--180,
  2005.

\bibitem{oikonomopoulos2008human}
A~Oikonomopoulos, M~Pantic, and I~Patras.
\newblock Human gesture recognition using sparse b-spline polynomial
  representations.
\newblock In {\em Proceedings of Belgium-Netherlands Conf. Artificial
  Intelligence (BNAIC 2008), Boekelo, The Netherlands}, pages 193--200, 2008.

\bibitem{papernot2018marauder}
Nicolas Papernot.
\newblock A marauder's map of security and privacy in machine learning.
\newblock {\em arXiv preprint arXiv:1811.01134}, 2018.

\bibitem{pappagari2021copypaste}
Raghavendra Pappagari, Jes{\'u}s Villalba, Piotr {\.Z}elasko, Laureano
  Moro-Velazquez, and Najim Dehak.
\newblock Copypaste: An augmentation method for speech emotion recognition.
\newblock In {\em IEEE International Conference on Acoustics, Speech and Signal
  Processing (ICASSP)}, 2021.

\bibitem{parada-cabaleiro2017the}
Emilia Parada{-}Cabaleiro, Alice Baird, Anton Batliner, Nicholas Cummins,
  et~al.
\newblock The perception of emotions in noisified nonsense speech.
\newblock In Francisco Lacerda, editor, {\em Interspeech 2017, 18th Annual
  Conference of the International Speech Communication Association, Stockholm,
  Sweden, August 20-24, 2017}. Isca, 2017.

\bibitem{paulmann2016psychological}
Silke Paulmann, Desire Furnes, Anne~Ming B{\o}kenes, and Philip~J Cozzolino.
\newblock How psychological stress affects emotional prosody.
\newblock {\em Plos one}, 11(11):e0165022, 2016.

\bibitem{pavlidis2002thermal}
Ioannis Pavlidis and James Levine.
\newblock Thermal image analysis for polygraph testing.
\newblock {\em IEEE Engineering in Medicine and Biology Magazine},
  21(6):56--64, 2002.

\bibitem{pavlidis2000thermal}
Ioannis Pavlidis, James Levine, and Paulette Baukol.
\newblock Thermal imaging for anxiety detection.
\newblock In {\em Proceedings IEEE Workshop on Computer Vision Beyond the
  Visible Spectrum: Methods and Applications (Cat. No. PR00640)}, pages
  104--109. Ieee, 2000.

\bibitem{pennebaker2001linguistic}
James~W Pennebaker, Martha~E Francis, and Roger~J Booth.
\newblock Linguistic inquiry and word count: Liwc 2001.
\newblock {\em Mahway: Lawrence Erlbaum Associates}, 71(2001):2001, 2001.

\bibitem{pepino2021emotion}
Leonardo Pepino, Pablo Riera, and Luciana Ferrer.
\newblock Emotion recognition from speech using wav2vec 2.0 embeddings.
\newblock In Hynek Hermansky, Honza Cernock{\'{y}}, Luk{\'{a}}s Burget, Lori
  Lamel, Odette Scharenborg, and Petr Motl{\'{\i}}cek, editors, {\em
  Interspeech 2021, 22nd Annual Conference of the International Speech
  Communication Association, Brno, Czechia, 30 August - 3 September 2021}.
  Isca, 2021.

\bibitem{piczak2015dataset}
Karol~J. Piczak.
\newblock Esc: Dataset for environmental sound classification.
\newblock In {\em Proceedings of the 23rd {Annual ACM Conference} on
  {Multimedia}}.

\bibitem{piersol2019pre}
Kurt~Wesley Piersol and Gabriel Beddingfield.
\newblock Pre-wakeword speech processing, January~29 2019.
\newblock US Patent App. 14/672,277.

\bibitem{poria2019emotion}
Soujanya Poria, Navonil Majumder, Rada Mihalcea, and Eduard Hovy.
\newblock Emotion recognition in conversation: Research challenges, datasets,
  and recent advances.
\newblock {\em IEEE Access}, 7:100943--100953, 2019.

\bibitem{Pruthi2022EvaluatingEH}
Danish Pruthi, Bhuwan Dhingra, Livio~Baldini Soares, Michael Collins,
  Zachary~Chase Lipton, Graham Neubig, and William~W. Cohen.
\newblock Evaluating explanations: How much do explanations from the teacher
  aid students?
\newblock volume~10, pages 359--375, 2022.

\bibitem{qiao2017transient}
Emilie Qiao-Tasserit, Maria~Garcia Quesada, Lia Antico, Daphne Bavelier, Patrik
  Vuilleumier, and Swann Pichon.
\newblock Transient emotional events and individual affective traits affect
  emotion recognition in a perceptual decision-making task.
\newblock {\em PloS one}, 12(2):e0171375, 2017.

\bibitem{querengasser2014sad}
Jan Quereng{\"a}sser and Sebastian Schindler.
\newblock Sad but true?-how induced emotional states differentially bias
  self-rated big five personality traits.
\newblock {\em BMC Psychology}, 2(1):14, 2014.

\bibitem{ren2021survey}
Pengzhen Ren, Yun Xiao, Xiaojun Chang, Po-Yao Huang, Zhihui Li, Brij~B Gupta,
  Xiaojiang Chen, and Xin Wang.
\newblock A survey of deep active learning.
\newblock {\em ACM computing surveys (CSUR)}, 54(9):1--40, 2021.

\bibitem{ringeval2013introducing}
Fabien Ringeval, Andreas Sonderegger, Juergen Sauer, and Denis Lalanne.
\newblock Introducing the recola multimodal corpus of remote collaborative and
  affective interactions.
\newblock In {\em Automatic Face and Gesture Recognition (FG), 2013 10th IEEE
  International Conference and Workshops on}, pages 1--8. Ieee, 2013.

\bibitem{rosenberg2012classifying}
Andrew Rosenberg.
\newblock Classifying skewed data: Importance weighting to optimize average
  recall.
\newblock In {\em Thirteenth Annual Conference of the International Speech
  Communication Association}, 2012.

\bibitem{rothkrantz2004voice}
Leon~JM Rothkrantz, Pascal Wiggers, Jan-Willem~A Van~Wees, and Robert~J van
  Vark.
\newblock Voice stress analysis.
\newblock In {\em International conference on text, speech and dialogue}, pages
  449--456. Springer, 2004.

\bibitem{rubo2018social}
Marius Rubo and Matthias Gamer.
\newblock Social content and emotional valence modulate gaze fixations in
  dynamic scenes.
\newblock {\em Scientific reports}, 8(1):3804, 2018.

\bibitem{ruder2017overview}
Sebastian Ruder.
\newblock An overview of multi-task learning in deep neural networks.
\newblock {\em arXiv preprint arXiv:1706.05098}, 2017.

\bibitem{russell2017emotion}
James~A Russell.
\newblock Emotion recognition: Is it universal?
\newblock 2017.

\bibitem{samson2016eliciting}
Andrea~C Samson, Sylvia~D Kreibig, Blake Soderstrom, A~Ayanna Wade, and James~J
  Gross.
\newblock Eliciting positive, negative and mixed emotional states: A film
  library for affective scientists.
\newblock {\em Cognition and emotion}, 30(5):827--856, 2016.

\bibitem{sarma2018emotion}
Mousmita Sarma, Pegah Ghahremani, Daniel Povey, Nagendra~Kumar Goel,
  Kandarpa~Kumar Sarma, and Najim Dehak.
\newblock Emotion identification from raw speech signals using dnns.
\newblock {\em Proc. Interspeech 2018}, pages 3097--3101, 2018.

\bibitem{scharenborg2018the}
Odette Scharenborg, Sofoklis Kakouros, and Jiska Koemans.
\newblock The effect of noise on emotion perception in an unknown language.
\newblock In {\em Speech Prosody 2018}. Isca, 2018.

\bibitem{Schick2021SelfDiagnosisAS}
Timo Schick, Sahana Udupa, and Hinrich Sch{\"u}tze.
\newblock Self-diagnosis and self-debiasing: A proposal for reducing
  corpus-based bias in nlp.
\newblock volume~9, pages 1408--1424, 2021.

\bibitem{schlotz2011perceived}
Wolff Schlotz, Ilona~S Yim, Peggy~M Zoccola, Lars Jansen, and Peter Schulz.
\newblock The perceived stress reactivity scale: Measurement invariance,
  stability, and validity in three countries.
\newblock {\em Psychological assessment}, 23(1):80, 2011.

\bibitem{schmidt2018adversarially}
Ludwig Schmidt, Shibani Santurkar, Dimitris Tsipras, Kunal Talwar, and
  Aleksander Madry.
\newblock Adversarially robust generalization requires more data.
\newblock In {\em Advances in NIPS}, 2018.

\bibitem{schmidt2019quantifying}
Philipp Schmidt and Felix Biessmann.
\newblock Quantifying interpretability and trust in machine learning systems.
\newblock {\em arXiv preprint arXiv:1901.08558}, 2019.

\bibitem{shinohara2016adversarial}
Yusuke Shinohara.
\newblock Adversarial multi-task learning of deep neural networks for robust
  speech recognition.
\newblock In {\em Interspeech}, pages 2369--2372. San Francisco, CA, USA, 2016.

\bibitem{Shparberg2021GoogleBN}
Anna~L. Shparberg.
\newblock Google books ngram viewer.
\newblock 2021.

\bibitem{siedlecka2019experimental}
Ewa Siedlecka and Thomas~F Denson.
\newblock Experimental methods for inducing basic emotions: A qualitative
  review.
\newblock {\em Emotion Review}, 11(1):87--97, 2019.

\bibitem{simon2017hand}
Tomas Simon, Hanbyul Joo, Iain Matthews, and Yaser Sheikh.
\newblock Hand keypoint detection in single images using multiview
  bootstrapping.
\newblock In {\em Cvpr}, 2017.

\bibitem{sohn2020fixmatch}
Kihyuk Sohn, David Berthelot, Chun-Liang Li, Zizhao Zhang, Nicholas Carlini,
  Ekin~D Cubuk, Alex Kurakin, Han Zhang, and Colin Raffel.
\newblock Fixmatch: Simplifying semi-supervised learning with consistency and
  confidence.
\newblock {\em arXiv preprint arXiv:2001.07685}, 2020.

\bibitem{soleymani2010crowdsourcing}
Mohammad Soleymani and Martha Larson.
\newblock Crowdsourcing for affective annotation of video: Development of a
  viewer-reported boredom corpus.
\newblock {\em SIGIR-Workshops}, 2010.

\bibitem{soto2009emotion}
Jose~Angel Soto and Robert~W Levenson.
\newblock Emotion recognition across cultures: The influence of ethnicity on
  empathic accuracy and physiological linkage.
\newblock {\em Emotion}, 9(6):874, 2009.

\bibitem{sperrle2019human}
Fabian Sperrle, Udo Schlegel, Mennatallah El-Assady, and Daniel Keim.
\newblock Human trust modeling for bias mitigation in artificial intelligence.
\newblock In {\em ACM CHI 2019 Workshop: Where is the Human? Bridging the Gap
  Between AI and HCI}, 2019.

\bibitem{staats1990paradigmatic}
Arthur~W Staats and Georg~H Eifert.
\newblock The paradigmatic behaviorism theory of emotions: Basis for
  unification.
\newblock {\em Clinical Psychology Review}, 10(5):539--566, 1990.

\bibitem{steeneken1999speech}
Herman~JM Steeneken and John~HL Hansen.
\newblock Speech under stress conditions: overview of the effect on speech
  production and on system performance.
\newblock In {\em 1999 IEEE International Conference on Acoustics, Speech, and
  Signal Processing. Proceedings. ICASSP99 (Cat. No. 99CH36258)}, volume~4,
  pages 2079--2082. Ieee, 1999.

\bibitem{stenback2016speech}
Victoria Stenb{\"a}ck.
\newblock {\em Speech masking speech in everyday communication: The role of
  inhibitory control and working memory capacity}, volume 1559.
\newblock Link{\"o}ping University Electronic Press, 2016.

\bibitem{Strapparava07}
Carlo Strapparava and Rada Mihalcea.
\newblock Semeval-2007 task 14: Affective text.
\newblock In {\em Proceedings of the 4th International Workshop on the Semantic
  Evaluations (SemEval 2007)}, Prague, Czech Republic, 2007.

\bibitem{stromfelt2017emotion}
Harald Str{\"o}mfelt, Yue Zhang, and Bj{\"o}rn~W Schuller.
\newblock Emotion-augmented machine learning: Overview of an emerging domain.
\newblock In {\em 2017 Seventh International Conference on Affective Computing
  and Intelligent Interaction (ACII)}, pages 305--312. Ieee, 2017.

\bibitem{subbaswamy2018learning}
Adarsh Subbaswamy, Peter Schulam, and Suchi Saria.
\newblock Learning predictive models that transport.
\newblock {\em arXiv preprint arXiv:1812.04597}, 2018.

\bibitem{Activity-Aware-Mental-Stress-Detection-Using-Physiological-Sensors}
Feng-Tso Sun, Cynthia Kuo, Heng-Tze Cheng, Senaka Buthpitiya, Patricia Collins,
  and Martin Griss.
\newblock Activity-aware mental stress detection using physiological sensors.
\newblock In Martin Gris and Guang Yang, editors, {\em Mobile Computing,
  Applications, and Services}, pages 282--301, Berlin, Heidelberg, 2012.
  Springer Berlin Heidelberg.

\bibitem{sundararajan2017axiomatic}
Mukund Sundararajan, Ankur Taly, and Qiqi Yan.
\newblock Axiomatic attribution for deep networks.
\newblock {\em arXiv preprint arXiv:1703.01365}, 2017.

\bibitem{Tausczik2010ThePM}
Yla~R. Tausczik and James~W. Pennebaker.
\newblock The psychological meaning of words: Liwc and computerized text
  analysis methods.
\newblock volume~29, pages 24--54, 2010.

\bibitem{thompson2013students}
E~Heather Thompson, Phyllis Robertson, Russ Curtis, and Melodie~H Frick.
\newblock Students with anxiety: Implications for professional school
  counselors.
\newblock {\em Professional School Counseling}, 16(4):2156759x150160402, 2013.

\bibitem{rmsprop}
T.~Tieleman and G.~Hinton.
\newblock Lecture 6.5---rmsprop: Divide the gradient by a running average of
  its recent magnitude.
\newblock COURSERA: Neural Networks for Machine Learning, 2012.

\bibitem{tulen1989characterization}
JHM Tulen, P~Moleman, HG~Van~Steenis, and F~Boomsma.
\newblock Characterization of stress reactions to the stroop color word test.
\newblock {\em Pharmacology Biochemistry and Behavior}, 32(1):9--15, 1989.

\bibitem{tull2007preliminary}
Matthew~T Tull, Heidi~M Barrett, Elaine~S McMillan, and Lizabeth Roemer.
\newblock A preliminary investigation of the relationship between emotion
  regulation difficulties and posttraumatic stress symptoms.
\newblock {\em Behavior Therapy}, 38(3):303--313, 2007.

\bibitem{Valente2021ImprovingTC}
Francisco Valente, Sim{\~a}o Paredes, and Jorge Henriques.
\newblock Improving the compromise between accuracy, interpretability and
  personalization of rule-based machine learning in medical problems.
\newblock pages 2132--2135, 2021.

\bibitem{valin2018hybrid}
Jean-Marc Valin.
\newblock A hybrid dsp/deep learning approach to real-time full-band speech
  enhancement.
\newblock In {\em 2018 IEEE 20th International Workshop on Multimedia Signal
  Processing (MMSP)}, pages 1--5. Ieee, 2018.

\bibitem{van2009immediacy}
Leaf Van~Boven, Katherine White, and Michaela Huber.
\newblock Immediacy bias in emotion perception: Current emotions seem more
  intense than previous emotions.
\newblock {\em Journal of Experimental Psychology: General}, 138(3):368, 2009.

\bibitem{varghese2015overview}
Ashwini~Ann Varghese, Jacob~P Cherian, and Jubilant~J Kizhakkethottam.
\newblock Overview on emotion recognition system.
\newblock In {\em 2015 international conference on soft-computing and networks
  security (ICSNS)}, pages 1--5. Ieee, 2015.

\bibitem{vastfjall2003subjective}
Daniel V{\"a}stfj{\"a}ll.
\newblock The subjective sense of presence, emotion recognition, and
  experienced emotions in auditory virtual environments.
\newblock {\em CyberPsychology \& Behavior}, 6(2):181--188, 2003.

\bibitem{LoVecchio2021UpdatingTO}
Nicholas~Lo Vecchio.
\newblock Updating the oed on the historical lgbtq lexicon.
\newblock 2021.

\bibitem{walker2009fading}
W~Richard Walker and John~J Skowronski.
\newblock The fading affect bias: But what the hell is it for?
\newblock {\em Applied Cognitive Psychology: The Official Journal of the
  Society for Applied Research in Memory and Cognition}, 23(8):1122--1136,
  2009.

\bibitem{wallace2019allennlp}
Eric Wallace, Jens Tuyls, Junlin Wang, Sanjay Subramanian, Matt Gardner, and
  Sameer Singh.
\newblock Allennlp interpret: A framework for explaining predictions of nlp
  models.
\newblock {\em arXiv preprint arXiv:1909.09251}, 2019.

\bibitem{wang2011emotion}
Manjie Wang and Kimberly~J Saudino.
\newblock Emotion regulation and stress.
\newblock {\em Journal of Adult Development}, 18(2):95--103, 2011.

\bibitem{wang2016twitter}
Wei Wang, Ivan Hernandez, Daniel~A Newman, Jibo He, and Jiang Bian.
\newblock Twitter analysis: Studying us weekly trends in work stress and
  emotion.
\newblock {\em Applied Psychology}, 65(2):355--378, 2016.

\bibitem{wegrzyn2017mapping}
Martin Wegrzyn, Maria Vogt, Berna Kireclioglu, Julia Schneider, and Johanna
  Kissler.
\newblock Mapping the emotional face. how individual face parts contribute to
  successful emotion recognition.
\newblock {\em PloS one}, 12(5):e0177239, 2017.

\bibitem{wei2016cpm}
Shih-En Wei, Varun Ramakrishna, Takeo Kanade, and Yaser Sheikh.
\newblock Convolutional pose machines.
\newblock In {\em Cvpr}, 2016.

\bibitem{werner2011assessing}
Kelly~H Werner, Philippe~R Goldin, Tali~M Ball, Richard~G Heimberg, and James~J
  Gross.
\newblock Assessing emotion regulation in social anxiety disorder: The emotion
  regulation interview.
\newblock {\em Journal of Psychopathology and Behavioral Assessment},
  33(3):346--354, 2011.

\bibitem{wolf2019huggingface}
Thomas Wolf, Lysandre Debut, Victor Sanh, Julien Chaumond, Clement Delangue,
  Anthony Moi, Pierric Cistac, Tim Rault, R{\'e}mi Louf, Morgan Funtowicz,
  et~al.
\newblock Huggingface's transformers: State-of-the-art natural language
  processing.
\newblock {\em ArXiv}, pages arXiv--1910, 2019.

\bibitem{xu2021head}
Mingke Xu, Fan Zhang, and Wei Zhang.
\newblock Head fusion: Improving the accuracy and robustness of speech emotion
  recognition on the iemocap and ravdess dataset.
\newblock {\em IEEE Access}, 9:74539--74549, 2021.

\bibitem{yang2011prediction}
Yi-Hsuan Yang and Homer~H Chen.
\newblock Prediction of the distribution of perceived music emotions using
  discrete samples.
\newblock {\em IEEE Transactions on Audio, Speech, and Language Processing},
  19(7):2184--2196, 2011.

\bibitem{yannakakis2017ordinal}
Georgios~N Yannakakis, Roddy Cowie, and Carlos Busso.
\newblock The ordinal nature of emotions.
\newblock In {\em Affective Computing and Intelligent Interaction (ACII), 2017
  Seventh International Conference on}, pages 248--255. Ieee, 2017.

\bibitem{yaribeygi2017impact}
Habib Yaribeygi, Yunes Panahi, Hedayat Sahraei, Thomas~P Johnston, and
  Amirhossein Sahebkar.
\newblock The impact of stress on body function: A review.
\newblock {\em EXCLI journal}, 16:1057, 2017.

\bibitem{you2016building}
Quanzeng You, Jiebo Luo, Hailin Jin, and Jianchao Yang.
\newblock Building a large scale dataset for image emotion recognition: The
  fine print and the benchmark.
\newblock In {\em Thirtieth AAAI Conference on Artificial Intelligence}, 2016.

\bibitem{Yovanoff2005GradeLevelIO}
Paul Yovanoff, Luke Duesbery, Julie Alonzo, and Gerald~A. Tindal.
\newblock Grade-level invariance of a theoretical causal structure predicting
  reading comprehension with vocabulary and oral reading fluency.
\newblock volume~24, pages 4--12, 2005.

\bibitem{yuan2011automatic}
Jiahong Yuan and Mark Liberman.
\newblock Automatic detection of ``g-dropping'' in american english using
  forced alignment.
\newblock In {\em 2011 IEEE Workshop on Automatic Speech Recognition \&
  Understanding}. Ieee.

\bibitem{yuan2015approach}
Xiaobu Yuan.
\newblock An approach to integrating emotion in dialogue management.
\newblock In {\em International Conference in Swarm Intelligence}, pages
  297--308. Springer, 2015.

\bibitem{zhang2017predicting}
Biqiao Zhang, Georg Essl, and Emily Mower~Provost.
\newblock Predicting the distribution of emotion perception: capturing
  inter-rater variability.
\newblock In {\em International Conference on Multimodal Interaction}, pages
  51--59, 2017.

\bibitem{zhang2014does}
Xuan Zhang, Hui~W Yu, and Lisa~F Barrett.
\newblock How does this make you feel? a comparison of four affect induction
  procedures.
\newblock {\em Frontiers in psychology}, 5:689, 2014.

\bibitem{zhao2019differential}
Jingwen Zhao, Yunfang Chen, and Wei Zhang.
\newblock Differential privacy preservation in deep learning: Challenges,
  opportunities and solutions.
\newblock {\em IEEE Access}, 7:48901--48911, 2019.

\bibitem{zhao2018personality}
Sicheng Zhao, Guiguang Ding, Jungong Han, and Yue Gao.
\newblock Personality-aware personalized emotion recognition from physiological
  signals.
\newblock In {\em Ijcai}, pages 1660--1667, 2018.

\bibitem{zuo2011cross}
Xin Zuo and Pascale Fung.
\newblock A cross gender and cross lingual study on acoustic features for
  stress recognition in speech.
\newblock In {\em ICPhS}, pages 2336--2339, 2011.

\end{thebibliography}
